\pdfoutput=1
\documentclass[12pt,a4paper]{report}

\usepackage{sty/packages}
\usepackage{sty/tocbibind}
\usepackage{sty/thesis_layout}
\usepackage{bm}
\usepackage{subfigure}
\renewcommand{\baselinestretch}{1.5}
\usepackage{float}

\usepackage{longtable}
\usepackage{pdfpages}
\usepackage{filecontents}

\usepackage{graphicx}

\begin{document}
\begin{titlepage}
\begin{center}

\mbox{}\vfill {\Large\bf\sc Distributed Navigation of Multi-Robot Systems for Sensing Coverage}

\vspace{6mm}

A Research Report written by

\vspace{3mm}

{\large {\bf  Waqqas Ahmad}}

\vspace{6mm}

\begin{figure}[H]
  \centering
  \includegraphics{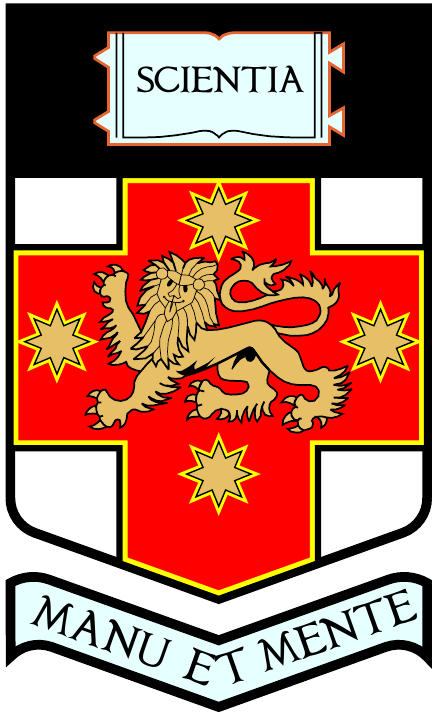}
\end{figure}
{\large School of Electrical Engineering and Telecommunications, \\
The University of New South Wales, Australia}

\vspace{6mm}

{\large {\bf 2016}}

\end{center}

\vfill

\end{titlepage}

\namechapter{Executive Summary}

A team of coordinating mobile robots equipped with operation specific sensors can perform different coverage tasks. If the required number of robots in the team is very large then a centralized control system becomes a complex strategy. There are also some areas where centralized communication turns into an issue. So, a team of mobile robots for coverage tasks should have the ability of decentralized or distributed decision making. This research investigates decentralized control of mobile robots specifically for coverage problems. A decentralized control strategy is ideally based on local information and it can offer flexibility in case there is an increment or decrement in the number of mobile robots. We perform a broad survey of the existing literature for coverage control problems. There are different approaches associated with decentralized control strategy for coverage control problems. We perform a comparative review of these approaches and use the approach based on simple local coordination rules. These locally computed nearest neighbour rules are used to develop decentralized control algorithms for coverage control problems.

We investigate this extensively used nearest neighbour rule based approach for developing coverage control algorithms. In this approach, a mobile robot gives an equal importance to every neighbour robot coming under its communication range. We develop our control approach by making some of the mobile robots playing a more influential role than other members in the team. We develop the control algorithm based on nearest neighbour rules with weighted average functions. The approach based on this control strategy becomes efficient in terms of achieving a consensus on control inputs, say heading angle, velocity, etc.

The decentralized control of mobile robots can also exhibit a cyclic behaviour under some physical constraints like a quantized orientation of mobile robot. We further investigate the cyclic behaviour appearing due to the quantized control of mobile robots under some conditions. Our nearest neighbour rule based approach offers a biased strategy in case of cyclic behaviour appearing in the team of mobile robots.

We consider a clustering technique inside the team of mobile robots. Our decentralized control strategy calculates the similarity measure among the neighbours of a mobile robot. The team of mobile robots with the similarity measure based approach becomes efficient in achieving a fast consensus like on heading angle or velocity. We perform a rigorous mathematical analysis of our developed approach. We also develop a condition based on relaxed criteria for achieving consensus on velocity or heading angle of the mobile robots. Our validation approach is based on mathematical arguments and extensive computer simulations.

\tableofcontents
\listoffigures
\listoftables

\chapter{Introduction}

In recent years, self-organizing systems are developed after biological inspiration. In such systems, the simple local interaction between agents collectively forms a group level behaviour. The group level behaviour is adaptive in the sense that it is purely constructed from the local level iteration. A local level change in the system can be adjusted without changing the whole group level behaviour. These systems have biological examples of animal behaviour like human swarm, schooling of fish, flocking of birds, etc.

\section{Animal Aggregation}
In Ethology, the animal behaviour is studied in a scientific manner to achieve an objective. 
%In the below mentioned figure, fish schooling behaviour has been shown. 
A fish interacts at local level and some adaptive properties emerge at school level \cite{lopez2012behavioural}.

%\begin{figure}[H]
%  \centering
%  \includegraphics[width=10cm]{Ch1/Fish_Schools.pdf}
%  \caption{Fish Schooling - Image Source \cite{lopez2012behavioural}}
%  \label{fig:Fish_Schools}
%\end{figure}

\begin{figure}[H]
  \centering
  \includegraphics[width=10cm]{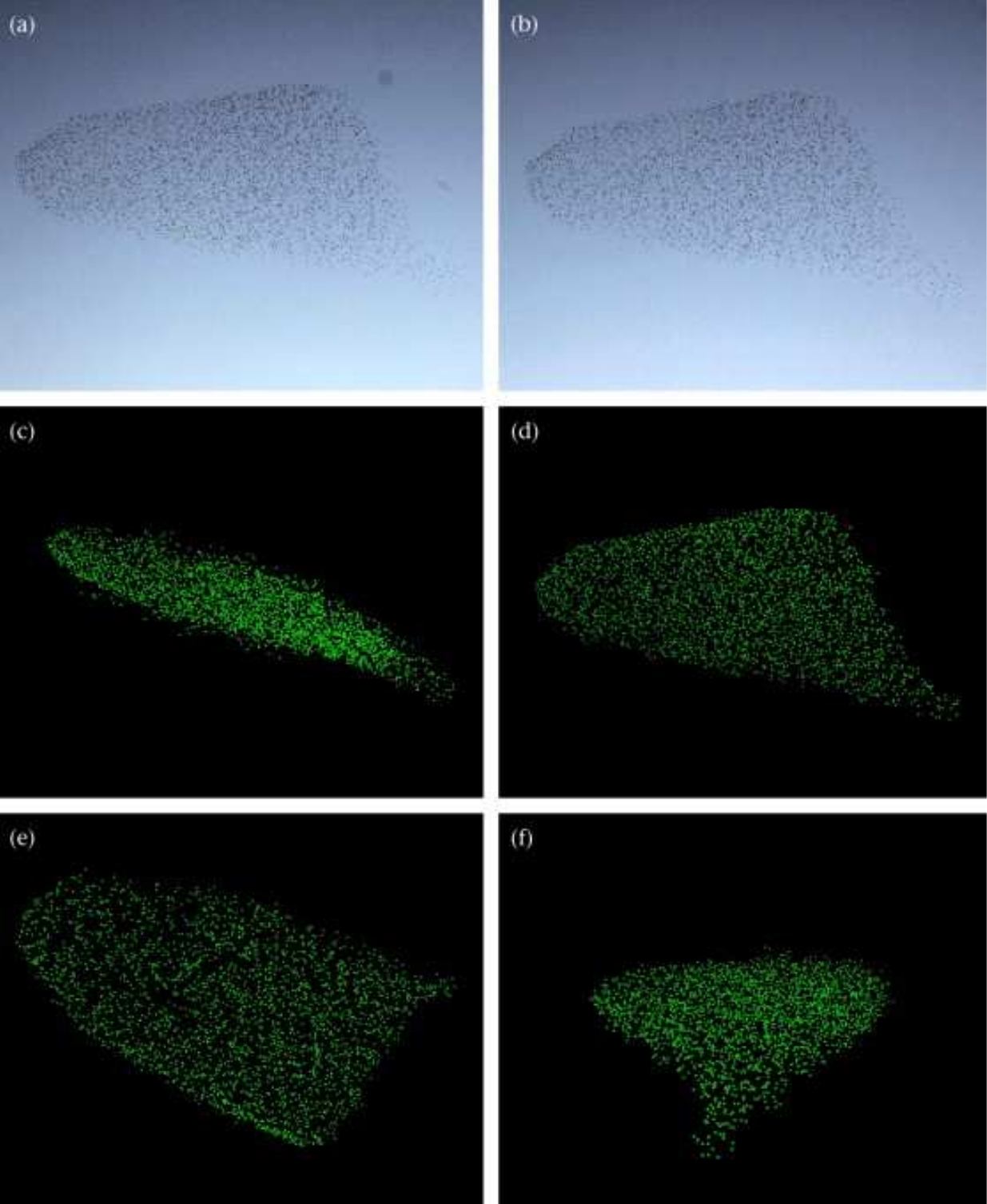}
  \caption{Flocking of Birds - Image Source \cite{ballerini2008empirical}}
  \label{fig:Flocking_of_Birds}
\end{figure}

 A similar phenomenon is observed in flock of birds. Figure \ref{fig:Flocking_of_Birds} from \cite{ballerini2008empirical} shows some examples of such a behaviour. Individual birds maintain a minimum distance from neighbours and collectively form a flocking level behaviour \cite{ballerini2008empirical}.

The above mentioned behaviour of fish schooling and bird flocking is also observed in different animals. The self-organization theory helps in understanding the group level behaviour. It is believed that the complex adaptive patterns at the group level are results of simple repeated interactions at the individual level \cite{sumpter2006principles}.

%and some of these patterns have been shown in Fig. \ref{fig:Animal_Behaviour}  of \cite{sumpter2006principles}.

%\begin{figure}[H]
%  \centering
%  \includegraphics[width=10cm]{Ch1/Animal_Behaviour.pdf}
%  \caption{Animal Behaviour - Image Source \cite{sumpter2006principles}}
%  \label{fig:Animal_Behaviour}
%\end{figure}

\section{Research Inspiration}

Some mathematical models have been developed after inspiring from the above mentioned behaviours. Our research is based on decentralized control of such a group level behaviour. We specifically develop algorithms based on local coordination rules among mobile robotic sensors. Mobile robots have physical constraints like heading angle, velocity, etc. Although the decentralized control strategy is inspired from animal aggregation, but its implementation with mobile robotic sensors is a quite challenging and emerging research area. The inspiration from animal aggregation can be used to develop different coverage algorithms. In this research, we investigate the decentralized control of mobile robotic sensors mainly focusing on coverage algorithms.

\section{Research Main Contributions}

A decentralized control of mobile robotic sensors for coverage problems can be achieved with different approaches. However, the main challenge is that the algorithm/ control law should ideally proceed without any global information and the decision making process should be performed in a decentralized manner. The control should also consider the physical constraints of a mobile robot. In this context, we perform a broad literature survey on decentralized control specifically for the coverage problems. We find different control and deployment approaches used in the literature. Wherever necessary, we review these approaches with an explicit comparison. Such a comparative review might help in clearly defining and evaluating coverage control problems. We find the practical and potential applications of the coverage control problems. We formulate the classification of these problems. We review each considered coverage control problem and the approaches used for its solution. Then, we review the advantages and disadvantages associated with a particular approach. We also list the considerations typically found in a coverage control problem. Such a qualitative literature survey leads us in identifying the research gaps in these problems.

We choose the approach based on nearest neighbour rules and we find this approach considers fully local information. With this approach, a global objective is achieved with fully local information exchange and a decentralized decision making is possible without the need of any central processor. This approach is also capable of locally incorporating any addition or subtraction in the number of mobile robots. So, the nearest neighbour rule based approach can deal a mobile robot failure during a coverage control mission. We also find the nearest neighbour rule based approach is widely used in coverage control problems. In this approach, a mobile robot gives an equal importance to a neighbour mobile robot coming under its communication range. We amend this approach as nearest neighbour rule with weighted average functions. Consequently, a mobile robot can give a higher importance to a neighbour mobile robot depending on some criteria. Such an amendment makes the coverage algorithms efficient in certain manners. Like, we validate one of our coverage control algorithm based on the amended approach and we perform a clear comparison with the existing approach considered in the literature. We find this novelty can reduce the number of linear iterations to achieve the coverage goal and it makes the algorithm quite efficient.

We consider the implementation of nearest neighbour rule based approach with a practical point of view. Like, the orientation or heading angle of a mobile robot can require a quantized control. Such a quantized control can lead to a cyclic behaviour. We consider the problem of this cyclic behaviour and we introduce a biased strategy. The algorithm based on this biased strategy can avoid the cyclic behaviour caused by a physical constraint. Thus, our algorithm can be implemented on a physical team of coordinating mobile robots.

We bring the concept of a similarity measure from social sciences into the team of mobile robots. We further develop our approach based on nearest neighbour rule with weighted average functions and we calculate the weighted average functions based on the similarity measure among the neighbours of a mobile robot. The incorporation of such a similarity measure among the team of mobile robots makes the coverage algorithm quite fast. The team of mobile robots can achieve the consensus; say on the heading angle or velocity, with a significantly reduced number of linear iterations. We validate our consideration mathematically and with extensive computer simulations. We compare our algorithm with the relevant coverage control problem and find our algorithm can behave fast due to an incorporation of the similarity measure.

Finally, we develop a relaxed control criteria which is based on a necessary and sufficient condition to validate whether a consensus (say on heading angle, velocity, etc.) is possible or not. There is already a necessary and sufficient condition on a strictly considered system available in the literature. But our condition is based on a mild criteria and it can be applied on certain coverage control problems. We analyse it mathematically similar to the existing one in the literature. In addition, we also validate it through extensive computer simulations proving the state convergence based on our relaxed criteria. Our considered criteria reduce the number of linear iterations to reach a state consensus value. A good example is that a consensus on the orientation of mobile robots can be achieved with reduced number of linear iterations and it can be applied in coverage control problems meeting certain criteria. These simple criteria can also be useful to design weighted average functions for the coverage control problems.

Our research report structure is based on chapters and we define a chapter problem with the respective problem statement. Each chapter shows its own introduction section, the related literature in the context of considered problem statement, our developed approach and the novelty of our approach with a clear comparison from the literature. Chapter 2 provides a comprehensive survey of the literature published for coverage control algorithms. We review different approaches to achieve a coverage objective. In Chapter 3, we make some of the individual robots more influential and investigate a group level behaviour. We specifically apply our approach on a coverage algorithm and validate its advantages in the algorithm. We address the physical constraint of a mobile robot in chapter 4. We review the local coordination model for its implementation on a team of mobile robots. We address the cyclic behaviour arising due to the physical constraints of mobile robots. In Chapter 5, we introduce a consideration of similarity measure among the individual mobile robots to collectively achieve a consensus typically required for a coverage control algorithm. The detailed mathematical analysis and criteria based condition has been provided in Chapter 6. Chapter 7 provides an over-all research based study conclusion and future research directions with some justified references from the literature.

\chapter{Literature Survey}

We survey decentralized control of mobile robotic sensors deployed for area coverage problems. The mobile robotic sensors are equipped with on-board sensing, communication and computation capability. The mobile robots are deployed in different configurations to sense an area. Such a control needs to be adaptive and it should be ideally based on the local information with decentralized decision making. The existing literature uses different approaches to develop such sort of algorithms or control laws. In this area, we investigate the classification of literature and provide a comparative review of the approaches used for certain coverage control problems. We also find the mathematical definition and formulation of the coverage problems. We survey the optimal coverage pattern during deployment of mobile robotic sensors. We classify the literature according to the coverage goal to be achieved. We further sub-classify a coverage goal on the grounds of an approach to be used in the existing literature. We also develop a comprehensive list of the considerations found in these problems. By using the itemized list of considerations, one can easily define the objective of the problem with a clear direction and evaluate the effectiveness of its solution by making a comparative review with other approaches considered in the literature.

\begin{flushleft}
%\section{\textbf{Introduction}}
\section{Introduction}
\end{flushleft}

The decentralized control of a team of autonomous mobile robots performing different tasks has been widely studied in the existing literature. In this control, a team member coordinates with other team members and makes an autonomous decision in a decentralized fashion. The main application of this sort of control is highly desirable where number of robots is very high and it becomes quite complex to achieve control from a central system. If there is a change in the central control unit, it would affect all the team members. There are also areas where centralized communication exchange is a problem and a decentralized control based on local communication becomes a justified solution. The decentralized control also offers flexibility in terms of addition of robots in an existing team and the reduction of team member(s) due to a failure. So, the decentralized control offers this sort of adaptability among team members.

We mainly focus our literature survey on area coverage with a team of mobile robotic sensors having different objectives. We present a few potential applications of some of these coverage problems in Section \ref{Applications}. We review the main focus of relevant literature surveys in Section \ref{Relevant_Literature} and the ways of literature classification performed in some of these surveys:

The survey \cite{younis2008strategies} categorises the nodes placement strategies into static and dynamic positioning schemes depending on whether the optimization is performed at the time of deployment or while the network is operational, respectively. The authors further classify the static deployment according to the deployment methodology, optimization objective and roles of the nodes. The deployment methodology can be either deterministic or random.

The coverage and connectivity is one of the fundamental considerations in deployment strategies. The work \cite{zhu2012survey} surveys the coverage and connectivity by considering deployment strategies, sleep scheduling mechanism and adjustable coverage radius. The authors with references therein classify the deployment strategy into static coverage and dynamic coverage. The static coverage is further sub-classified into efficient coverage area, k-coverage and path coverage. The authors classify the dynamic coverage based on the approaches: virtual force, graph based and repair policies of coverage hole.

The survey \cite{zhu2014survey} covers the communication and data management issues in mobile wireless sensor networks. The work compares the literature with respect to topology control methods, coverage methods, localization methods, target tracking methods, data gathering methods and data replication methods. Specifically, the coverage is achieved by two methods: self-deployment method and relocation method. This survey with references therein further categorises the self-deployment method into movement-assisted methods, potential-field methods and virtual-force methods. In relocation method, the redundant nodes are relocated to fill the positions of the failed nodes. The survey further categorises the relocation method as grid-quorum method, zone flooding method, and mesh-based method.

The work \cite{sangwan2015survey} surveys coverage strategies (with references therein) as grid strategy, computational geometry, target coverage, virtual force strategy, k-coverage, path coverage, three-dimensional coverage and network lifetime maximization. The authors classify the literature with coverage type as area coverage/ barrier coverage/ target coverage and coverage algorithm characteristics as centralized/distributed/ localized.

In this survey, we classify the literature on the grounds of coverage objective to be achieved by the team of mobile robots. Such an objective could be to achieve Blanket, Barrier, Sweep, Encircling, Three-dimensional or Dynamic Coverage. This classification has been reviewed in detail under Section \ref{Classifcation}. We further sub-classify the coverage objective on the basis of type of environment for which a control strategy has been considered. Such a sub-classification has been explained under the relevant coverage problems. The control strategy for these coverage problems may be achieved with different approaches. So, we sub-classify the coverage problems on the grounds of approaches used in the literature. We also consider these approaches and review it in a comparative manner. Our classification of literature and comparative review of the approaches for the coverage problems are different from the relevant literature surveys presented in Section \ref{Relevant_Literature}.

The remainder of the literature survey is organized section-wise. We survey the problem of Blanket Coverage under Section \ref{Blanket_Coverage}, Barrier Coverage under Section \ref{Barrier_Coverage}, Sweep Coverage under Section \ref{Sweep_Coverage}, Heuristic Coverage under Section \ref{Heuristic_Coverage}, Dynamic Coverage (Search and Rescue by Multi-robots) under Section \ref{Dynamic_Coverage}, Three-dimensional (3D) Coverage under Section \ref{3D_Coverage}, coverage based on Formation Building under Section \ref{Formation_Building} and Encircling Coverage part of Mobile Actuator Sensor Network under Section \ref{MAS-net}. Then, we briefly describe the robot kinematic models under Section \ref{Implementation_Robots}. We also review the optimal deployment pattern (Section \ref{optimal_grid_pattern}) and develop an overall comprehensive list of the considerations (Section \ref{Conclusion_Future}) found in these decentralized/ distributed coverage control problems.

 \newpage

\subsection{Applications of Coverage}\label{Applications}
%A number of mobile robots cooperating in a team\footnote{The terms, "agent" or "sensor" or "the mobile robotic sensor" or simply, "the mobile robot" will be used throughout this paper for an autonomous mobile robot having an on-board computation, operation-specific sensing, nearest neighbour position sensing and communication capability.} with limited local information has got different applications like surveillance of an area, reconnaissance, maintenance job and inspection in hazardous areas [21-22].

The mobile robotic sensors \footnote{The terms, "agent" or "sensor" or "the mobile robotic sensor" or simply, "the mobile robot" will be used throughout this chapter for an autonomous mobile robot having an on-board computation, operation-specific sensing, nearest neighbour position sensing and communication capability.} with different coverage types have got practical applications: surveillance of an area, reconnaissance, maintenance job, inspection in hazardous areas, mine deployment, mine sweeping, surveillance, sentry duty, maintenance inspection, ship hull cleaning, communications relaying \cite{gage1992command}, boarder patrolling \cite{kumar2005barrier}, environmental studies, detecting and localizing the origin of hazardous chemicals leakage or vapour emission, finding sources of pollutants and plumes, environmental monitoring of disposal sites on the deep ocean floor \cite{jeremic1998design}, sea floor surveying for hydrocarbon exploration \cite{borhaug2007straight}, ballistic missile tracking, bush fire monitoring, oil spill detection at high seas, environmental extremum seeking \cite{cochran2009nonholonomic,matveev2011navigation,zhu2014cooperative}, environmental filed level tracking \cite{matveev2012method}, target capturing \cite{zakhar2015distributed} and many others. A good example of hazardous areas coverage is mine sweeping, which is an extremely challenging and dangerous task \cite{gage1992command,cassinis1999strategies,acar2003path}.

%Other potential applications of similar sweeping are boarder patrolling \cite{kumar2005barrier}, environmental monitoring of disposal sites on the deep ocean floor \cite{jeremic1998design}, and sea floor surveying for hydrocarbon exploration \cite{borhaug2007straight}, robot navigation for environmental extremum seeking \cite{cochran2009nonholonomic,matveev2011navigation} and environmental filed level tracking \cite{matveev2012method}.

We provide the following table listing above mentioned potential applications according to the coverage type. However, we explain the detail of each coverage type in the subsequent sections.

\begin{table}[!htbp]
\begin{center}
    \begin{tabular}{ |  p{4.5cm} | p{8.5cm} |}
    \hline
   \textbf{Coverage Type}                                    &   \textbf{Potential Applications}                                                                                     \\ \hline
    Blanket Coverage                                         &    Border Protection, Communications Relay, Ship Hull Cleaning                                 \\ \hline
    Barrier Coverage                                         &    Mine Deployment, Sentry Duty, Border Surveillance                                                                  \\ \hline
    Sweep Coverage                                           &  Multi-robotic Mine Sweeping, Reconnaissance,
                                                                Maintenance Inspection, Carrier Deck FOD Disposal,
                                                                Ship Hull Cleaning, Environmental Extremum Seeking, Environmental Field Level Seeking, Monitoring of Disposal Sites on the Deep Ocean Floor, Sea Floor Surveying for Hydrocarbon Exploration, Border Patrolling                                                                                                      \\ \hline
    Heuristic Coverage                                        & Heuristic Coverage Algorithms, Heuristic Sweeping on the Floor                                                                                                                       \\ \hline
    Dynamic Coverage (Search and Rescue by Multi-robots)      & Search and Rescue Missions                                                                                             \\ \hline
    Three-dimensional Coverage                                & Three-dimensional Ocean Space Coverage, Formation based Coverage of UAVs in Three-dimensional Space                                                                                                                        \\ \hline
    Formation Building                                        & Mine Sweeping, Border Patrolling, Environmental Monitoring of Disposal Sites on the Deep Ocean Floor, Sea Floor Surveying for Hydrocarbon Exploration                                                                                                                       \\ \hline
    Mobile Actuator and Sensor Network (Encircling Coverage)  &    Detecting and Localizing Hazardous Chemical Leakage/ Oil Spill/ Vapour Emission, Sources of Pollutants and Plumes, Environmental Studies, Hydrothermal Vents, Disposal Sites on the Deep Ocean Floor, Sea Floor Surveying for Hydrocarbon Exploration                                                                    \\  \hline

    \end{tabular}
    %\caption{Voronoi Partitioning}
%\label{fig:vor}
\end{center}
\caption{Coverage Classification and Applications}
\end{table}

\newpage

\subsection{Related Survey Papers}\label{Relevant_Literature}

In this section, we provide a brief outline of the related literature surveys as under:

\begin{longtable}{|l|p{10cm}|}
\caption{A Brief Overview of Related Surveys}\\
\hline
\textbf{Literature Survey} & \textbf{Survey Focus} \\
\hline
\endfirsthead
\multicolumn{2}{c}
{\tablename\ \thetable\ -- \textit{Continued from previous page}} \\
\hline
\textbf{Literature Survey} & \textbf{Survey Focus} \\
\hline
\endhead
\hline \multicolumn{2}{r}{\textit{Continued on next page}} \\
\endfoot
\hline
\endlastfoot

    \cite{fiorini2000cleaning}-2000                   &    Technological approaches towards cleaning robots                                   \\ \hline
    \cite{choset2001coverage}-2001                    &    Coverage path planning algorithms                                                  \\ \hline
    \cite{ahmed2005holes}-2005                        &    Coverage holes: types, characteristics, effects                                    \\ \hline
    \cite{skoglar2007uav}-2007                        &    Planning and control approaches for optimal estimation, search and exploration     \\ \hline
    \cite{gajbhiye2008survey}-2008                    &    Network architectures and deployment strategies                                    \\ \hline
    \cite{younis2008strategies}-2008                  &    Sensor node placement strategies and techniques                                    \\ \hline
    \cite{ghosh2008coverage}-2008                     &    Algorithms and techniques to address the coverage and connectivity issues          \\ \hline
    \cite{daneshfar2009multi}-2009                    &    Control engineering applications with multi-agent systems                          \\ \hline
    \cite{chen2009deployment}-2009                    &    Comparison among random, incremental and movement assisted deployment algorithms   \\ \hline
    \cite{leitner2009multi}-2009                      &    Multi-robot cooperation techniques for space applications                          \\ \hline
    \cite{wang2009survey}-2009                        &    Optimizing network coverage                                                        \\ \hline
    \cite{brown2010region}-2010                       &    Sensors positioning for coverage and protection                                    \\ \hline
    \cite{mulligan2010coverage}-2010                  &    Coverage issues, approaches and literature comparison                              \\ \hline
    \cite{fan2010coverage}-2010                       &    Design considerations for coverage problems                                        \\ \hline
    \cite{argany2010voronoi}-2010                     &    Voronoi diagram and Delaunay triangulation for geosensor network optimization      \\ \hline
    \cite{norouzi2011integrated}-2011                 &    Network architecture and power management                                          \\ \hline
    \cite{guvensan2011coverage}-2011                  &    Categorizing coverage optimization solutions                                       \\ \hline
    \cite{wang2011movement}-2011                      &    Mending, defense, sweeping barriers and a review of movement strategies            \\ \hline
    \cite{zhu2012survey}-2012                         &    Coverage and connectivity issues with respect to deployment, sleep scheduling and coverage radius                                                            \\ \hline
    \cite{wang2012three}-2012                         &    Deployment, localization, topology and position based routing for 3D ocean sensor networks                                                                    \\ \hline
    \cite{li2013survey}-2013                          &    Topology control techniques and its classification                                 \\ \hline
    \cite{zhang2013survey}-2013                       &    Techniques for multiple unmanned vehicles formation control and coordination       \\ \hline
    \cite{galceran2013survey}-2013                    &    Coverage path planning approaches                                                  \\ \hline
    \cite{yan2013survey}-2013                         &    Multi-robot coordination                                                           \\ \hline
    \cite{mamatha2013deployment}-2013                 &    Classification of node deployment algorithms and comparison of its deployment techniques                                                                         \\ \hline
    \cite{liu2013robotic}-2013                        &    Control methodologies for robotic urban search and rescue mission                  \\ \hline
    \cite{zhu2014survey}-2014                         &    Communication and data management issues                                           \\ \hline
    \cite{khoufi2014survey}-2014                      &    Coverage and connectivity issues in deployment algorithms                          \\ \hline
    \cite{chen2014coverage}-2014                      &    Classification of coverage algorithms and issues in a UAV network                  \\ \hline
    \cite{hoy2015algorithms}-2015                     &    Review and classification of approaches for collision-free navigation in unmanned vehicles                                                                       \\ \hline
    \cite{sangwan2015survey}-2015                     &    Classification of coverage problem and strategies used for coverage                 \\ \hline
    \cite{robin2015multi}-2015                        &    Classification of target management and approaches used for its detection or tracking\\ \hline
    \cite{tan2015survey}-2015                         &    Modelling methods for swarm robotics and swarm robotic algorithms for flocking, navigating and searching applications                                              \\ \hline
\end{longtable}

\section{Classification of Coverage Control Problems}\label{Classifcation}

Coverage can be achieved with the static arrangement of mobile robotic sensors or with the group motion of the mobile robotic sensors. Gage \cite{gage1992command} has classified the coverage problem mainly into three configurations Blanket, Barrier and Sweep Coverage. But coverage can be of different types, like in \cite{savkin2013algorithm}, an encircling coverage has been introduced. So, we classify the below mentioned coverage problems on the basis of an objective to be achieved.

\begin{itemize}
  \item Blanket Coverage
  \item Barrier Coverage
  \item Sweep Coverage
  \item Heuristic Coverage
  \item Dynamic Coverage (Search and Rescue by Multi-robots)
  \item Three-dimensional (3D) Coverage
  \item Formational Building Coverage
  \item Mobile Sensor Actuator Network (Encircling Coverage)
\end{itemize}

The above coverage types may be further classified on the basis of considered coverage environment and sub-classified on the basis of approaches used to achieve this objective.

\section{Blanket Coverage}\label{Blanket_Coverage}
If the region of interest is covered by static arrangement of mobile robotic sensors to maximize the detection rate of an intruder, such sort of coverage is known as Blanket Coverage \cite{gage1992command}.

In \cite{wang2008coverage}, the authors present a distributed algorithm to optimally position the mobile robotic sensors. This work considers a network for static and mobile robotic sensors. In order to reduce the cost, this hybrid network structure considers higher number of static sensors and lower number of mobile sensors.

\subsection{Blanket Coverage Classification}
We sub-classify the problem of blanket coverage on the basis of the type of region to be covered.
\subsubsection{Blanket Coverage over Boundary Line Environment}
In this environment, the coverage is considered between a corridor defined by boundary lines. The boundary lines may also be arbitrary in nature.

In \cite{cheng2013decentralized}, the authors consider a circular model for sensing range $R_{s}$ and communication range $R_{c}$. The communication range and sensing range is related as $R_{c} \geq \sqrt{3} R_{s}$. These sensors are driven to achieve blanket coverage in a two-dimensional corridor defined between the two boundary lines. The work considers deployment of sensors achieving the triangular lattice pattern, which is considered to be optimal with minimum number of sensors required for full coverage of a bounded set  \cite{kershner1939number}. Both achieving full coverage and maintaining connectivity among nodes are important issues in wireless sensor networks \cite{ghosh2008coverage}. When $R_{c} \geq \sqrt{3} R_{s}$, the triangular lattice pattern provides not only 1-coverage, but also 6-connectivity among the sensor nodes (except the boundary sensors). The 1-coverage means that every point in the region falls under the sensing range ($R_{s}$) of at least one node. The 6-connectivity provides a communication of every sensor node (except boundary sensor nodes) with six neighbouring sensors. The work deploys sensor nodes with the triangular lattice pattern as shown in the below mentioned figure. We also explain the triangular lattice pattern under the subsection \ref{optimal_grid_pattern}.

\begin{figure}[H]
  \centering
  \includegraphics[width=10cm]{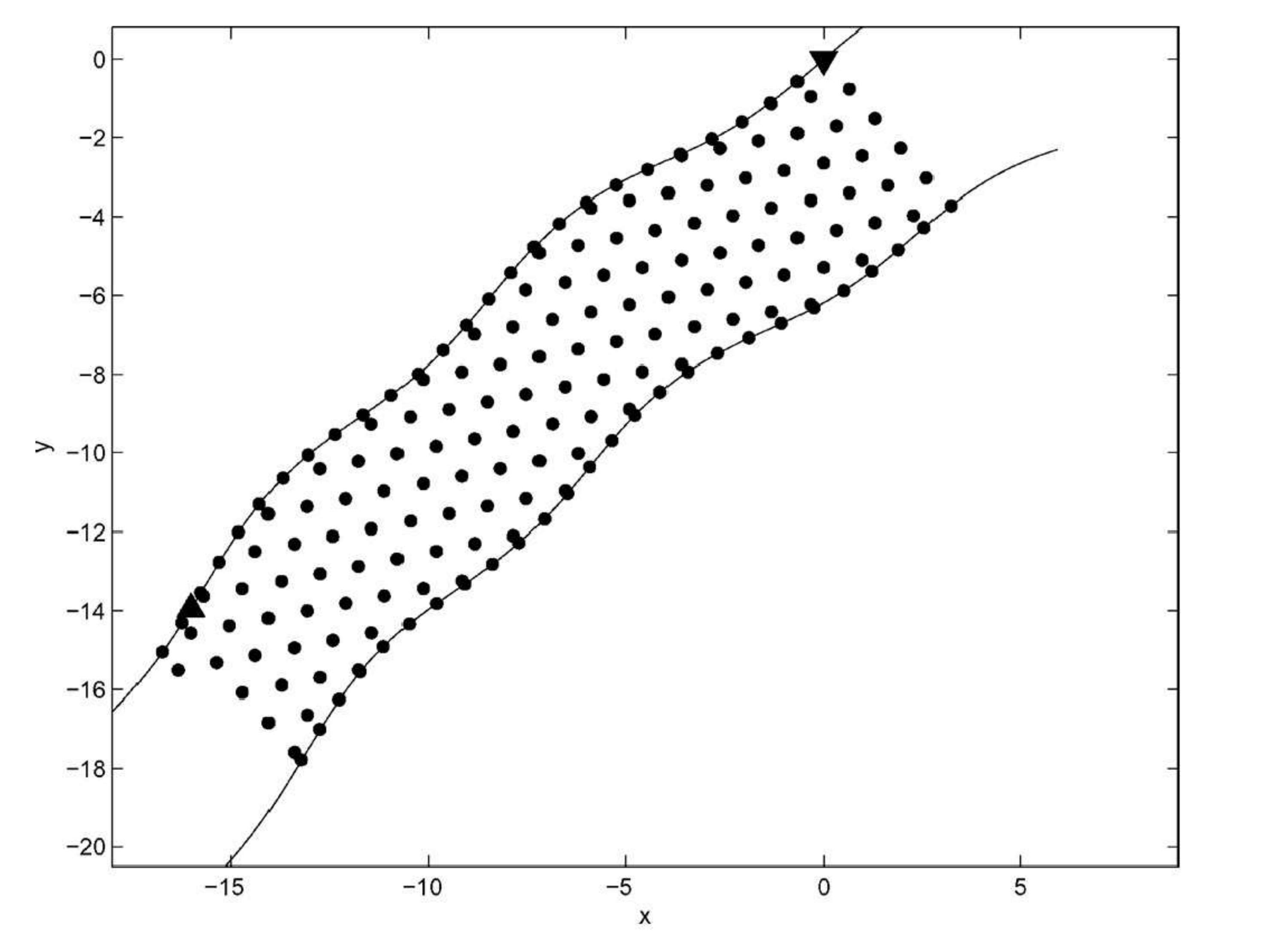}
  \caption{Blanket coverage between two boundaries \cite{cheng2013decentralized}}
  \label{fig:Blanket_Boundary}
\end{figure}

Now, we formally define the problem of blanket coverage over the boundary line environment as stated in \cite{cheng2013decentralized}:

Let $m(R_{s})$ be the number of sensors that covers a bounded region $R \subset \mathbb{R}$ and $P(R_{s}) = {h_{i, j(R_{s})}}$ be the corresponding set of desired sensor locations in $\mathbb{R}$. Let $N(R_{s})$ be the minimum number of sensors to cover $\mathbb{R}$. The locations $P(R_{s})$ is said to be asymptotically optimal for covering $\mathbb{R}$ if for all $p \in \mathbb{R}^2$, there exists $h_{i, j}(R_{s}) \in P(R_{s})$ such that

\begin{equation}\label{eq:Blanket_Boundary}
 \parallel p - h_{i,j}(R_{s}) \parallel \leq R_{s} \\
\end{equation} and

\begin{equation}\label{eq:Blanket_Boundary_Min_Sen}
 \lim_{R_{s}\to \infty } R_{s}^2 m(R_{s}) = \lim_{R_{s}\to 0 } R_{s}^2 N(R_{s})\\
\end{equation}

\begin{equation}\label{eq:Blanket}
 \lim_{k\to \infty } \parallel p_{z}(kT) - h_{i,j} \parallel=0\\
\end{equation}

The authors finally introduce a set of decentralized control laws for heading angle $\theta_{i}(kT)$ and velocity $v_{i}(kT)$. The decentralized control laws prove the blanket coverage between two straight lines, $L_{1}$ and $L_{2}$. Next, the authors extend the work between smooth curves instead of straight lines. All the homogeneous sensors in terms of $R_{s}$ and $R_{c}$ have been considered. There might be some consideration for a static or moving obstacle. A similar boundary line environment has been considered in \cite{santoso2011decentralised}.

\subsubsection{Blanket Coverage over Convex/ Arbitrary Region}

The coverage area can be convex or arbitrary in nature. A similar coverage problem for a bounded and connected two-dimensional region has been considered by a number of researchers (see e.g. \cite{cheng2010decentralizedregion,santoso2010sub,savkin2012optimal,ahmad2012decentralized}). We formally define the blanket coverage problem for a region \cite{cheng2010decentralizedregion} as under:

\begin{figure}[H]
  \centering
  \includegraphics[width=10cm]{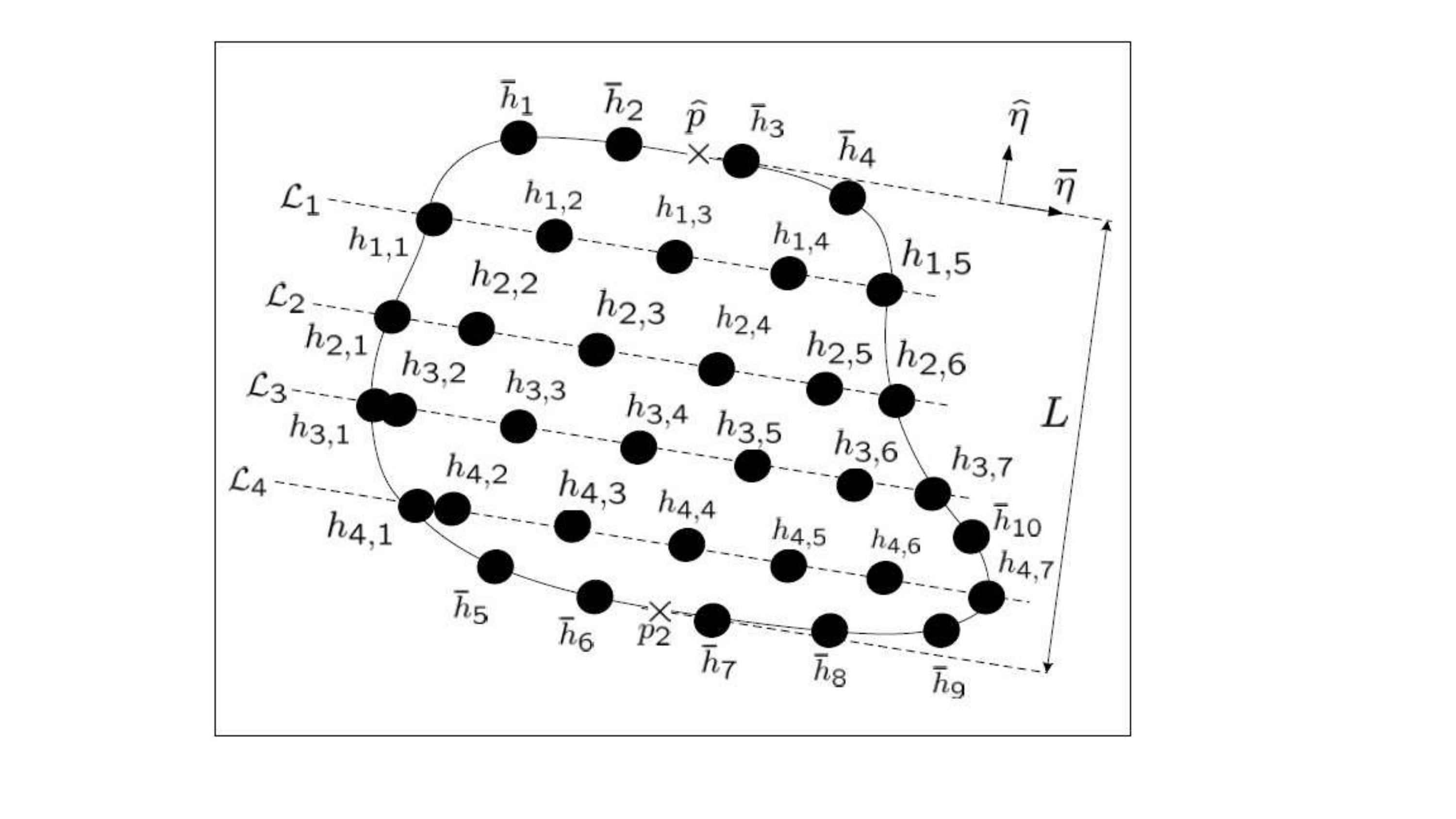}
  \caption{Blanket coverage over a region \cite{cheng2010decentralizedregion}}
  \label{fig:Blanket_Region}
\end{figure}

Given a region $R$ and $n$ mobile sensors, a set of distributed control laws is said to be a triangular blanket coverage control for the network of mobile sensors in $R$ if for almost all initial sensor positions, and for each location $h_{i,j}$ with $i \in {1, 2, . . . ,K}$ and $j \in {1, 2, . . . , n}$, there exists a unique index $z \in {1, 2, . . . , n}$ such that:

\begin{equation}\label{eq:Blanket}
 \lim_{k\to \infty } \parallel p_{z}(kT) - h_{i,j} \parallel=0\\
\end{equation}

\subsection{Blanket Coverage Approaches}
We categorise the problem of blanket coverage on the basis of approaches used in the literature.
\subsubsection{Blanket Coverage based on Nearest Neighbour Rule}\label{BC:NNR}

In this local rule, a variable is updated on the average of its own plus the variables of its neighbours. The authors in \cite{vicsek1995novel} proposed a discrete time system of $1,2,...,n$ autonomous agents moving in a plane with the same speed, but having different heading angles. A further theoretical explanation of the behaviour observed in \cite{vicsek1995novel} has been provided in \cite{jadbabaie2003coordination}. An agent's heading angle is updated as the average of its own heading plus the headings of its neighbours. The $j$ neighbours are defined as the agents which come under circular communication range of agent $i$. Let ${\cal N}_{i}(kT)$ be the set of neighbours of $i$ at time $kT$. Then, this nearest neighbour rule is written as:

\begin{equation}\label{eq:jadbabie}
 {\Theta}_{i}(kT) :=\dfrac{1}{|1+{\cal N}_{i}(kT)|}\left(\phi_i(kT)+\sum_ {j\in {\cal N}_{i}(kT)}\phi_{j}(kT)\right)
\end{equation}

for $i=1,2,3,\ldots, n$, where $\phi_i(kT)$ denotes agent $i$ heading angle at a particular time $kT$.

Then, an agent updates its heading angle using (\ref{eq:jadbabie}).

\begin{equation}\label{eq:phi_update}
 \theta_{i}((k+1)T):= \Theta_{i}(kT)
\end{equation}

The motivation from nearest neighbour rule has been taken to develop control algorithms for blanket coverage (see, e.g.\cite{cheng2013decentralized,santoso2011decentralised,cheng2010decentralized,santoso2010sub,ahmad2012decentralized,santoso2015range,ahmad2015decentralized}). One can notice the nearest neighbour rule is based on local information. Thus, the algorithms based on this approach are decentralized or distributed in nature.

\subsubsection{Blanket Coverage based on Artificial Potential Field (APF)}
In this approach, the mobile sensors are subjected to a virtual potential field, which causes attraction of a sensor towards a goal and a repulsion from the obstacles and other nodes. This way, the network of mobile robots is scattered itself to cover an area. Such artificial potential field based algorithms are distributed. The potential field based approach is easy for implementation, but it becomes computationally complex for a large sensor network. The potential functions based approach is being used in a team of mobile robots (see e.g. \cite{howard2002mobile,popa2004robotic,tanner2005towards,song2002potential,wachter2008potential,gayle2009multi}).

In the context of mobile robots application, the philosophy of an artificial potential field is introduced in \cite{khatib1986real} - where a manipulator travels in a field of forces, the goal to be achieved is an attractive pole for the end effector and considered obstacles are causing repulsive forces on the manipulator parts. In the context of blanket coverage, a self-deployment algorithm based on artificial potential fields is considered in \cite{poduri2004constrained} - which is distributed, scalable and does not need any information about the environment a priori.

Mathematically, we describe the potential field based approach of \cite{poduri2004constrained} as under:\\
A sensor node "i" experiences a force $F$, which is a gradient of scalar potential field $U$, i.e.
\begin{equation}\label{eq:Force_pot}
F= - \Delta U
\end{equation}
Let $F_{cover}(i,j)$ represents force of repulsion to increase coverage and $F_{degree}(i,j)$ denotes force of attraction to constrain degree of a mobile node. $F_{cover}(i,j)$ and $F_{degree}(i,j)$ are constructed as inversely proportional to the square of Euclidean distance (say $d_{ij}$) between nodes $i$ and $j$, and these forces have the following extreme conditions:\\

\begin{itemize}
  \item When the distance between nodes is zero then to avoid collision $\parallel F_{cover} \parallel \rightarrow \infty$.
  \item When the distance between nodes is $R_{c}$ then to avoid loss of connectivity $\parallel F_{degree} \parallel  \rightarrow \infty$.
\end{itemize}
Let $p_{i}$ and $p_{j}$ be the positions of sensors $i$ and $j$, respectively. So, $d_{ij}=\parallel p_{i}-p_{j} \parallel$.
Then,
\begin{equation}\label{eq:Force_cover}
F_{cover}(i,j)= (\frac{-K_{cover}}{d_{ij}^2})(\frac{p_{i}-p_{j}}{d_{ij}})
\end{equation}

%$\hat{n}_{ij}$

\begin{equation}\label{eq:Force_degree}
F_{degree}(i,j)=\begin{cases}
    \frac{K_{degree}}{(d_{ij}-R_{c})^2}(\frac{p_{i}-p_{j}}{d_{ij}}), & \text{ for critical connection}\\
    0, & \text{otherwise}\\
  \end{cases}
\end{equation}

The force constants are represented by $K_{cover}$ and $K_{degree}$. The resultant force between nodes $i$ and $j$ can be written as:

\begin{equation}\label{eq:Force_Resultant}
F_(i,j)=F_{cover}(i,j)+F_{degree}(i,j)
\end{equation}

Finally, a sensor $i$ experiences a net force as under:

\begin{equation}\label{eq:Force_Net}
F_i=\sum_{\text{all j neighbours}} F_(i,j)
\end{equation}

The sensor $i$ can follow the below mentioned equation of motion:

 \begin{equation}\label{eq:Motion_Eq}
\ddot{x}_i(t)=\frac{(F_i-\nu\dot{x}_i)}{(m)}
\end{equation}

where $\nu$ represents a damping factor and $m$ is the virtual mass (assumed to be 1) of the sensor.

 In the above algorithm, each sensor uses a combination of $F_{cover}$ and $F_{degree}$ to maximize the coverage and maintaining connectivity with at least $K$ neighbours. If a sensor has more than $K$ neighbours (critical neighbours) then it will repel its neighbours till only $K$ are left. When the distance between a sensor and its critical neighbours increases then $\parallel F_{cover} \parallel$ decreases and $\parallel F_{degree} \parallel$ increases. Hence, the net force (equation: \ref{eq:Force_Resultant}) becomes zero at some distance $\eta R_{c}$, where $0 < \eta < 1$. So, the sensors and its neighbours develop equilibrium with respect to each other at the distance, $\eta R_{c}$. The computational details of $\nu, \eta, K_{cover}, \text{ and } K_{degree}$ can be found in \cite{poduri2004constrained}.

%\Cite{Khatib1986Real,Howard2002Mobile,,Poduri2004Constrained}---4

%\cite{poduri2004constrained,howard2002mobile,olfati2002distributed,ogren2004cooperative}---5

%\cite{olfati2002distributed,ogren2004cooperative}---Formation Control
%\cite{howard2002mobile,poduri2004constrained,khatib1986real}---6

%\cite{heo2005energy},\cite{wang2006movement}---Voronoi based---7

\subsubsection{Blanket Coverage based on Virtual Force Field (VFF)}

The virtual force approach is based on an artificial force of attraction and repulsion introduced among the mobile robotic sensors. These forces of attraction and repulsion are used to maximize the coverage on the sensing field of interest. This approach can spread the mobile robotic sensors after initial random deployment and it has got different military applications.

In \cite{howard2002mobile}, a potential-field based approach has been used for deployment. A sensor node exerts a force of repulsion to other sensors nodes and obstacles. Thus, the mobile nodes are scattered to maximize the coverage in a distributed and scalable manner. In \cite{locatelli2002packing}, the problem of packing "n" equal circles into the unit square has been considered. This problem is to select the "n" positions inside the unit square in a manner that the minimum pairwise distance between two points is maximized. A similar motivation is taken in virtual force based coverage approach. A virtual force field is created around the mobile robot itself (see e.g. \cite{wang2006variable,liu2006force,wang2007pso,rout2016dynamic}).

Mathematically, we describe the virtual force model of \cite{zou2003sensor} as under:

Let sensor "i" experiences a total attractive force $\overrightarrow{F_{ia}}$ caused by the preferential coverage areas, and a repulsive force $\overrightarrow{F_{ir}}$ due to obstacles. If the force between sensors $i$ and $j$ is $\overrightarrow{F_{ij}}$, then the total force $\overrightarrow{F_{i}}$ experienced by sensor "i" can been expressed as:

\begin{equation}\label{eq:total_force}
\overrightarrow{F_{i}}=\overrightarrow{F_{ia}}+\overrightarrow{F_{ir}}+ \sum_ {j=1,j\neq1}^{k}\overrightarrow{F_{ij}}
\end{equation}

Let,\\
$d_{ij}$ = Euclidean distance between sensor $i$ and $j$,\\
$d_{th}$ = threshold distance between sensor $i$ and $j$,\\
$\alpha_{ij}$ = angle of line segment from sensor $i$ to sensor $j$,\\
$w_{a}$ = measure of attractive force,\\
$w_{r}$ = measure of repulsive force\\

Then, $\overrightarrow{F_{ij}}$ can be expressed in polar coordinates form as under:

\begin{equation}\label{eq:total_force_polar}
\overrightarrow{F_{ij}}=\begin{cases}
    (w_{a}(d_{ij}-d_{th}),\alpha_{ij}), & \text{if $d_{ij}>d_{th}$}\\
    0, & \text{if $d_{ij}=d_{th}$}\\
    (w_{r}\frac{1}{d_{ij}},\alpha_{ij}+\pi) & \text{if $d_{ij}<d_{th}$}
  \end{cases}
\end{equation}

The work \cite{zou2003sensor} provides some simulation results showing that virtual force based approach can achieve the area coverage after an initial random sensor deployment. This virtual force based approach takes negligible computation time and a one-time repositioning of sensor nodes. The algorithm also offers flexibility for the desired field coverage and model parameters. The authors have also shown how a probabilistic localization method can be used along with the force-directed sensor placement and it can reduce the energy consumption for target detection and location. An extended version of traditional virtual force based approach has been considered in \cite{li2012extended} to overcome the connectivity maintenance and nodes stacking problems. The work \cite{chen2007novel} presents improved virtual force based algorithms providing better performance in coverage rate, moving energy consumption, convergence, etc. However, the sensors in \cite{zou2003sensor,chen2007novel} might cause mutual collision due to the instability at the desired threshold distance and these algorithms are also based on the cluster head, which might be subjected to a single point failure \cite{zhu2014survey}.

In \cite{tan2009connectivity}, an enhancement of traditional virtual force based deployment approach has been presented and it is termed as Connectivity-Preserved Virtual Force (CPVF) scheme. In this scheme, the coverage is maximized and it also guarantees connectivity for a network of sensor nodes with arbitrary communication or sensing range, while adjusting a cost of small moving distance. In CPVF, the sensor nodes move in a greedy manner and it results in an arbitrary overlaps of the sensing ranges. When $R_{c}/R_{s}$ is small, the sensors experience a lack of information, which is required to maximize the coverage. So, the CPVF performs well in restricted scenarios and the authors also present a second floor-based scheme, which divides the sensing field into a floors of common height $2R_{s}$ and makes the sensors to stay at the central floor lines of those floors. Hence, the sensors are separated by floors, the overlapped sensing area is reduced and the global network coverage is improved. Another virtual force based distributed deployment approach has been considered in \cite{bartolini2010push}. The algorithm is named as Push and Pull and it ensures full coverage with triangular lattice pattern subject to meeting some necessary assumptions like there are enough number of sensors in the network and $R_{c}\geq \sqrt{3} R_{s}$. The work \cite{wang2007improved} presents a virtual force directed co-evolutionary particle swarm optimization (VFCPSO) algorithm, which is a combination of co-evolutionary particle swarm optimization (CPSO) with the virtual force (VF) algorithm. The simulation results show a better performance with respect to computation time and effectiveness than the VF, PSO and VFPSO algorithms. In \cite{mougou2012redeployment}, a distributed virtual force algorithm (DVFA) is presented and its simulation results are compared with a centralized virtual force algorithm (CVFA) algorithm \cite{kribi2009redeploying}. However, the comparison shows that the distance travelled by each sensor node in DVFA is larger than that found in CVFA. The work \cite{mahfoudh2013relocation} enhances the DVFA to consider obstacles.

We highlight an explicit comparison between APF and VFF based approaches as mentioned in the relevant literature:

\begin{table}[!htbp]
\begin{center}
    \begin{tabular}{ |p{6.5cm}| p{6.5cm} |}
    \hline
   \textbf{Artificial Potential Field (APF)}                                   &   \textbf{Virtual Force Field (VFF)}                                       \\ \hline
    APF is considered in the environment                                       &   VFF is considered around each robot                                      \\ \hline
    APF does not change in the same environment                                &   VFF changes in the same environment                                   \\ \hline                                                                                                                                                                                  APF does not depend on robot status                                        &   VFF depends on robots status (travelling speed, dimension, priority, location, environmental factors, etc)                                          \\ \hline
    Driving force derived from the artificial potential field                  &   Driving force directly calculated                                           \\ \hline
    APF does not consider any physical constraints of a robot                  &   VFF considers physical constraints of a robot                               \\ \hline
    Mathematically simple and computationally efficient                        &   Mathematically simple and computationally more efficient                         \\ \hline
    Local Minima problem might occur                                           &   Local Minima problem might also occur                                       \\ \hline
    Unstable Oscillatory movements might occur                                 &   Unstable Oscillatory movements might also occur                             \\ \hline

    \end{tabular}
\end{center}
\caption{Artificial Potential Field (APF) Versus Virtual Force (VFF) based Approach}
\end{table}

\subsubsection{Blanket Coverage based on Voronoi Diagram (VD)}

In Voronoi partitioning, a given region is divided into Voronoi cells equal to a given number of sensor nodes and each point in a Voronoi cell becomes associated with its closest sensor. The cells adjacent to the $ith$ sensor's cell are known Voronoi neighbours of sensor $i$. So, a Voronoi tessellation is generated by the number of given sensors. A Voronoi tessellation having Voronoi cells coinciding with the mass centroid of the Voronoi cell is known as Centroidal Voronoi Tessellation (CVT).  A Voronoi Diagram is considered to be one of the most fundamental geometric structures \cite{aurenhammer1991voronoi} in computational geometry.\\
Let $p_{i}$ for all  $i={1,2,3,..,n}$ be a set of points in a plane. Then, a Voronoi cell $V(p_{i})$ corresponding to $p_{i}$ can be formally defined as under:

\begin{equation}\label{eq:Voronoi}
V(p_{i})= \{ p:|p_{i}-p| \leq |p_{j}-p| \text{ for all } j \neq i \}
\end{equation}
i.e., a Voronoi cell $V(p_{i})$ associated with $p_{i}$ is the set of all points whose distance to $p_{i}$ is not greater than their distance to the other points $p_{j}$. The points $p_{i}$ are known as
generators or sites and the resulting tessellation is known as Voronoi Tessellation or Voronoi diagram. In the figure, a Voronoi diagram has been generated with $i=8$.

\begin{figure}[H]
\begin{center}
   \includegraphics[scale=0.5]{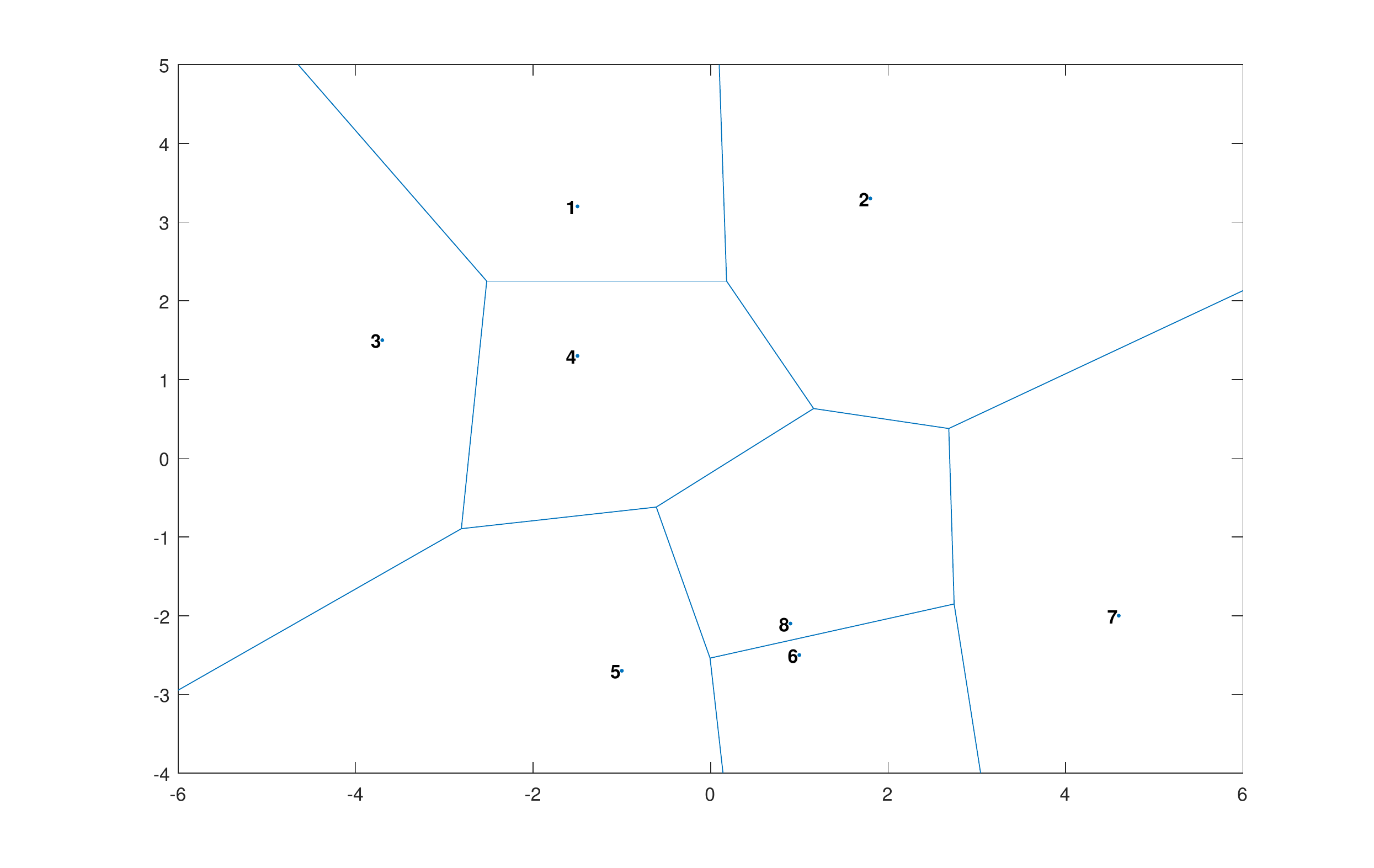}
  % \label{fig:theFig}
    \caption{A Voronoi Diagram of eight sites}
\end{center}
\end{figure}

The work \cite{megerian2005worst} uses the properties of Voronoi diagram for the maximal breach path, which is a metric for the worst-case coverage. The authors define maximal breach path between two arbitrary points of a bounded sensing field as a path where the distance between any point of the path and a sensor node is maximum. The properties of Voronoi Diagram can be used to limit the search for an optimal path.

The sensors are evenly distributed when they form Centroidal Voronoi Tessellation (CVT) \cite{du1999centroidal} and it is known as Gresho's conjecture \cite{gersho1979asymptotically} for a given area and a set of sensors.

Mobile robotic sensors can achieve blanket coverage by Voronoi partitioning techniques. A number of such algorithms have been presented in \cite{cortes2009distributed}. However, the computation cost is increased as the mobile robotic sensors (with limited computation capability) need to solve the geometry problems. We summarise these algorithms (presented in \cite{cortes2009distributed}) in the coming sub sections:\\
In all of these algorithms, the agents have a similar information exchange structure, i.e. an agent transmits its position and receives every neighbour's position. Then, it calculates a notion of the geometric centre of its own Voronoi cell. The Voronoi cell is based on some notion of partition of the environment. Then, each agent moves towards the calculated geometric centre during a communication round.
These laws have been categorised based on the networks, $S_{D}$ and $S_{LD}$.

%\cite{ingle2011energy}

\textbf{Geometric Centre Laws}:

The Geometric Centre Laws are based on Voronoi partition of the environment $Q$ and are defined on the network $S_{D}$.

\begin{enumerate}
  \item Voronoi-centroid control and communication law: This distributed algorithm has been introduced by \cite{cortes2002coverage} and it is denoted as $V_{rn-cntrd}$. The $V_{rn-cntrd}$ law calculates the notion of the centroid of a Voronoi cell. This law adopts a gradient ascent strategy and it monotonically optimizes a multicentric function $H_{dist}$.
  \item Voronoi-centroid law on planar vehicles: This law is denoted by $V_{rn-cntrd-dynmcs}$ and it has been presented by \cite{cortes2002coverage}. The $V_{rn-cntrd-dynmcs}$ law presents the same centroid strategy on a network of planar vehicles.
  \item Voronoi-circumcentre control and communication law:
This law is denoted as $V_{rn-crcmcntr}$ and it has been presented by  \cite{cortes2005coordination} for the network $S_{D}$. This law, $V_{rn-crcmcntr}$, calculates the notions of the circumcenter of a Voronoi cell. This law optimizes the disk-covering multi-centre function $H_{d}$.

\item Voronoi-incentre control and communication law: This law is denoted as $V_{rn-ncntr}$ and it has been presented by  \cite{cortes2005coordination} for the network $S_{D}$. The law, $V_{rn-ncntr}$, calculates the notions of the incentre of a Voronoi cell. This law optimizes the sphere-packing multi-centre function $H_{sp}$.
\end{enumerate}

\textbf{Geometric Centre Laws with Range Limited Interactions}:

The Geometric Centre Laws are based on Voronoi partition of the environment $Q$ and are defined on the network $S_{LD}$.

\begin{enumerate}
  \item Limited-Voronoi-normal control and communication law: This law is denoted by $L_{mtd-Vrn-nrml}$ and it has been presented by \cite{cortes2005spatially}. This adopts a geometric centring strategy for each robot and it optimizes the area multi-centre function $H_{area, r/2}$.

  \item Limited-Voronoi-centroid control and communication law: This law is denoted by $L_{mtd-Vrn-cntrd}$ and it has been presented by \cite{cortes2005spatially}. This adopts a geometric centering strategy for each robot and it optimizes the area multi-centre function $H_{dist-area, r/2}$.

\end{enumerate}

In \cite{arslan2015voronoi}, the authors introduce power diagrams to ensure a collision free navigation of multi robots towards centroid of Voronoi partition. A constrained optimization framework is also introduced to combine area coverage and collision avoidance. The authors also propose a heuristic congestion manager to speed up the convergence and a lift of the point particle controller to the more practical differential drive kinematics.

The work \cite{wang2006movement} considers a distributed deployment of the sensors based on Voronoi diagrams. A mobile sensor calculates Voronoi polygons based on the received neighbour information. Then, the existence of a coverage holes is determined. In general, the algorithm considers pushing the sensors from a dense area, moving the sensors to a sparsely covered area and move the sensors towards the centroid of the calculated polygon.

The work \cite{song2014distributed} presents a new algorithm for distributed energy-efficient self-deployment (DEED) in mobile sensor networks. A widely used distributed algorithm for the construction of Centroidal Voronoi Tessellations (CVTs) is Lloyd's algorithm \cite{lloyd1982least}. In Lloyd's algorithm, the initial locations of all the sensors and boundaries of the region are known a priori. This work improves the energy efficiency of the iterative Lloyd's method by considering two metrics: the travelling distance of the sensors and the number of deployment steps. The mechanical movement of sensors is one of the major source of energy consumption \cite{sibley2002robomote}. The simulation results in \cite{song2014distributed} show that the new algorithm requires 54 percent less travelling distance and 46 percent less energy consumption than Lloyd's method \cite{lloyd1982least}. However, this algorithm still requires energy efficient coverage with obstacle avoidance technique.

%\subsubsection{\textbf{Blanket Coverage with Delaunay Triangulation}}
%\cite{}
\subsubsection{Blanket Coverage based on Delaunay Triangulation}

A triangulation of a set of points (say $i=1,2,3,...,n$) in a plane is called a Delaunay triangulation if none of the points (vertices) lies inside the circumcircle of each of its triangles. A Delaunay Triangulation of the same points used in the above mentioned Voronoi Diagram has been shown in the figure. The work \cite{megerian2005worst} uses the properties of Delaunay triangulation for the maximal support path, which is a metric for the best coverage. The authors define maximal support path between two arbitrary points of a bounded sensing field as a path where the distance between any point of the path and a sensor node is minimum. The properties of Delaunay triangulation can be used to limit the search for the optimal path. A Delaunay triangulation method is computationally expensive method and it requires centralized information in most of the cases.

\begin{figure}[H]
\begin{center}
   \includegraphics[scale=0.5]{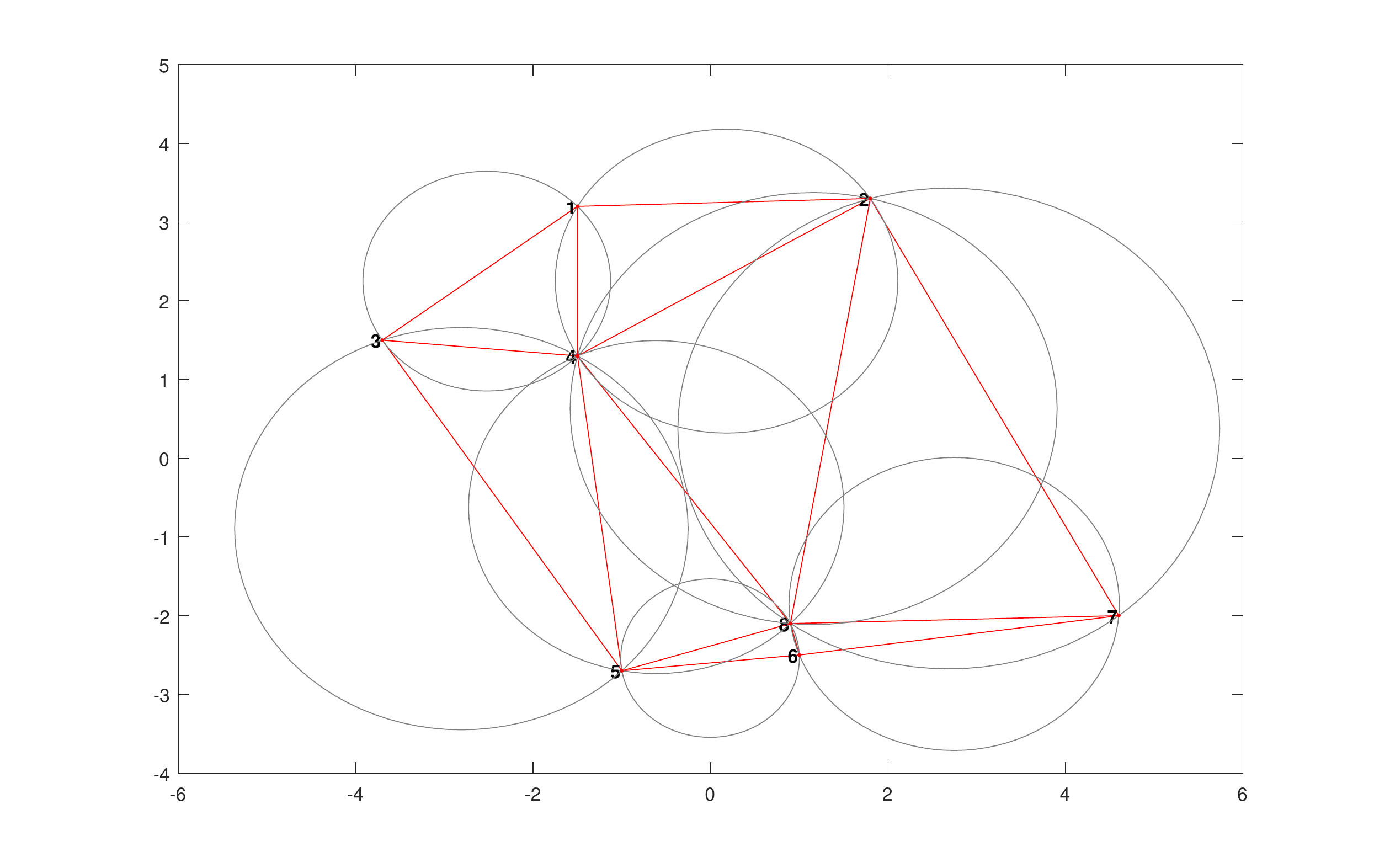}
  % \label{fig:theFig}
    \caption{Delaunay Triangulation}
\end{center}
\end{figure}

One can create Delaunay Triangulation of the point set by first creating the Voronoi Diagram of points ($i=1,2,3,...,n$) and subsequently creating the dual graph of this diagram as shown in the figure. In the below mentioned figure, we can notice that a Delaunay triangulation can be drawn by connecting only those points, which share a common edge in the Voronoi diagram of all the points.

\begin{figure}[H]
\begin{center}
   \includegraphics[scale=0.5]{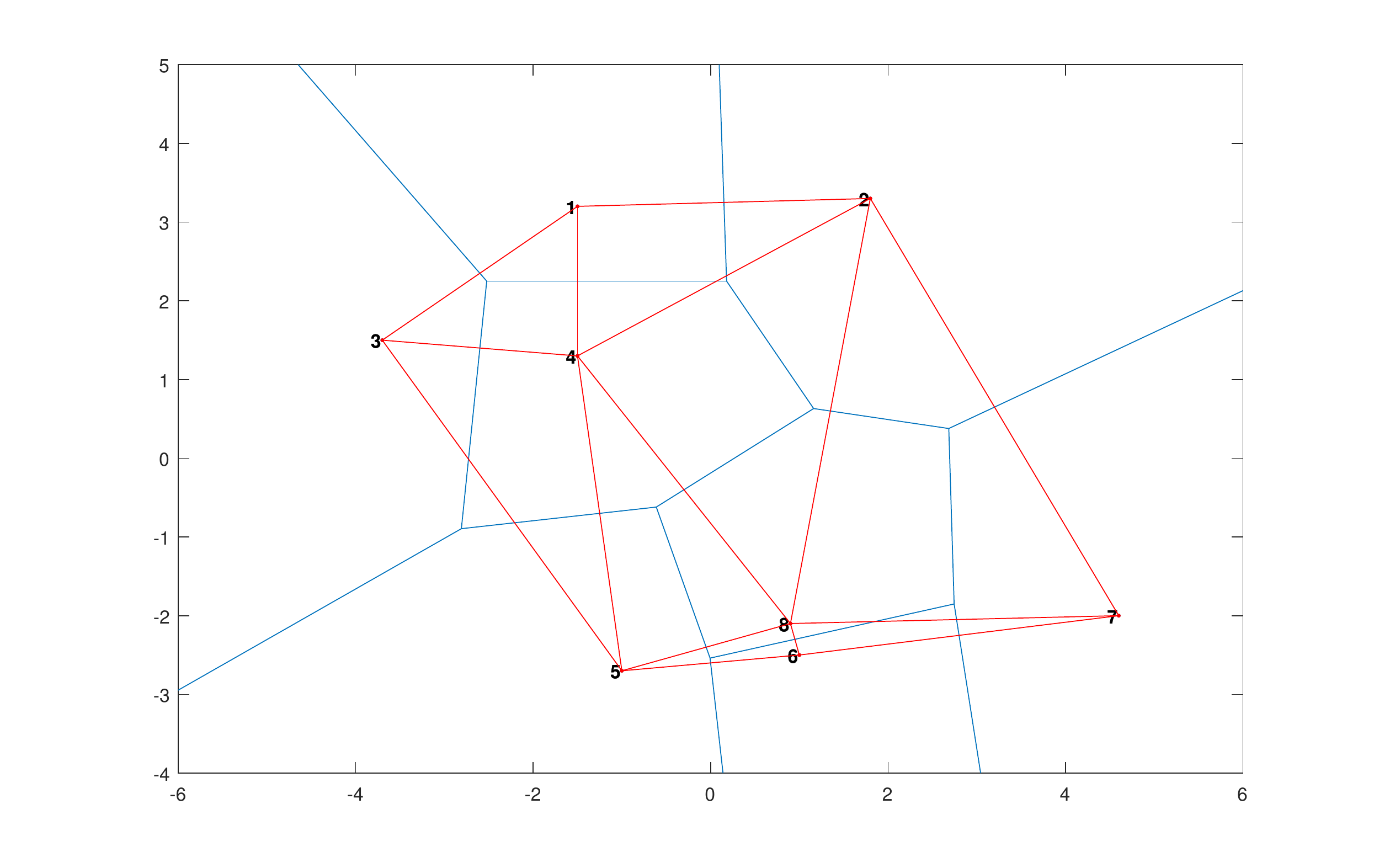}
  % \label{fig:theFig}
     \caption{Superimposing Delaunay Triangulation over Voronoi Diagram}
\end{center}
\end{figure}

The work \cite{wu2007delaunay} uses a contour based deployment to avoid coverage holes around the boundary of the area of interest and the obstacles. In the next phase, it uses a Delaunay triangulation based method for rest of the area. The authors compare the work with the grid based \cite{dhillon2003sensor} and randomized deployment strategy. The comparison suggest that the algorithm is scalable and outperforms the grid based and randomized strategy. However, the considered algorithm is centralized and deterministic in nature.

In \cite{rahman2007probabilistic}, the authors use an approach to point the least covered region in wireless sensor network, where further sensor nodes are desired. The authors use a clustering algorithm which is based on Delaunay triangulated sensor nodes.

In \cite{fazli2010multi}, the constrained Delaunay triangulation (CDT) has been used to address the area coverage problem. The constrained Delaunay triangulation \cite{chew1989constrained} is the triangulation of $n$ vertices in the plane along with a set of non-crossing, straight line edges and it offer two properties: the pre-specified edges are part of the triangulation and it is as close to the Delaunay triangulation as possible. However, the work \cite{fazli2010multi} uses a centralized information for the area map along with static obstacles.

In (\cite{tan2010arbitrary}), the authors present a full coverage method to find coverage holes and place sensors efficiently for arbitrary regions and obstacles. In fact, the authors use Delaunay triangulation based technique over the vertices of the hole for further coverage.

We present a comparative table \ref{tab:Voronoi_Delaunay} of Voronoi diagram and Delaunay triangulation based approaches.\\

\begin{table}[!htbp]
\begin{center}
    \begin{tabular}{ |p{6.5cm}| p{6.5cm} |}
    \hline
   \textbf{Voronoi Diagram (VD)}                                               &   \textbf{Delaunay Triangulation (DT)}                                         \\ \hline
    A maximal breach path (Worst case coverage) can be obtained                &   A maximal support path (Best case coverage ) can be obtained                 \\ \hline
    Polygon edges equidistant from neighbouring nodes                          &   Triangle edges connecting neighbouring nodes                                 \\ \hline
    Focuses on sensing range addressing coverage issues                        &   Focuses on communication range addressing connectivity issues                \\ \hline
    Empty Circle property does not exist                                       &   Empty Circle property exists                                                 \\ \hline
    Computationally complex                                                    &   Computationally complex                                                      \\ \hline
    Centralized information required in most cases                             &   Centralized information required in most cases                               \\ \hline
    \end{tabular}
\end{center}
\caption{Voronoi Diagram versus Delaunay Triangulation based Approach}
\label{tab:Voronoi_Delaunay}
\end{table}

\newpage

\subsection{Blanket Coverage: Deployment Approaches}
We explain one of the few fundamental deployment approaches found in the literature survey.
\subsubsection{Randomized Algorithms}

In randomized coverage algorithms, the localization of sensors is not predetermined before the final coverage. The movement of the sensors is also not predetermined. A randomized coverage algorithm is preferred in a large scale mobile sensor network, where the appropriate positions and number of sensors cannot be predetermined. It is also suitable in the cases where the terrain information is very uncertain. The main challenge in randomized coverage algorithms is to maximize the coverage and minimize the energy consumed by the sensors.

The work \cite{gage1994randomized} suggests randomized search strategy rather coordinated search strategy because of two reasons: the effectiveness of a coordinated search strategy decreases as the probability of target detection decreases and the cost of navigating a coordinated search strategy may be prohibitive as compared to the cost of a less capable search element. So, a careful consideration is required before deciding a strategy required for uniform area coverage. However, a random deployment might leave a coverage hole or mobile robotic sensors might be denser in some parts of the area and leaving the other parts without any coverage. We present some of the algorithms \cite{savkin2012optimal,nasimov2014suboptimal} having a characteristic of randomized movement of mobile robotic sensors for the full coverage.

In \cite{savkin2012optimal}, the authors consider a distributed random algorithm to deploy mobile robotic sensors in a bounded unknown region. The unknown region is considered to be 2-dimensional and arbitrary in shape. The considered region is not linearly connected and it can have holes in it. The algorithm considers an agents communication range ($R_{c}$) and sensing range $R_{s}$ by the relation, $R_{c} \geq \sqrt{3} R_{s}$. The algorithm deploys mobile robotic sensors making equilateral triangular grids pattern, which is considered to be an optimal pattern for covering the region \cite{kershner1939number}. The algorithm mathematically proves asymptotic optimality and convergence with probability 1. This blanket coverage algorithm is based on the probabilistic arguments. The algorithm uses two stages (say Stage-A and Stage-B). In Stage-A, the algorithm  drives all the mobile robotic sensors to the vertices of some triangular covering set. To fully comply with the definition of blanket coverage, the algorithm switches to the Stage-B to ensure a static arrangement of the mobile robotic sensors while occupying the vertices of the triangular covering set. Some necessary assumptions are required like mobile robotic sensors are connected and there is enough number of mobile robotic sensors to cover the planar region. We briefly explain this coverage algorithm as under:\\
Let $\theta_{i}(kT)$ be a consensus variable characterising one of the three lines of a triangular grid, and $q_{i}(kT)$ characterises two-dimensional coordinates of a vertex of the grid. A sensor "i" have initial values of $\theta_{i}(0)$ and $q_{i}(0)$. Similar to the nearest neighbour rule \ref{eq:jadbabie} and the update law \ref{eq:phi_update}, it guarantee that these initial values ($\theta_{i}(0)$ and $q_{i}(0)$) will eventually converge to a common triangular grid (say $\theta_{0}$ and $q_{0}$, respectively) following some necessary assumptions.

%\begin{equation}\label{eq:theta_random}
% {\theta}_{i}((k+1)T) =\dfrac{1}{|1+{\cal N}_{i}(kT)|}\left(\theta_i(kT)+\sum_ {j\in {\cal N}_{i}(kT)}\theta_{j}(kT)\right)
%\end{equation}
%
%for $i=1,2,3,\ldots, n$.
%
%\begin{equation}\label{eq:q_random}
% {q}_{i}((k+1)T) =\dfrac{1}{|1+{\cal N}_{i}(kT)|}\left(q_i(kT)+\sum_ {j\in {\cal N}_{i}(kT)}q_{j}(kT)\right)
%\end{equation}

Then, a sensor $i$ is able to update its coordinates as under:

\begin{equation}\label{eq:C_random}
 {p}_{i}((k+1)T) = C[q_i(kT),\theta_i(kT)]p_i(kT)
\end{equation}
where $C[q_i(kT),\theta_i(kT)](p_{i}(kT))$ represents the vertices of triangular covering set $\bar{V} [q,\theta]$ closest to $p$. In case of more than one vertex in $\bar{V} [q,\theta]$, the algorithm decides any one of them. In the Stage-A of \cite{savkin2012optimal}, one can have a detailed look on the proof of theorem sating how the nearest neighbour rule and \ref{eq:C_random} under some necessary assumptions guarantee that a sensor's coordinates achieve the vertices of some triangular covering set. However, the Stage-A does not guarantee that the sensors will occupy the vertices of the triangular covering set.

In Stage-B, we assume that the sensors are at the vertices of a triangular covering set and move from vertex to vertex, i.e. $p_{i}(kT)\in \bar{V}$ for all $i$, $kT$. Let $S(p_i(kT))$ is the set consisting of $v=p_{i}(kT)$ and all its unoccupied neighbours in the triangular covering set. Then, one can notice that $1 \leq |S(p_i(kT))| \leq 7$ as $v \in \bar{V}$. The below mentioned random algorithm stops the sensors when all the vertices of the triangular covering set are occupied. This follows a necessary assumption that there is enough number of mobile robotic sensors in the network.

\begin{equation}\label{eq:q_probab}
 p_{i}((k+1)T) = s \text{ with probability } \dfrac{1}{|S(p_i(kT))|}\\
  \forall s \in S(p_i(kT))
\end{equation}

\begin{figure}[H]
  \centering
  \includegraphics[width=10cm]{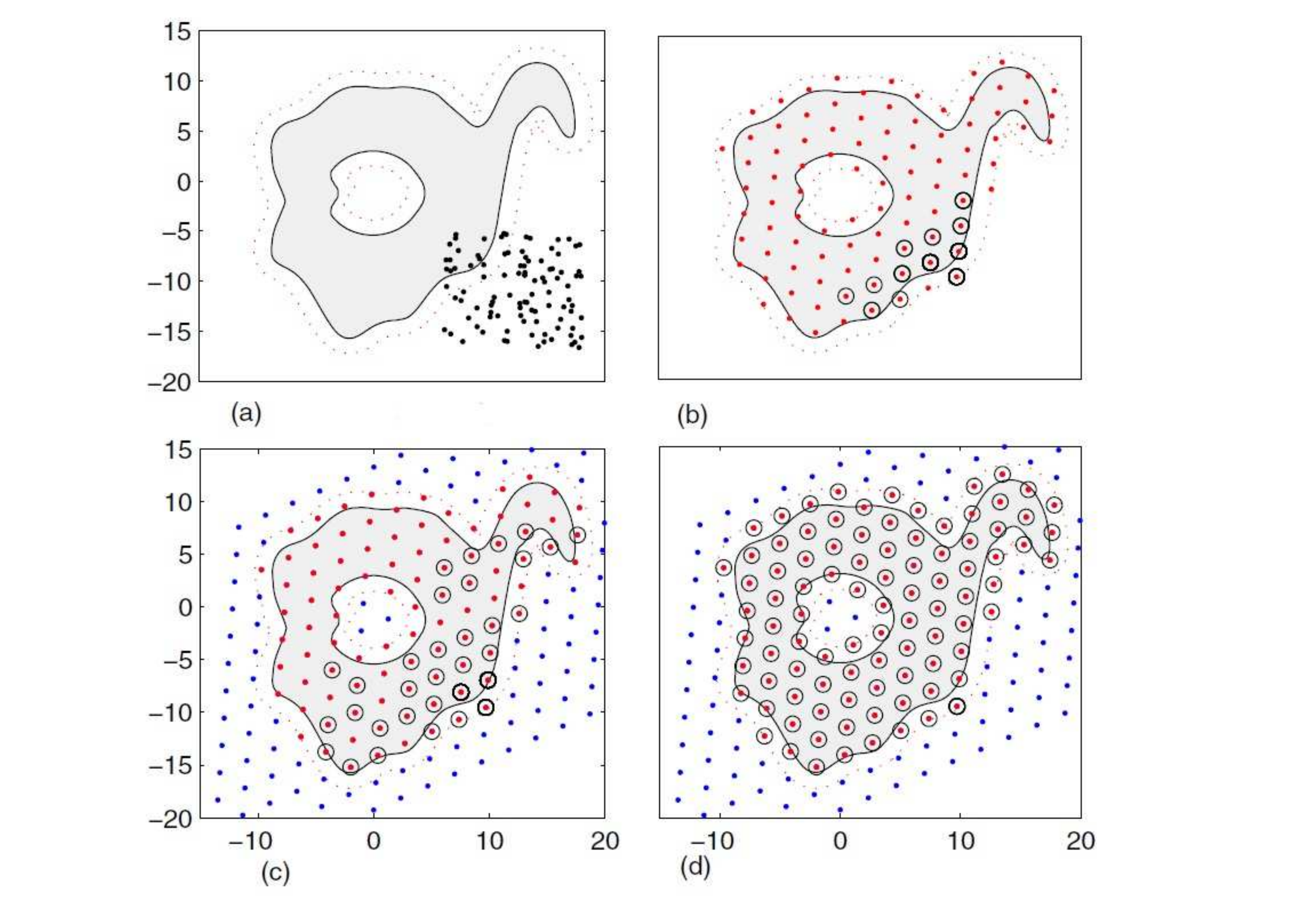}
  \caption{Blanket coverage over a region - a randomized approach \cite{savkin2012optimal}}
  \label{fig:Randomized_Blanket}
\end{figure}

Similar to \cite{savkin2012optimal}, another blanket coverage algorithm has been presented in \cite{nasimov2014suboptimal}. This algorithm \cite{nasimov2014suboptimal} differs from \cite{savkin2012optimal} in two ways: Firstly the above mentioned Stage-A and Stage-B is executed in parallel and secondly it gives a rigorous mathematical proof that the full blanket convergence will be achieved within finite time. However, this algorithm considers a slightly excessive number of mobile robotic sensors and it solution is considered to be sub-optimal.

In \cite{hefeeda2007randomized}, a distributed randomized k-coverage algorithm has been presented. In this work, the goal is to choose a minimum set of sensors to activate from a pre-deployed set of sensors such that all locations are k-covered. The algorithm is motivated from the approximation algorithm for the optimal hitting set problem. This work is compared with \cite{yang2006connected,zhou2004connected} and the algorithm is found to offer faster convergence, activating near-optimal number of sensors and it consumes less energy. A similar problem has been considered in \cite{zhou2004connected}, where also a small subset of over-deployed sensors is kept active to achieve k-coverage.

\subsubsection{Deterministic Algorithms}

The algorithms which deploy the mobile robotic sensors to a predefined coverage pattern are called to posses a deterministic characteristic. Most of the algorithms in the literature are deterministic in nature. Especially, if the terrain information is available a priori then sensors can be placed in an effective manner using deterministic characteristic considered in a deployment algorithm. One can also consider a higher weighting factor to some high priority part of the area. A military surveillance area problem might choose some region critical than others and the deployment approach could use denser coverage in those areas rather than a sparse coverage. However, the deterministic deployment characteristic is hard to implement when the area information is very uncertain. The deterministic deployment is suitable for a small to medium sensor network.

In \cite{howard2002incremental}, the authors present an incremental self-deployment algorithm with one senor deployment at a time in an unknown environment and the next node uses the information gathered by the previous node to determine its location. The algorithm also exhibits deterministic policies. Another good example of deterministic deployment is a grid-based sensors deployment and the grids can be of different shapes like: triangular lattice, square, hexagon/ honeycomb, diamond, etc. In \cite{bai2010optimal}, the authors propose a Diamond pattern to achieve 4-connectivity and 1-coverage. The authors also prove this pattern to be asymptotically optimal pattern for 4-connectivity and 1-coverage, when $R_{c} > \sqrt{2} R_{s}$. Some other algorithms having deterministic characteristic and deploying sensors with a grid based pattern has been presented above (see, e.g. \cite{cheng2013decentralized,santoso2011decentralised,cheng2010decentralized,santoso2010sub}, \cite{ahmad2012decentralized}, \cite{santoso2015range,ahmad2015decentralized}). A strip based sensor deployment is also an example of deterministic deployment (see e.g. \cite{kar2003node}). Similarly, the work \cite{wang2005efficient} deterministically partition the coverage area into smaller sub-regions depending on the shape of the field and then sensors are deployed into the sub-regions.

Finally, we compare the randomized and deterministic deployment approaches as under:

\begin{table}[!htbp]
\begin{center}
    \begin{tabular}{ |p{6.5cm}| p{6.5cm} |}
    \hline
   \textbf{Deterministic}                                &   \textbf{Randomized}                                                \\ \hline
    Terrain information available a priori               &    Uncertain terrain information                                     \\ \hline
    Grid based deployment                                &    Random deployment                                                 \\ \hline
    Region sub-division predetermined                    &    No prior sub-division of the region                               \\ \hline
    Predetermined number of sensors at certain locations &    Unknown number of sensors at certain locations                    \\ \hline
    Full coverage guaranteed                             &    Coverage hole might exist                                         \\ \hline
    Suitable for small to medium scale sensor network    &    Suitable for large scale network                                  \\ \hline
    Easier coverage scheme                               &    Comparatively harder coverage scheme                              \\ \hline
    Coverage and connectivity easy to control            &    Coverage and connectivity difficult to control                    \\ \hline
    Optimizing number of deployed nodes                  &    Optimization objective hard to achieve                            \\ \hline
    Suitable for expensive sensor nodes                  &    Suitable for cheap sensor nodes especially for harsh environment  \\ \hline
    Recommended for static nodes deployment              &    Recommended for dynamic nodes deployment                          \\ \hline
    \end{tabular}
\end{center}
\caption{Randomized versus Deterministic Deployment Approaches}
\end{table}

\subsubsection{Asymptotically Optimal Grid Pattern}\label{optimal_grid_pattern}

One of the most important deployment approaches in a blanket coverage algorithm is its optimality in terms of minimum number of deployed mobile sensors. In \cite{kershner1939number}, it has been shown that the triangular lattice pattern is optimal in terms of minimum number of sensors required for a complete coverage of a bounded set.

If the communication range ($R_{c}$) and sensing range ($R_{s}$) of an agent is related as $R_{c}\geq \sqrt{3}(R_{s})$ then the deployment of mobile robotic sensors using triangular lattice pattern provides 1-coverage and 6-connectivity \cite{bai2006deploying}. Mathematically, we can write the definition of optimal blanket coverage as stated in \cite{cheng2010decentralized,cheng2013decentralized}:

Let $m(R_{s})$ be the number of sensors that covers a bounded region $R \subset \mathbb{R}^2$ and $P(R_{s}) = {h_{i, j(R_{s})}}$ be the corresponding set of desired sensor locations in $\mathbb{R}^2$. Let $N(R_{s})$ be the minimum number of sensors to cover $R$. The locations $P(R_{s})$ is said to be asymptotically optimal for covering $R$ if for all $p \in R$, there exists $h_{i, j}(R_{s}) \in P(R_{s})$ such that

\begin{equation}\label{eq:Blanket_Boundary}
 \parallel p - h_{i,j}(R_{s}) \parallel \leq R_{s} \\
\end{equation} and

\begin{equation}\label{eq:Blanket_Boundary_Min_Sen}
 \lim_{R_{s}\to \infty } R_{s}^2 m(R_{s}) = \lim_{R_{s}\to 0 } R_{s}^2 N(R_{s})\\
\end{equation}

The condition (\ref{eq:Blanket_Boundary_Min_Sen}) can be obtained from the below mentioned main result of Kreshner's theorem \cite{kershner1939number}.

\begin{equation}\label{eq:Kreshner}
 \lim_{R_{s}\to 0} \pi R_{s}^2 N(R_{s})=\lim_{R_{s}\to 0} \pi R_{s}^2 m(R_{s})\\
 =(2\pi\sqrt{3}/9)A(R)
\end{equation}, where $A(R)$ denotes area of $R$.

The work (see e.g. \cite{cheng2010decentralized,savkin2012optimal,ahmad2012decentralized,cheng2013decentralized}) has developed decentralized control laws driving the mobile robotic sensor to achieve asymptotically optimal blanket coverage with triangular lattice pattern - which provides 1-coverage and 6-connectivity (except the boundary sensors).

\section{Barrier Coverage}\label{Barrier_Coverage}

If the static arrangement of mobile robotic sensors is forming a Barrier and it minimizes the probability of undetected intruder passing through the arrangement, such sort of coverage is known as Barrier Coverage \cite{gage1992command}. In ancient times, there was protection of castles using moats, so that any sort of intrusion/ obstacle from the enemy could be faced \cite{cheng2011decentralized}. Now, the research has been established to replace such barriers with the mobile robots carrying sensors. If the sensors are placed randomly and they could rearrange themselves while covering a certain corridor, it is said that there has been Barrier Coverage by the mobile robotic sensors against any intruder.

\subsection{Barrier Coverage Classification}
We classify the problem of barrier coverage on the basis of the type of region.

\subsubsection{Barrier Coverage along a Landmark}
The problem of barrier coverage can be formulated along a line or on a point on it. In \cite{cheng2011decentralizedsweep}, the problem of sweep coverage has been studied along a line $W$. However, the autonomous sensors are required to form a barrier along the line first. In \cite{cheng2011decentralizedsweep}, the problem of barrier coverage along line $W$ with direction $\bar{\theta}$ is formulated as under:\\

\begin{equation}\label{eq:Line_Sweep}
W:= \{ p \in \mathbb{R}^2:u^Tp:=d_{1} \}, \bar{\theta}:=\beta+\pi/2
\end{equation},
where $u=[\cos(\theta) \sin(\theta)]^T$ is a unit vector with a given $\beta \in [-\pi/2 \text{ } \pi/2)$ measured with respect to x-axis and $d_{1}$ is a given scalar associated with $W$. The mobile robots are supposed to make a barrier of length $L$ from $W$. Ideally, the sensors are to be evenly deployed maximizing the length $L$.

\begin{figure}[H]
  \centering
  \includegraphics[width=10cm]{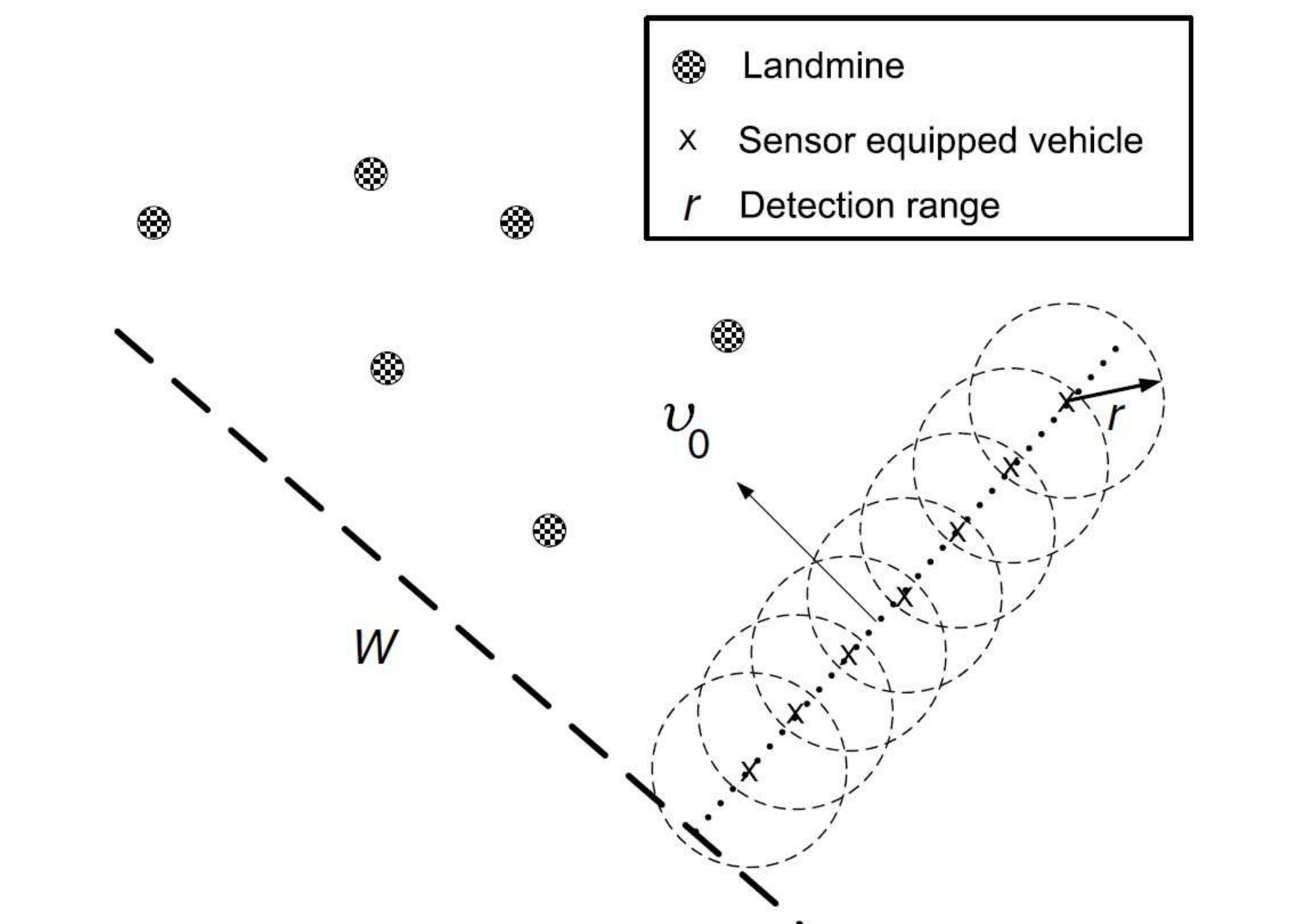}
  \caption{Barrier coverage along a line \cite{cheng2011decentralizedsweep}}
  \label{fig:sweep_line}
\end{figure}

\subsubsection{Barrier Coverage between Two Landmarks}
In this problem, a barrier of sensors is supposed to ensure coverage between two landmarks. The problem of barrier coverage between two landmarks (say $L_{1}$ and $L_{2}$) is formulated in (see e.g. \cite{cheng2009distributed}) as under:\\
Let $u$ be a unit vector associated with $L_{1}$ and $L_{2}$.

\begin{equation}\label{eq:l_unit vector}
u=\frac{(L_{2}-L_{1})}{\|(L_{2}-L_{1})\|}
\end{equation}
 The unit vector $u$ characterizes the bearing of $L_{2}$ relative to $L_{1}$ and it can be written as $u=[\cos(\beta) \sin(\beta)]^T$ for some $\beta \in [-\pi/2 \text{ } \pi/2)$. An associated scalar is also defined as $\bar{\theta}:=\beta+\pi/2$. A line $L$ using $L_{1}$ and $L_{2}$ is defined as:\\

\begin{equation}\label{eq:l_unit vector}
L:= \{ p \in \mathbb{R}^2:(L_{2}-L_{1})^T u_{\perp}=0\},
\end{equation},
where $u_{\perp}$ is normal vector to $L$ such that $u^Tu_{\perp}=0$. Then, the authors define $n$ number of points (say $h_{i}$) on L as under:\\

\begin{equation}\label{eq:n_points}
h_{i}:= L_{1}+i    \Bigg(\frac{\|L_{2}-L_{1}\|}{n+1}\Bigg)u,i=1,2,...,n
\end{equation}
The decentralized control laws are supposed to drive the sensors to the above mentioned points. The authors formally define this problem of barrier coverage as under:

Let there are $n$ mobile sensors and two distinct landmarks $L_{1}$ and $L_{2}$. Then, a decentralized control law for barrier coverage between the landmarks is formulated if for almost all initial sensor positions, there exists a permutation ${z_{1}, z_{2}, . . . , z_{n}}$ of the set ${1, 2, . . ., n}$ such that the following condition holds:\\

\begin{equation}\label{eq:Barrier}
 \lim_{k\to \infty } \parallel p_{z_{i}}(kT) - h_{i} \parallel=0,\text{for all $i={1, 2, . . ., n}$}
\end{equation}

\begin{figure}[H]
  \centering
  \includegraphics[width=10cm]{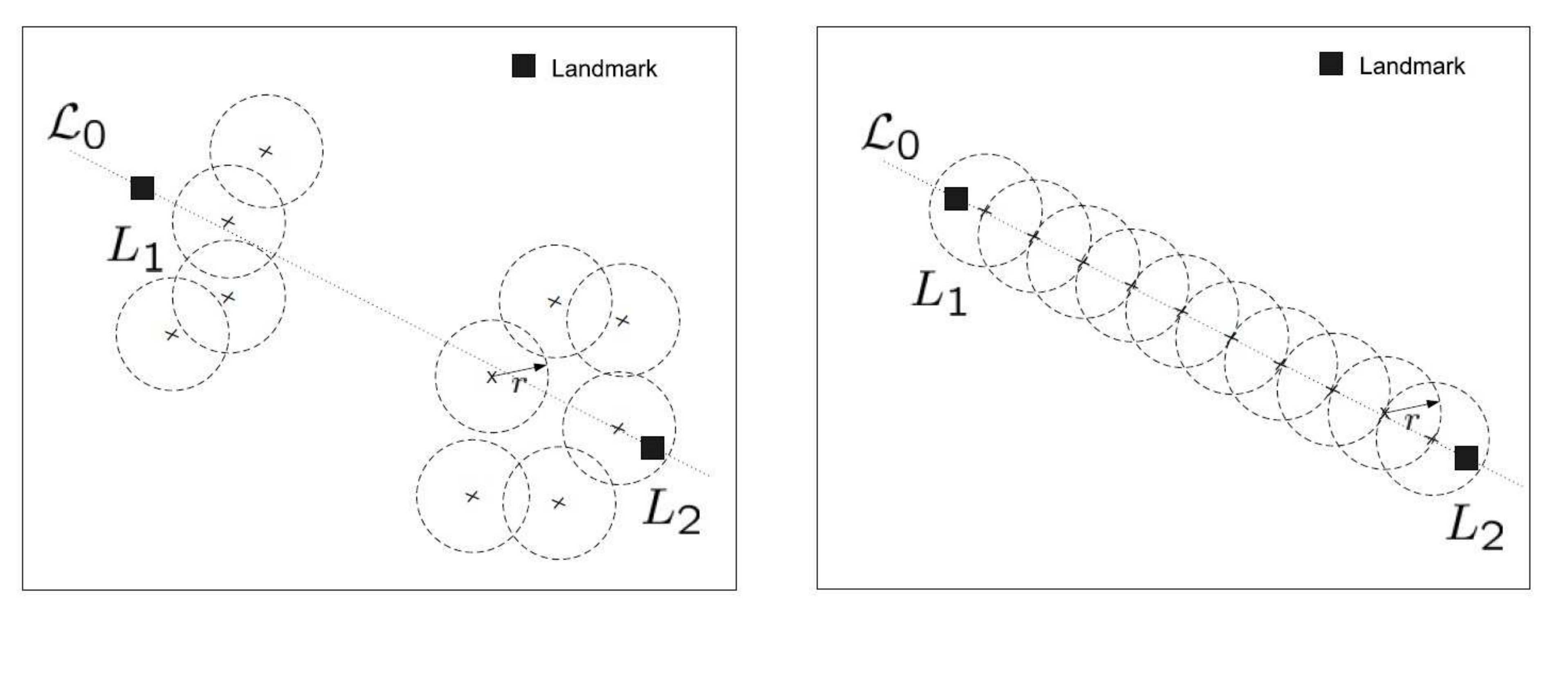}
  \caption{Barrier coverage between two landmarks\cite{cheng2011decentralizedsweep}}
  \label{fig:test}
\end{figure}

Similar problem of barrier coverage has been considered in (see e.g. \cite{chen2007designing,chen2010local,cheng2014density,kloder2007barrier,kumar2005barrier,shen2008barrier}).

%\begin{figure}[h]
%\begin{center}
%   \includegraphics[width=\linewidth]{BarrierLandmarks.jpg}
%  % \label{fig:theFig}
%    \caption{A Voronoi Diagram of 8 sites}
%\end{center}
%\end{figure}

%\begin{figure}[h]
%\centering
%\includegraphics[width=2.9in]{Figure/Barrier_Landmarks}
%%\caption{Plots of $O_r(t), O_e(t)$  and $O_et(t)$ ($O{e,Dm}(t)$ with $D_m$ and $O_{e,CM}(t)$ with $C_M$) for S1($f_c$=30 Hz, $\lambda_p$=532 nm, $\lambda_e$=1205 nm) }
%%\label{fig13}
%\end{figure}

%\cite{cheng2011decentralized}

\subsection{Barrier Coverage Approaches}
We categorise the problem of barrier coverage on the basis of approaches/ methodologies used in the literature.

\subsubsection{Barrier Coverage based on Nearest Neighbour Rule}

The nearest neighbour rule based on the local information has been introduced in the section \ref{BC:NNR}. One can see the work (e.g. \cite{cheng2009problem,cheng2011decentralized}), where the nearest neighbour rule has been used as one of the fundamentals to calculate the set of decentralized control laws for the control inputs: velocity $v_{i}(kT)$ and heading angle $\theta_{i}(kT)$ of a mobile robot.

\subsubsection{Barrier Coverage based on Artificial Potential Field}

In \cite{razafindralambo2011connectivity}, the authors present localized deployment algorithm and it is based on artificial potential field theory. A sensor node is considered as a particle and its movements are based on communication with its neighbouring node. A mobile sensor travels a constrained distance by the connectivity of the node to its neighbours in a connected subgraph like the relative neighbourhood graph. This approach also leads to a barrier coverage algorithm with the preserved connectivity. The algorithm spreads out the mobile nodes between the starting points and the points of interest. Once a node is reached at the point of interest, it spreads a flooding message to the whole network. However, a straight line barrier of mobile nodes is formed and an obstacle(s) avoidance technique remains as a future consideration.

%\cite{hynes2004multi}

\subsubsection{Barrier Coverage based on Virtual Force Field}
The concept of virtual attractive force and repulsive force can be used to form barrier coverage. In \cite{shen2008barrier}, the below mentioned virtual force model is used for the problem of barrier coverage nearby a line from the left boundary to the right boundary of a given area having length $L$ and width $W$.

Let $x_{i},y_{i}$ be the initial coordinates of a sensor $i$ and $N_{i}$ be the set neighbours coming under its communication range $R_{c}$.

The repulsive force (say $\overrightarrow{F_{r_{ij}}}$) between sensor $i$ and $j$ is calculated to distribute sensors uniformly on the x-axis between $[0,L]$.

\begin{equation}\label{eq:repulsive_force_barrier}
\overrightarrow{F_{r_{ij}}}=\begin{cases}
    \frac{\alpha}{(x_{i}-x_{j})}, & \text{$ if j \in N_{i}$} \\
    0, & \text{$ if j \notin N_{i}$}
    \end{cases}
\end{equation}
where $\alpha > 0$ is a constant to normalize the repulsive force.

The left or right boundaries also exert repulsive forces on a mobile sensor if the mutual distance between a sensor and the boundary is less than the communication range, $R_{c}$. The authors model this repulsive force as under:

\begin{equation}\label{eq:repulsive_force_boundary}
\overrightarrow{F_{r_{ib}}}=\begin{cases}
    0, & \text{$ if    R_{c} \leq  x_{i} \leq (L-R_{c})$} \\
    \frac{-1}{x_{i}}, & \text{$ if x_{i} < R_{c}$} \\
     \frac{1}{(x_{i}-L)}, & \text{$ if  x_{i} > (L-R_{c})$}
    \end{cases}
\end{equation}

Then, the total repulsive force on sensor $i$ to orientate it along the x-axis is as under:

\begin{equation}\label{eq:repulsive_force_total}
\overrightarrow{F_{r_{i}}}=\overrightarrow{F_{r_{ib}}}+ \sum_ {j\in N_{i}}\overrightarrow{F_{r_{ij}}}
\end{equation}

The attractive force (say $\overrightarrow{F_{a_{ij}}}$) between sensor $i$ and $j$ is calculated to relocate sensors nearby the y-axis.
\begin{equation}\label{eq:attractive_force_barrier}
\overrightarrow{F_{a_{ij}}}=\begin{cases}
    \beta(y_{j}-y_{i}), & \text{$ if j \in N_{i}$} \\
    0,                  & \text{$ if j \notin N_{i}$}
    \end{cases}
\end{equation}
where $\beta>0$ is a constant to normalize the repulsive force.

Then, the total attractive force on sensor $i$ to orientate it along the y-axis is as under:

\begin{equation}\label{eq:attractive_force_total}
\overrightarrow{F_{a_{i}}}=\sum_ {j\in N_{i}}\overrightarrow{F_{a_{ij}}}
\end{equation}

A similar concept of virtual forces has been considered in \cite{jinlin2013barrier} to save the mobile sensor network energy and improve the barrier coverage adapting to the environment. But this work is compared with the centralized barrier coverage algorithm of \cite{shen2008barrier}.

%\subsubsection{\textbf{Barrier Coverage with Voronoi Diagram}}
%
%
%
%The combination of Voronoi diagram and graph search algorithms has been used in \cite{megerian2005worst} to develop an optimal polynomial time algorithm for the best and worst-case formulations for isotropic sensor coverage in wireless ad hoc sensor networks. In maximal breach path, a path in an area is defined to connect the initial and final positions.
%
%%\cite{meguerdichian2001exposure}
%%\cite{li2003coverage}
%
%\subsubsection{\textbf{Barrier Coverage with Delaunay Triangulation}}
%The formulation of Delaunay Triangulation has been used in \cite{jiang2008double} for Barrier coverage formulation.

\subsection{Barrier Coverage: Deployment Approaches}
In this section, we categorize the deployment approaches.

\subsubsection{Randomized Barrier Coverage}

%A randomized sensor deployment can be used to act as a barrier for intrusion detection. In \cite{keung2010intrusion}, the problem of k-barrier coverage is considered by a sensor mobility pattern in a belt region. The work computes the probability $Pr(\Lambda \geq k)$ for a given mobile sensor network and its network parameters to achieve k-barrier coverage. The symbol $\Lambda$ represents the cumulative coverage count by mobile sensors. The considered problem is similar to a problem in classical kinetic theory of gas molecules in physics. The sensor mobility pattern is based on independent movement of every sensor and without any mutual coordination. A sensor randomly chooses a direction $\theta \in [0,2\pi)$ according to the distribution with probability density function $P_{\Theta}(\theta)$. A sensor also randomly chooses speed $v \in [0,v_{max}] $ according to a probability density function $P_{V}(v)$, where $v_{max}$ is the maximum sensor speed. If a sensor reaches the boundary of the region then it bounds back by choosing another angular direction. The authors with the formulation of the uncovered distance and the sensor coverage rate obtain the below mentioned probability of k-barrier coverage:
%
%\begin{equation}\label{eq:Probability_Coverage}
%Pr(\Lambda \geq k)= 1- \frac{\Gamma(k,\Theta_{v}.\tau)}{\Gamma(k)}
%\end{equation}

The barrier coverage can be achieved with randomized sensor deployment following some model, but this sort of coverage is guaranteed on some probabilistic grounds.

\begin{figure}[H]
  \centering
  \includegraphics[width=10cm]{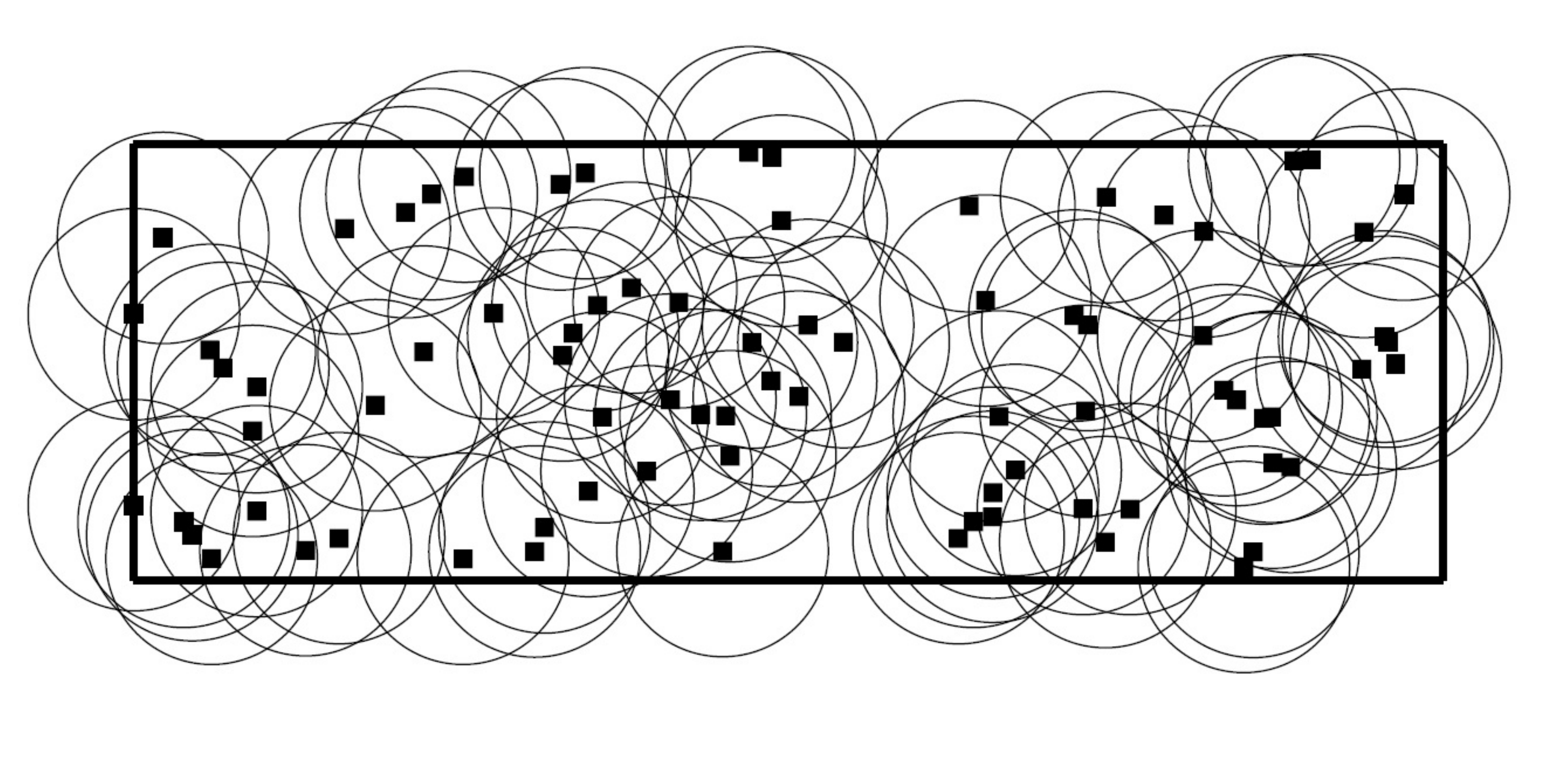}
  \caption{Barrier Coverage with randomized deployment \cite{kumar2005barrier}}
  \label{fig:test}
\end{figure}

The work \cite{kumar2005barrier} considers randomized barrier coverage problem with high probability. In \cite{kumar2004k}, a randomized independent sleeping (RIS) scheme has been introduced. In this scheme, time is divided into periods and each node independently decides to be awake (with probability $p$) or sleep (with probability $1-p$) at the start of a period. If in a Poisson distributed sensor network with rate $n$, each sensor sleeps according to the scheme \cite{kumar2004k} then the distribution of the active sensors achieves Poisson distribution of rate $n \times p$ \cite{ross2000introduction}. The work \cite{kumar2005barrier} establishes a critical condition for a belt region with sensors deployment following a Poisson distribution with rate $n \times p$. The developed condition allows computation of the number of sensors necessary to ensure weak k-barrier coverage of the region with high probability. It has been proved in (Page 39 of \cite{hall1988introduction}) for a region of unit area when $n$ becomes larger and larger  Poisson distribution of sensors with rate $n$ becomes equivalent to random uniform distribution of $n$ sensors, which means each sensor has an equal likelihood of staying at any location within the deployed region, independently of the other sensors. However, it has been mentioned that the randomized barrier coverage needs approximately $\log(n)$ more sensors than that required with the deterministic deployment approach. As an example of \cite{kumar2005barrier}, a deterministic deployment needs 500 sensor to achieve 1-barrier coverage and the randomized deployment will need 6200 sensors to achieve 1-barrier coverage with high probability.

Another example of randomized sensors deployment forming a barrier has been considered in \cite{saipulla2009barrier}. The sensors are deployed along lines with normally distributed random offsets. The authors show when the variance of the random offset in the line-based normal random offset distribution (LNRO) is relatively small compared to the sensor's sensing range, then the barrier coverage of LNRO outperforms the Poisson model. The probability of intrusion detection is dependent on the model under which nodes are randomly deployed.

%\newpage

\subsubsection{Deterministic Barrier Coverage}

A pre-defined mobile sensor deployment forming a barrier is commonly considered in the literature. An evenly distributed grid or strip based senor deployment is one of the examples of deterministic barrier coverage. The work already explained (see e.g. \cite{cheng2011decentralizedsweep,cheng2009problem,cheng2009distributed}) considers deterministic barrier coverage.

\subsubsection{Optimal Barrier Coverage}
The optimization of Barrier Coverage might be taken in terms of battery energy saving of the mobile sensor and/ or in terms of the strength of the mobile robotic sensors, which is based on the number of mobile sensors in a certain area. The problem of optimal barrier coverage can be categorized depending on the objective function subject to some constraints.

In \cite{jia2014autonomous}, line-based barrier coverage has been considered with the minimum moving distance. The work provides a theoretical optimization analysis on optimal sensor layout to achieve a line based barrier coverage with minimum moving distance. The work \cite{czyzowicz2010minimizing} considers optimization objective as minimizing the sum of distances travelled by all the sensors from initial to final positions forming a line barrier.

In \cite{bar2015green}, the problem of energy efficient barrier coverage has been considered. The authors consider two objective functions: minimizing the sum of the energy spent by all mobile sensors (i.e. minimize $ \sum_{\substack{i=1}}^n E_{i}(y,r)$) and minimizing the maximum energy consumed by any mobile sensor (i.e. minimize $ max_{i} E_{i}(y,r)$), where $y$ is the deployment vector and $r$ is the sensing radii vector of mobile sensors. The energy consumption model has been considered from a single battery source for the mobile sensors movement and sensing field. So, the authors consider two variants with fixed and variable sensing radii of mobile sensors. However, the problem has been considered for a straight-line barrier.

The work \cite{cheng2014density} considers the problem of strong barrier coverage problem in a given two-dimensional plane. The work considers sensor density requiring a minimized moving distances of all participating mobile sensors and using hole-handling mechanism to prolong the network lifetime. In \cite{deng2014barrier}, the barrier lifetime maximisation (BLM) has been defined as the optimization objective and the sensors can have different sensing ranges, while constructing a sensor barrier.

In \cite{kumar2007optimal}, the problem of maximizing the network lifetime for barrier coverage has been considered. The work presents optimal solutions to the sleep wakeup problems for the model of barrier coverage with sensors having homogeneous and heterogeneous lifetimes. It has been shown that the network lifetime is six times longer than that achieved with Randomized Independent Sleeping algorithm (RIS) of \cite{kumar2005barrier}. In \cite{chen2007designing}, the network lifetime is maximized with a sleep-wakeup algorithm known as Localized Barrier Coverage Protocol (LBCP). The work provides near optimal enhancement in network lifetime while providing global barrier coverage most of the time. This work also outperforms the RIS of \cite{kumar2005barrier} by up to six times.

In contrast to the previous work \cite{cheng2009distributed,cheng2010distributed,bhattacharya2008optimal,flocchini2008self,liu2008strong,cheng2009decentralized,saipulla2010finding,tafa2011algorithms}, the work \cite{tafa2012towards} considers the improvement of barrier coverage in a network where some mobile sensors are deployed after the initial deployment. The work analyses the most efficient way for a distributed algorithm to find and fill the barrier gaps in a network, where the sensors from the accessible sites can be guided after the initial deployment.

\subsubsection{k-Barrier Coverage}

If every crossing path through the width of a belt region is intersected by at least k distinct sensors then the region is called to be k-barrier covered. The problem of k-Barrier coverage is closer to the blanket coverage. It is in contrast with the k-blanket coverage problem, where every point of the region is covered by at least k distinct sensors. This notion of k-barrier coverage is first defined by \cite{kumar2005barrier}.

We provide the formal definition of k-barrier coverage as described in \cite{cheng2012self}:\\
A set of distributed control laws is known to be k-barrier coverage coordinated control law between two landmarks (say $L_{1}$ and $L_{2}$) if for almost all initial mobile sensor positions, there exists permutation of the set $2,3,...,K$ mentioned by ${x_{1},x_{2},...,x_{K-1},}$ such that:

\begin{equation}\label{eq:k-Barrier_def1}
 \lim_{k\to \infty } \parallel p_{x_{i},1}(kT) - h_{(i+1),1} \parallel=0,\text{for all $i={1, 2, . . ., K-1}$}
\end{equation}

and for each group $x_{i}$, there exists a permutation of the set $2,3,...,n$ as ${y_{1}^{x_{i}},y_{2}^{x_{i}},...,y_{n-1}^{x_{i}}}$ such that:

\begin{equation}\label{eq:k-Barrier_def2}
 \lim_{k\to \infty } \parallel p_{x_{i},y_{j}^{x_{i}}}(kT) - h_{i,j} \parallel=0,\text{for all $j={2,3, . . ., n}$}
\end{equation}

\begin{figure}[H]
  \centering
  \includegraphics[width=10cm]{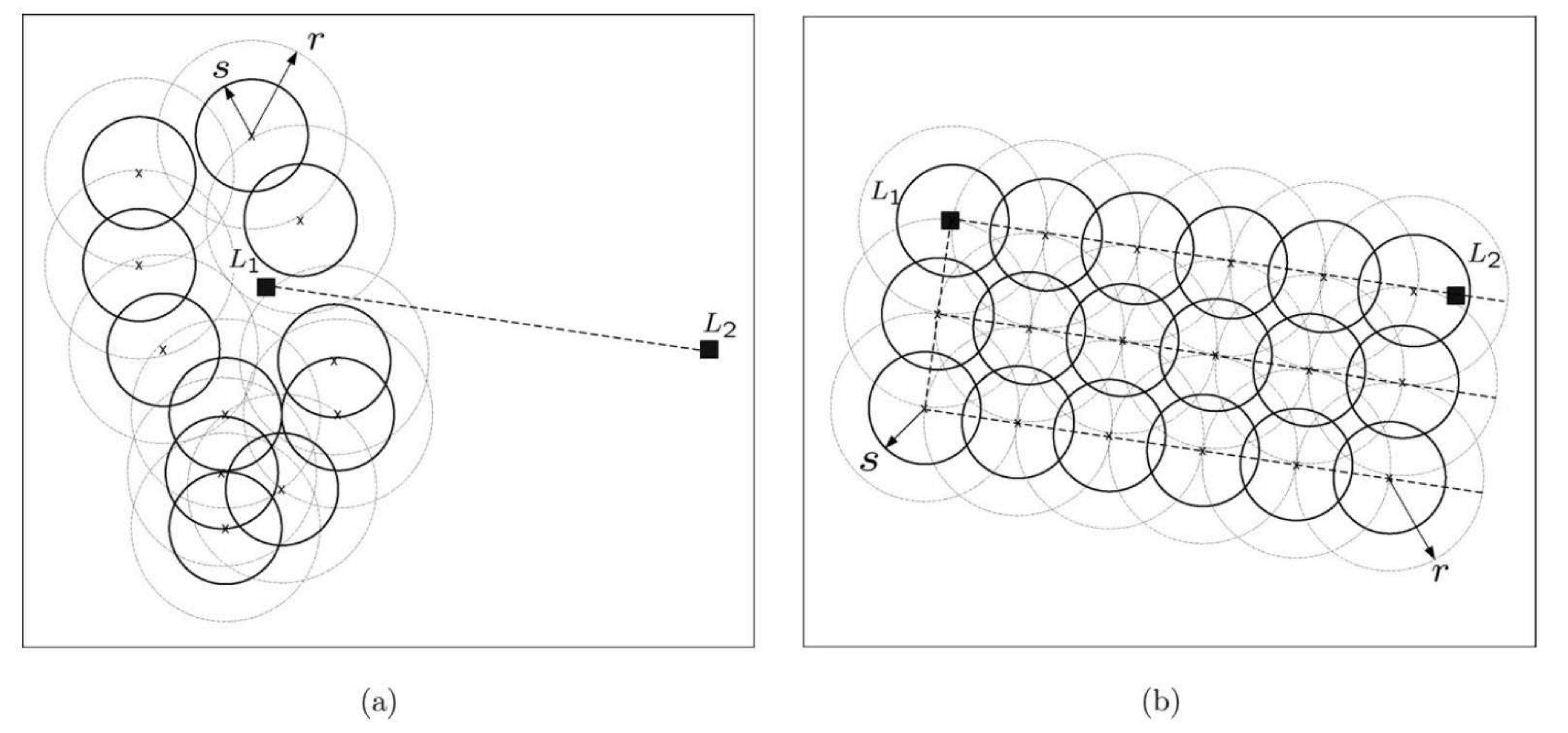}
  \caption{Multilevel Barrier \cite{cheng2012self}}
  \label{fig:Multilevel_Barrier}
\end{figure}

In \cite{lin2009accumulation}, the deployment of sensors to achieve k-barrier coverage has been considered with accumulation point model (APM). The simulation results show that the network lifetime with APM is higher than that of random point barrier coverage model. The APM also offers a comparatively better fault tolerance than independent belt barrier coverage model.

In \cite{nezhad2012novel}, the authors present a fault tolerant k-barrier coverage protocol. The simulations results show that this protocol enhances network lifetime in comparison with the randomized independent sleeping (RIS) method of \cite{kumar2004k}. A concept of localised barrier coverage protocol (LBCP) has been presented in \cite{chen2010local} and this protocol provides near optimal enhancement in the network lifetime, while achieving global barrier coverage most of the time. This protocol outperforms the RIS \cite{kumar2004k} by up to six times.

%Based on the strength of the sensing fields and the possibility of the intruder detection, the Barrier Coverage can be considered as weak and strong. The formal definition in terms of mathematical formulation has been given in \cite{cheng2011decentralized}. Some of the developed conditions \cite{cheng2011decentralized} can also define the quality of surveillance in terms of weak, stronger and full coverage. In a general consideration, the weak coverage is to be employed with the minimum number of mobile sensors covering a certain area. The weak coverage is useful where the cost is of prime importance. It does not consider the failure/ malfunctioning of the mobile sensors and the intruder might remain undetected. However, the strong coverage is achieved with the pre-defined critical condition(s) meant for the worst case scenario. Such sort of problems are still under investigation \cite{cheng2011decentralized}.

%\cite{jia2014autonomous}
%\cite{santoso2010sub}
%\cite{cheng2011decentralized}
%\cite{}
%\subsubsection{Single-line Barrier Coverage}
%\subsubsection{Multi-level Barrier Coverage}
%\cite{cheng2010distributed}
%
%
%\cite{kumar2005barrier}

\section{Sweep Coverage} \label{Sweep_Coverage}

If the mobile robotic sensors move together in an area such that there is a specific balance between maximizing the number of detections per unit time and minimizing the number of missed detections per area, such sort of moving arrangement of mobile robotic sensors is known as Sweep Coverage - which can also be exhibited by a moving Barrier Coverage \cite{gage1992command}.

The sweep coverage can be defined as if the formed Barriers move together with the constant speed, while maintaining an equal distance between each other. More formally, let us assume that there are mobile robotic sensors placed randomly to monitor the points of Interest (POIs) say in a particular region (straight, circular or curvilinear). Let us also assume that all the mobile sensors will move with the same sweeping speed. We can say that POI is sensed during sweep covered by mobile robotic sensor(s) if and only if the POI has once fallen under the sensing range of mobile robotic sensor sweeping with speed at a specific time instance. It is assumed that all the sensors are placed on mobile robots, so that the points which are not covered by the stationary sensors could be covered during sweeping phenomenon. In Barrier coverage as the mobile sensors are static so the predefined corridor is covered at all times. However, in case of mobile sensors the POI is considered for a certain time interval depending on the control variable and the constant speed of the mobile robotic sensors.

\subsection{Sweep Coverage Classification}
We classify the problem of sweep coverage on the basis of the type of environment.

\subsubsection{Sweep Coverage along a Line}

In \cite{cheng2011decentralizedsweep}, the problem of sweep coverage along a line is formulated as under:

First, a moving line $L(kT)$ is defined.

\begin{equation}\label{eq:moving line}
L(kT):= \{ p \in \mathbb{R}^2:p^Tl=F+kTv_{0} ,k=0,1,2,...\},
\end{equation}

\begin{equation}\label{eq:moving points}
h_{i}(kT)=\begin{cases}
    h_{0}(kT)+(s \times i)l,  & \text{if $B \subset C_{1}$}\\
    h_{0}(kT)-(s \times i)l, & \text{if $B \subset C_{2}$}\\
   \end{cases}
\end{equation}

\begin{figure}[H]
  \centering
  \includegraphics[width=10cm]{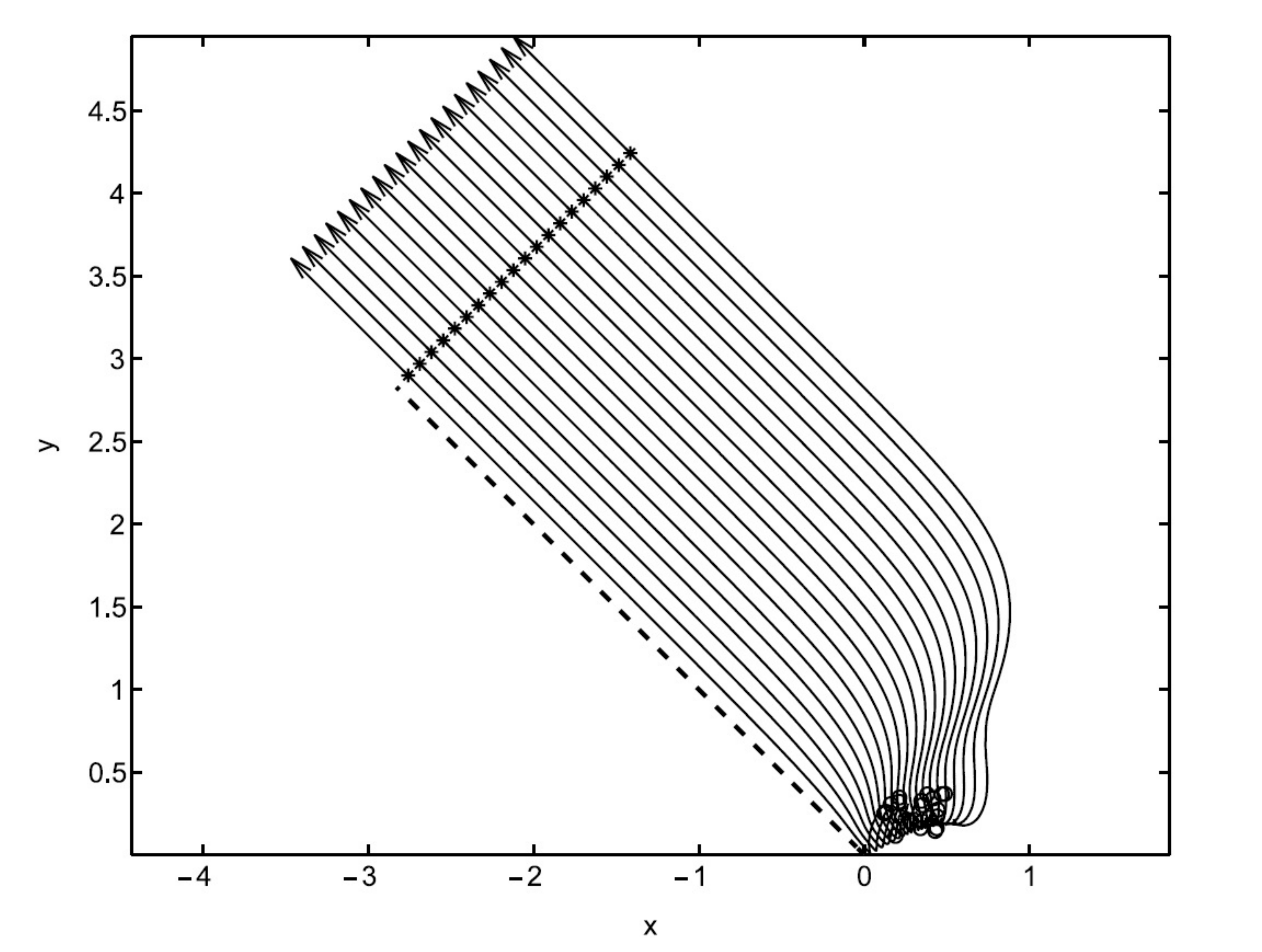}
  \caption{Sweep coverage along a line \cite{cheng2011decentralizedsweep}}
  \label{fig:sweep_line}
\end{figure}

The control for sweep coverage problem has been studied in \cite{cheng2011decentralizedsweep} and its formal definition is as under:\\
Given $n$ autonomous mobile robots, a line $W$ and scalars $s > 0$, $v \neq 0$, and $\phi_{W}$ that is associated with the line $W$. A set of decentralized control laws is said to be a sweep coverage coordinated control with sweeping speed $v_{0}$ along the line $W$ in the direction of $\phi_{W}$ and with the equidistant of s between vehicles if for almost all initial vehicle positions, there exists a permutation of the set ${1, 2, . . . , n}$ denoted by ${z_{1}, z_{2}, . . . , z_{n}}$ such that the following condition holds:

 \begin{equation}\label{eq:Sweep_line}
 \lim_{k\to \infty } \parallel p_{z_{i}}(kT) - h_{i}(kT) \parallel=0 \text{ for all $i={1, 2, . . ., n}$}\\
\end{equation}

\subsubsection{Sweep Coverage along a Corridor}
In the corridor sweep coverage problem, the mobile robots are required to proceed through a corridor as a moving barrier with a desired sweeping speed. Mathematically, the authors \cite{cheng2011decentralized} formulate a corridor as under:\\

Let $C \in \mathbb{R}^2$ be a corridor formed by two-dimensional region between two parallel lines (say $W_{1}$ and $W_{2}$). Let $l=[l_{1} \text{ } l_{2}]^T$ be a given vector, and let $d_{1}$ and $d_{2}$ be given scalars associated with the lines $W_{1}$ and $W_{2}$, respectively. If $d_{1}>d_{2}$ then a corridor $C$ formed by the intersection of two regions defined by the lines $W_{1}$ and $W_{2}$ can be written as:

\begin{equation}\label{eq:Corridor}
C:= \{ (x,y) \in \mathbb{R}^2:l^T[x \text{ } y]^T \leq d_{1}\}\\
\bigcap \{ (x,y) \in \mathbb{R}^2:l^T[x \text{ } y]^T \geq d_{2}\}
\end{equation}

Let $B \subset C$ be a line segment connecting $W_{1}$ and $W_{2}$ and $\theta_{0} \in [0,\pi)$ be the angle of the corridor with respect to x-axis. Then, the authors \cite{cheng2011decentralized} formulate the sweep coverage problem by defining a moving line (say $L(kT)$) with a desired sweeping speed $v_{0}$ meeting some necessary assumptions.

\begin{equation}\label{eq:moving line_corridor}
L(kT):= \{ p \in \mathbb{R}^2:x\cos(\theta_{0})+y\sin(\theta_{0}) =\\
   F+kTv_{0} \text{ ,$k=0,1,2,...$ }
\}
\end{equation}
where $F$ is some scalar. Next, the authors \cite{cheng2011decentralized} define $n$ points (say $h_{i}(kT)$) on the above mentioned line.

\begin{equation}\label{eq:points_corridor}
h_{i}(kT):= h_{0}(kT)+i  (\frac{d_{1}-d_{2}}{n+1})l  \text{ , for $i=1,2,...,n$ and $k=0,1,2,...$ }
\end{equation}

where,

\begin{equation}\label{eq:reference_point_corridor}
h_{0}(kT):= L(kT) \cap  \{p \in \mathbb{R}^2: \sin(\theta_{0})x- \cos(\theta_{0})y=d_{2}\}
\end{equation}

\begin{figure}[H]
  \centering
  \includegraphics[width=10cm]{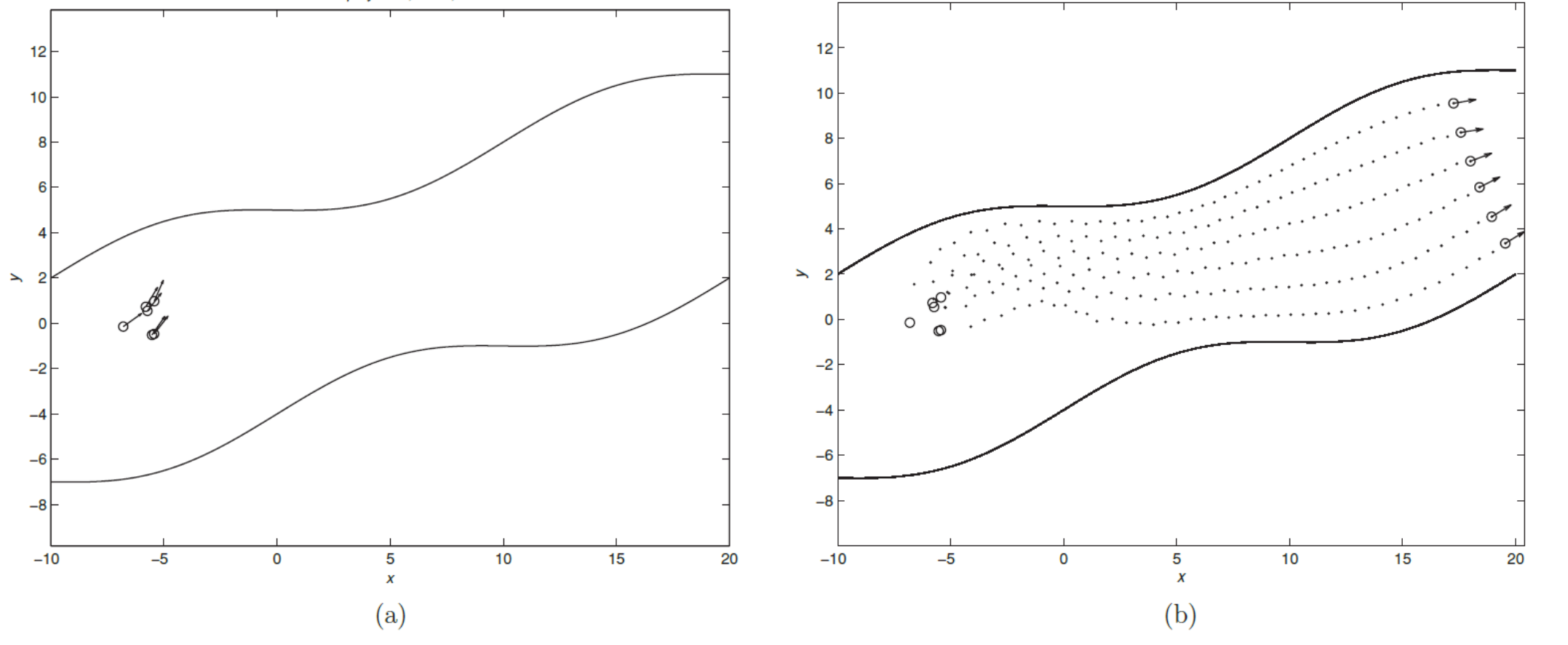}
  \caption{Sweep coverage along a corridor \cite{cheng2011decentralized}}
  \label{fig:sweep_line}
\end{figure}

Then, the work \cite{cheng2011decentralized} formally define the corridor sweep coverage problem as under:\\
Let $C$ be a given corridor with angle $\theta_{0}$ and $v_{0}$ be the desired sweeping speed of the mobile sensors. Then, the mobile robots are called to be driven by an optimal corridor sweep coverage control law with a decentralized strategy if for almost all initial mobile sensor positions, a permutation of the set ${1, 2, . . . , n}$ denoted by ${z_{1}, z_{2}, . . . , z_{n}}$ exists such that the following condition holds:

 \begin{equation}\label{eq:Sweep_corridor}
 \lim_{k\to \infty } \parallel p_{z_{i}}(kT) - h_{i}(kT) \parallel=0 \text{ for all $i={1, 2, . . ., n}$}\\
\end{equation}

\subsection{Sweep Coverage Approaches}
We categorise the problem of sweep coverage on the basis of approaches/ methodologies used in the literature.

%
%\cite{cheng2008sweep}

\subsubsection{Sweep Coverage based on Nearest Neighbour Rule}

The nearest neighbour rule has been described in the section \ref{BC:NNR}. This locally calculated rule has been used in the literature to solve the problem of sweep coverage with decentralized control laws. One can see the work (e.g. \cite{cheng2011decentralizedsweep,cheng2011decentralized}), where this rule has been used as a fundamental to calculate the control inputs: velocity $v_{i}(kT)$ and heading angle $\theta_{i}(kT)$ of a mobile robot.

%\subsubsection{\textbf{Sweep Coverage with virtual force field}}
%\subsubsection{\textbf{Sweep Coverage with artificial potential field}}
%\subsubsection{\textbf{Sweep Coverage with Voronoi diagram and Delaunay triangulation}}
%\subsection{\textbf{Sweep Coverage: Deployment approaches}}

%\cite{ahmad2013decentralized,cheng2009decentralized}

%\subsubsection{Sweep Coverage along a Line}
%
%\subsubsection{Sweep Coverage along a Corridor}

%\section{Barrier and Sweep Coverage}
%
%The research paper \cite{cheng2011decentralized} is providing the Barrier and Sweep coverage algorithm by mobile robotic sensors. The mobile robotic sensors are driven through decentralized control law to achieve both barrier and sweep coverage.

%\section{\textbf{Search Algorithms}}

%\section{Encircling Coverage}
%
%In this coverage, a network of mobile robotic sensors is deployed to encircle a planar region. In \cite{savkin2013algorithm}, the authors deploy the sensors in the region $R(d_{min},d_{max})$ such that all the points between $d_{min}$ and $d_{max}$ are sensed by at least one mobile robotic sensor. This coverage is inspired from the well-known Hannibal double envelopment maneuver during the Battle of Cannae \cite{goldsworthy2012fall} %\cite{chen2014distributed}.
%
%
%
%\begin{figure}[h]
%  \centering
%  \includegraphics[width=10cm]{Encircling_Coverage.eps}
%  \caption{A region encircled by mobile sensors \cite{savkin2013algorithm}}
%  \label{fig:sweep_line}
%\end{figure}

\section{Heuristic Coverage Algorithms}\label{Heuristic_Coverage}

Heuristic algorithms can be used to minimize the coverage time. Some of the Heuristic coverage control algorithms have been compared by \cite{savkin2015decentralized}. For example, the authors \cite{zheng2005multi} have shown that NP-complete algorithm can be used to minimize coverage time, and the authors have also introduced polynomial-time multi-robot coverage heuristic. The mobile robots can achieve triangular lattice pattern by the heuristic algorithm as proposed in \cite{chang2009obstacle}. However, the validation has only been performed by the simulation study. Another heuristic algorithm was presented in \cite{ma2007adaptive}, but the algorithm can only achieve an equilateral triangular lattice pattern. We present some other well-known heuristic coverage algorithms as under:

\subsection{Ant-Like Algorithms}

The work \cite{fankhausersafe} reviews that Ant-like robots can also be used for coverage algorithms. Ant like robots cannot keep maps in the memory and are not capable to perform convectional path planning due to limited sensing and computation capability, a heuristic coverage algorithm is a good choice for these sort of robots to achieve coverage \cite{koenig2001efficient}. These sort of heuristic algorithms do not need any localization or any information about the area. Some research work \cite{wagner2008cooperative} has been done to achieve heuristic sweeping on the floor.

\subsection{Physicomimetics}

The work \cite{spears2006physics} considers the swarm of robots as gas and each individual robot is considered as a gas particle. This method can be used for sweeping on a bounded region and there is no information required a priori. However, this algorithm always require plenty of mobile robots in order to maximize the sweep coverage in the unknown region.

\section{Dynamic Coverage (Multi-robot Search and Rescue) Algorithms}\label{Dynamic_Coverage}

%\subsection{\textbf{Dynamic Coverage with virtual force}}
%\subsection{\textbf{Dynamic Coverage with Voronoi diagram}}

%\cite{baranzadeh2015distributed}

In this type of coverage algorithms, the mobile robotic sensors dynamically search a given region such that each point in the region is sensed for a certain preset level. The mobile robotic sensors in dynamic coverage algorithms require consideration for mutual collision avoidance and preserving mutual communication linkages. As compared with static deployment strategy, the number of mobile robotic sensors is significantly reduced with dynamic sensor networks. This type of coverage has special importance in search and rescue missions.

In (\cite{wagner1996cooperative,wagner1999distributed}), the authors consider the ability of a group of robots leaving chemical odour traces for communication purpose and the group's ability to perform the task of cleaning the floor of an un-mapped building or any other task which requires the traversal of an unknown network. This work has been compared with references therein (\cite{russell1995laying,russell1993mobile,wagner1997cooperative}) for cleaning task by multiple robots with some sort of guidance. The described algorithms are decentralized in nature, which make connected robots as adaptive to complete the traversal of the graph even if some robots fail or the graph changes during the execution.

The work \cite{wagner1998efficiently} describes a method for searching an undirected connected graph by the use of Vertex-Ant-Walk (VAW) like algorithm, where a robotic sensor walk along the edges of a graph, while occasionally leaving "pheromone" traces at nodes. These traces are used to assist in exploration, while offering a trade-off between random and self-avoiding walks, as it forces a lower priority for repeated visit to neighbours. The trace-oriented search has also been performed in (\cite{kozen1978power},\cite{blum1977capability}), where pebbles are used to guide the search. The authors with references therein compare the work with other graph search methods based on deterministic (\cite{algorithms1979computer,fraenkel1971economic,hopcroft1973algorithm,tarjan1972depth,tarry1895probleme}), random (\cite{aleliunas1979random,barnes1993short,broder1994trading}) and semi-random covering (\cite{gal1990search}). This method \cite{wagner1998efficiently} also exhibits properties of modularity and a possible convergence to a limit cycle. The work \cite{wagner2000ants} further investigates performance of the VAW method on dynamic graphs, where edges may be added or subtracted during the search process.

In \cite{wagner2000mac}, three methods have been considered to explore a continuous planar region by a group of mobile robots having limited sensors and the team of robots is without any explicit communication. The deterministic mark and cover (MAC) algorithm has guaranteed coverage within time and aware of the completion, but it might have problems: vanishing of the traces with time and sensory measurement errors. The randomised probabilistic covering (PC) method is free of sensor dependency (with an exception to sense collision), but it has no awareness of the completion. There has been shown that a third hybrid algorithm by combining the two methods can be used to better trade-off between the performance of first method and the robustness of the second method. However, all the methods have been considered in a two-dimensional space with an obstacle free environment.

In \cite{wagner2008cooperative}, the authors consider cooperative robots for a cleaning problem in a two-dimensional non-convex region. The dirt on the floor has been used as the main means of the inter-robot communication.

The work \cite{altshuler2005swarm} considers cooperative cleaner problem of \cite{wagner1997cooperative}, but in an environment where changes can occur regardless of an agent's activity. The authors compare the work with multi agents robotics of (\cite{wagner1997cooperative,wagner1998efficiently,arkin1998cooperative}), where a static environment has been considered. The work \cite{altshuler2005swarm} considers a grid, part of which is 'dirty', where the 'dirty' part consists of connected region of the grid. The multi-agents can move on the dirty grid region to clean it. The problem involves a deterministic evolution of the environment. However, this work has been considered in two-dimensional obstacle free environment.

The work \cite{altshuler2008swarm} considers a centralized optimal algorithm for dynamic cooperative cleaners problem of \cite{altshuler2005swarm}. A similar work has been presented in (\cite{polycarpou2001cooperative,koenig2001terrain,rekleitis2004limited}). The authors assume a connected region as grids part of which is "dirty". There is a deterministic evolution of the environment. The authors compare the work with the SWEEP protocol of \cite{altshuler2005swarm}. The simulation results show that the performance of SWEEP protocol  \cite{altshuler2005swarm} comes closer to the optimal algorithm \cite{altshuler2008swarm} as the problem gets harder.

The survey paper \cite{choset2001coverage} focuses on coverage path planning algorithms for mobile robots constrained to operate in two-dimensional space. The work classifies the path planning algorithms as either heuristic or complete. Some of the dynamic coverage algorithms are based on ant like algorithms (see e.g. \cite{wagner1998efficiently,wagner2000mac,wagner2008cooperative,wagner1997cooperative,wagner1996smell,wagner2000ants,wagner1996cooperative,feliudistributed,yamakita2004formation,altshuler2005cooperative,altshuler2008efficient,altshuler2006swarm,altshuler2011multi,altshuler2006shape,osherovich2008robust,altshuler2005swarm,altshuler2009collaborative}).

In case of dynamic network coverage, the authors \cite{gabriele2008area} have shown a comparison by considering the percentage of the covered area as a function of time. Based on some necessary assumptions, it has been numerically shown that a decrease in number of mobile robotic sensors increases the time required to fully cover an area.

In \cite{hussein2006effective}, a given number of mobile robotic sensors dynamically cover a known two-dimensional region. A deterministic approach has been used and each agent has access to the state histories of all other agents in the team. The authors in \cite{hussein2007effective} modify dynamic coverage control strategies of \cite{hussein2006effective} and include flocking of the agents along with guaranteed collision avoidance. The flocking of the agents has been considered to guarantee reliable communications links among the agents. In \cite{hussein2007effective}, the authors further modify the control law to guarantee that a partially connected group of agents also achieves the coverage goal.

\begin{figure}[H]
  \centering
  \includegraphics[width=10cm]{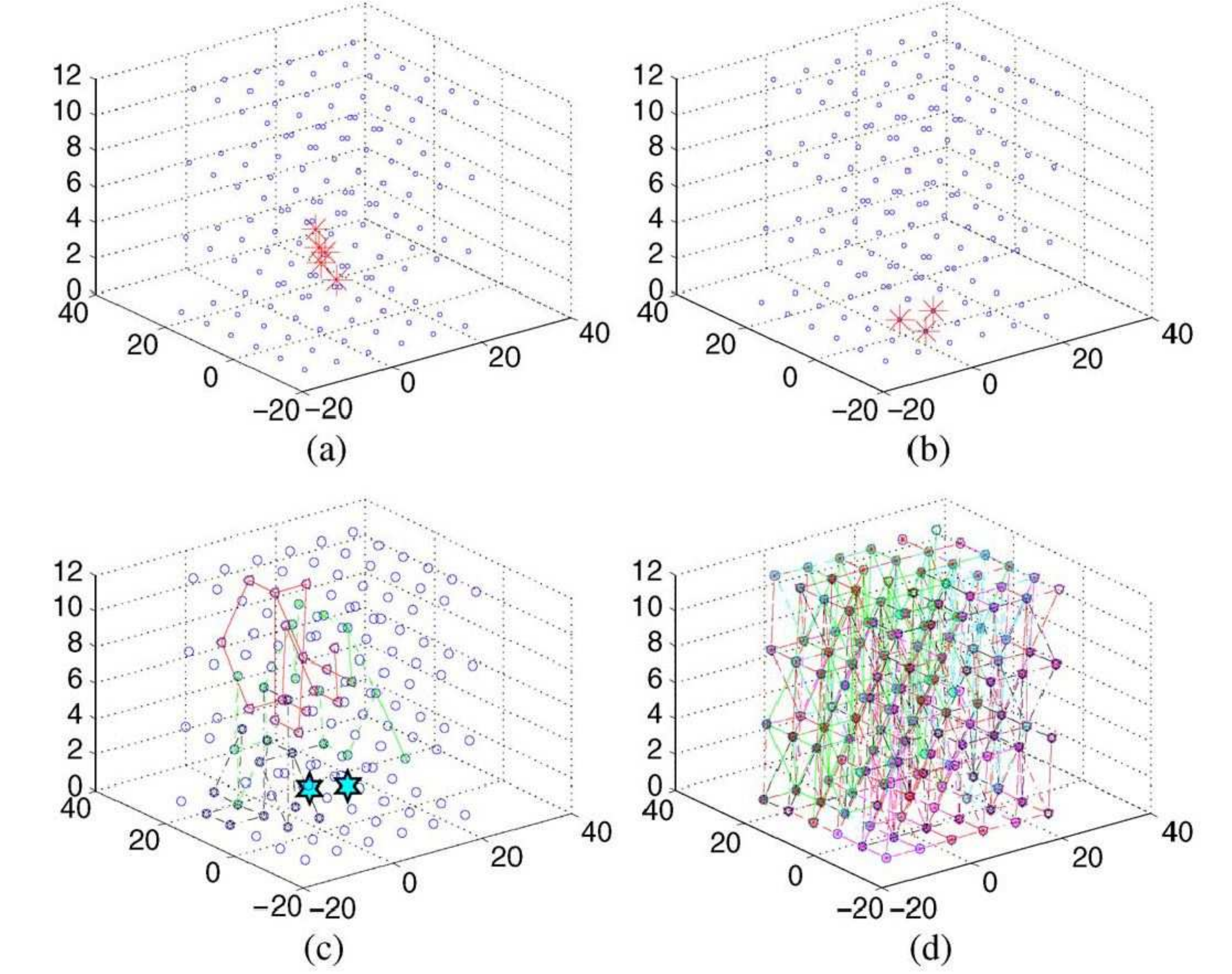}
  \caption{Dynamic search coverage in three-dimensional space \cite{nazarzehi2015distributed}}
  \label{fig:3D_Dynamic_Coverage}
\end{figure}

The work \cite{nazarzehi2015distributed} considers the problem of dynamic search coverage in three-dimensional space. A randomized deployment approach has been considered with no prior information about the three-dimensional region. However, the mobile robots are capable of detecting the boundary of the bounded region. The mobile robots follow the vertices of a Cubic and Truncated Octahedron grid based patterns. The authors show that the Truncated Octahedron grid based search minimizes the coverage time than the one performed by Cubic grid based pattern. The convergence of the proposed algorithm has been mathematically proved with probability '1'. As shown in the below mentioned figure (Fig. \ref{fig:3D_Dynamic_Coverage} from \cite{nazarzehi2015distributed}), the First scenario (Fig. \ref{fig:3D_Dynamic_Coverage} c) considers stopping the search after a known number of targets have been found. The Second scenario (Fig. \ref{fig:3D_Dynamic_Coverage} d) with no prior information about the number of targets continues the search unless all parts of the region are fully searched. A mobile robot uses the nearest neighbour rule based approach to calculate its coordinates in three-dimensions.

\section{Three-Dimensional (3D) Coverage} \label{3D_Coverage}

%\cite{nazarzehi2015distributed}
%\cite{nazarzehidecentralized}
%The work \cite{bai2010optimal} presents optimal deployment pattern achieving four-connectivity and full coverage under different ratios of the sensor's communication range ($R_{c}$) to the sensing range $R_{s}$. The authors propose a Diamond pattern, which has been proved asymptotically optimal when $\frac{R_{c}}{R_{s}} > \sqrt{2}$.

The optimal sensor deployment strategy in three-dimensional space has been analyzed in \cite{watfa2006optimal}. The authors have compared the results of random deployment strategy with that of a body centred cubic (bcc) lattice pattern. It has been shown that the bcc lattice deployment strategy requires minimum number of sensor nodes to cover a three-dimensional space. The distributed coverage problem in a three-dimensional space has been considered in \cite{watfa20063}. The issue of a three-dimensional coverage with minimum number of sensors has not been considered with the related references therein ( see e.g. \cite{gupta2006connected,cardei2002wireless,slijepcevic2001power,guha1996approximation,chen2002approximating,srikant2003unreliable,lieska1998radio,huang2005dynamic,patwari2003relative,gage1992command,howard2002mobile,zhang2005maintaining,heinzelman2000energy,zheng2005pmac,wang2003integrated}). In this work, a distribution of sensor nodes is given and a subset of active nodes is selected for full coverage. The optimum three-dimensional coverage of a given region is achieved by the minimum number of sensors. However, the obstacle avoidance technique still remains as the future consideration for this problem.

The network design principle for ocean sensors has been surveyed in \cite{wang2012three}. The survey suggests that there are only a few ocean sensor networks deployed in the bottom \cite{akyildiz2006state} or at the surface (\cite{guo2008oceansense}) of the ocean. The survey mainly emphasises on deployment, localization, topology design, and position-based routing \cite{abdallah2006randomized} specifically for three-dimensional ocean spaces. We briefly describe these key considerations of \cite{wang2012three} with references there in:

The work classifies the three-dimensional ocean deployment of sensors based on the static deployment, semi-mobile deployment and mobile deployment. In static deployment, the ocean sensors float at different depths \cite{cayirci2006wireless,pompili2009three,akyildiz2005underwater}. The semi-mobile deployment strategy requires the ocean sensors to adjust underwater positioning \cite{howe2004sensor,glenn2006leo,detweiler2012autonomous}. In mobile deployment, the ocean sensors can move vertically and horizontally \cite{zhang2012marine}.

%\subsection{Localization}

In underwater networks, GPS signal does not pass and RF signal is absorbed by water. The acoustic signal works underwater, but it has limitations due to low bandwidth, high propagation delay and high bit error rate. The velocity of acoustic signal can also change with salinity, pressure and temperature. There is also requirement of more anchor nodes to locate ocean nodes. The anchor nodes have been categorised as static or mobile (see e.g. \cite{erol2007localization,erol2007auv,erol2008multi}). The localization algorithms are also designed centralized or decentralized/ distributed. Thus, localization becomes one of the challenging tasks in the ocean sensor networks (see e.g. (\cite{erol2011survey,tan2011survey,han2012localization}).

%\subsection{Topology Design}

%The network topology can be static or dynamic to achieve the power efficiency, fault tolerance and throughput maximization. The authors classify the literature (with references therein) into four categories: topology design for coverage (\cite{akkaya2009self,cayirci2006wireless,pompili2009three,watfa20063,watfa2006optimal}), topology design for connectivity and coverage (\cite{bai2009full,bai2009low,alam2006coverage,ammari2010study,aslam2010distributed}), topology control for connectivity and power efficiency (\cite{wang2008energy,junseok2010adaptive,bahramgiri2006fault,ghosh2007efficient,poduri2009using,li2012localized}), handling fault tolerance and shadow zone(\cite{wang2009self,bahramgiri2006fault,preisig2007acoustic,domingo2009topology}). The underwater topology can be selected depending on the coverage objectives.

The network topology can be static or dynamic to achieve the power efficiency, fault tolerance and throughput maximization. The authors classify the literature into four categories: topology design for coverage, topology design for connectivity and coverage, topology control for connectivity and power efficiency, handling fault tolerance and shadow zone. The underwater topology can be selected depending on the coverage objectives.

In three-dimensional ocean coverage, routing is a way of delivering data packets from a source node to destination node via multihop relays. The authors further classify 3D Position based routing literature with references therein as greedy routing, routing via mapping and projection, randomised greedy routing, greedy routing over constructed structures and hybrid greedy routing.

The work \cite{pompili2009three} considers a deployment analysis for two-dimensional and three-dimensional underwater acoustic sensor networks. The authors present related work (see e.g. \cite{akyildiz2005underwater,hsin2004network,ravelomanana2004extremal,shakkottai2005unreliable,zou2003sensor}) - which is different from the current work mainly focusing on the issues of underwater acoustic sensor networks. The main objective of this work is to determine the minimum number of sensors achieving: application dependent target sensing and communication coverage, how to select the deployment surface area for a given target region, robustness in case of network nodes failure and providing an estimate of required number of redundant sensors.

In \cite{xiao2014novel}, the regular polyhedron models in three-dimensional scene is analysed and the minimum number of sensor nodes are deployed based on the subdivided 3D grids. The authors also derive the relationship between the coverage region and the sensing radius of the sensor nodes. However, the monitored region is considered to be a cuboid.

In \cite{shah2014survey}, the reliable communication in three-dimensional space has been considered. The authors have worked to identify the unique properties of communication environments. However, the covered region and control of agents dynamics for deployment is not part of the survey.

The sensing range of mobile robots is modelled as space tessellating polyhedral in \cite{mishra2010optimal}. The authors have tested cube, hexagonal prism, rhombic dodecahedron, and truncated octahedron based on the number of nodes needed for tessellation, overlapped region and symmetry in the lattice pattern. The authors have also considered a trade-off ratio between the amount of overlapping achieved and the number of deployed sensor nodes. However, the work is not focused on agents dynamics and the algorithm (control laws) to archive a pattern in the three-dimensional space.

In the paper \cite{fang2014coverage}, the authors consider the deployment strategy to a target field which can be 1-dimensional, 2-dimensional or 3-dimensional. In fact, the movement-assisted sensor deployment has been considered with different optimization objectives to enhance the coverage. The decentralized or distributed control of the considered robotic dynamics is not part of the work.

In \cite{lee2010three}, the problem of self-deployment in Three-dimensional space has been considered. The deployment pattern is forming a regular tetrahedron. The swarm of robots using local rules can span a network of regular tetrahedrons. This self-deployment method is based on the formation control of mobile robots in a distributed manner.

The formation of UAVs in 3-dimensional space has been considered in \cite{hattenberger2007formation}. However, the formation planning is done off-line on the basis of prior knowledge of the environment.

The cost of validating a control strategy in a Three-dimensional space is high. In \cite{wong2004autonomous}, there has been some research on developing a Three-dimensional graphical simulation test-bed environment. This test bed is to facilitate the design and testing of multi agent control algorithms specifically in Three-dimensional space. The work also highlights a potential application of inspection and repair of propulsion systems with a team of mobile robots operating in a Three-dimensional environment.

In \cite{anderson2008uav}, the authors present a decentralized control law for preserving the planar formation of UAVs in a Three-dimensional space. It has been considered that the shape of formation can be maintained if the distance between a certain numbers of the agent pairs is maintained. This will eventually result in the distance between all pairs being constant. However, this work is focusing on planar formation of UAVs and considers a constant speed in the UAV model.

The authors in \cite{moshtagh2005vision} also consider group agents in a Three-dimensional space. The proposed control law is distributed and it eventually ensures that velocity vectors converge to a constant heading angle and the mutual distance between agents is stabilised.

In \cite{alam2006coverage}, the authors consider the coverage of 3-dimensional space by placing the optimal number of nodes. The authors have shown using Kelvin’s
conjecture that the truncated octahedral tessellation of 3-dimensional space is the best strategy. The authors define a metric called volumetric quotient (ratio of the volume of a polyhedron to the volume of its circumsphere), which is a measure of the quality of the competing space-filling polyhedrons. The authors have found truncated octahedron with volumetric quotient of 0.68329, both optimized hexagonal prism (considered in \cite{decayeux2004new}) and rhombic dodecahedron (considered in \cite{carle2001basis}) with volumetric quotient of 0.477 and cube with that of 0.36755. The cells created by the Vornoi tessellations of 3-dimensional space decide the number of nodes required for full coverage of that space. The number of required nodes is smaller if the shape of each cell is a space-filling polyhedron with higher volumetric quotient. So, the truncated octahedron is the best choice with a much better volumetric quotient than that of other possible choices. The work also investigates maintaining the connectivity issue and the authors conclude that the truncated octahedron placement strategy requires the transmission range to be at least 1.7889 times the sensing range of the nodes. The results are based on extensive simulations and it opens guidance for developing the distributed/ decentralized control laws driving the nodes to the optimal placement strategy. However, the nodes are considered to be homogeneous in terms of sensing and communication range. The authors also remark that any rigorous proof of the discussed conjecture is very difficult to be found.

The work \cite{alam2008coverage} investigates the coverage and connectivity in three-dimensional under water sensor network system. The authors investigate the minimum number of node placement strategy providing full coverage and full connectivity in a three-dimensional space with all values of the ratio between a nodes communication range ($R_{c}$) and sensing range ($R_{s}$). The authors also find the node placement strategy providing full coverage and 1-connectivity with minimum number of nodes in a three-dimensional network with all values of the ratio, $\frac{R_{c}}{R_{s}}$  . The same problem for two-dimensional networks has been investigated in \cite{bai2006deploying,zhang2005maintaining,kershner1939number}. As stated above, the related work in three-dimensional space has considered a Voronoi cell as hexagonal prism \cite{decayeux2004new} and as rhombic dodecahedron (in \cite{carle2001basis}). However, the authors \cite{alam2008coverage} show that if $\frac{R_{c}}{R_{s}}\geq \frac{4}{\sqrt{5}}$, the truncated octahedron used to model the cell reduces the required number of nodes by $43.25$ percent as compared with the cell model of presented in \cite{decayeux2004new} (hexagonal prism) and \cite{carle2001basis}) (rhombic dodecahedron). The aforementioned work does not deal with the problem when $\frac{R_{c}}{R_{s}} < \frac{4}{\sqrt{5}}$. As shown in \cite{barnes1983optimal}, the body centred cubic (BCC) lattice structure has the smallest mean squared error of any lattice quantizer in three-dimensional space. Hence, it implies that the BCC lattice structure pattern deploys minimum number of nodes achieving full coverage, but it does not address the nodes connectivity issue. The authors extend their previous work \cite{alam2006coverage} presented in the above mentioned paragraph to discuss the coverage and connectivity issue when $\frac{R_{c}}{R_{s}} < \frac{4}{\sqrt{5}}$. So, the authors solve the problem for all values of $\frac{R_{c}}{R_{s}}$ in \cite{alam2008coverage}. The authors considering a three-dimensional space find the best strategy for full coverage and full connectivity with all geographically neighbouring nodes by placing minimum number of sensor nodes among the four considered models: Cube, Hexagonal Prism, Rhombic Dodecahedron and Truncated Octahedron. The results are listed as under based on $\frac{R_{c}}{R_{s}}$ of a node:

\begin{enumerate}
  \item If $\frac{R_{c}}{R_{s}} \geq 1.587401$ then select modified Truncated Octahedron placement strategy.
  \item If $1.211414 \leq \frac{R_{c}}{R_{s}} < 1.587401 $ then select modified Hexagonal Prism placement strategy.
  \item If $\frac{R_{c}}{R_{s}} < 1.211414$ then select modified Cube placement strategy.
\end{enumerate}

The authors also provide a solution of full coverage with 1-connectivity (instead of maintaining a full connectivity) with fewer number of nodes in case $\frac{R_{c}}{R_{s}}$ is very small. However, the rigorous proof of the work is still to be analysed. A sphere based sensing and communication model considering all the nodes to be the homogenous has been assumed in deciding the deployment strategies.

The work \cite{alam2006coverage,alam2008coverage} assumes that nodes can be deployed and maintained at the desired locations in three-dimensional space, but this assumption is less practical in deploying larger number of underwater sensor nodes. The work \cite{alam2015coverage} is different from \cite{alam2006coverage,alam2008coverage} in the sense that the authors consider no control over a node movement and the main goal is to find distributed scalable scheme dynamically determining the subset of active nodes. In fact, node position can be random and a large number of redundant nodes ensure that every point of the network is within the sensing range of at least one node. The authors also present some related work \cite{basagni2004sensor,chen2002span,ye2003peas,zhang2005maintaining}, where only a subset of nodes is kept active for a two-dimensional sensor network. Based on the sensing and communication range of sensor nodes, this scheme divides the three-dimensional network space into identical regions. Then, one sensor node is dynamically and locally selected (e.g. \cite{lynch1996distributed}) to sense its region, while maintaining its connectivity with the active nodes of the neighbouring region. A similar idea for two-dimensional network has been used in \cite{xu2001geography}, but this work use truncated octahedral tessellations of three-dimensional space to find the best possible division minimizing the number of active nodes at any time. In this work, the constraints are that the circumspect of each region cannot be greater than the sensing range ($R_{s}$) and the maximum distance between two furthest points of the neighbouring region cannot be greater than the communication range ($R_{c}$). However, the authors assume a sphere based sensing and communication model of a node and all the nodes are considered to be homogeneous.

The work \cite{feng2014coverage} considers a three-dimensional environment with complex terrain in the real world applications. The authors compare this work with \cite{huang2004coverage,watfa2006optimality}, where three-dimensional space is considered as ideal. The deterministic sensor-deployment problem has been proved (\cite{zhao2009surface,o1992computational}) as nondeterministic polynomial time complete (NP-complete), which means that it cannot be done in polynomial time and it has only an approximate solution. The authors compare the work with two-dimensional coverage (\cite{hoffmann1991art,shermer1992recent,dhillon2003sensor,liu2005mobility,lazos2006stochastic,howard2002mobile,alam2006coverage,zhang2005maintaining,wang2005efficient,huang2005coverage}) and three-dimensional coverage (\cite{alam2006coverage,bai2009full,oktug20083d,zhao2009surface,liu2012coverage,jin2012optimal}). A brief review of the related work shows that the coverage holes of the whole wireless sensor network have not been considered. So, this algorithm takes the coverage parameters (cost factor, reduction parameter and perceived probability parameters) to complete the transformation of points from three-dimensional space to the two-dimensional plane via dimensionality reduction method. The dimensionality reduction method can maintain the topology characteristic of the non-linear terrian. The measure to evaluate the coverage is based on the detection probability in the optimal breath path. The work presents simulation results showing full coverage area with fewer sensors. However, this algorithm still considers obstacle avoidance as the future problem.

The work \cite{boufares2015three} focuses on redeploying the randomly deployed nodes to cover a three-dimensional space and maintain connectivity of the nodes as well. In fact, the mobile nodes are randomly scattered within the region to be covered. These randomly scattered nodes neither have full coverage of the region nor maintain at least one connectivity path among them. The authors use the constraint $R_{c}\leq R_{s}$ to ensure coverage and connectivity at the same time \cite{bai2006deploying}. This distributed redeployment algorithm is based on virtual forces-based strategy \cite{zou2003sensor}. This algorithm ensures the nodes provide full coverage of the region and it also maintains connectivity among them. The authors work is different from the related work with references therein (see e.g., \cite{poduri2006sensor,ravelomanana2004extremal,huang2007coverage,huang2004coverage,aslam2010optimizing,alam2006coverage,bai2009full,bai2009low,li2012deploying}). The simulation results show coverage and connectivity in three-dimensional space. However, the obstacle avoidance technique stays as one of the future considerations.

In \cite{nazarzehi2015decentralized}, a decentralized random algorithm has been developed to drive a team of mobile sensors on the vertices of a truncated octahedral grid ensuring complete coverage of a bounded three-dimensional region. The algorithm is based on the nearest neighbour rule. In addition, the mobile robots can also make a desired three-dimensional geometric shape as shown in the below mentioned simulation results.

\begin{figure}[H]
  \centering
  \includegraphics[width=10cm]{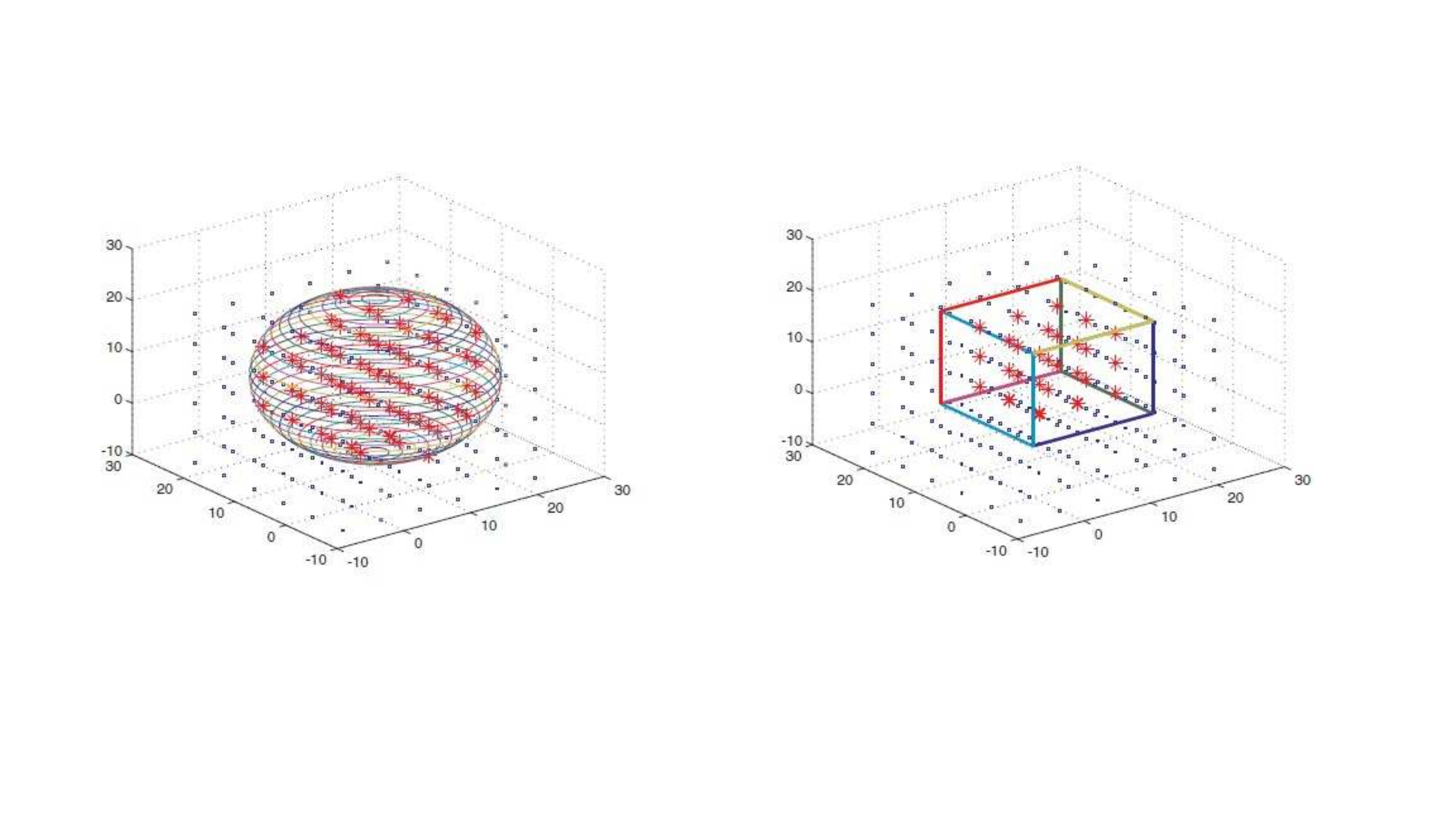}
  \caption{Complete three-dimensional coverage forming spherical and cuboid pattern \cite{nazarzehi2015decentralized}}
  \label{fig:3D_Coverage}
\end{figure}

%The work \cite{bai2010optimal} presents optimal deployment pattern achieving four-connectivity and full coverage under different ratios of the sensor's communication range ($R_{c}$) to the sensing range $R_{s}$. The work presented in \cite{kershner1939number} states that regular triangle lattice pattern is asymptotically optimal in terms of covering a planar area with minimum number of circles. The same result was proved in \cite{zhang2005maintaining}. The authors present a four connectivity and full coverage pattern which is different from the related work \cite{wang2005efficient,iyengar2005low,bai2006deploying,hochbaum1985approximation,melissen1996improved,nurmela2000covering,gonzalez1991covering,clarkson2007improved,toth2005covering,slijepcevic2001power,tsai2007coverage,abrams2004set}. The authors propose a Diamond pattern, which has been proved asymptotically optimal when $\frac{R_{c}}{R_{s}} > \sqrt{2}$.

\section{Formation Building}\label{Formation_Building}

The decentralized/ distributed control to drive the multiple autonomous agents into a desired pattern is an active research area. The main challenge in developing the decentralized/ distributed control law is to control the adaptive formation building from the random initial conditions.

In \cite{savkin2009decentralized}, different formations (like the right flank, square, wedge, etc) are attained with the decentralized control law. The authors consider two cases: leaderless and with a leader. In leaderless case, the vehicles attain same heading angle with a constant speed. In the second step, there is a leader and the leader direction is coordinated to the local neighbours. Eventually, all the connected vehicles attain the same heading angle as that of the leader. The control law has been developed by considering the vehicles physical constraints on linear and angular velocities. However, this formation building has been considered in plane.

In \cite{jia2015leader}, the authors present cohesive and formation flocking on the kinematic model of the robotic fish with the associated nonholonomic constraints. The flocking is based on a leader follower scenario. There has been some potential function methodology for causing attraction or repulsion among the fish. The main objective of this work is to make followers reaching leader's speed and to maintain an equilibrium distance between any two neighbours while avoiding collision. However, it has been considered that the external input to the leader is zero. The arbitrarily shaped flocking remains as future extension of this work. This work has considered the robotic fish modeled in a two-dimensional Euclidean space.

In \cite{jia2014distributed}, the authors consider a distributed leader-follower adaptive flocking problem. This work considers adaptive flocking which is the main difference from \cite{jia2015leader}. The leader's speed can vary within a bounded set. Similar to \cite{jia2015leader}, authors have proposed a combination of consensus and attraction/repulsion functions to respectively solve the cohesive flocking problem and the formation flocking problem. However, this work asymptotically tracks the leader’s varying velocities and maintains the equilibrium distances among the neighbours. The stability of the system has been analysed on LaSalle-Krasovskii invariance principle and computer simulations have also been provided. The problem has been considered in a two-dimensional Euclidian space.

The authors \cite{jia2014experimental} combine consensus protocol with attraction/ repulsion potential functions. The main objective is to make agents swim together with consistent velocities and maintain an equilibrium distance between any two neighbours. The LaSalle-Krasovskii invariance principle, computer simulations and practical experimental results have been obtained to validate the control methodology. This flocking algorithm is different from \cite{jia2015leader,jia2014distributed} as no leader follower scenario has been considered and there has been performed a practical experiment using three robotic fish inside a swimming pool. The authors have remarked that the water wave disturbances and a bounded experimental field (swimming pool) make the problem difficult. However, these disturbances are a good source to test the capability of the algorithm against the field limitations and uncertain disturbances.

In \cite{savkin2011method}, the authors consider a bounded planar region with a finite number of obstacles. The authors propose a decentralized control algorithm to drive the randomly placed and connected mobile robots on the vertices of square grids. The length of the side of the square grid is known to each sensor a priori. The region divided in square grids is unknown and it has unknown but finite number of obstacles. The authors also consider a formation pattern like circle, ellipse, rectangle and ring. The computer simulations and mathematically rigorous proof with probability '1' has also been considered. The algorithm has convergence based on the probabilistic arguments.

The authors \cite{savkin2010decentralized} consider a bio-inspired decentralized navigation law which guarantees that all the mobile robots will eventually move in the same direction and with the equal speed. The control is based on the local information and it considers the physical constraints on the linear velocity and angular velocity of the robotic vehicle. The minimum turning radius of the robotic vehicle is also considered as the constraint while developing the control law. The quantized consensus on speed and heading angle is achieved in the defined discrete set. The authors have presented a rigorous mathematical analysis of the control algorithm and computer simulations have also been provided. However, collision avoidance from a moving or stationary obstacle still remains an open area for this control law.

The rectangular formation is considered with a set of decentralized control law in \cite{cheng2011decentralizedpattern}. The authors have developed a two stage algorithm. In the first stage, the autonomous mobile robot using local information aligns to form a line. In the second stage, agents start forming parallel lines after getting unique identity number and attain a rectangular lattice pattern. However, the rectangular lattice pattern is planar. The issues with collision avoidance and obstacle (moving or stationary) avoidance are still an open problem for this formation.

\begin{figure}[H]
  \centering
  \includegraphics[width=10cm]{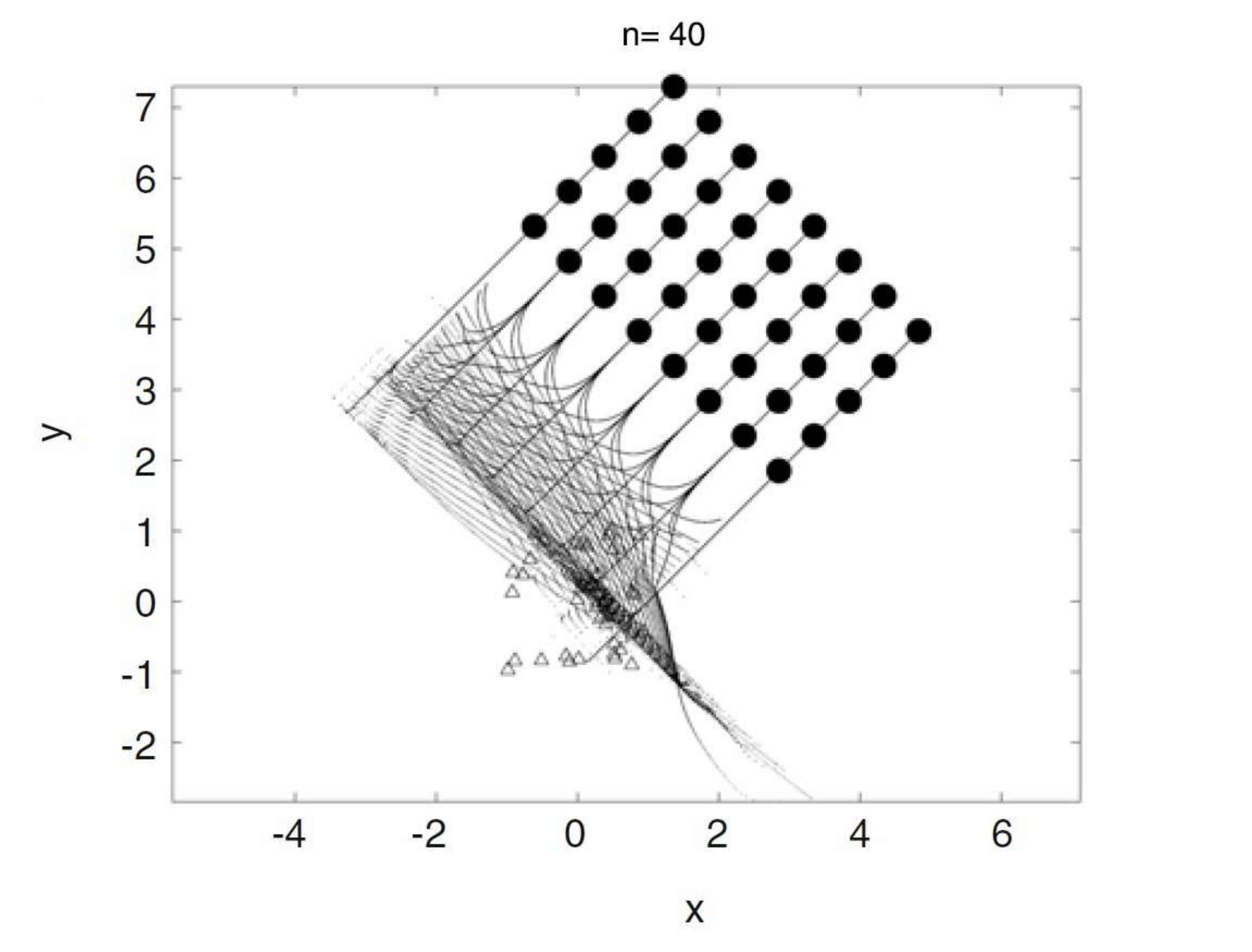}
  \caption{Rectangular Formation with 40 agents \cite{cheng2011decentralizedpattern}}
  \label{fig:Formation_Given_Pattern}
\end{figure}

In \cite{savkin2016distributed}, the authors consider a desired pattern formation with decentralized control algorithm applicable on unicycle like robotic dynamics. This work also considers physical constraints on linear velocity, angular velocity and minimum turning radius of the mobile robot. The control algorithm has been validated with computer simulations and real robots. The distributed motion coordination control algorithm drives the mobile robots in a desired geometric pattern from initial random conditions. The mathematical proof of the algorithm guarantees convergence with probability '1'. This algorithm is different from \cite{cheng2011decentralizedpattern} as it does not involve any line forming mode.

\begin{figure}[H]
  \centering
  \includegraphics[width=10cm]{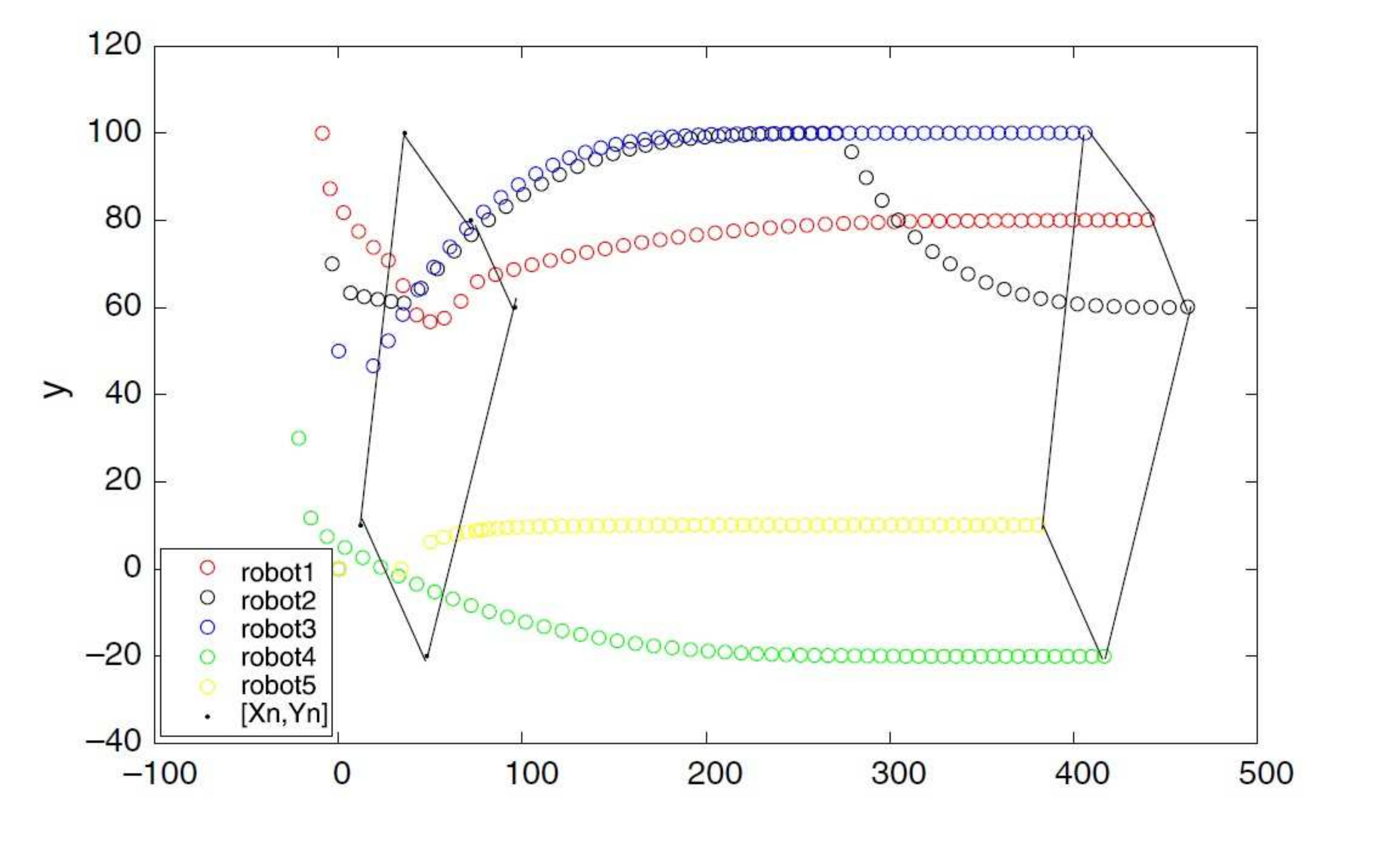}
  \caption{Robots form a random shape with index permutation applied \cite{savkin2016distributed}}
  \label{fig:sweep_line}
\end{figure}

The problem of velocity alignment among multiple agent have been considered in \cite{dehghanipour2014velocity}. This work considers a second order dynamics and it investigates velocity alignment on the directed networks of agents. The authors propose a set of sufficient conditions to guarantee the velocity alignment with the desired $H_{\inf}$. The computer simulations have shown that the agents can form a cohesive flock while withstanding against the external disturbances. The proposed velocity alignment algorithm is also capable of collision avoidance by the multiple agents.

In \cite{nazarzehi2014decentralized}, the authors consider a kinematic model with three degree of freedom. The authors also consider constraints on the linear and angular velocities. The control law builds formation control in 3-dimensional space. The geometric pattern is known a priori. However, the avoidance from the moving or static obstacle still remains an open area for this formation building problem.

%The work \cite{hsieh2008decentralized,lee2009decentralized,xiao2009finite,panyakeow2013optimal,dong2016time} is still under review.

\section{Mobile Actuator and Sensor Networks}\label{MAS-net}
%to be written

Mobile Actuator and Sensor Network (MAS-net) is a project that adds node mobility and close-loop control concept into the field of Wireless Sensor Network. In this network, a large number of robots can act as actuators and sensors at the same time. A mobile sensor actuator network has mobile nodes with sensor / actuator combined with a closed loop control system. Such sort of a platform is presented in \cite{wang2004masmote}. This platform has mobile robots, test-bed, pseudo-GPS system, fog generator and host workstation. The platform is compact, low cost, structurally robust and modular in software configuration. This platform has application in diffusion process boundary tracing and it is suitable for experimental investigation of mobile actuator sensor network in a laboratory environment. The main objective of the MAS-net is to monitor and control a diffusion process. However, the platform only offers experimental investigation with the released fog.

\begin{figure}[H]
  \centering
  \includegraphics[width=10cm]{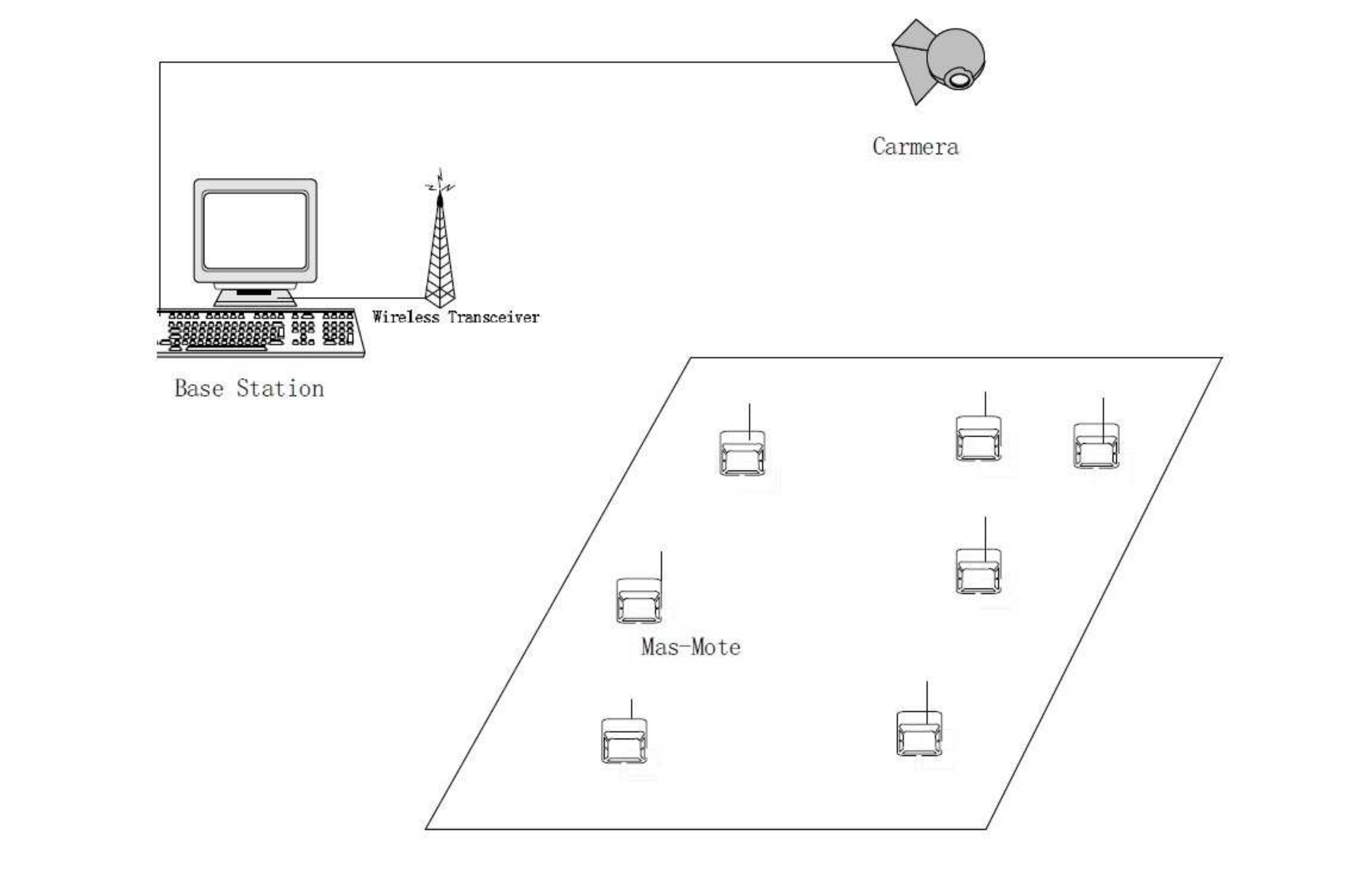}
  \caption{Components of a MAS-net System \cite{wang2004masmote}}
  \label{fig:3D_Coverage}
\end{figure}

In \cite{wang2005formation}, the authors review how the pattern formation can be achieved on the MAS-net platform presented in \cite{wang2004masmote}. The geometric pattern formation is important to obtain accurate mesh measurements. The accuracy and reliability of measurement or control of the diffusion process is expected by the coordination of mobile robots. In pattern formation, the authors discuss the design considerations like stability of the formation, controllability of different geometric patterns, safety and uncertainties in formations. The authors present the references for formation control with leader-follower strategy \cite{scharf2004survey}, virtual structure approach \cite{cowan2003vision} and behaviour based method \cite{ren2004decentralized}. However, the authors classify the formation control as formation regulation control and formation tracking control. The authors review the formation control in the context of MAS-net project and present the references for the leader-follower approach\cite{desai1998controlling,takahashi2004autonomous}, the generalized coordinates approach \cite{spry2004formation,zhang2003control,yamakita2004formation} and the virtual structure method \cite{tan1996virtual,beard2000feedback,beard2001coordination,ren2004decentralized}. In the leader-follower approach, only the local information is gathered by each mobile robot. The formation with the generalized coordinates approach uses a reference point to characterise a vehicle's location, orientation and its shape. The virtual structural method first defines the desired dynamics of the virtual structure. Then, it translates the desired motion of the virtual structure into desired motions for each agent. Finally, each agent is tracked by an individual tracking controller. The authors also present a new controller design for the leader-follower approach. The experiment has been performed with one leader and three followers. However, the considered kinematic model of the wheeled robots is in two-dimensional space.

The paper \cite{wang2005distributed} further presents the application of MAS-net project with a model-free approach based on the centroidal Voronoi diagram. The thesis considers centroidal Voronoi tessellations (CVT) based 2-level hierarchical heterogeneous MAS-net. This approach achieves an effective distributed control of a diffusion process. This simulation experiment has also been performed in two-dimensional space.

%//from Book
The MAS-net has different applications to provide efficient and effective monitoring and control of industrial and environmental processes. As explained above, a MAS-net platform has mobile robots (e.g. ground vehicles, underwater vehicles, unmanned aerial vehicles), which act as sensors and actuators. The actuators are capable to implement control objectives. Some of the theoretical research can be found in \cite{chao2006study,demetriou2010guidance} with references therein.
%///

In \cite{moore2004diffusion}, the authors presents some preliminary results related to path-planning for diffusion process. In the experiment, the authors use mobile actuator sensor network to collect samples of the distribution of interest. Then, a predictive model of the process is generated from the collected samples. This prediction is used to find the new sampling locations. In the experimental work, the mobile actuator sensor network collects samples of the fog and it is shown how concentration of fog is evolved as time progresses. This concept has a vision to derive the paths of the robots based on the distribution of the concentration of fog. The authors have presented some preliminary results in a two dimensional environment.

Similar to \cite{moore2004diffusion}, the authors extend their work \cite{chen2004diffusion} to determine the boundary of a diffusion chemical and form the networked actuators releasing a neutralizing chemical agent in a way to control the shape of the polluted zone. The authors use a model based distributed control approach to achieve the goals of diffusion boundary estimation and zone control. However, the problem has been investigated considering a two dimensional diffusion process and a team of ground mobile robots has been used to track the diffusion boundary.

The above mentioned work \cite{moore2004diffusion} and \cite{chen2004diffusion} with some other application scenarios has also been presented in \cite{moore2004model}.

In \cite{chen2005patternexperiment,chen2005patternthesis}, the authors use a mobile actuator sensor network platform presented in \cite{wang2004masmote}. The authors investigate the pattern formation of mobile actuator network in leader-follower configuration. The followers calculate their desired relative position based on the periodically received message from the leader. This pattern formation problem has been considered in a two dimensional environment.

Apart from the above mentioned work, the research work \cite{ren2007experimental,ren2008experimental} experimentally implements and validates the distributed consensus algorithms on a low cost mobile actuator and sensor network platform. The authors mainly targets three applications namely, rendezvous, axial alignment and formation manoeuvring. The authors study the rendezvous application where multiple mobile robots have to coordinate and simultaneously reach a common target location which is unknown a priori. The authors study rendezvous of four robots under time-invariant and dynamic interaction topologies.
In the case of axial alignment, multiple mobile robots coordinate and align their ﬁnal positions along a straight line. The axial alignment experiment has been performed on four robots evenly distributing under time-invariant interaction topology.
If multiple mobile robots form a rigid geometric shape and manoeuvre as a group with a given group velocity, this sort of application is considered as manoeuvring application. The authors perform the manoeuvring application on five robots, who maintain desired V-shaped formation geometry under time invariant information exchange topology. The main objective to perform these experiments on mobile actuator sensor network platform is to learn practical issues like physical limitations of the platform, communication data packet loss, information delay, etc.

In \cite{bourgeous2007swarm}, the authors propose a general engineering approach to synthesis of a mobile sensor network for a cooperative phototaxis task. In this Phototaxis, the robots move towards a directional light stimulus. The team of mobile robots cooperatively climb the gradient to rendezvous at the light source simultaneously. The experimental validation is based on simulations and implementation on the MAS-net platform with stationary and slowly moving light source. However, the experiments have been performed considering a two-dimensional environment.

Similar to \cite{wang2005distributed}, the authors \cite{rounds2009dynamic} use centroidal Voronoi tessellations (CVTs) based approach on MAS-net to create the desired static and dynamic formations. The formation performance is also analysed based on the initial positions and number of robots. The authors perform experiments in simulation and on an actual MAS-net platform. The experimental results are based on the formations: Rendezvous, Ellipse, Line and V-Shape. The authors show the flexibility and robustness of the CVT based formation control methodology on MAS-net project. The authors have found that the CVT based approach is beneficial to formation control applications not requiring a precise positioning. However, the experiments have been performed considering a two-dimensional space.

Similar to \cite{bourgeous2007swarm}, the research work \cite{rounds2009cooperative} uses centroidal Voronoi tessellations (CVT) based approach to achieve cooperative phototaxis. The array of stationary sensors characterizes the environment creating a CVT and the actuators move according to the tessellation achieving a cooperative phototaxis. In fact, this work combines swarm intelligence with CVT localizing and tracking a dynamic moving light source. The experiments have been performed with simulations and on the MAS-net platform. This algorithm has also been implemented on a two-dimensional environment.

The MAS-net project has also been considered in encircling coverage and termination of a moving deformable region \cite{savkin2013algorithm}. This algorithm is partially inspired from the well-known Hannibal double envelopment manoeuvre during the Battle of Cannae \cite{goldsworthy2012fall}. %\cite{chen2014distributed}.
This work first deploys the sensors to encircle a moving and possibly deformable bounded two dimensional region, $R(k)\subseteq \mathbb{R}^2$. In the region $R(k,d_{min},d_{max})$, the points between the distances $d_{min}$ and $d_{max}$ are sensed by at least one mobile robotic sensor. The region may be moving and changing shapes. The mobile robotic sensors with some necessary assumptions should follow the region. The work proposes a decentralized control algorithm with randomized deployment approach. The algorithm drives the sensors to cover $R(d_{min},d_{max})$ with equilateral triangular lattice pattern.

\begin{figure}[H]
  \centering
  \includegraphics[width=10cm]{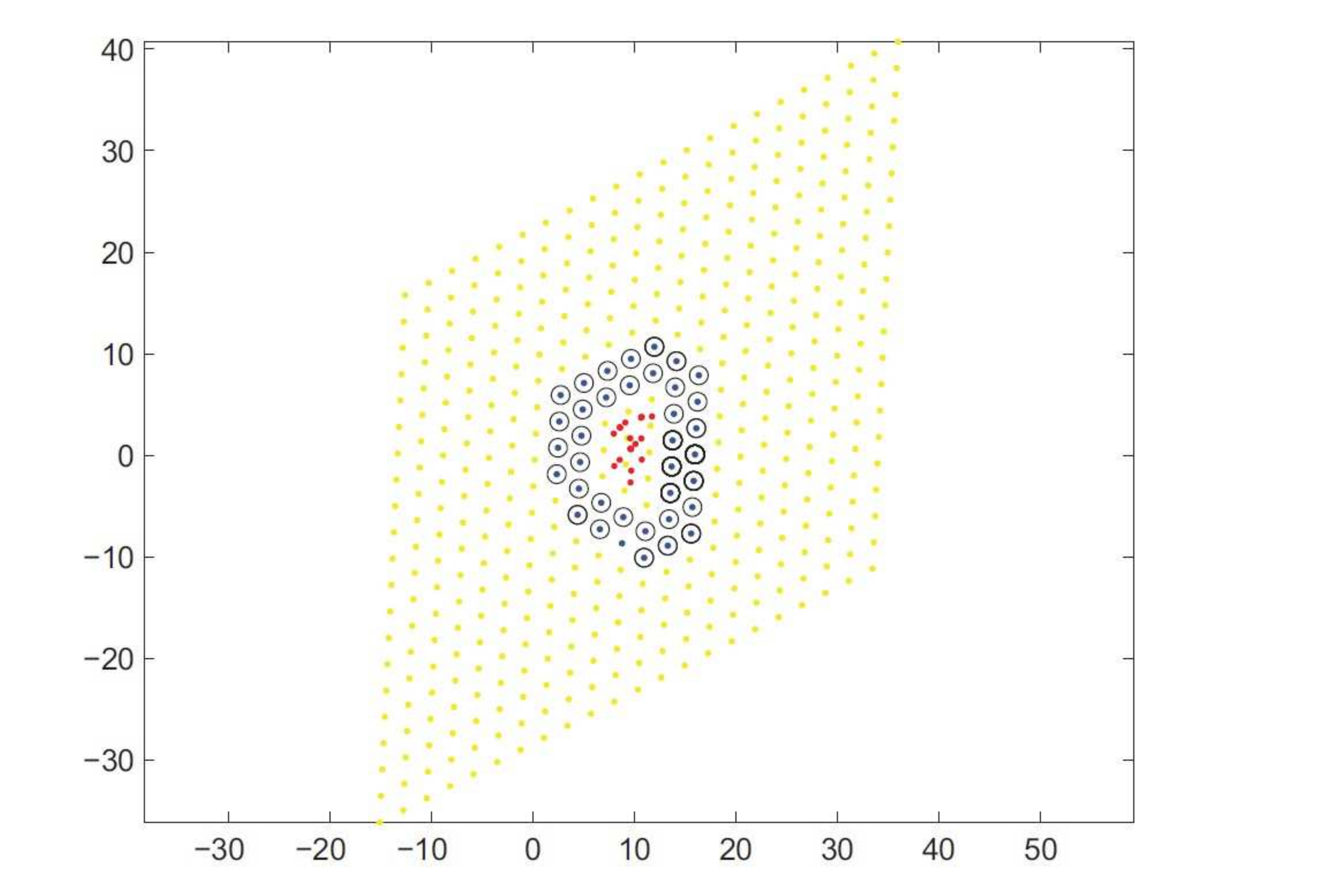}
  \caption{A region encircled by mobile sensors \cite{savkin2013algorithm}}
  \label{fig:Encircling_Coverage}
\end{figure}

The work also considers termination of a moving region by mobile actuator/ sensor network.

\begin{figure}[H]
  \centering
  \includegraphics[width=10cm]{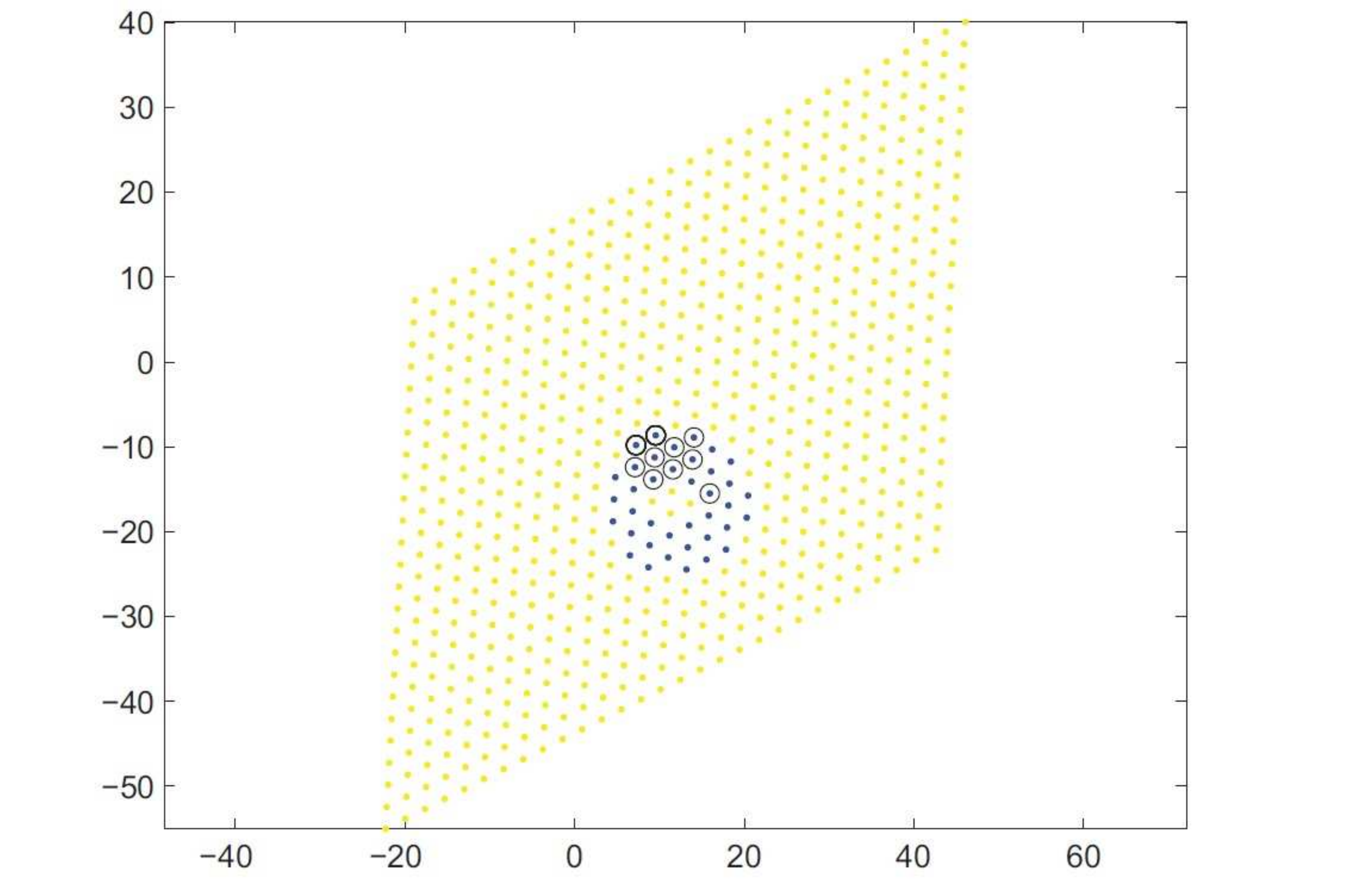}
  \caption{A region terminated by mobile sensors \cite{savkin2013algorithm}}
  \label{fig:Encircling_Termination}
\end{figure}

The moving region might be due to an oil spill, hazardous chemical or biological field. In this termination problem, the mobile robotic sensors are assumed to be equipped with actuators capable of releasing a neutralizing chemical. This way the parts of the moving region are terminated. This work also provides a mathematically rigorous proof of convergence with probability 1 from any initial positions of the mobile robotic sensors. The work has also been verified by computer simulations (see e.g. Fig: \ref{fig:Encircling_Coverage} and \ref{fig:Encircling_Termination}). However, the developed algorithm is for a two-dimensional region and an obstacle avoidance strategy remains as future consideration.

\begin{figure}[H]
  \centering
  \includegraphics[width=10cm]{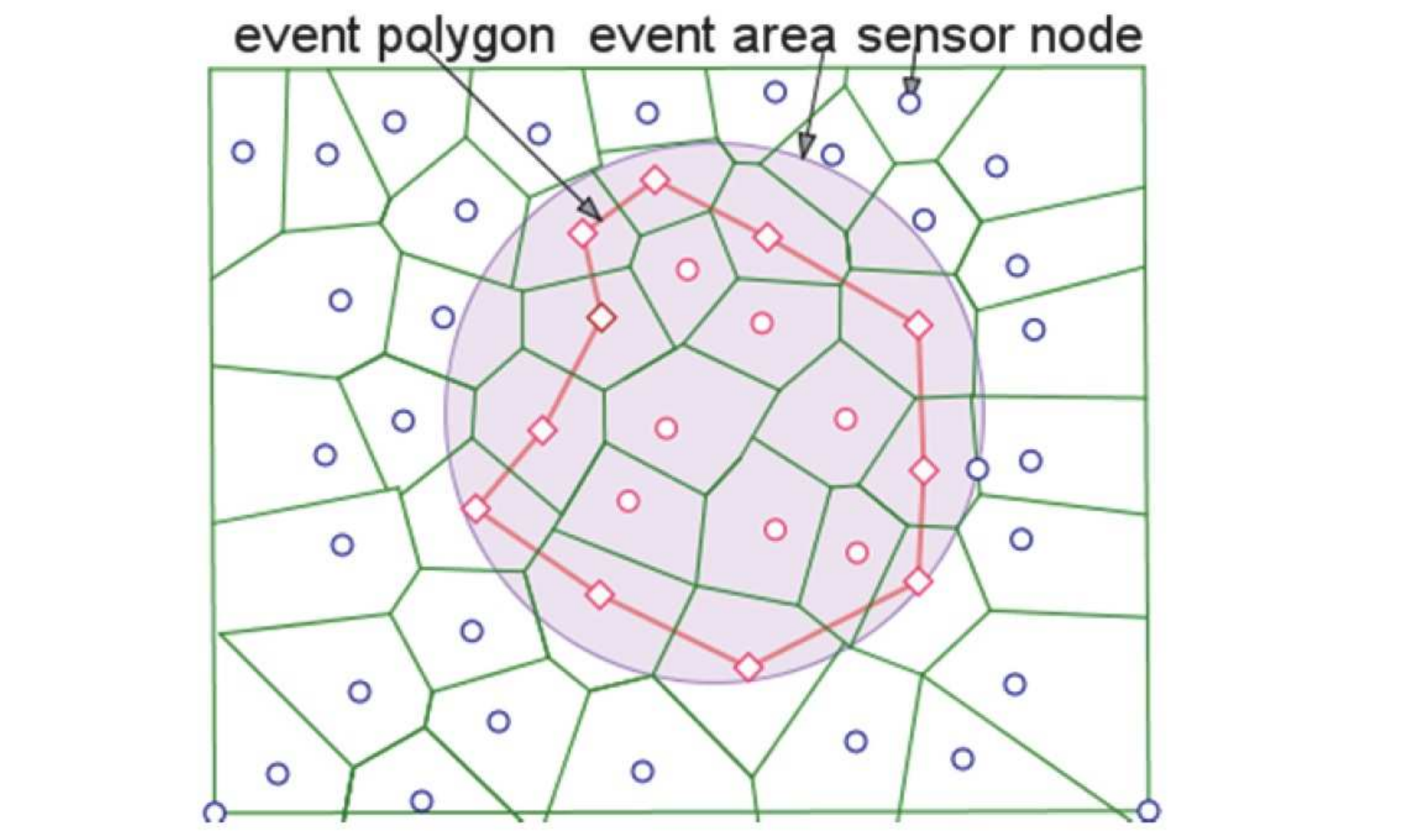}
  \caption{Determining boundary of the event region using Voronoi diagram \cite{lalouani2014effective}}
  \label{fig:Encircling_Polygon}
\end{figure}

A similar problem considered in \cite{lalouani2014effective} has been called as Sensor-Actuator Coordination for Handling Spreading events (SACHS). This work operates in three steps: boundary recognition of the event, coverage of the event region, and task assignment to the actuators. The simulation results of SACHS have been compared with \cite{melodia2010handling} and \cite{savkin2013algorithm}. However, the approach \cite{lalouani2014effective} is based on Voronoi diagram to determine the boundary of the event region as shown in figure \ref{fig:Encircling_Polygon}.

\section{Implementation Robots}\label{Implementation_Robots}

The kinematic models of implementation robots for coverage problems can be Unicycle, Bicycle, Tricycle, Differential Wheel Drive, Omnidirectional, etc.

In the coming sub-sections, we briefly describe the kinematic models of the mobile robots considered for coverage control problems. These kinematics models have also been reviewed in (see e.g., \cite{hoy2015algorithms,gracia2007kinematic,gracia2007kinematicslip,kozlowski2006robot,micaelli1993trajectory}).

\subsection{Holonomic}
%\This section is under review \cite{richards2006robust}.
A robot is considered to be Holonomic in nature if all the mechanically present degree of freedoms is controlled. In a Two-dimensional space, a mobile robot with three omnidirectional wheels in a triangular configuration would be considered as Holonomic type.

\subsection{Unicycle}
%\This section is under review \cite{manchester2004circular,manchester2006circular}
The unicycle vehicles are associated with a particular angular orientation. The angular orientation determines the direction of the velocity vector. However, the orientation change has a constraint of turning rate. Unicycle kinematic models can be used to describe different vehicles like differential drive wheeled mobile robots and fixed wing aircrafts. The following kinematic models of vehicles have been used by a number of researchers (see e.g. \cite{cheng2009distributed,cheng2011decentralizedsweep,manchester2004circular,manchester2006circular}).

\begin{equation}\label{eq:Unicycle_Model}
\begin{split}
\dot{x_{i}}(t)=v_{i}(t)\cos(\theta_{i}(t)) \\
\dot{y_{i}}(t)=v_{i}(t)\sin(\theta_{i}(t))  \\
\dot{\theta_{i}}(t)=\omega_{i}(t)
\end{split}
\end{equation} for $i=1,2,...,n$.

The state variables: $x \in \mathbb{R}$, $y \in \mathbb{R}$ are the Cartesian coordinates and $\theta \in \mathbb{S}$ is the orientation of the vehicle. The control inputs: $v_{i}(t)$ is the forward linear velocity and $\omega_{i}(t)$ is the angular velocity of the vehicle. In case of unicycle mobile robots:  $v_{i}(t) \in [-1,1]$ and $\omega_{i}(t) \in [-1,1]$.
In \cite{hoy2015algorithms}, it has been reviewed that the model (\ref{eq:Unicycle_Model}) has many mechanical applications as in case of wheeled robots, aerial vehicles, missiles, etc (see e.g. \cite{fossen1994guidance,low2007biologically}, and references therein).

\subsection{Bicycle}

%\This section \cite{matveev2013nonlinear,bicchi1996planning} is under review.
 This kinematics has as a steerable front wheel, which is separated from a fixed rear wheel. In this model, the maximum turning rate is proportional to the robots speed. This constraint has as an absolute bound on the curvature of any trajectory followed by the robot regardless of speed.

\section{Chapter Conclusion and Future Research Directions}\label{Conclusion_Future}

\subsection{Chapter Conclusion}
We have performed a qualitative survey of the literature primarily focusing on decentralized control of mobile robotic sensors for coverage problems. Such a literature can be classified based on the type of coverage. The coverage problem is further sub-classified based on the environment and approaches used to solve the problem. We have also performed an explicit comparison of these approaches to clarify the advantages and disadvantages associated with a particular control methodology. We have formulated a list of considerations to be included in a coverage control law/ algorithm.

The literature classification and its comparative study based on the approaches can be useful in formulating/ evaluating the coverage control law/ algorithm.

This overall literature survey creates the clarity in the literature and the coverage control problems may be considered or evaluated based on a combination of the major considerations listed in the below mentioned table:

\begin{longtable}{|l|p{8cm}|}

\caption{List of Considerations}\\
\hline
%\textbf{Problem Consideration} & \textbf{Type} \\
\textbf{Problem Consideration} & \textbf{Type        } \\
\hline
\endfirsthead
\multicolumn{2}{c}
{\tablename\ \thetable\ -- \textit{Continued from previous page}} \\
\hline
\textbf{Problem Consideration} & \textbf{Type        } \\
\hline
\endhead
\hline \multicolumn{2}{r}{\textit{Continued on next page}} \\
\endfoot
\hline
\endlastfoot

    Coverage Type                     &  Blanket, Barrier, Sweep, Encircling, Heuristic, Dynamic                                            \\  \hline
    Region Information                &  Known, Unknown, Partially Known                                                                    \\  \hline
    Sensors Initial Location          &  Known, Unknown, Partially Known                                                                    \\  \hline
    Coverage Dimension                &  1D, 2D, 3D                                                                                         \\  \hline
    Region Shape                      &  Convex, Arbitrary                                                                                  \\  \hline
    Region Environment                &  Static, Dynamic                                                                                    \\  \hline
    Robot Environment                 &  UAVs, UGVs, UWVs                                                                                   \\  \hline
    Algorithm                         &  Decentralized/ Distributed, Centralized                                                            \\  \hline
    Deployment                        &  Optimal, Sub-optimal, Ordinary                                                                     \\  \hline
    Coverage and Connectivity         &  k-coverage, k-connectivity                                                                         \\  \hline
    Sensor Grid Pattern               &  Triangular, Hexagonal, Square, Honeycomb,etc                                                       \\  \hline
    Coverage Hole                     &  Complete Coverage, Approximate Coverage                                                            \\  \hline
    Robotic Trajectory                &  Optimal, Sub-optimal, Ordinary                                                                     \\  \hline
    Robot Kinematics                  &  Unicycle, Bicycle, Tricycle, Omnidirectional, Differential Wheel Drive, Fish Type, Ant Type, etc   \\  \hline
    Sensing Range                     &  Homogeneous, Heterogeneous                                                                         \\  \hline
    Communication Range               &  Homogeneous, Heterogeneous                                                                         \\  \hline
    Sensing Range Model               &  Convex, Non-convex                                                                                 \\  \hline
    Communication Range Model         &  Convex, Non-convex                                                                                 \\  \hline
    Convergence Speed                 &  Optimal, Sub-optimal, Ordinary                                                                     \\  \hline
    Obstacle Avoidance                &  Dynamic, Static, Not Considered                                                                    \\  \hline
    Collision Avoidance               &  Considered, Not Considered                                                                         \\  \hline
    Geometric Constraints             &  Considered, Not Considered                                                                         \\  \hline
    Dynamics Constraints              &  Considered, Not Considered                                                                         \\  \hline
    Communication Constraints         &  Considered, Not Considered                                                                         \\  \hline
    Communication Topology            &  Time Varying, Time-Invariant                                                                       \\  \hline
    Information Exchange              &  Asynchronous, Synchronous                                                                          \\  \hline
    Algorithm Approach                &  Deterministic, Probabilistic/ Random, Semi-random, Hybrid                                          \\  \hline
    Algorithm Validation              &  Mathematical, Experimental Platform, Computer Simulation                                           \\  \hline
\end{longtable}

The above mentioned considerations in the problem can quickly assist us in designing the control law for a certain application. Like we can consider the application(s) and decide from the list of above mentioned items to know what sort of considerations is required for the control law or algorithm.

\subsection{Future Research Directions}
We recommend the considerations to quantitatively evaluate the coverage approaches. Such a consideration could highlight efficient approach for a problem. There could be a separate survey for coverage problems considering obstacles.

\chapter{Nearest Neighbour Rule with Weighted Average Functions}

In the previous chapter, we have performed a comprehensive literature survey on coverage control problems. We found the approach based on nearest neighbour ru.le is decentralized in nature and it requires local information. Consequently, the control strategy based on nearest neighbour rule offers adaptability in the algorithm. However, we amend this rule as nearest neighbour rule with weighted average functions. So, this chapter is based on nearest neighbour rule with weighted average functions. We apply this approach specifically on sweep coverage algorithm to validate our amendment in the rule.

\section{Introduction}

This chapter addresses the decentralized control of mobile robotic sensors performing sweep coverage along an arbitrary boundary. Current sweep coverage control does not exhibit a smooth trajectory for a mobile robotic sensor when following an abrupt change in the direction of an arbitrary boundary. Simulation results of the recent sweep coverage problem reveal that the control needs to be more sensitive for some of the mobile robotic sensor(s) after detection of an arbitrary boundary. The goal is to attain smooth sweep coverage along an arbitrary boundary by a swarm of mobile robots-which are driven under decentralized control strategy. We introduce normalised weighting factors in nearest neighbour rules to achieve consensus among mobile robotic sensors. Current simulation results demonstrate that the nearest neighbour rules with weighted average functions give rise to a smooth sweeping trajectory for each mobile robot. The concept of these weighted average functions also make some of the mobile robotic sensor(s) more sensitive in following an abrupt change in the direction of an arbitrary boundary.

\section{Related Work}

Decentralized coordination of groups of mobile sensors is an active area of research in robotics\footnote{The terms, "sensor" or "the mobile robotic sensor" or simply, "the mobile robot" will be used throughout this research report for mobile robotic sensor having an on-board computation, boundary detection and communication capability.}. In this chapter, vision based sensors are considered, which provide the position of neighbouring robotic sensors together with adjacent obstacle boundaries. Possible applications include surveillance, reconnaissance, maintenance, inspection and training (see e.g. \cite{gage1992command,gage1995many}).

In Chapter 1, we have shown that a swarm of mobile robotic sensors is basically inspired from flocks of birds and schools of fish, who organise themselves with mutual coordination (see e.g. \cite{reynolds1987flocks,savkin2004coordinated}). In Chapter 2, we have shown how Gage has classified the coverage of the mobile robots into three basic patterns: Blanket Coverage, Barrier Coverage and Sweep Coverage \cite{gage1992command} \footnote{ The terms, "sweep coverage" or simply "sweeping of the mobile robots" will be used throughout this chapter to address a sweep coverage problem.}.
%We can define these coverage patterns as follows \cite{gage1992command}:
%
%\begin{itemize}
%\item Blanket Coverage occurs when the mobile robots form a static arrangement to cover a region such that any intruding objects has probability of being detected for maximum number of times.
%\item Barrier Coverage occurs when the mobile robots arrange themselves in the form of a static barrier, which minimizes the probability of undetected intruders passing through it. Fig. 1(ii) is a good example of a Barrier Coverage demonstrated by five mobile robotic sensors, each one being represented by a black arrow as its heading angle and a red circle as its sensing range. Fig. 1(ii) shows a virtual boundary moving mobile robot with a yellow arrow as its heading angle. The purpose of a virtual mobile robot is to share the information of the boundary with a neighbour mobile robot(s).
%\item If the formed static barrier starts moving such that every mobile robot inside it maintains an equal distance from its neighbour(s) and gains the same speed as that of its neighbours, then this sort of collective behaviour is known as the Sweep Coverage.
%\end{itemize}

  Many researchers (\cite{cheng2009distributed,cheng2009decentralized,cheng2010decentralizedregion,cheng2011decentralized,cheng2012self,cheng2013decentralized,cheng2011decentralizedsweep,savkin2012optimal}) have developed decentralized control algorithms based on Gage's classification \cite{gage1992command} for coverage problems.

\begin{figure}[H]
\begin{center}
   \includegraphics[width=0.8\columnwidth]{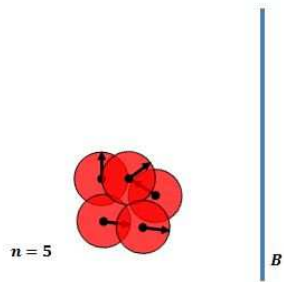}
  % \label{fig:theFig}
\caption{Objective - Initial Deployment}\label{fig:objective_Initial}
\end{center}
\end{figure}

Our main objective is to develop an improved decentralized control, which could smoothly drive the randomly placed mobile robots (shown in Fig. \ref{fig:objective_Initial}), to form a sweep coverage and to also sweep smoothly across an arbitrary boundary (as shown in \ref{fig:objective_Desired_Sweepin}). In Chapter 2, we can remember that the sweep coverage has potential applications for minesweepers \cite{acar2003path,cassinis1999strategies}, patrolling borders \cite{kumar2005barrier}, environment monitoring of the deep ocean floor \cite{jeremic1998design} and underwater oil exploration \cite{borhaug2007straight}. While centralized systems can also be used to accomplish the aforementioned tasks, there is high complexity associated with each mobile robot's communication with a central system. A decentralized control strategy for the mobile robots to accomplish sweep coverage is considered a cost effective solution.

The coverage path planning problem has been addressed by a number of researchers where the environment is assumed to be known (see e.g. \cite{choset2001coverage,kurabayashi1996cooperative,garcia2004mobile}). Other research has achieved path planning by introducing autonomous and cooperative behaviour within the algorithm ( see e.g. \cite{min1998decentralized,butler2000complete}).

An alternative to path planning uses simple average functions to provide a fully decentralized sweep coverage along an arbitrary boundary \cite{cheng2011decentralizedsweep}. Our current research extends this idea to ensure a smooth and an efficient movement especially when there is an abrupt change required in the direction of the sweeping mobile robots. In order to make our sweep coverage smooth and sensitive to an abrupt change in the direction of an arbitrary boundary, we introduce weighted average functions for the calculation of nearest neighbour rules. Our approach broadens nearest neighbour rules specifically for the heading angle and velocity of the mobile robots sweeping across an arbitrary boundary. A fully decentralized control strategy with weighted average nearest neighbour rules has been adopted throughout the control design. Simulation results with weighted average functions are presented in order to compare the improvement on recent work \cite{cheng2011decentralizedsweep}.

\section{Problem Statement}

The sweep coverage problem studied in \cite{cheng2011decentralizedsweep} may be stated as follows:\\
Let $s_{i}(kT)$ be a position and $\theta_{i}(kT)$ be a heading angle of the mobile robotic sensor measured counter clockwise from the $X$-axis. Let a mobile robot $i$ has a linear velocity $v_{i}(kT)$.

Let a unit vector function $u(\gamma)$ be such that for any $\gamma\in [-\pi/2, \pi/2)$ measured with respect to the X-axis. It follows that:

\begin{equation}\label{eq:Input_Vector}
 u(\gamma)= [\cos(\gamma)\text{ }\sin(\gamma)]^{T}
\end{equation}

Let $B$ be a boundary line with a direction $\phi_{b}$, and it has an associated scalar $b_{1}$. It can be mathematically defined as follows:

\begin{gather}
B:=\{p\in{\mathbb{R}}^2:u^Tp=b_{1}\}\\
\phi_{b}:=\gamma +\pi/2
\end{gather}

Let $B_{1}$ and $B_{2}$ be two sets defined as follows:

\begin{gather}
B_{1}:=\{p\in{\mathbb{R}}^2:u^Tp>b_{1}\}\\
B_{2}:=\{p\in{\mathbb{R}}^2:u^Tp<b_{1}\}
\end{gather}

\begin{figure}[H]
\begin{center}
   \includegraphics[width=0.8\columnwidth]{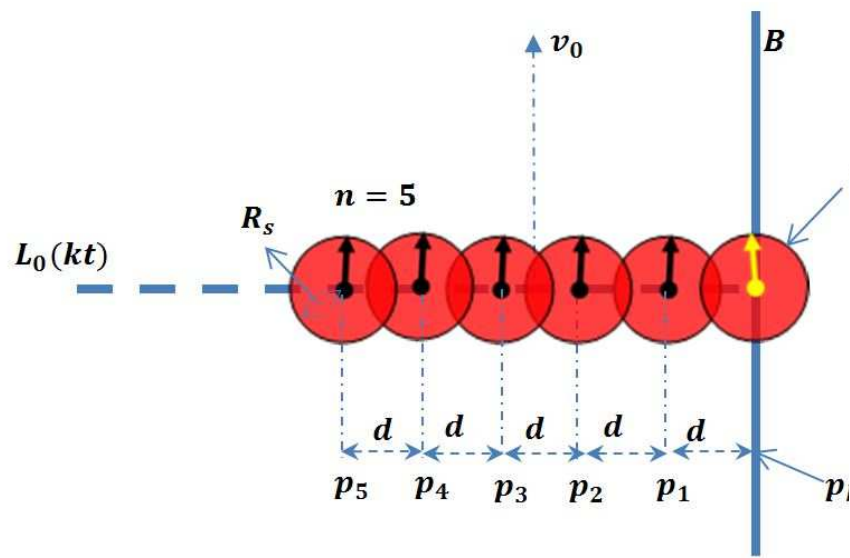}
  % \label{fig:theFig}
    \caption{Objective - Desired Sweeping}\label{fig:objective_Desired_Sweepin}
\end{center}
\end{figure}

Next, we define a below mentioned moving line $ L_{0}(kT)$ with points $p_{i}$ (as shown in Fig. \ref{fig:objective_Desired_Sweepin}) for $i=1,2,...,n$:

\begin{equation}\label{eq:L0_Eqn}
 L_{0}(kT):=\{p\in{\mathbb{R}}^2:p^Tu=c_{0}+kTv_{0}\}\\
\end{equation}
for $k=0,1,2,...$, where $c_{0}$ is a scalar and $v_{0}$ is the desired sweeping speed. The desired points denoted by $p_{i}$ (as shown in Fig. \ref{fig:objective_Desired_Sweepin}) on $ L_{0}(kT)$ can be mathematically described as follows:

\begin{equation}\label{eq:points_on_L0}
\begin{split}
  &p_{i}(kT):= \begin{cases}
 p_{b}(kT)+(d \times i)u\text{   }if\text{ }{\cal{P}} \subset B_{1}\\
 p_{b}(kT)-(d \times i)u\text{   }if\text{ }{\cal{P}} \subset B_{2}\\
 \end{cases}
\end{split}
\end{equation}
for $i=1,2,...,n$ and $k=0,1,2,...$, where $ p_{b}(kT):= L_{0}(kT) \cap B$ and ${\cal{P}}\in{\mathbb{R}}^2$ is a bounded set with Lebsegue measure for initial headings of the mobile robotic sensors.

The discrete-time position, $s_{i}(kT)\in{\mathbb{R}}^2$, of the mobile robotic sensor with control inputs, velocity $v_{i}$ and heading angle $\theta_{i}$, can be written as follows:
\begin{equation}\label{eq:smooth_Dynamics}
 s_{i}((k+1)T)= s_{i}(kT) + Tv_{i}(kT)u(\theta_{i}(kT))
\end{equation} for $i=1,2,\ldots,n$, and $k=0, 1, 2,\ldots$.

There are some physical constraints associated with the mobile robot, that is, $|v_{i}(t)|\le v_{\max}$ for $i=1, 2, \ldots, n$ and $t\ge0$. The initial heading of each mobile robotic sensor meets the condition: $\theta_{i}(0) \in [0, \pi)$ for $i=1, 2, \ldots, n$.

Mathematically, a control law is said to be a sweep coverage with sweeping speed $v_{0}$ along boundary $B$ for $n$ mobile robotic sensors maintaining mutual distance $d$ if for almost all initial positions of the mobile robotic sensors, there exists permutation $\{{z_{1},z_{2},...,z_{n}}\}$ of set $\{{1,2,...,n}\}$ such that the following condition is satisfied:

\begin{equation}\label{eq:smooth_sweep_def}
 \lim_{k\to\infty}||s_{z_{i}}(kT) - p_{i}(kT) ||=0, i=1,2,3,\ldots,n.
\end{equation}

Hence, the main objective is to develop a decentralized control law to sweep the mobile robots (meeting assumptions) smoothly along an arbitrary boundary $B$. The control action should produce a smooth trajectory for each mobile robot especially facing a sudden change in the direction of an arbitrary boundary. The control action should adjust a weighted average movement (not just an equally averaged movement) for each mobile robotic sensor after the detection of an arbitrary boundary, which is an improvement on recent research \cite{cheng2011decentralizedsweep}.

\subsection{Assumptions}\label{smooth_Assumptions}

We make the following necessary assumptions for control development similar to \cite{cheng2011decentralizedsweep}.

\begin{enumerate}
 \item Define a disk of radius $R_{c}>0$ for $t\in[kT, (k+1)T)$ and $k=0, 1, 2,\ldots$ as follows:
\begin{equation}
C_{i,R_{c}}(kT):=\{p\in{\mathbb{R}}^2:||p - s_{i}(kT)|| \leq R_{c}\}
\end{equation}
where $\|\cdot\|$ denotes the Euclidean norm. So, any mobile robotic sensor with a disk of radius $R_{c}$ has the capability to communicate with its neighbour(s) within this range, as long as the condition, $R_{c}< v_{max}T/\sqrt{2}$, remains valid for it.

  \item A mobile robotic sensor can detect any operation specific target within a disk of radius $R_{s}>0$ for $t\in[kT, (k+1)T)$ and $k=0, 1, 2,\ldots$ defined as follows:

\begin{equation}
S_{i,R_{s}}(kT):=\{p\in{\mathbb{R}}^2:||p - s_{i}(kT)|| \leq R_{s}\}
\end{equation}
where $\|\cdot\|$ denotes the Euclidean norm.
Here, we also assume that any mobile robotic sensor $i$ can detect and find the slope of an arbitrary boundary $B$ within a range $R_{b}$ such that the condition, $ R_{b}/ R_{c}> \sqrt{2}$, holds.

  \item There are $n$ number of the mobile robots satisfying the below mentioned condition to form a sensor barrier of length $D$ from boundary $B$.\\

\begin{equation}
(n+1)R_{s}> D
\end{equation}
for $R_{s} \in (0,R_{c})$.\\
  \item The physical parameters (position, heading angle and the desired sweeping speed) of the mobile robotic sensor have the constraints as follows:\\

The initial randomly placed mobile robotic sensors come in a bounded set ${\cal{P}} \in {\mathbb{R}}^2$ with Lebesgue measure. Initially, each randomly placed mobile robotic sensor satisfies $\theta_{i}(0) \in [0, \pi)$ for $i=1, 2, \ldots, n$. The desired sweeping speed satisfies the condition as follows:\\

\begin{equation}
0<  |v_{0}|  \leq 1/T min\{(v_{max}T-R_{c}\sqrt{2}),(R_{b}-R_{c}\sqrt{2})\}
\end{equation}
  \item There exists an infinite sequence of contiguous, non-empty, bounded, time intervals $[k_{i},k_{i+1}]$ for $i=0, 1, 2,\ldots,n$ and $k_{0}=0$, such that for all $[k_{i},k_{i+1}]$ the graph from the union of the collection $G(kT)\in{\cal P}$ for $kT\in [k_{i},k_{i+1}]$ is connected.
\end{enumerate}

\begin{figure}[H]
\begin{center}
   \includegraphics[width=0.8\columnwidth]{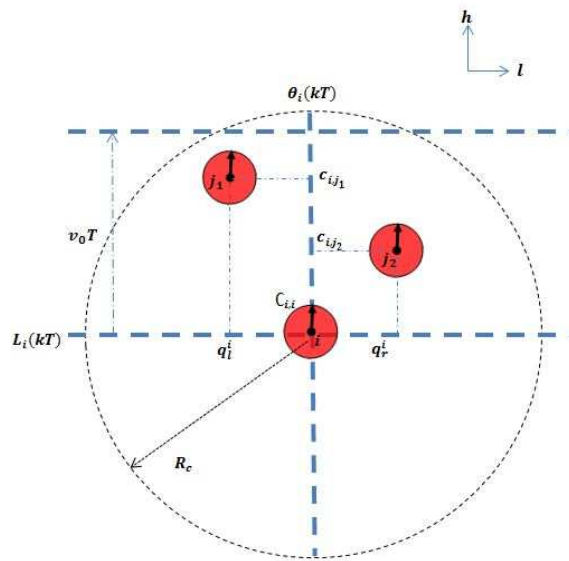}
  % \label{fig:theFig}
    \caption{Objective - Defined Variables}\label{fig:Movement_Projection}
\end{center}
\end{figure}

\section{Nearest Neighbour Rules with Weighted Average Functions}
We mathematically amend the variables for the nearest neighbour rules of recent sweep coverage problem \cite{cheng2011decentralizedsweep}. We define the weighted average functions which replaces simple average functions as used by \cite{cheng2009distributed,cheng2009decentralized,cheng2010decentralizedregion,cheng2011decentralized,cheng2012self,cheng2013decentralized,cheng2011decentralizedsweep,savkin2012optimal,savkin2004coordinated,savkin2010decentralized} for the decentralized coordinated control.

We consider the very first sensor detecting the slope of a boundary switches its heading angle by considering a weighting factor. Similarly, each mobile robot inside the rest of the barrier also gets a weighted motivation to adjust its heading angle. Formally, we first define a weighted average variable for the heading angle update of the mobile robot:

\begin{equation}\label{smooth_Theta}
 {\Theta}_{i}(kT) :=w_{ii}\phi_i(kT)+\sum_ {j\in {\cal F}_{i}(kT)}w_{ij}\phi_{j}(kT)
\end{equation}
for $i=1,2,3,\ldots, n$, where ${\cal F}_i(kT)$ denotes the set of neighbours of the mobile robotic sensor $i$ at a particular time $kT$. We have defined a non-negative weighting factor, $w_{ii}$, for the mobile robotic sensor itself and a non-negative weighting factor, $w_{ij}$, for its neighbour(s) at a particular time $(kT)$ such that the following condition is satisfied.
\begin{equation}\label{smooth_w_norm}
 w_{ii}+\sum_ {j\in {\cal F}_{i}(kT)}w_{ij}=1
\end{equation}
The above equation means that we have used normalised weighting factors for a mobile robot to decide its heading angle. The mobile robotic sensor decides these weighting factors by considering the highest value for the neighbour mobile robot nearest to the boundary, whereas, the lowest value is given to the neighbour mobile robot farthest from the boundary. Hence, this approach makes the mobile robots nearer to the boundary to be more sensitive in adjusting individual heading angle, and the adjustment of individual heading angle becomes less sensitive for the mobile robots sweeping farther from the boundary. For example, if a mobile robotic sensor (say $i$) nearest to the boundary has two neighbours ($|{\cal F}_i(kT)|=2$), then $w_{ii}=2/6$, $w_{ij}=1/6$ for the neighbour farthest from the boundary and $w_{ij}=3/6$ for the neighbour (virtual mobile robotic sensor) based on the boundary. Similarly, if this mobile robotic sensor has just one neighbour, the weighting factors will be $w_{ii}=1/3$ for the mobile robot itself and $w_{ij}=2/3$ for its neighbour (virtual mobile robotic sensor). However, if any of the mobile robotic sensor has not detected the boundary, then the control law \cite{cheng2011decentralizedsweep} does not need any modification for the sweep coverage problem, that is:
\begin{equation}
 w_{ii}=w_{ij} \text{ }for \text{ }B \cap R_{b} \ne \emptyset.
\end{equation}

The above condition means an equal weighting factor (that is, for one neighbour: $w_{ii}=w_{ij}=1/2$ or for two neighbours: $w_{ii}=w_{ij_{1}}=w_{ij_{2}}=1/3$) is given to a mobile robot itself and to its neighbours in the absence of boundary $B$. If two or more than two mobile robots detect the boundary at the same time, then the weighting factors with different value can be decided in the desired direction of sweeping.

The heading angle of the mobile robot is our one control input, which is being calculated by normalised weighted average function in a decentralized way.

We define the coordination variable $\phi_{i}(kT)$ for $ i=1,2,...,n$ and $k=0,1,2,...$ to be updated using the following rule:
\begin{equation}\label{eq:smooth_phi_update}
 \phi_{i}((k+1)T)= \Theta_{i}(kT)
\end{equation}

By using $\Theta_{i}(kT)$, we define a scalar at time $kT$ ($k=0,1,2,...$) for individual sensor $i$ itself:
\begin{equation}\label{eq:smooth_c}
 c_{ii}(kT) = s_{ii}^{T}(kT)u({\Theta}_{i}(kT))
\end{equation}
Similarly, we define a scalar for each neighbour $j$ of mobile robot $i$:
\begin{equation}\label{eq:smooth_cij}
 c_{ij}(kT) = s_{ij}^{T}(kT)u({\Theta}_{i}(kT))
\end{equation}
Similar to the heading angle of the mobile robot, there is a requirement that its velocity component along the direction of the boundary should also be calculated in a weighted average fashion. So, we again define a non-negative weighting factor $m_{ii}$ for the mobile robotic sensor itself and a non-negative weighting factor $m_{ij}$ for its neighbour(s), ${\cal F}_{i}(kT)$, such that the following condition is satisfied:
\begin{equation}\label{smooth_m_norm}
 m_{ii}+\sum_ {j\in {\cal F}_{i}(kT)}m_{ij}=1
\end{equation}

For example, if mobile robotic sensor $i$ has two neighbours (i.e., $|{\cal F}_i(kT)|=2$) and it is closest to the boundary, then $m_{ii}=2/6$, $m_{ij}=3/6 $ for the neighbour farthest from the boundary and $m_{ij}=1/6$ for the neighbour nearest to the boundary (which could be the virtual boundary based mobile robotic sensor). Similarly, if a mobile robotic sensor has just one neighbour, then the weighting factors will be $m_{ii}=2/3$ for itself and $m_{ij}=1/3$ for its neighbour.

However, it should be noted that:

\begin{equation}
 m_{ii}=m_{ij} \text{ }for \text{ }B \cap R_{b} \ne \emptyset.
\end{equation}

The above condition again states that an equal weight (that is, for one neighbour: $m_{ii}=m_{ij}=1/2$ or for two neighbours: $m_{ii}=m_{ij_{1}}=m_{ij_{2}}=1/3$) is given to mobile robot $i$ itself and to its neighbours, unless boundary $B$ is out of boundary detection range $R_{b}$ of the mobile robot(s). As mentioned earlier, if two or more than two mobile robots detect boundary $B$ at the same time, then the weighting factors can be decided in the desired direction of sweeping.

Now, we can amend the scalar variable used in recent control \cite{cheng2011decentralizedsweep} to calculate velocity component along the direction of the mobile robot. So, we define this scalar variable as the weighted average function (not a simple average function) at a particular time $(kT)$:

\begin{equation}\label{eq:smooth_C_avg}
 {\cal C}_{i}(kT) := m_{ii}c_{ii}(kT)+\sum_{j\in {\cal F}_{i}(kT)}m_{ij}c_{ij}(kT)
\end{equation}

Thus, the update rule for $c_{ii}(kT)$ is as follows:

\begin{equation}\label{eq:smooth_cii_update}
 c_{ii}(k+1)T ={\cal C}_{i}(kT)
\end{equation}

\subsection{Control Action}

By utilising the above mentioned nearest neighbour rules with weighted average functions, we can write the control action of recent research \cite{cheng2011decentralizedsweep} as follows:

The line to be followed by the mobile robot $i$ can be mathematically defined with the help of ${\Theta}_{i}(kT)$:

\begin{equation}\label{eq:smooth_L_Eqn}
 L_{i}(kT) =\{(x,y)\in{\mathbb{R}}^2:\\
 x\cos({\Theta}_{i}(kT))+y\sin({\Theta}_{i}(kT))\\
 = {\cal C}_{i}(kT)\}
\end{equation}
for $ i=1,2,...,n$ and $k=0,1,2,...$.

Let the projection of mobile robot $i$ and the projection of its neighbouring mobile robot $j$ on $L_{i}(kT)$ at time $kT$ be $q^{i}_{i}(kT)$ and $q^{i}_{j}(kT)$, respectively:

 \begin{equation}\label{eq:smooth_qij}
q^{i}_{j}(kT) = [\sin({\Theta}_{i}(kT)) -\cos({\Theta}_{i}(kT))]s_{j}^{T}(kT)
\end{equation}
for $k=0,1,2,...$.

Similarly, we define $q^{i}_{l}(kT)$ and $q^{i}_{r}(kT)$ be the projections of the neighbouring mobile robots to the left and to the right of mobile robot $i$ such that the following condition is satisfied:
\begin{equation}
 q^{i}_{l}(kT) < q^{i}_{i}(kT) < q^{i}_{r}(kT).
\end{equation}

In order to keep the mobile robots away from the boundary, we introduce a variable $q^{i}_{b}$.\\
First, a variable $b_i$ is defined:
\begin{equation}
 b_{i}(kT)= L_{i}(kT) \cap B
\end{equation}
Then it follows that:
\begin{equation}
q^{i}_{b}=  [\sin({\Theta}_{i}(kT)) -\cos({\Theta}_{i}(kT))]b_{i}(kT)
\end{equation}
Thus we can limit the mobile robots as follows:
\begin{equation}
\begin{split}
q^{i}_{r}(kT)=  q^{i}_{b},{\text{ }if\text{ }q^{i}_{b} \le q^{i}_{i}(kT)}\text{ }and\text{ }0 \notin {\cal F}_{i}(kT) \\
q^{i}_{l}(kT)=  q^{i}_{b}, {\text{ }if\text{ }q^{i}_{b} \ge q^{i}_{i}(kT)}\text{ }and\text{ }0 \notin {\cal F}_{i}(kT)
\end{split}
\end{equation}

The coordinates of the mobile robot are updated as follows:

\begin{equation}
\begin{split}
  &{\cal Q}_{i}(kT):= \begin{cases}
 (q_{l}^{i}(kT)+q^{i}_{r}(kT))/2\text{  }{if\text{ }l\text{ }and\text{ }r\text{ }exist}\\
 (q_{l}^{i}(kT)+q^{i}_{i}(kT)+d)/2\text{  }{if\text{ }only\text{ }l\text{ }exists}\\
 (q_{r}^{i}(kT)+q^{i}_{i}(kT)-d)/2\text{  }{if\text{ }only\text{ }r\text{ }exists}\\
 q^{i}_{i}(kT)\text{  }{if\text{ }l\text{ }and\text{ }r\text{ }do\text{ }not\text{ }exist}\\
 \end{cases}
\end{split}
\end{equation}

The projection of mobile robotic sensor $i$ itself on $ L_{i}(kT)$ is updated as follows:

\begin{equation}\label{smooth_qi_update}
q_{i}^{i}(k+1)T= {\cal Q}_{i}(kT)
\end{equation}
for $ i=1,2,...,n$ and $k=0,1,2,...$.

We can define the velocity components of mobile robot $i$ along $L_{i}(kT)$ and ${\Theta}_{i}(kT)$ as follows:

\begin{equation}\label{smooth_vi_barhat}
\begin{split}
\bar{v}_{i}(kT) & :=  ({\cal Q}_{i}(kT)-q_{i}^{i}(kT))/T \\
\hat{v}_{i}(kT) & :=  ({\cal C}_{i}(kT)-c_{ii}(kT)+v_{0}T)/T
\end{split}
\end{equation}

for $ i=1,2,...,n$ and $k=0,1,2,...$.

Hence, a set of control inputs can be written as follows:
\begin{equation}\label{eq:smooth_control}
\begin{split}
 v_{i}(kT) &= \sqrt{\bar{v}_{i}(kT)^{2}+\hat{v}_{i}(kT)^{2}} \\
 \theta_{i}(kT) &=
\begin{cases}
 \Theta_{i}(kT) + \beta_{i}(kT) - \pi/2, & \text{ if } \hat{v}_{i}(kT) \ge 0\\
 \Theta_{i}(kT) - \beta_{i}(kT) - \pi/2, & \text{ if } \hat{v}_{i}(kT) < 0
\end{cases}
\end{split}
\end{equation}
where $\beta_{i}(kT) := \cos^{-1}(\bar{v}_{i}(kT)/v_{i}(kT))$, for $i=1, 2,\ldots, n$ and $k = 0, 1, 2, \ldots.$

\subsection{Theorem}

If $n$ autonomous mobile robots governed by dynamics (~\ref{eq:smooth_Dynamics}) meeting the assumptions (\ref{smooth_Assumptions}) are supposed to sweep across line $B$ with angle $\phi_{b}(kT)$, then control law (\ref{eq:smooth_control}) is a decentralized control with sweeping speed $v_{0}$ along $B$ and with the equidistant of $d$ between vehicles. \\
\textbf{Proof:}
Our proof is similar to \cite{cheng2011decentralizedsweep} and we also incorporate the weighted average functions to show how this approach still achieves the solution.

First, we write the coordinates of a mobile robot as under:

\begin{equation}\label{smooth_coord}
  s_{i}(kT)=
  \begin{pmatrix}
    x_{i}(kT) \\
    y_{i}(kT)
  \end{pmatrix}=
    \begin{pmatrix}
     \bar{x_{i}}(kT)+ \hat{x}(kT)\\
     \bar{y_{i}}(kT)+ \hat{y}(kT)
  \end{pmatrix}
 \end{equation}
for $i=1,2,3,\dots,n$ and $k=0,1,2,\dots$. Where,

\begin{equation}\label{smooth_coord_cap}
  \begin{pmatrix}
    \hat{x}((k+1)T) \\
    \hat{y}((k+1)T)

  \end{pmatrix}=
    \begin{pmatrix}
         \hat{x}(kT)+v_{0}T\cos(\phi_{b}) \\
         \hat{y}(kT)+v_{0}T\sin(\phi_{b})
  \end{pmatrix}
 \end{equation} with $\hat{x}(0)=\hat{y}(0)=0$
 and
 \begin{equation}\label{smooth_coord_bar}
  \begin{pmatrix}
    \bar{x}_{i}((k+1)T) \\
    \bar{y}_{i}((k+1)T)

  \end{pmatrix}=
    \begin{pmatrix}
         \bar{x}_{i}(kT)+ T\cos(\theta_{i}(kT))\sqrt{\bar{v}_{i}(kT)^2+(\hat{v}_{i}(kT)-v_{0})^2}    \\
         \bar{y}_{i}(kT)+ T\sin(\theta_{i}(kT))\sqrt{\bar{v}_{i}(kT)^2+(\hat{v}_{i}(kT)-v_{0})^2}
  \end{pmatrix}  +  w_{i}(kT)
 \end{equation}

 where

\begin{equation}\label{smooth_w_i}
w_{i}(kT):=
 \begin{pmatrix}
    v_{i}(kT) - \sqrt{\bar{v}_{i}(kT)^2+(\hat{v}_{i}(kT)-v_{0})^2})T\cos(\theta_{i}(kT))  \\
    v_{i}(kT) -   \sqrt{\bar{v}_{i}(kT)^2+(\hat{v}_{i}(kT)-v_{0})^2})T\sin(\theta_{i}(kT))

  \end{pmatrix}-
   \begin{pmatrix}
    {v}_{0}T \cos(\phi_{b}) \\
    {v}_{0}T \sin(\phi_{b})

  \end{pmatrix}
 \end{equation}
 with $\bar{x}_{i}(0)=x_{i}(0)$, $\bar{y}_{i}(0)=y_{i}(0)$

Next, we consider Eq. (\ref{smooth_coord_bar}). By using Eq. (\ref{smooth_vi_barhat}), we show that there exists a line as:

\begin{equation}\label{eq:L0_Proof}
  \bar{L}_{0} =\{(x,y) \in \mathbb{R}^2 : x \cos(\phi_{b})+y \sin(\phi_{b})=\bar{c}_{0} \}
\end{equation}

for some $\bar{c}_{0}$ such that
\begin{equation}\label{eq:smooth_s_L0}
 \lim_{k\to\infty}\parallel \bar{s}_{i}(kT)-\bar{L}_{0}\parallel= 0 \text{, for }i=1,2,3,\ldots,n.
\end{equation}
where $\bar{s}_{i}(.):=[\bar{x}_{i}(.) \text{ } \bar{y}_{i}(.)]^T$. We also show that for a give $d>0$, there exists a set $\{z_{1},z_{2},\dots,z_{n}\}$ that is a permutation of the set $\{{1,2,...,n}\}$ such that

\begin{equation}\label{smooth_barrier}
  \lim_{k\to\infty} \parallel \bar{s}_{z_{i}}(kT) - p_{i}       \parallel=0, \text{ i=1,2,\dots,n}
\end{equation}

where $p_{i}$ is defined as:

\begin{equation}\label{eq:smooth_points}
\begin{split}
  &p_{i}:= \begin{cases}
 p_{b}+(d \times i)u\text{   },if\text{ }{\cal{P}} \subset B_{1}\\
 p_{b}-(d \times i)u\text{   },if\text{ }{\cal{P}} \subset B_{2}\\
 \end{cases}
\end{split}
\end{equation}
 and $ p_{b}:= \bar{L}_{0} \cap B$.

We have defined our heading update rule as under:

\begin{equation}\label{smooth_Theta}
 {\Theta}_{i}(kT) :=w_{ii}\phi_i(kT)+\sum_ {j\in {\cal F}_{i}(kT)}w_{ij}\phi_{j}(kT)
\end{equation}

and

\begin{equation}\label{eq:smooth_phi_update_repeated}
 \phi_{i}((k+1)T)= \Theta_{i}(kT)
\end{equation}

We compare our weighting factors with the nearest neighbour rule based approach used in \cite{cheng2011decentralizedsweep} and we notice that:

\begin{equation}\label{smooth_w_eq}
  w_{ii}=w_{ij}=\frac{1}{1+|{\cal F}_{i}(kT)|}
\end{equation}

where ${\cal F}_{i}(kT)$ are number of neighbours of a mobile robotic sensor $i$ at a particular time $kT$. So, our considered weighting factors are also non-negative and normalised to 1.

If we have a line $L$ and initial conditions of a mobile robotic sensor as\\
$(\bar{x}_{i}(0),\bar{y}_{i}(0),\theta_{i}(0) \in \cal{P})$ for $i=1,2,\dots,n$ then we consider boundary moving virtual robot as the leader. Using assumption (\ref{smooth_Assumptions}-5), $(w_{ii} \text{ \& } w_{ij}) \in (0,1)$, the condition of Eq.(\ref{smooth_w_norm}) and Eq. (\ref{eq:smooth_phi_update_repeated}), we can see the consensus can be achieved on the heading angle using the result from \cite{jadbabaie2003coordination} and all the mobile robotic sensors can follow the boundary moving mobile robotic sensor's heading angle as under:

\begin{equation}\label{eq:smooth_theta_consensus}
 \lim_{k\to\infty}\theta_{i}(kT)= \phi_{b} \text{, for }i=1,2,3,\ldots,n.
\end{equation}

Now Eqs. ((\ref{eq:smooth_control}), (\ref{eq:smooth_theta_consensus})) and $ \lim_{k\to\infty}\phi_{i}(kT)= \phi_{b}$

\begin{equation}\label{eq:smooth_v_bar}
 \Rightarrow \lim_{k\to\infty}\bar{v}_{i}(kT)= 0 \text{, for }i=1,2,3,\ldots,n.
\end{equation}, as \text{ $v_{i}(kT)\neq 0$ $\forall$ $k \geq 0$}.

 i.e. the velocity component of a mobile robotic sensor along line $L_{i}(kT)$ becomes zero.

Eq. (\ref{eq:smooth_theta_consensus}) and Eq. (\ref{eq:smooth_v_bar}) guarantee:

\begin{equation}\label{eq:smooth_w_i}
 \Rightarrow \lim_{k\to\infty}\parallel w_{i}(kT)\parallel= 0 \text{, for }i=1,2,3,\ldots,n.
\end{equation}

Using Eq. (\ref{eq:smooth_w_i}):

\begin{equation}\label{eq:smooth_F_i}
 \Rightarrow \lim_{k\to\infty}c_{ii}(kT)= \bar{c}_{0} \text{, for }i=1,2,3,\ldots,n.
\end{equation}

By the definition of $c_{ii}$ and the property $\lim_{k\to\infty}\phi_{i}(kT)=\phi_{b}$, the condition in Eq. \ref{eq:smooth_s_L0} holds. Obviously, $\bar{p_{0}} \in \bar{L}_{0}$.

Now, we show the condition mentioned as Eq. (\ref{smooth_barrier}). We define the largest distance, $d_{max}(kT)$, from the line $\bar{L}_{0}$ to $\bar{s}_{i}(kT)=[\bar{x}_{i}(kT) \text{ } \bar{y}_{i}(kT)]^T$ as under:

\begin{equation}\label{eq:smooth_dmax}
d_{max}(kT):=  \max_{i=1,2,\dots,n}\parallel \bar{s}_{i}(kT)-\bar{L}_{0}\parallel
\end{equation}

For a given $\epsilon > 0$, condition (\ref{eq:smooth_s_L0}) implies that there exists $J \geq 0$ such that $d_{max}(kT) < \epsilon $, for all $k\geq J$, and there also exists a set $\{z_{1},z_{2},\dots, z_{n}\}$ that is a permutation of the set $\{{1, 2,\dots,n}\}$ such that the projections of the positions of vehicles $z_{1},z_{2},\dots, z_{n}$ on the line $\bar{L}_{0}$ satisfy one of the following conditions:

\begin{equation}\label{eq:smooth_proj_left}
q_{b} < q_{z_{1}}(kT)< q_{z_{2}}(kT) < \dots <q_{z_{n}}(kT) \text{ ,if } {\cal{P}} \subset B_{1}
\end{equation}

 and

\begin{equation}\label{eq:smooth_proj_right}
q_{b} > q_{z_{1}}(kT)> q_{z_{2}}(kT) > \dots > q_{z_{n}}(kT) \text{ ,if } {\cal{P}} \subset B_{2}
\end{equation} for all $k \geq J$, where

 \begin{equation}\label{smooth_qz}
   q_{z_{i}}(kT):=[\sin(\phi_{b})\text{ }-\cos(\phi_{b})]^T \times s_{z_{i}}(kT)
 \end{equation}

  \begin{equation}\label{smooth_qb}
   q_{b}:=[\sin(\phi_{b})\text{ }-\cos(\phi_{b})]^T \times p_{b}
 \end{equation}
 for $i=1,2,\dots,n$. The conditions (\ref{eq:smooth_proj_left}-\ref{eq:smooth_proj_right}) hold for all initial vehicle positions in $\cal{P}$.

With the update law (\ref{smooth_qi_update}), one can claim (see appendix of \cite{cheng2011decentralizedsweep} for proof of claim 2).

\begin{equation}\label{eq:smooth_qzi}
\begin{split}
  &q_{z{_{i}}}(kT):= \begin{cases}
 q_{b}+(d \times i)\text{   }if\text{ }{\cal{P}} \subset B_{1}\\
 q_{b}-(d \times i)\text{   }if\text{ }{\cal{P}} \subset B_{2}\\
 \end{cases}
\end{split}
\end{equation}
for $i=1,2,...,n$. As per our definition(\ref{eq:smooth_points}), Eq. (\ref{eq:smooth_qzi}) implies that

\begin{equation}\label{eq:smooth_qzlh}
 \lim_{k\to\infty} | q_{z_{i}}(kT) - l^T p_{i} | = 0 \text{, for }i=1,2,3,\ldots,n.
\end{equation}

Eq. (\ref{eq:smooth_s_L0}) and Eq. (\ref{eq:smooth_qzlh}) give Eq. (\ref{smooth_barrier}),
as $\parallel \bar{s}_{z_{i}}(kT) - p_{i} \parallel  \leq \parallel \bar{s}_{i}(kT) - L_{0} \parallel+|q_{z_{i}}(kT)-u^T p_{i}|$. So, we can say: $\bar{s}_{i}(kT)\text{ for } i=1,2,\dots,n$ converges to the line $\bar{L}_{0}$ and $\bar{s}_{z_{i}}(kT)$ converges to the point $p_{i}$.

We conclude our theorem using (\ref{smooth_barrier}). Eqs. (\ref{eq:L0_Eqn}-\ref{eq:points_on_L0}), (\ref{eq:L0_Proof}) and (\ref{eq:smooth_points}), we have

\begin{equation}\label{eq:pi_found}
p_{i}(kT)=p_{i}+kTv_{0} \times [\cos(\bar{\phi}) \text{ } \sin(\bar{\phi})]^T ,i=1,2,\dots,n.
\end{equation}

Now, Eqs. (\ref{eq:pi_found}), (\ref{smooth_coord}-\ref{smooth_coord_cap}) and (\ref{smooth_barrier}):

\begin{equation}\label{eq:smooth_sweep_def_proved}
\Rightarrow  \lim_{k\to\infty}||s_{z_{i}}(kT) - p_{i}(kT) ||=\lim_{k\to\infty}||\bar{s}_{z_{i}}(kT) - p_{i}(kT) ||=0, i=1,2,3,\ldots,n,
\end{equation}, as $[\hat{x}_{i}(kT) \text{ } \hat{y}_{i}(kT)]^T = kTv_{0} \times [\cos(\bar{\phi}) \text{ } \sin(\bar{\phi})]^T$. Hence, we meet the sweep coverage definition (\ref{eq:smooth_sweep_def}) - which completes the proof of the theorem. We have also demonstrated above that the nearest neighbour rule with weighted average functions still completes the proof of the theorem similar to \cite{cheng2011decentralizedsweep} and there is no effect on the convergence conditions.

\section{Simulation Results}

We obtain Figs. (3.4-3.7) and Figs. (3.8-3.11) for $n=6$, $R_{c}=2$, $R_{b}=2$, $R_{s}=2$, $v_{0}=0.05$ and ${\phi}_{b}=\pi/2$. We represent the moving mobile robot with a blue circle and its heading angle with a red arrow. We divide the simulations to make a clear comparison of the results obtained from recent research \cite{cheng2011decentralizedsweep} with that of the current control law using weighted average functions.

\begin{figure}[H]
\begin{center}
   \includegraphics[width=1.0\columnwidth]{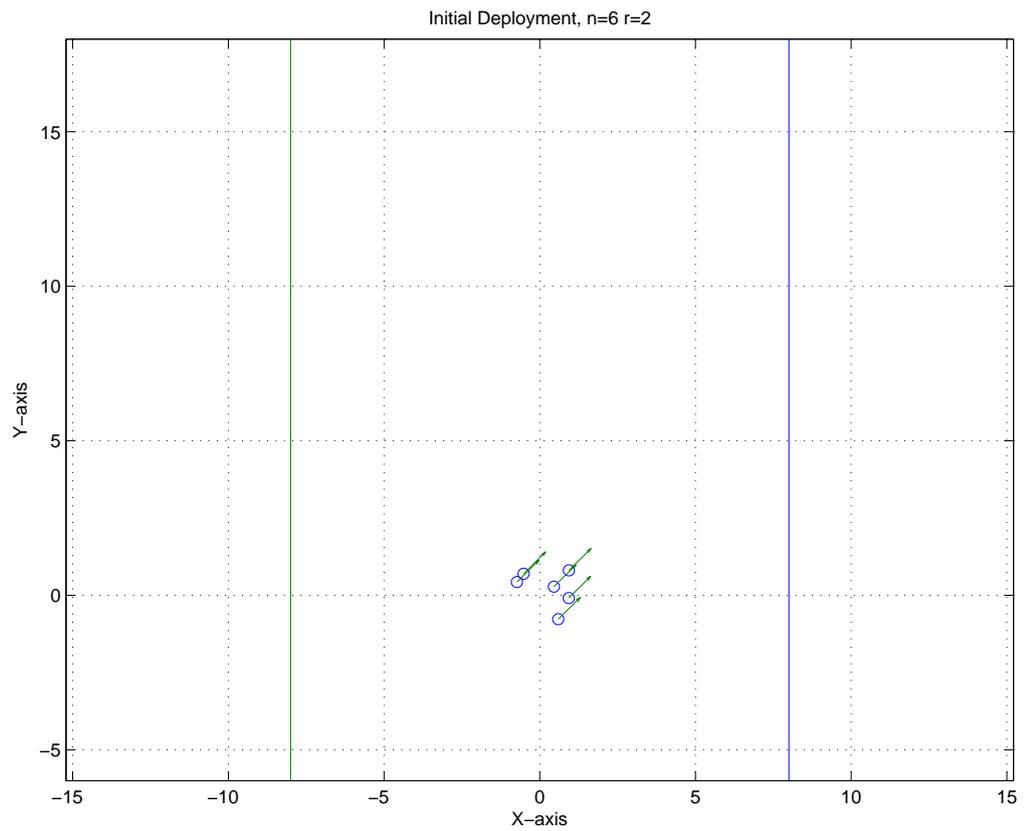}
    \caption{Simple Average Functions - Initial Deployment}
\end{center}
\label{fig:SAF_Initial_Deployment}
\end{figure}

\begin{figure}[H]
\begin{center}
   \includegraphics[width=1.0\columnwidth]{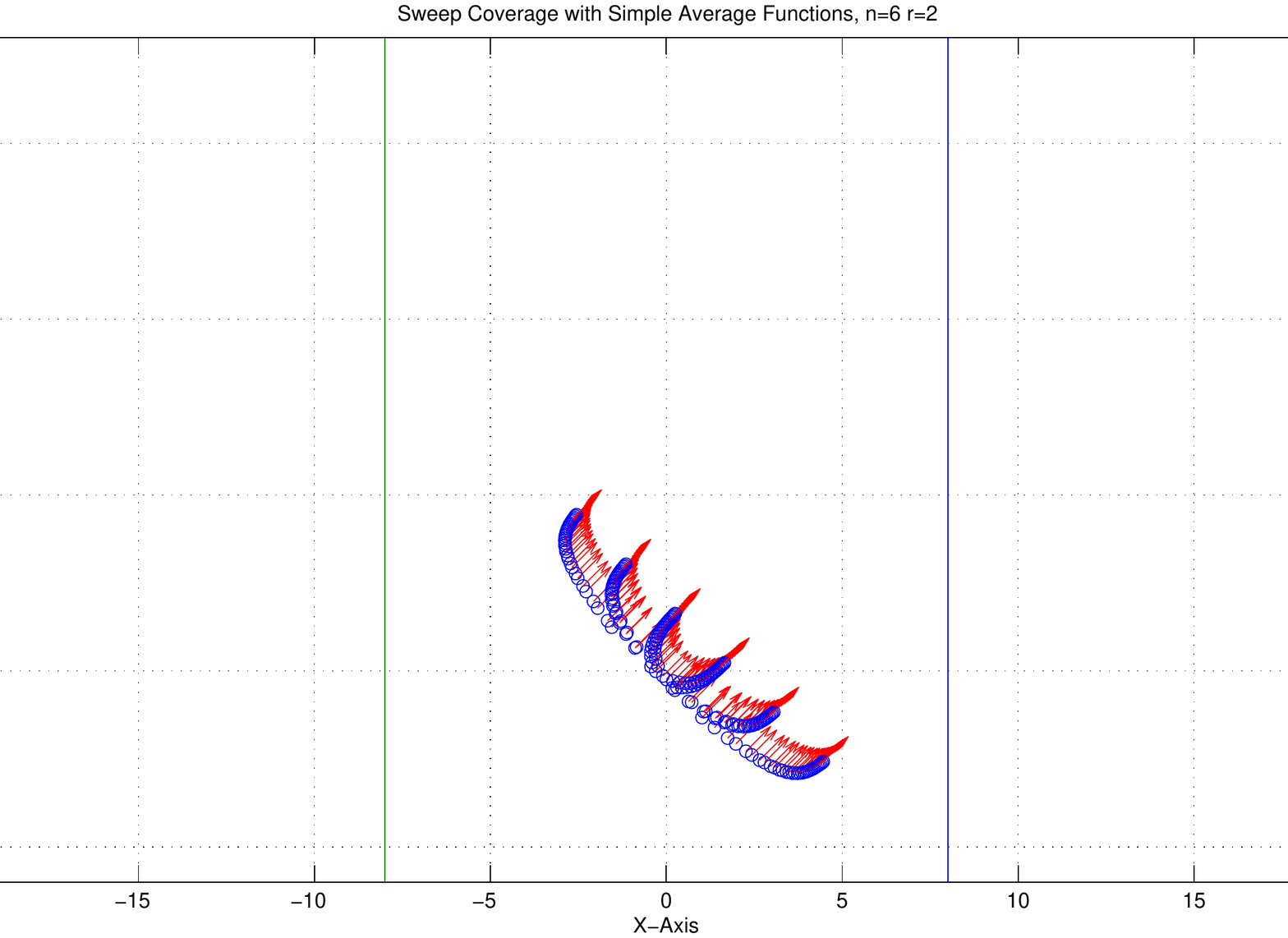}
  % \label{fig:theFig}
    \caption{Simple Average Functions - Sweeping Mobile Robots}
\end{center}
\label{fig:SAF_Sweeping}
\end{figure}

\begin{figure}[H]
\begin{center}
   \includegraphics[width=1.0\columnwidth]{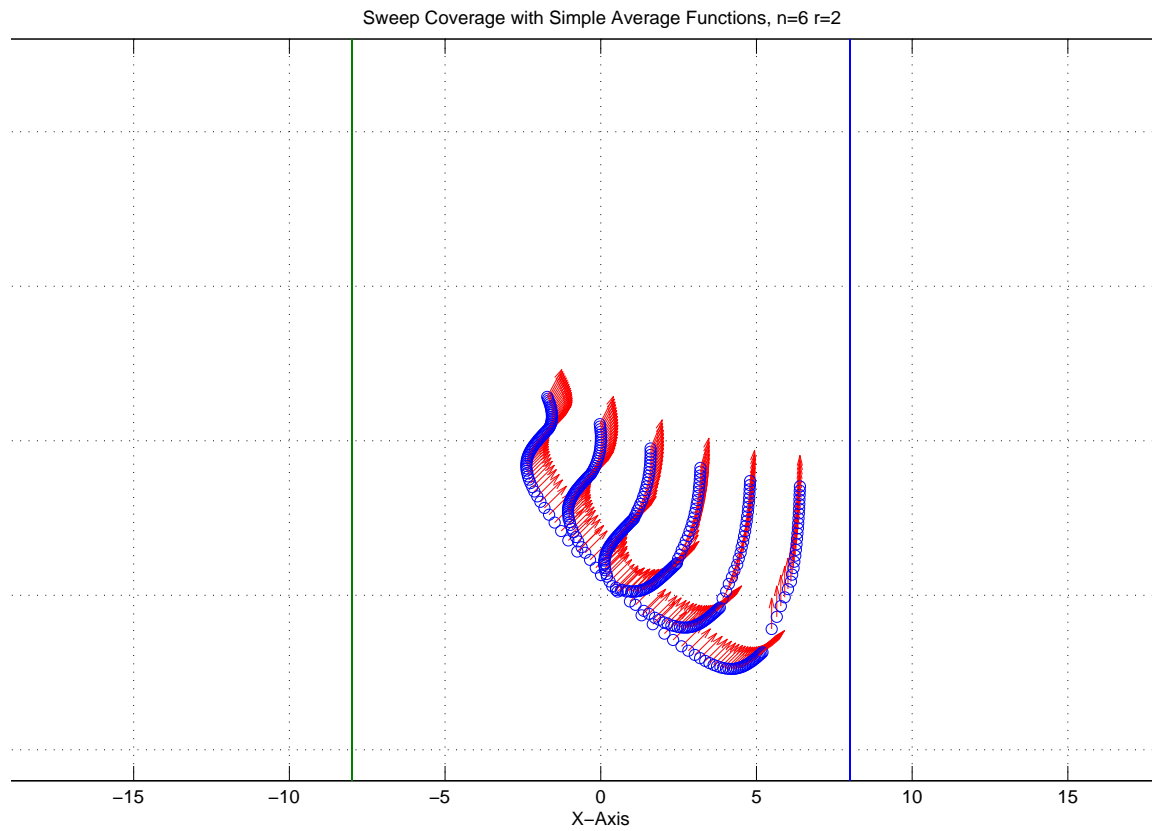}
  % \label{fig:theFig}
    \caption{Simple Average Functions - Turning Mobile Robots}
\end{center}
\label{fig:SAF_Turning}
\end{figure}

\begin{figure}[H]
\begin{center}
   \includegraphics[width=1.0\columnwidth]{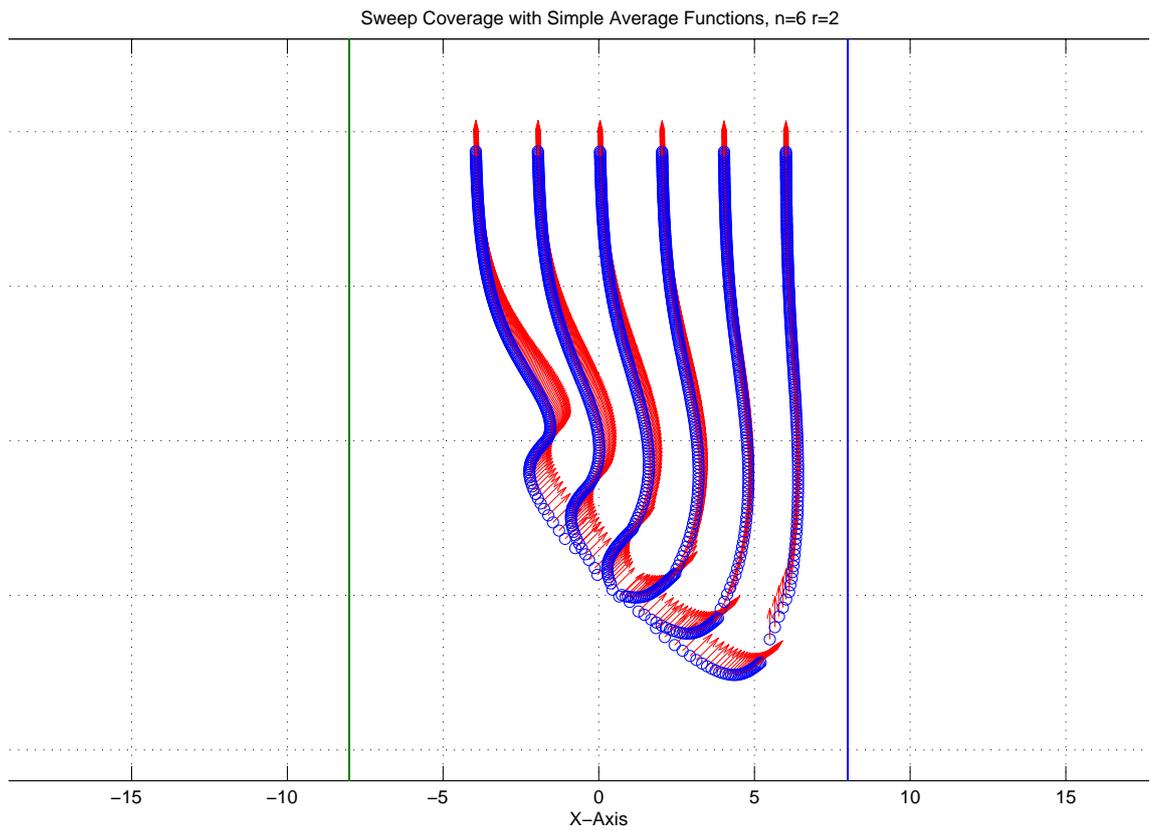}
  % \label{fig:theFig}
    \caption{Simple Average Functions - Final Sweeping Mobile Robots}
\end{center}
\label{fig:SAF_Finally_Sweeping}
\end{figure}

%///////////////////////////////////////////////////////////////////////////////////////

\begin{figure}[H]
\begin{center}
   \includegraphics[width=1.0\columnwidth]{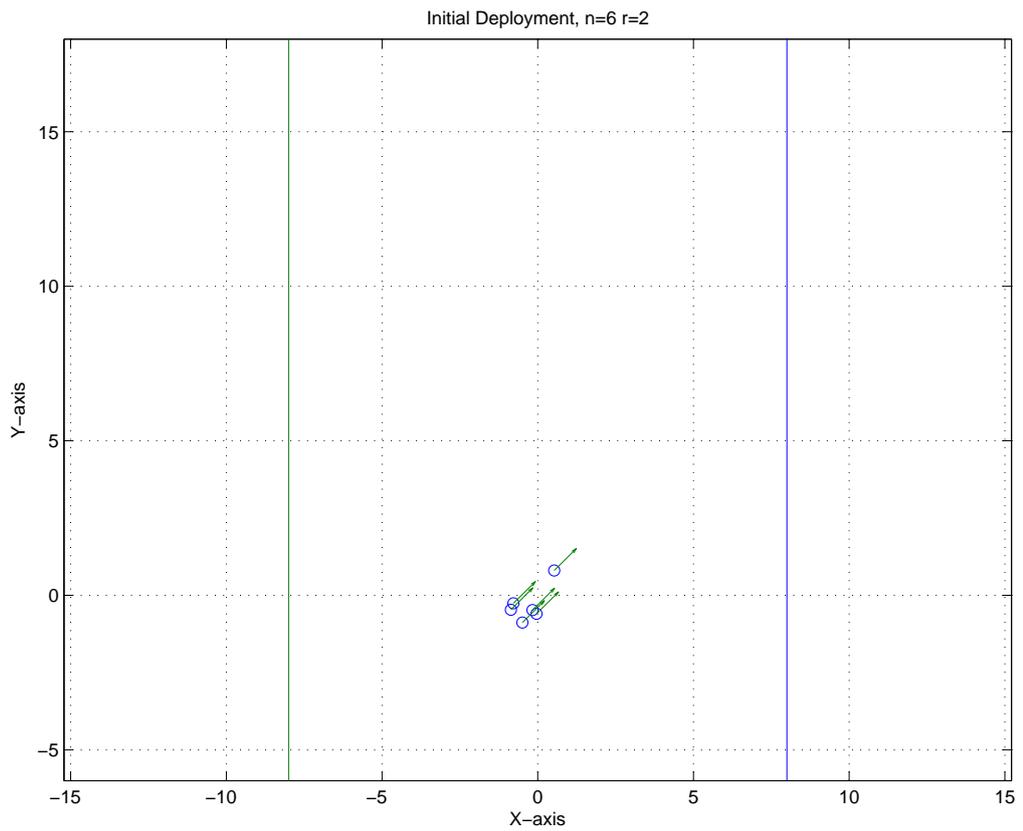}
  % \label{fig:theFig}
    \caption{Weighted Average Functions - Initial Deployment}
\end{center}
\label{fig:WAF_Initial_Deployment}
\end{figure}

\begin{figure}[H]
\begin{center}
   \includegraphics[width=1.0\columnwidth]{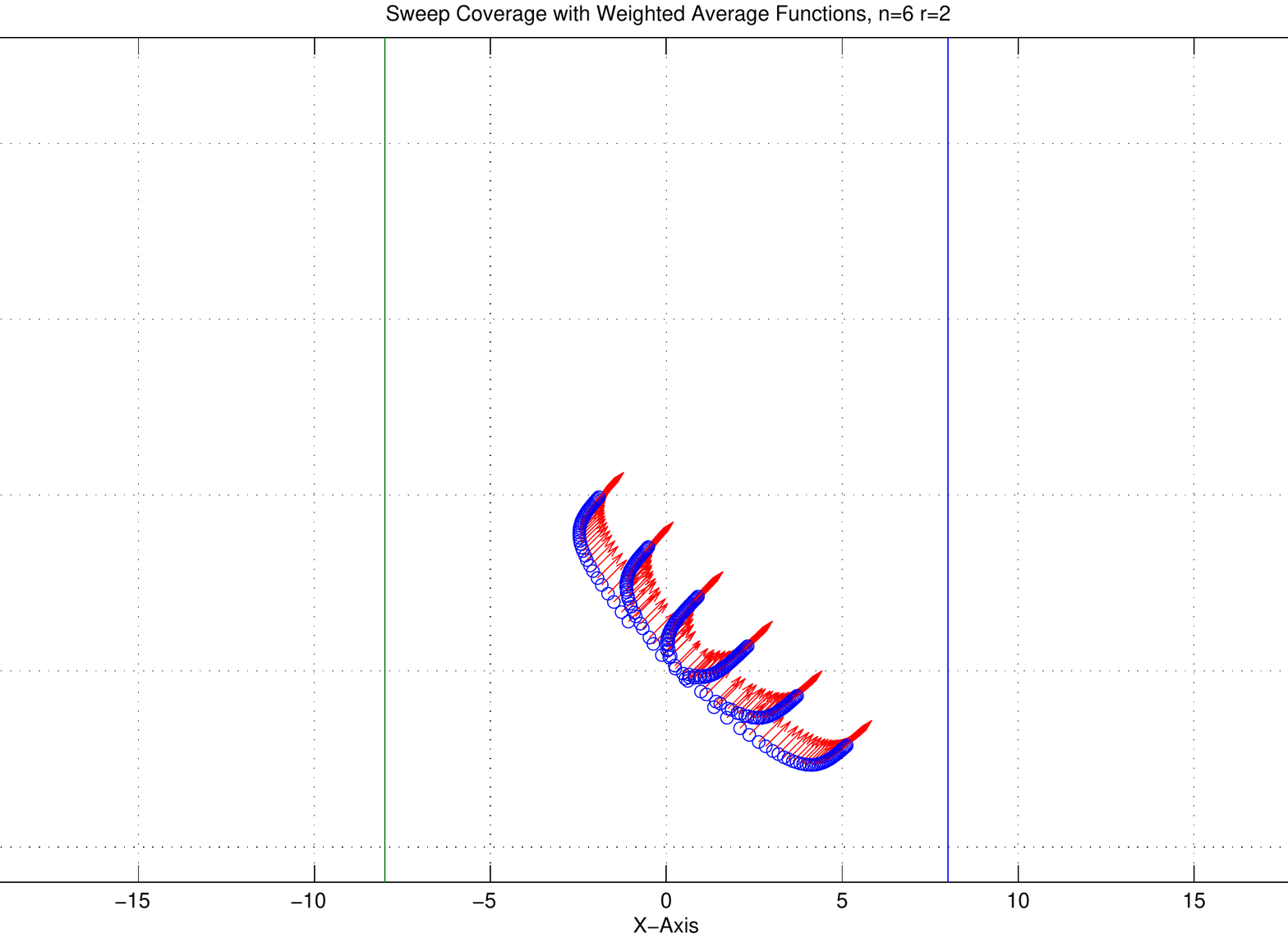}
  % \label{fig:theFig}
    \caption{Weighted Average Functions - Sweeping Mobile Robots}
\end{center}
\label{fig:WAF_Sweeping}
\end{figure}

\begin{figure}[H]
\begin{center}
   \includegraphics[width=1.0\columnwidth]{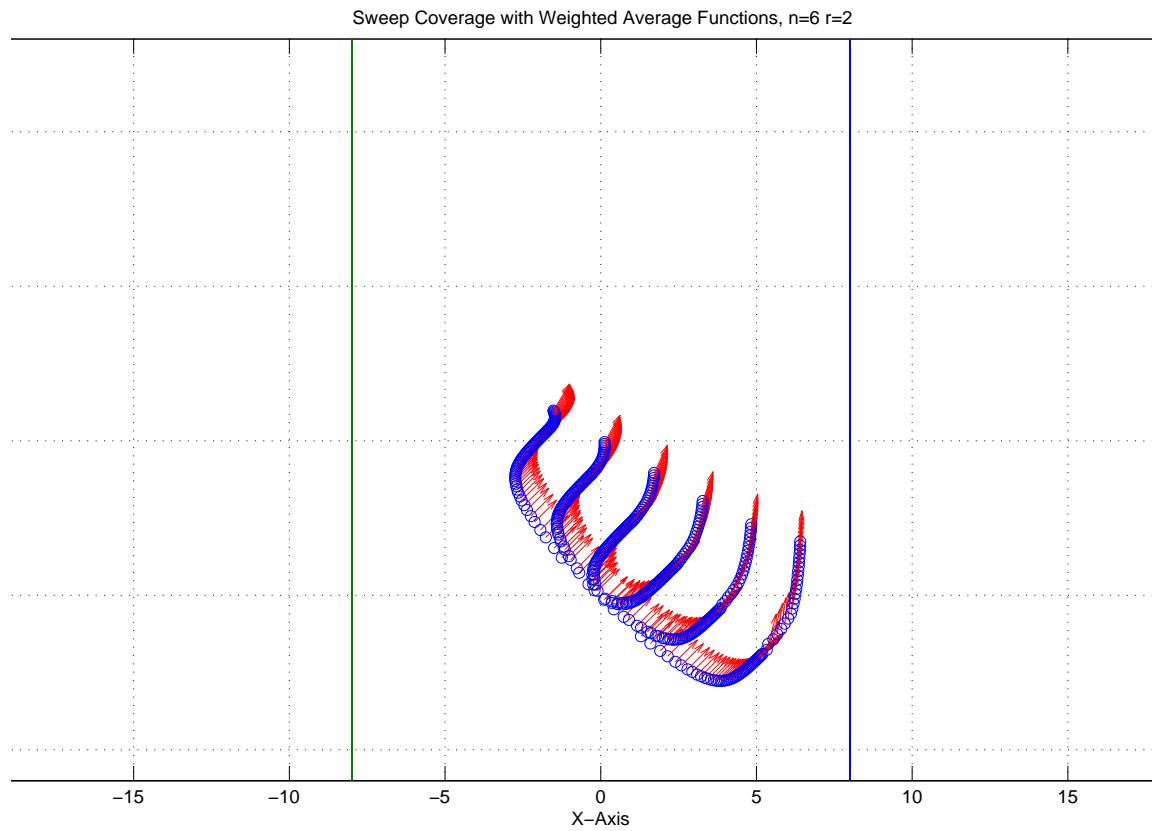}
  % \label{fig:theFig}
    \caption{Weighted Average Functions - Turning Mobile Robots}
\end{center}
\label{fig:WAF_Turning}
\end{figure}

\begin{figure}[H]
\begin{center}
   \includegraphics[width=1.0\columnwidth]{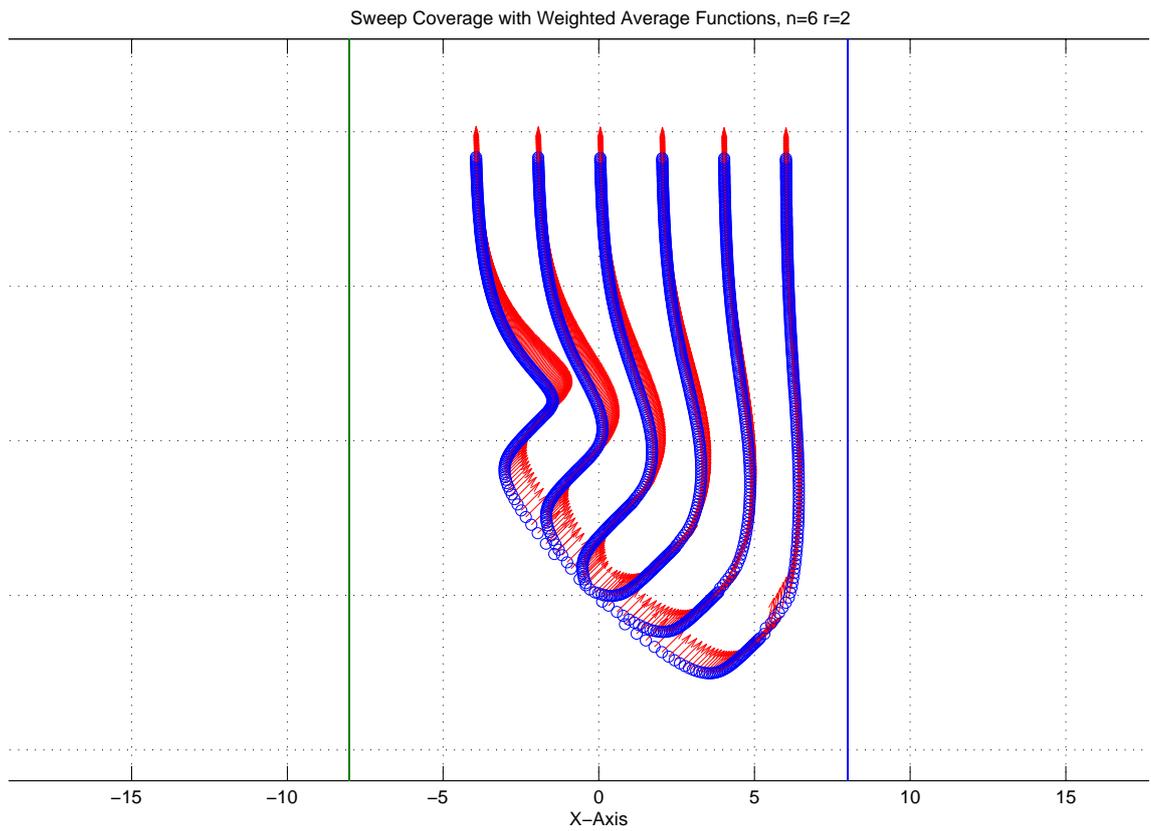}
  % \label{fig:theFig}
    \caption{Weighted Average Functions - Finally Sweeping Mobile Robots}

\end{center}
\label{fig:WAF_Finally_Sweeping}
\end{figure}

%///////////////////////////////////////////////////////////////////////////////////////

Figs. (3.12-3.15) and Figs. (3.16-3.19) also show stage-wise simulations for a worst case scenario in order to make a full comparison of recent research \cite{cheng2011decentralizedsweep} with the current work.

%\begin{figure}[h]
%\begin{center}
%   \includegraphics[width=1.0\columnwidth]{Ch3/SIM21.eps}
%  % \label{fig:theFig}
%    \caption{(i) Initial deployment}
%\caption{Sweep Coverage with Simple Average Functions for a Worst Boundary Angle}
%\end{center}
%\end{figure}

\begin{figure}[H]
  \centering
  \includegraphics[width=1.0\columnwidth]{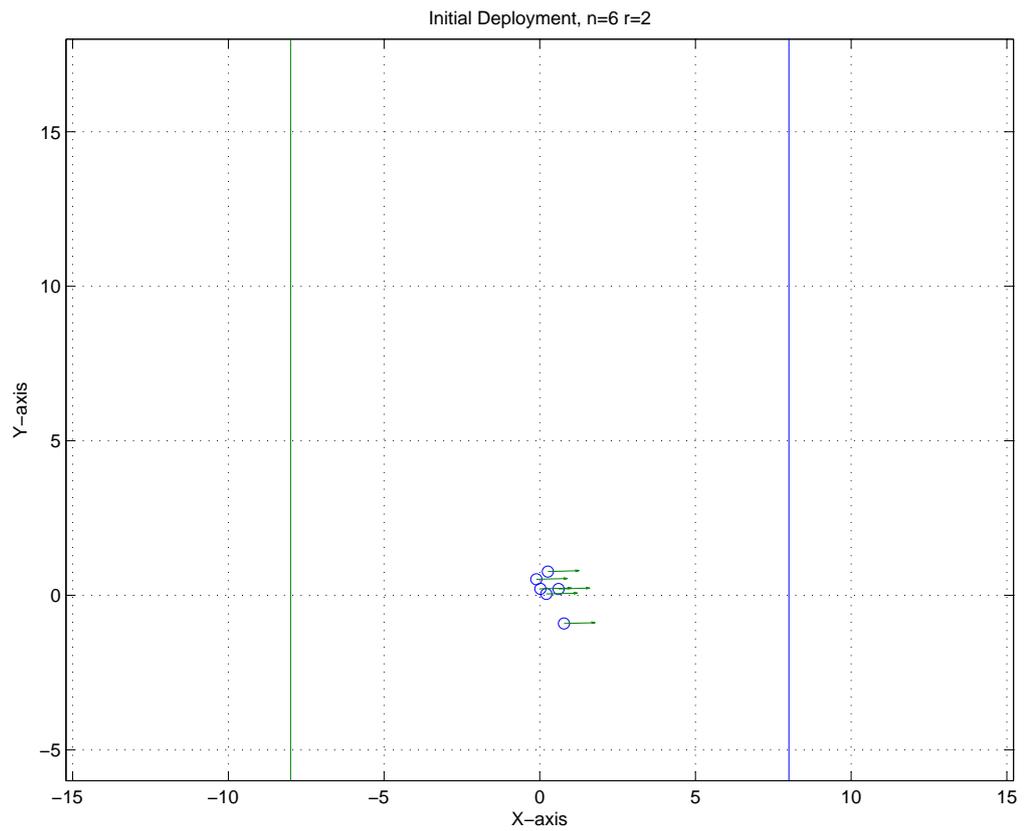}
  \caption{Simple Average Functions for a Worst Boundary Angle - Initial Deployment}\label{SAF_WBA_Initial_Deployment}
\end{figure}

\begin{figure}[H]
  \centering
  \includegraphics[width=1.0\columnwidth]{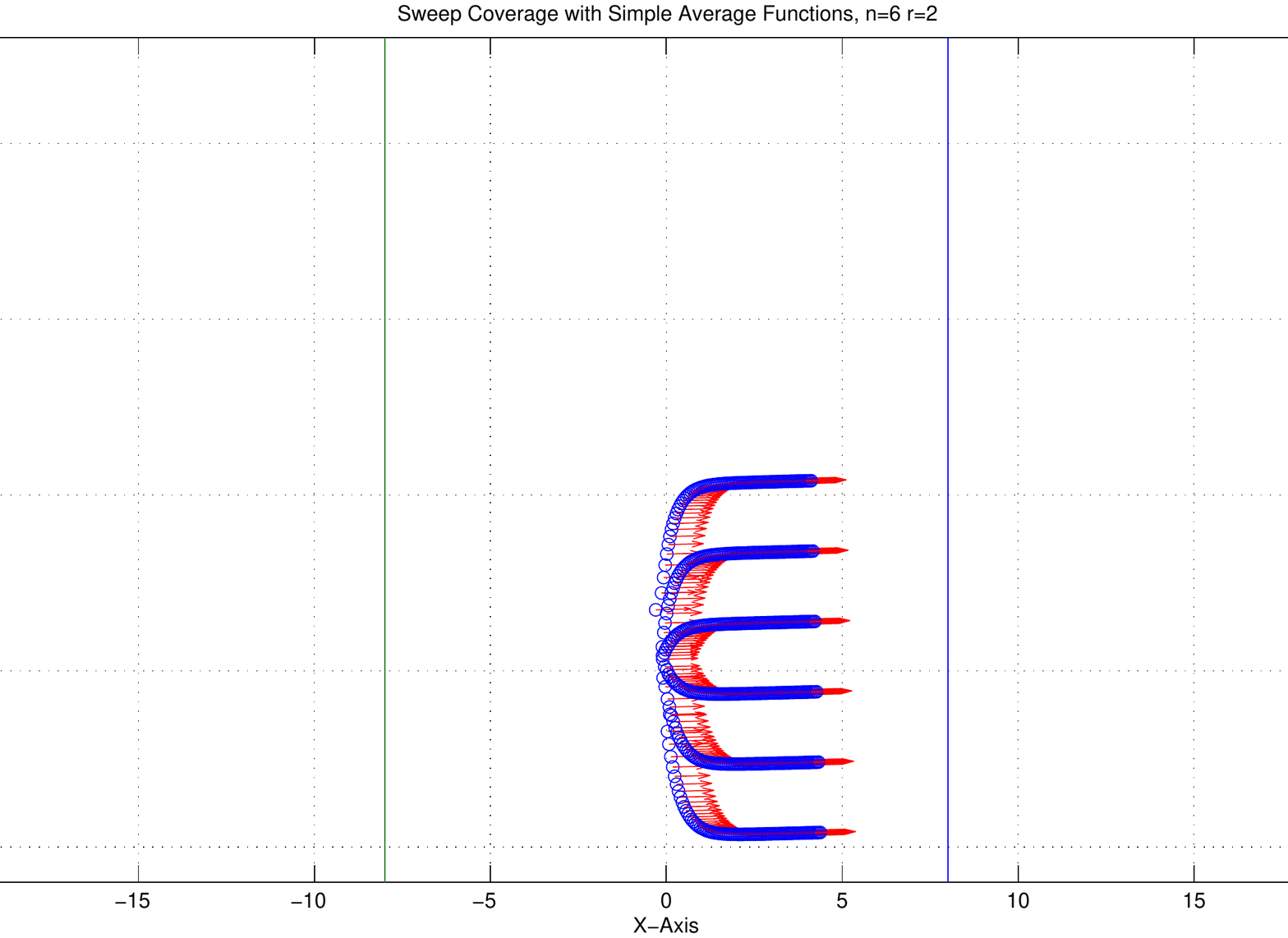}
  \caption{Simple Average Functions for a Worst Boundary Angle - Sweeping Mobile Robots}\label{SAF_WBA_Sweeping}
\end{figure}

\begin{figure}[H]
  \centering
  \includegraphics[width=1.0\columnwidth]{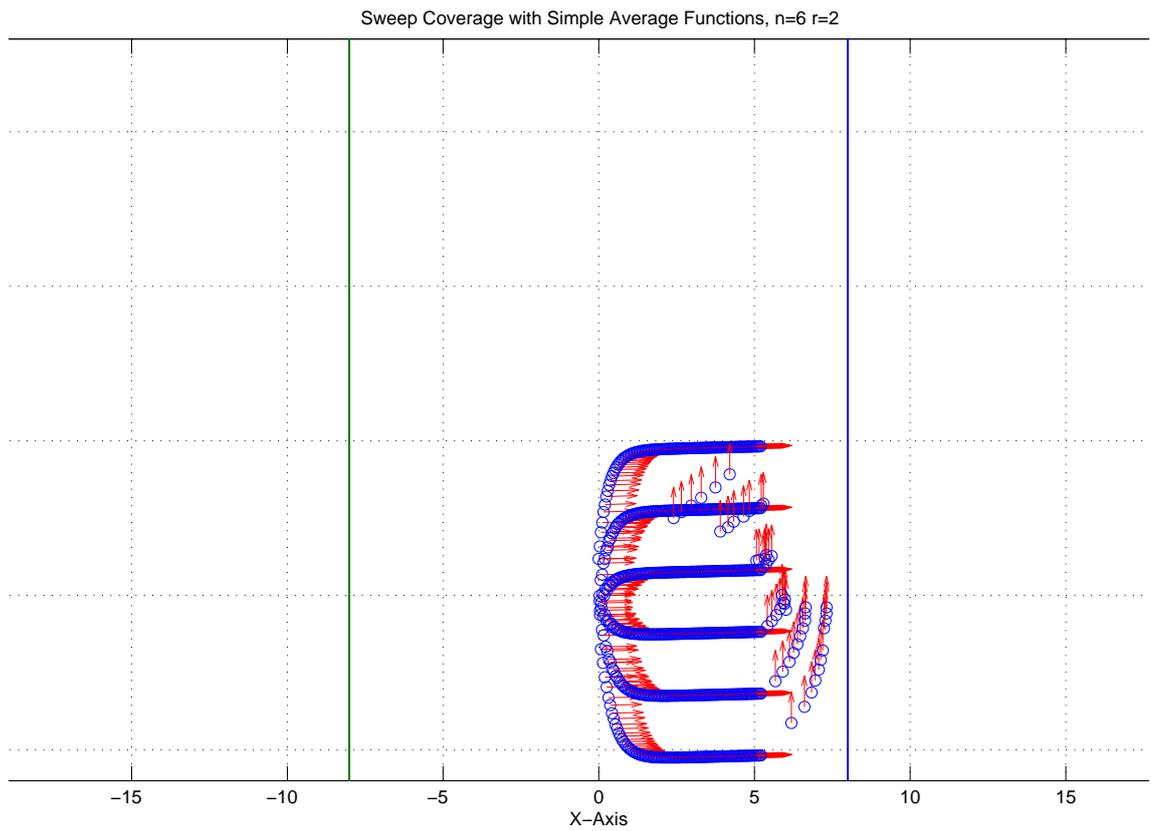}
  \caption{Simple Average Functions for a Worst Boundary Angle - Turning Mobile Robots}\label{SAF_WBA_Turning}
\end{figure}

\begin{figure}[H]
  \centering
  \includegraphics[width=1.0\columnwidth]{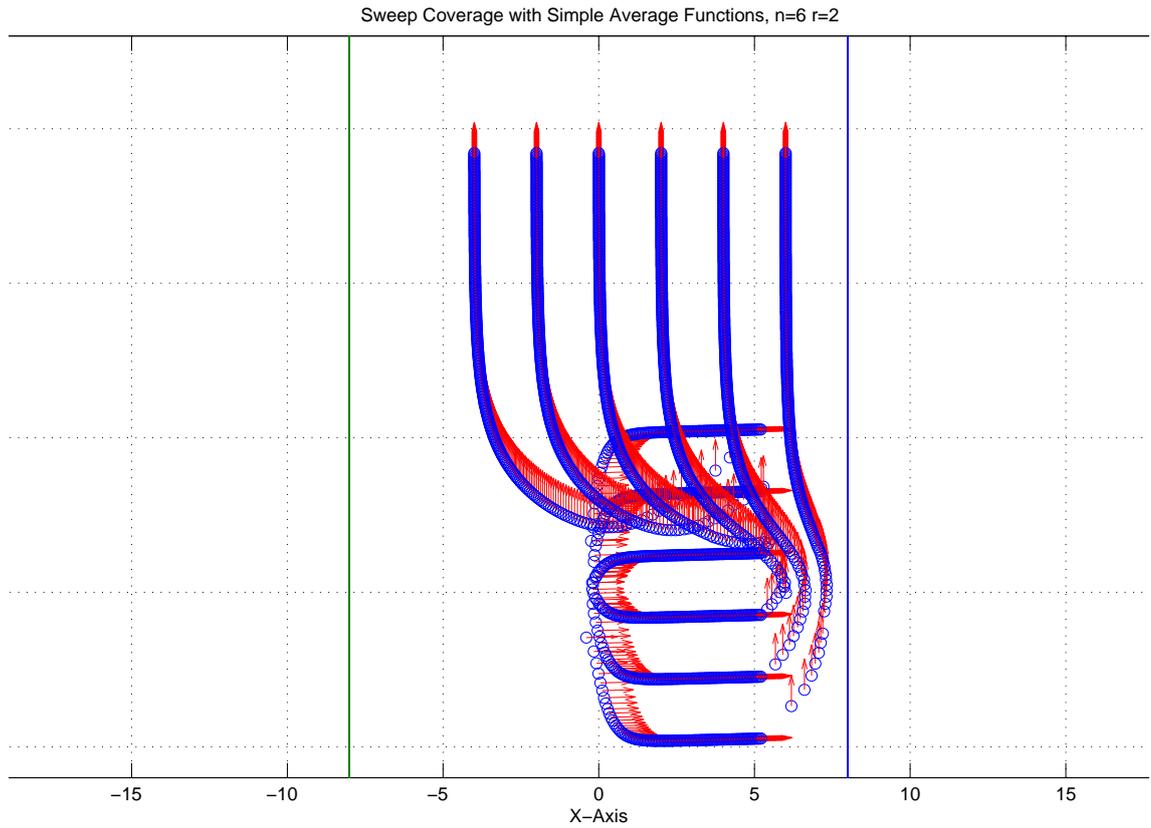}
  \caption{Simple Average Functions for a Worst Boundary Angle - Finally Sweeping Mobile Robots}\label{SAF_WBA_Final}
\end{figure}

\begin{figure}[H]
  \centering
  \includegraphics[width=1.0\columnwidth]{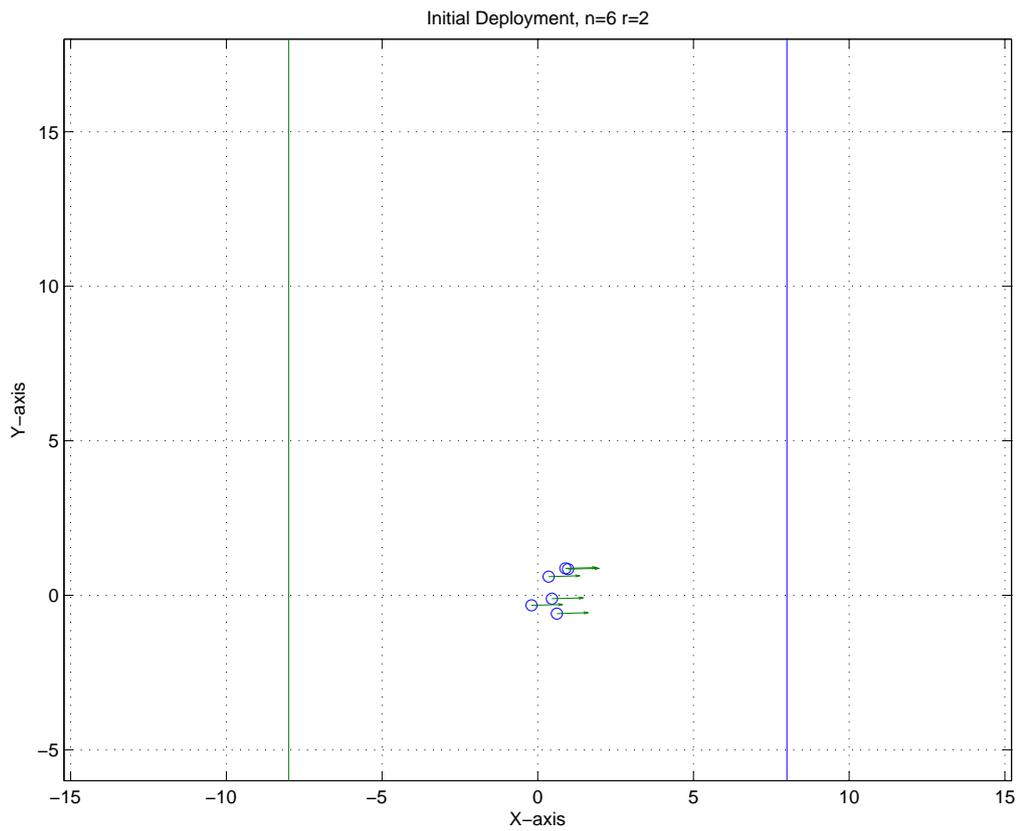}
  \caption{Weighted Average Functions for a Worst Boundary Angle - Initial Deployment}\label{WAF_WBA_Initial_Deployment}
\end{figure}

\begin{figure}[H]
  \centering
  \includegraphics[width=1.0\columnwidth]{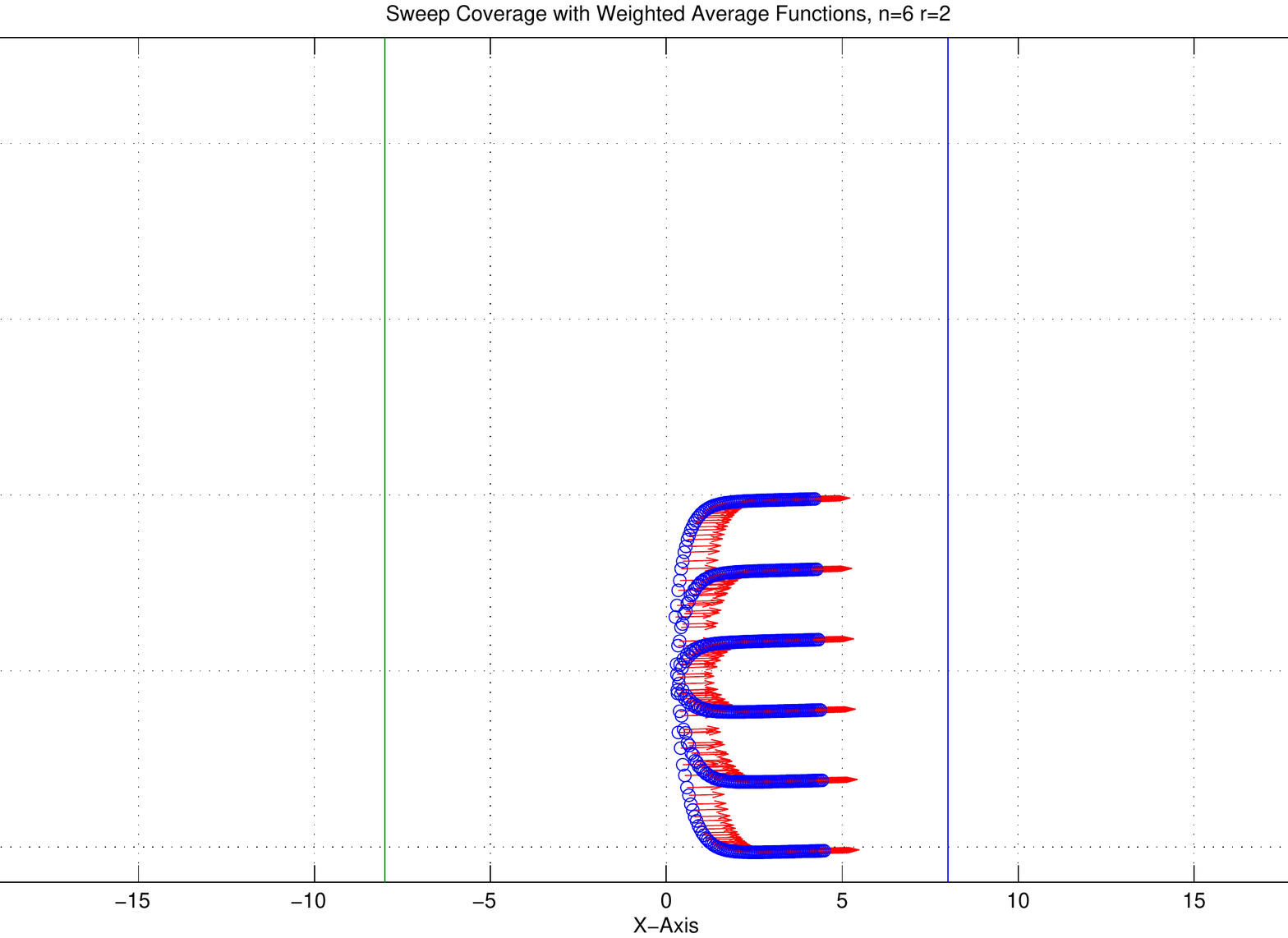}
  \caption{Weighted Average Functions for a Worst Boundary Angle - Sweeping Mobile Robots}\label{WAF_WBA_Sweeping}
\end{figure}

\begin{figure}[H]
  \centering
  \includegraphics[width=1.0\columnwidth]{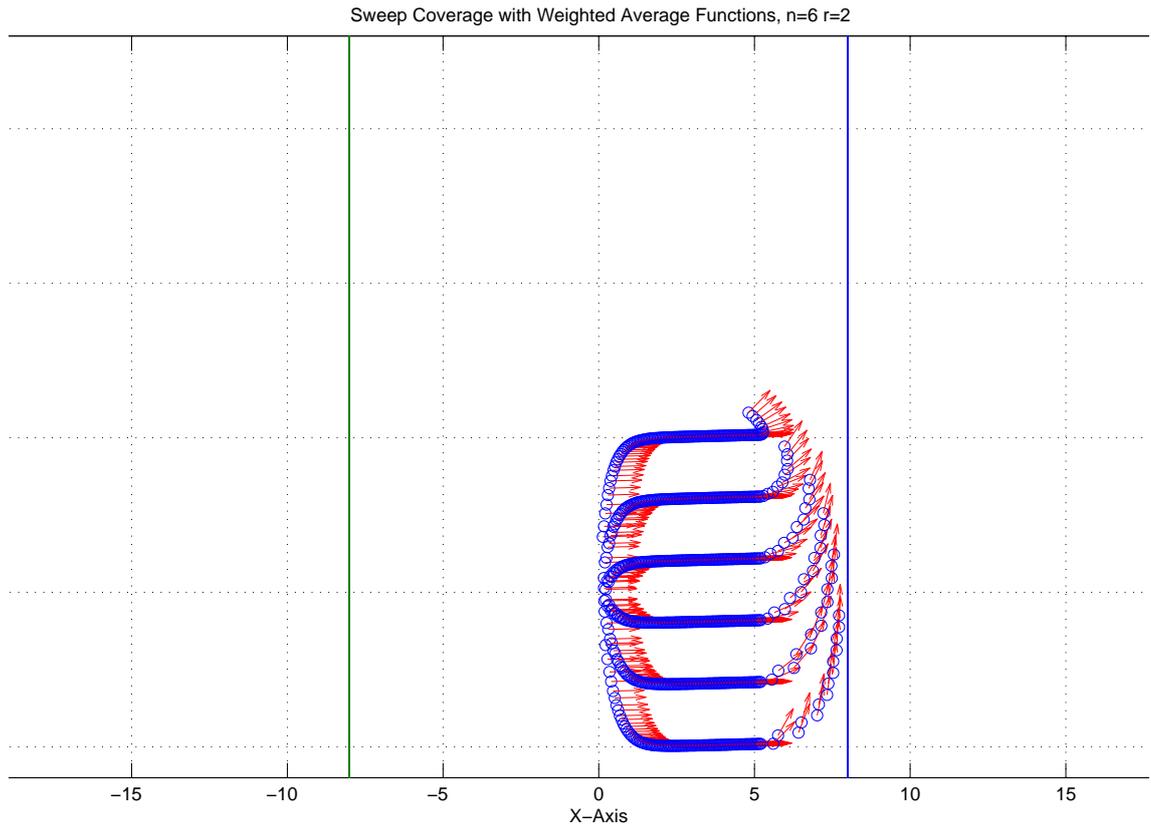}
  \caption{Weighted Average Functions for a Worst Boundary Angle - Turning Mobile Robots}\label{WAF_WBA_Turning}
\end{figure}

\begin{figure}[H]
  \centering
  \includegraphics[width=1.0\columnwidth]{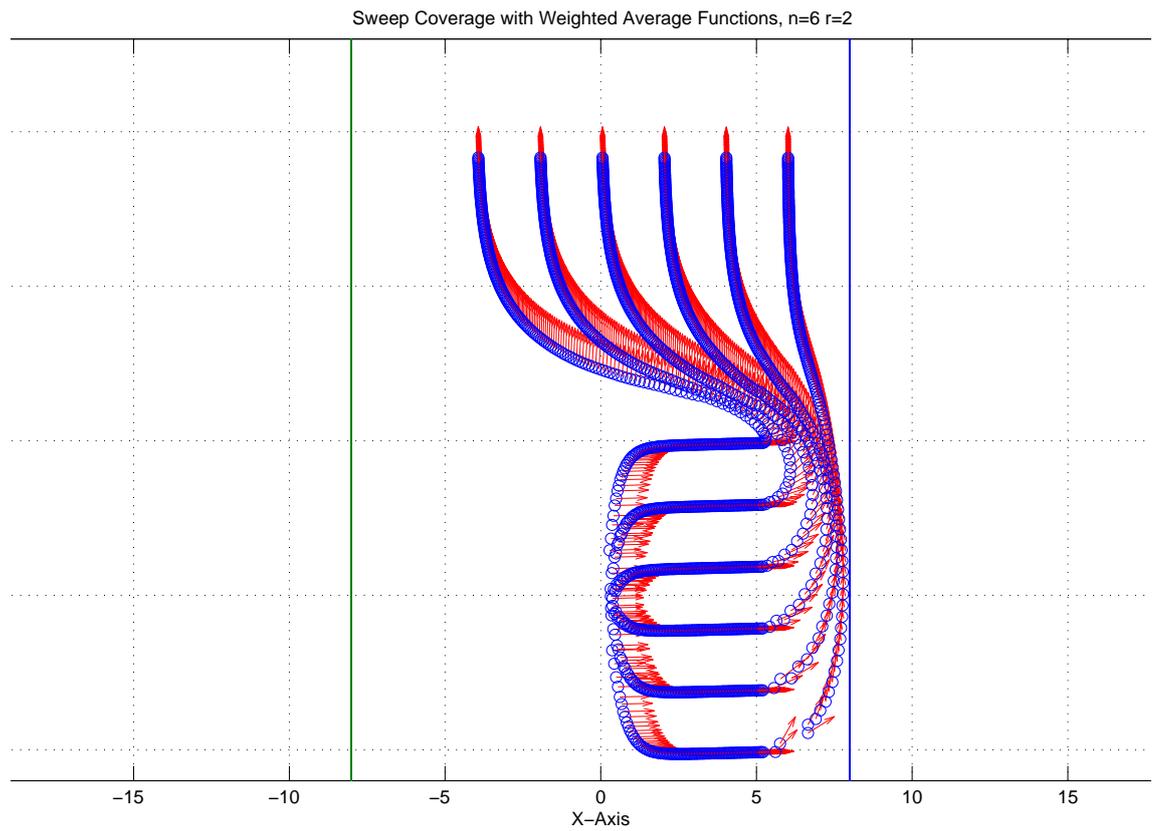}
  \caption{Weighted Average Functions for a Worst Boundary Angle - Finally Sweeping Mobile Robots}\label{WAF_WBA_Final}
\end{figure}

\section{Chapter Conclusions and Future Research Directions}

\subsection{Chapter Conclusions}

We can compare simulation results obtained from recent research \cite{cheng2011decentralizedsweep}, Fig. (3.6-3.7) with that of current research, Fig. (3.10-3.11). After detection of the boundary shown in Fig. (3.10-3.11), the mobile robot nearest to the boundary is making a smooth trajectory, and consequently neighbouring mobile robot is more sensitive in adjusting its velocity and position. Note that, there is an agreement between Fig. 3.5 and  Fig. 3.9, because our revised nearest neighbour rules give equal weight (i.e. $w_{ii}=w_{ij}$ and $m_{ii}=m_{ij}$) during calculation of velocity and heading angle in the absence of detected boundary. So, our broadened nearest neighbour rules with weighted average functions can behave well in both manners, that is, an equal normalised weighting value is considered without the detection of a boundary (which has already been demonstrated by recent research \cite{cheng2011decentralizedsweep} and a normalised but an unequal weighting value for the calculation of control inputs (velocity and heading angle) is given to each mobile robot after the detection of a boundary. Remember that, recent sweep coverage algorithm \cite{cheng2011decentralizedsweep} always gives a normalised equal weighting value to each mobile robot for the calculation of its own control inputs.

The advantage of weighted average functions can be further noticed, when there is a worst boundary angle in the path of slightly inclined (less than 2 degree) sweeping mobile robots. In Fig. \ref{SAF_WBA_Turning}, we can see the upper mobile robots are unable to make a smooth turn and these mobile robots had to proceed backwards in order to adjust the respective trajectory, but clearly in Fig. \ref{WAF_WBA_Turning} all the mobile robots can make a smooth turn in order to adjust individual trajectory against a worst angle boundary coming ahead. We can also see Fig. \ref{SAF_WBA_Final} is not showing a smooth individual trajectory for each mobile robot against a worst angle boundary, whereas Fig. \ref{WAF_WBA_Final} demonstrates a smooth and an efficient trajectory for each mobile robot against the same boundary.

\subsection{Future Research Directions}
Future work could incorporate the tools for state estimation and control in case of missing data with limited communication (see e.g. \cite{savkin1997robust,matveev2004analogue,savkin1995recursive,petersen1999robust,savkin1998robust,pathirana2004location,pathirana2005node}) among the mobile robotic sensors.

%\section*{Acknowledgment}
%This work was supported in part by the Australian Research Council (ARC) with an allocated grant (DP130103898). The main objective of this grant is to conceptually develop new design rules in the area of robust control of mobile networked systems.

\chapter{Vicsek's Model Revisited and Cyclic Behaviour}

In the previous chapter, we have based our control methodology on the nearest neighbour rule and we applied it specifically for sweep coverage problem. The nearest neighbour rule is based on Vicsek's model \cite{vicsek1995novel}. This model can offer some cyclic behaviour under physical constraints of mobile robots. The detailed analysis of Vicsek's model has been performed in \cite{savkin2004coordinated}. In this chapter, we address this cyclic behaviour and our approach is based on a biased control strategy. We validate our approach to be used for decentralized control of mobile robots.

%%%%%%%%%%%%%%%%%%%%%%%%%%%%%%%%%%%%%%%%%%%%%%%%%%%%%%%%%%%%%%%%%%%%%%%%%%%%%%%%

\section{Introduction}
We address the discrete time control of a team of mobile robots reaching a consensus in terms of the same heading angle. In fact, our discrete time control strategy is inspired from Vicsek's model for decentralized coordination of a team of agents. Some researchers have already shown appearance of a cyclic behaviour in Vicsek's model. The existence of a cyclic behaviour prevents the mobile robots to reach a constant heading angle. We define simple rules for the quantized control of mobile robots to reach a constant heading. The simulation study has revealed that a cyclic behaviour of discrete time Vicsek's model can be avoided by the use of an alternate biased strategy bringing the calculated heading angle of a mobile robot to a quantized level. We update the discrete time individual heading angle of a mobile robot based on the self-biased or neighbour biased strategy. A number of numerical simulations have been performed to check the avoidance of a cyclic behaviour in the discrete time swarm of mobile robots meeting constraints.

%%%%%%%%%%%%%%%%%%%%%%%%%%%%%%%%%%%%%%%%%%%%%%%%%%%%%%%%%%%%%%%%%%%%%%%%%%%%%%%%

\section{Related Work}
%
%A team of mobile robots\footnote{The terms, "agent" or "sensor" or "the mobile robotic sensor" or simply, "the mobile robot" will be used throughout this chapter for an autonomous mobile robot having an on-board computation, operation-specific boundary detection and communication capability.} with decentralized control has got different applications like surveillance, reconnaissance, maintenance, inspection and training \cite{gage1995many,gage1992command}.

%The team of mobile robots reaching a constant heading angle is inspired from flock of birds. The mathematical analysis of Vicsek's model \cite{savkin2004coordinated} already demonstrates the decentralized coordination of flock of birds heading in the same direction. It is believed that an agent makes a decision to update its state based on the average of its own plus its neighbours \cite{savkin2004coordinated}. A number of researchers \cite{santoso2010sub,cheng2009decentralized,teimoori2010biologically,cheng2009distributed,cheng2011decentralized,savkin2010decentralized,cheng2011decentralizedsweep,bullo2009distributed,savkin2012optimal,cheng2010decentralizedregion} have developed decentralized control laws based on Vicsek's model \cite{vicsek1995novel} to update the heading angle of each mobile robot participating in the team.

Vicsek's model \cite{vicsek1995novel} provides computer simulation of autonomous agents moving in a plane with the same heading. But in real time applications, a team of mobile robots\footnote{The terms, "agent" or "sensor" or "the mobile robotic sensor" or simply, "the mobile robot" will be used throughout this chapter for an autonomous mobile robot having an on-board computation, operation-specific boundary detection and communication capability.} to reach the same heading angle needs a discrete time model addressing the physical constraints like its quantized orientation. Some researchers \cite{jadbabaie2003coordination,jadbabaie2003distributed,tahbaz2007recurrence,li2004multi} have provided conditions showing that all agents reach the same heading provided existence of a mutual linkage. In research work \cite{savkin2004coordinated}, a rigorous qualitative analysis of Vicsek's model has been addressed. It has been mathematically shown that each mobile robot meeting some assumptions eventually reach a constant heading angle \cite{savkin2004coordinated}. A remark regarding counter-intuitive consequences has been given for the use of the averaging rule mentioned in \cite{jadbabaie2003coordination}. Furthermore, some specific examples of \cite{savkin2008decentralised} demonstrate that a cyclic dynamics of Vicsek's model \cite{vicsek1995novel} could also exist if initial discrete time heading angles are not limited to two adjacent quadrants. So, an assumption (assumption 3.1 in \cite{savkin2008decentralised}) has been considered to upper bound the heading angle of each mobile robot in order to eventually reach a constant heading angle.

We extend the concept of two quadrant operation of \cite{savkin2008decentralised} to four quadrants. Hence, we develop a strategy for the calculation of heading angle of each mobile robot, so that a cyclic behaviour (as mentioned in \cite{savkin2008decentralised}) of Vicsek's model \cite{vicsek1995novel} could be avoided and there could be no limitation on the assumption of upper bounding a heading angle of each mobile robot.

In the upcoming sections, we formally define the objective of this chapter, necessary assumptions to be considered, our control strategy, conclusion stating the achievement of this chapter and some future research directions.

\section{Problem Statement}

Let $x_{i}(kT)$ and $y_{i}(kT)$ be the  Cartesian coordinates ($s_{i}(kT)$), $v$ be the constant speed and $\theta_{i}(kT)$ be the heading of a mobile robotic sensor measured counter clockwise from
the $X$-axis.

\begin{equation}\label{eq:Revisit_sys1}
 x_{i}(k+1)T= x_{i}(kT) + Tv\cos(\theta_{i}(kT))\\
\end{equation}
\begin{equation}\label{eq:Revisit_sys2}
 y_{i}(k+1)T= y_{i}(kT) + Tv\sin(\theta_{i}(kT))\\
\end{equation}for $i=1,2,\ldots,n$, and $k=0, 1, 2,\ldots$

Our control input in the above mentioned dynamics is the heading angle, $\theta_{i}$, which is to be selected from a set of discrete time values.

The objective is to develop a real time decentralized control strategy, so that each mobile robot in the team could eventually reach a constant heading angle. In developing the control strategy, we specifically consider how the cyclic behaviour \cite{savkin2008decentralised} of Vicsek's model \cite{vicsek1995novel} could be avoided and how the limitation to bound the team of mobile robots in two adjacent quadrants (or half plane) operation could be extended to four quadrants.

\subsection{Assumptions}

\begin{enumerate}\label{Assumptions}
  \item We assume each mobile robot has on-board computation capability.
  \item Each mobile robot has the ability to communicate with the neighbour(s) within a disk of radius $R_{c}>0$ for $t\in[kT, (k+1)T)$ and $k=0, 1, 2,\ldots$. The communication range, $R_{c}$, basically describes the edge of an undirected graph. We mathematically define disk of communication range $R_{c}$ as under:

\begin{equation}
C_{i,R_{c}}(kT):=p\in R^2:||p - s_{i}(kT)|| \leq R_{c}
\end{equation}
where $\|\cdot\|$ denotes the Euclidean norm.

  \item We consider each mobile robot as a vertex of an undirected graph. We assume that the graph is connected, i.e. there exists an infinite sequence of contiguous, non-empty, bounded, time intervals $[k_{i},k_{i+1}]$  where $i=0, 1, 2,\ldots,n$ with $k_{0}=0$; such that for all $[k_{i},k_{i+1}]$ the graph from the union of the collection $G(kT)\in{\cal P}$ for all $kT\in [k_{i},k_{i+1}]$, where $i=0, 1, 2,\ldots,n$ is connected.
\end{enumerate}

\section{Control Strategy}

We define a control algorithm, where each mobile robot calculates its discrete time heading angle based on its own value and the values obtained from the neighbours. In case of a cyclic behaviour, the mobile robot adapts itself to a self-biased or neighbour biased strategy for the calculation of its quantized heading angle.

Let $M$ be the number of steps per revolution of a mobile robot. Then,

\begin{equation}
 Q:=\dfrac{2\pi}{M}
\end{equation}

We define a discrete set of heading angles as under:

\begin{equation}\label{eq:Vicsek_Revisited_Theta_discrete}
 \Theta :=\{0,Q,2Q,...,(M-1)Q\}
\end{equation}
In fact, we consider that each mobile robot has the physical constraint (e.g. a stepper motor with limited number of steps per revolution) to select an orientation from the above mentioned set of quantized heading angles.

Let $n$ be number of mobile robots in the team. Next, we define a local variable for each mobile robot to select the self-biased or neighbour biased strategy during a cyclic behaviour.

\begin{equation}
 {\Delta}_{i}(kT):=([\theta_{i}(kT)]-\dfrac{1}{|{\cal N}_{i}(kT)|}\sum_{j\in {\cal N}_{i}(kT)}[\theta_{j}(kT)])/Q
\end{equation}
for $i=1,2,3,\ldots, n$, where ${\cal N}_i(kT)$ is the set of neighbours of mobile robotic sensor $i$ and $|{\cal N}_i(kT)|$ denotes the number of its neighbours at a particular time $kT$.

Normally ($|{\Delta}_{i}(kT)|>1$), the discrete time heading angle is calculated as under:\\
$[{\Theta}_{i}(kT)]:=$\\
\begin{equation}
 Q\lfloor(\dfrac{1}{(1+|{\cal N}_{i}(kT)|)}([\theta_{i}(kT)]+\sum_{j\in {\cal N}_{i}(kT)}[\theta_{j}(kT)]))/Q\rfloor\\
\end{equation}

In case of cyclic behaviour, our biased algorithm switches the heading angle calculation as under:\\

IF $\text{} {\Delta}_{i}(kT) \neq 0 \text{ and } -1\leq {\Delta}_{i}(kT) \leq 1$ THEN\\
\begin{equation}\label{Heading_Floor}
[{\Theta}_{i}(kT)]:= Q\lfloor(\dfrac{1}{|{\cal N}_{i}(kT)|}\sum_{j\in {\cal N}_{i}(kT)}[\theta_{j}(kT)])/Q\rfloor\\
\end{equation}
OR\\
IF $\text{}{\Delta}_{i}(kT) \neq 0 \text{ and } -1 \leq {\Delta}_{i}(kT)\leq 1$ THEN\\
%$\text{}{\Delta}_{i}(kT)\neq \dfrac{-1}{|{\cal N}_{i}(kT)|} \text{ AND } |{\cal N}_{i}(kT)| \in {{2z+1; \forall z\in \mathbb{Z}}}\\$ THEN
\begin{equation}\label{Heading_Ceil}
[{\Theta}_{i}(kT)]:= Q\lceil(\dfrac{1}{|{\cal N}_{i}(kT)|}\sum_{j\in {\cal N}_{i}(kT)}[\theta_{j}(kT)])/Q\rceil\\
\end{equation}

OTHERWISE
\begin{equation}\label{Self_Biased}
[{\Theta}_{i}(kT)]:=   [\theta_{i}(kT)]
\end{equation}

The choice of Eq. (\ref{Heading_Floor}) or Eq. (\ref{Heading_Ceil}) can be made depending on the available data. If a mobile robot follows equation Eq. (\ref{Heading_Floor}) or Eq. (\ref{Heading_Ceil}), we call it as adapting to a neighbour biased strategy. Otherwise, Eq. (\ref{Self_Biased}) defines self-biased strategy for a mobile robot. One can notice that introduction of the combination of self-biased and neighbour biased strategy helps in breaking the mentioned cyclic behaviour. Next, a mobile robot updates its heading angle as under:

\begin{equation}\label{eq:phi_update}
 [\theta_{i}(k+1)T)]:= [\Theta_{i}(kT)]
\end{equation}

We present the main result (Theorem 3.2) of \cite{savkin2008decentralised} as under:

\subsection{Theorem}

Consider $n$ autonomous mobile robots governed by dynamics, Eq. (\ref{eq:Revisit_sys1}-\ref{eq:Revisit_sys2}), meeting the assumptions (\ref{Assumptions}) are supposed to reach a consensus in terms of quantized heading angle. Then, there exists a constant $c>0$ such that for any $k \leq c$ there exists a heading $\bar{\theta} \in \Theta $ and an integer $T_{h}>0$ such that

\begin{equation}\label{eq:Theor_update}
 [\theta_{i}(kT)]:=\bar{\theta} \text{ }\forall\text{ }T \geq T_{h}
\end{equation}
for all $i=1,2,3,\ldots, n$.

\textbf{Proof:}
If we make two assumptions:  i.e., a necessary assumption that the graph stays connected and the heading upper bound satisfies, say $u<\pi$, then the main result from \cite{savkin2008decentralised} proves the below mentioned discrete time consensus on the heading angle:

\begin{equation}\label{eq:Theor_update}
 [\theta_{i}(kT)]:=\bar{\theta} \text{ }\forall\text{ }T \geq T_{h}
\end{equation} for all $i=1,2,3,\ldots, n$.

The second assumption limits the mobile robot to an upper bound ($u<\pi$) and this assumption has been proved crucial in \cite{savkin2008decentralised}. It has been proved that the multi-robotic system can go under cyclic behaviour without this assumption even though the mobile robots stay connected. It has also been mentioned that the multi-robots can oscillate in a bounded region of the plane over infinite time. Obviously, such a bounded oscillation is due to a difference in quantization level. Let $k \in \mathbb{Z}$ be an integer defining a quantization level. Then, $k=0,1,2,\dots,(M-1)$ and the bounded cyclic oscillation can be in between: say $kQ$ and $(k-1)Q$. In order to eliminate this difference in the quantization level, we simply define a biased strategy. Thus, a discrete time consensus can be achieved by simply adapting to a biased strategy in the algorithm.

\section{Simulation}

We run these simulations on MATLAB R2012b. We represent the mobile robotic sensor with a green circle and its heading angle with a green arrow. We have used ceil or floor functions to bring the calculated heading angle to a quantized level (Q). A rounding function (upto 4 decimal places) has been used for comparative statements. In these simulation examples, we have used the following parameters:\\
$R_{c}=2.0, M=20, v=0.8$.

\begin{figure}[h]
\begin{center}
   \includegraphics[width=1.0\columnwidth]{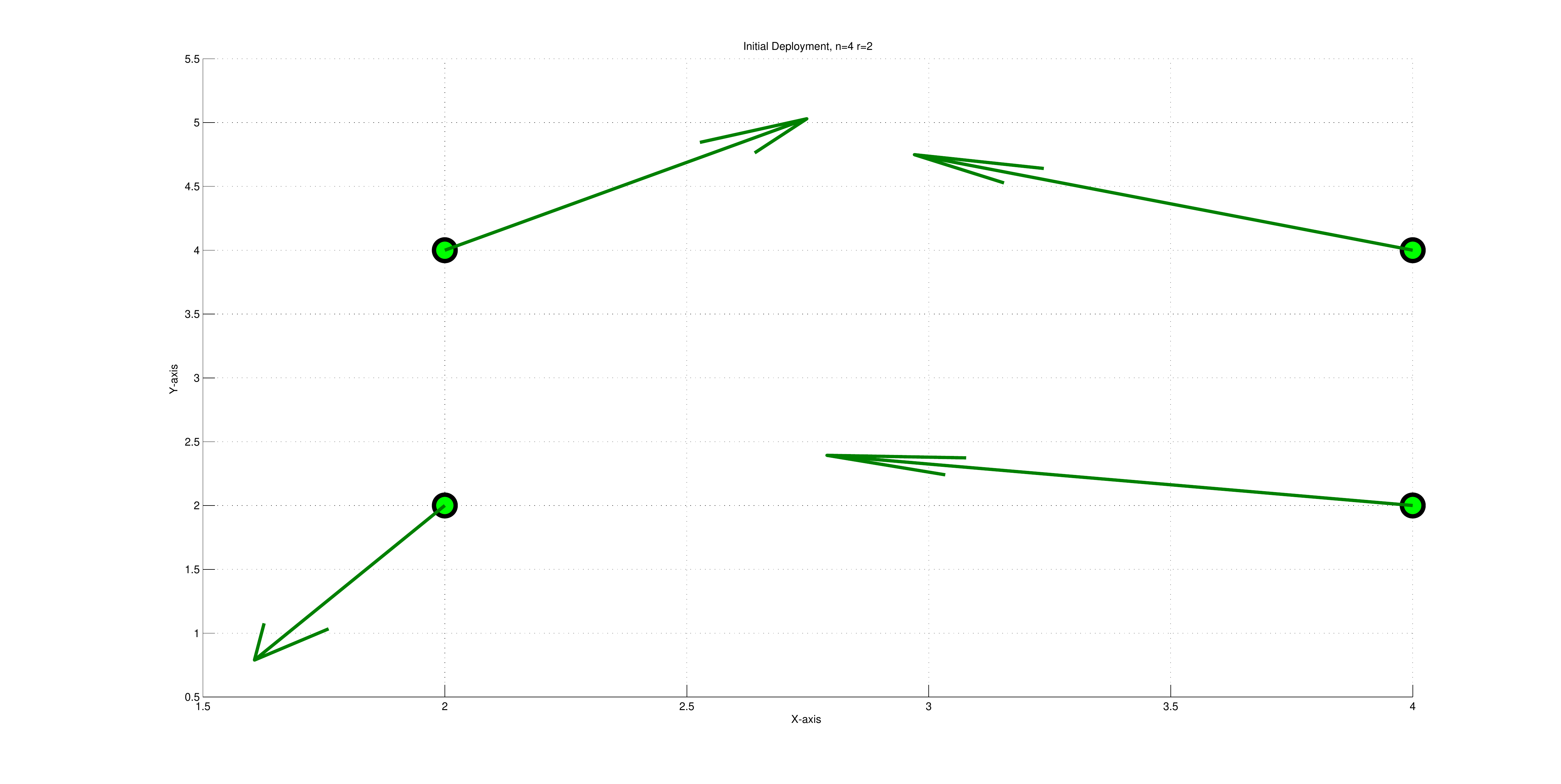}
  % \label{fig:theFig}
    \caption{Mobile Robots on the Vertices of Regular Polygon - Initial Random Headings}\label{Polygon_Initial_Random}
\end{center}
\end{figure}

\begin{figure}[h]
\begin{center}
   \includegraphics[width=1.0\columnwidth]{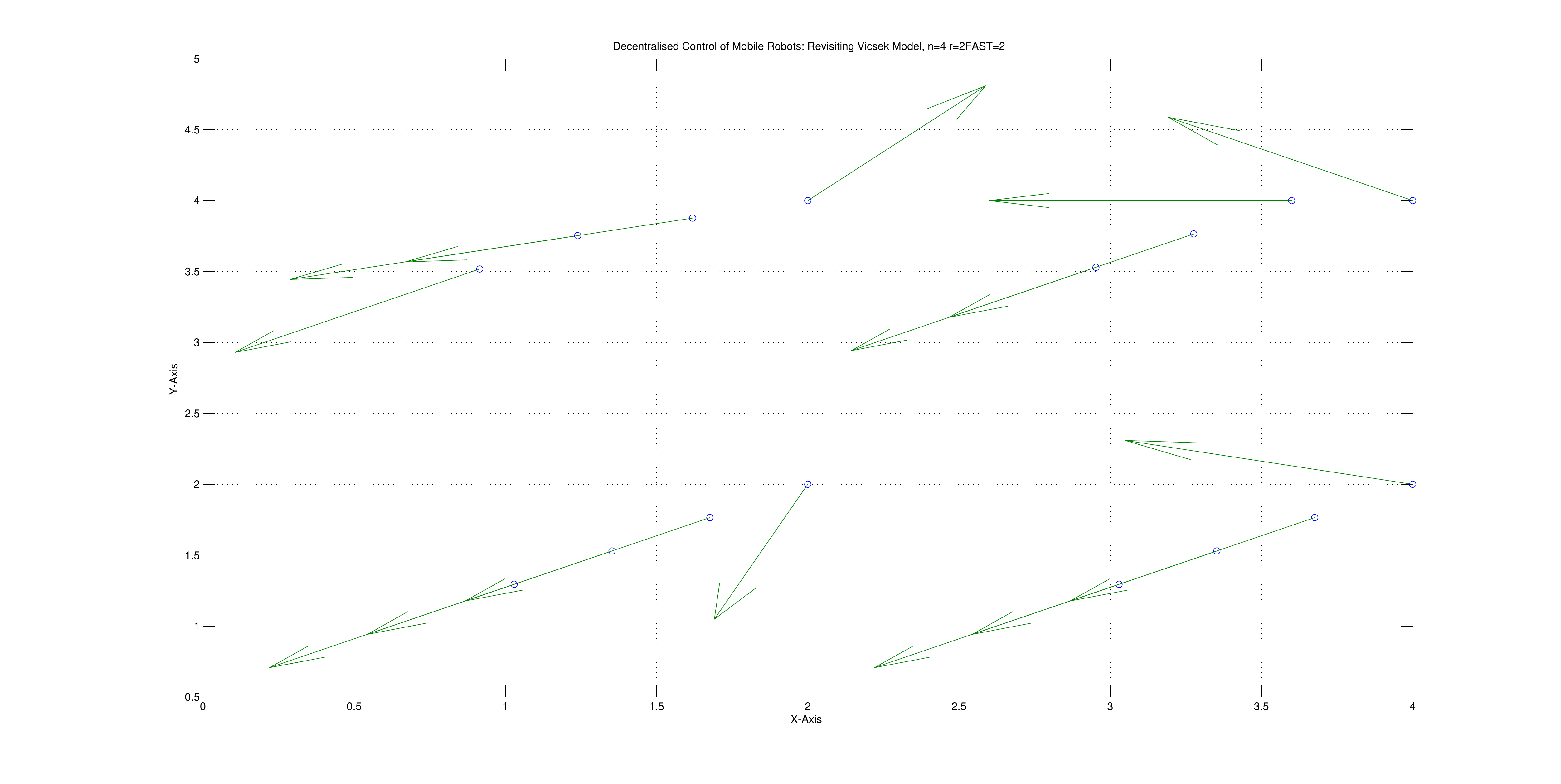}
  % \label{fig:theFig}
    \caption{Mobile Robots on the Vertices of Regular Polygon - Mobile Robots Trajectory}\label{Polygon_Robots_Trajectory}
\end{center}
\end{figure}

\begin{figure}[h]
\begin{center}
   \includegraphics[width=1.0\columnwidth]{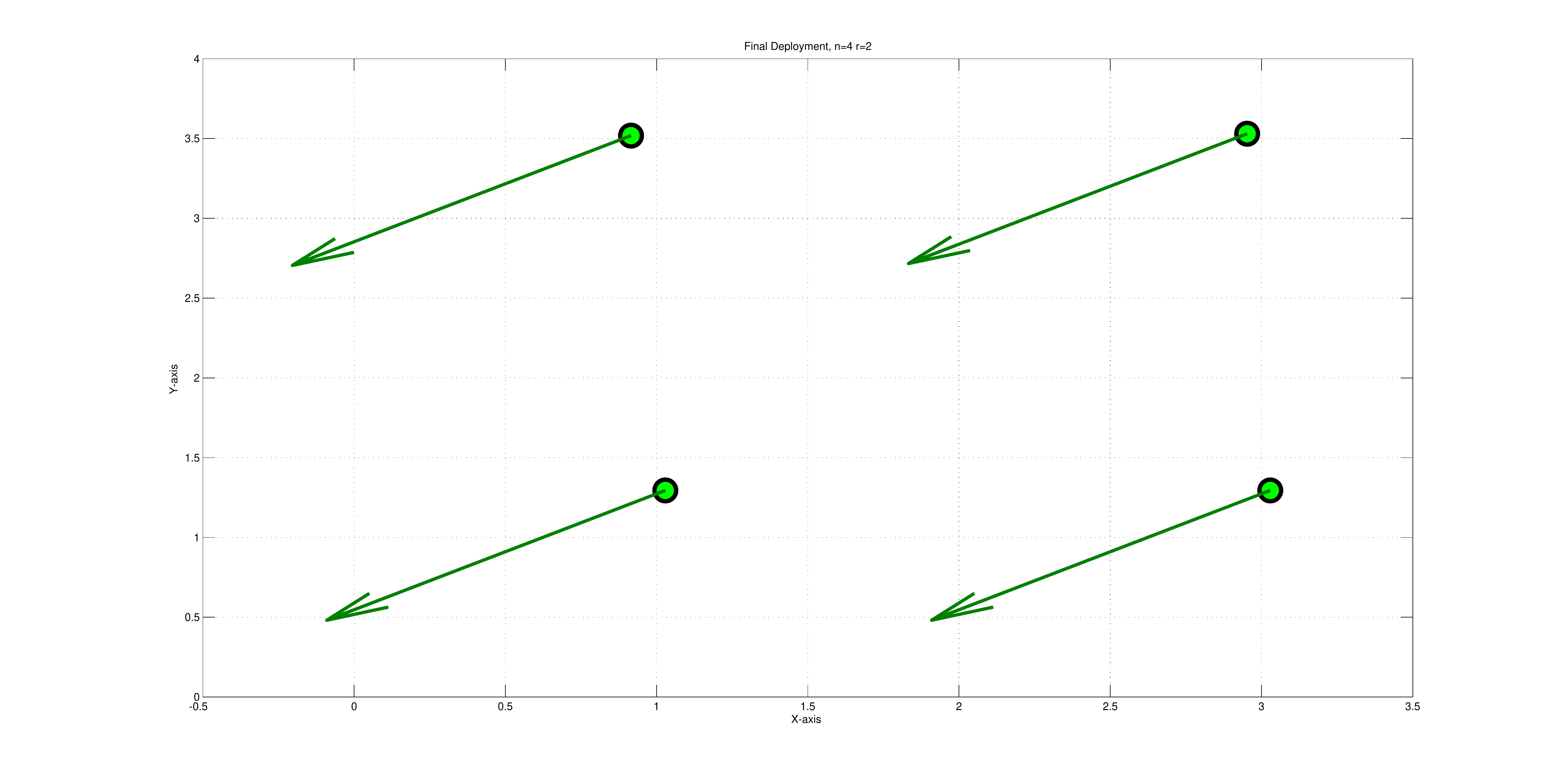}
  % \label{fig:theFig}
     \caption{Mobile Robots on the Vertices of Regular Polygon - Mobile Robots Final Heading}\label{Polygon_Robots_Final}
\end{center}
\end{figure}

In Fig. \ref{Polygon_Initial_Random}, four mobile robots with random initial heading angles are placed on the vertices of a regular polygon. The concept of regular polygon has been used as described in the example of cyclic behaviour \cite{savkin2008decentralised}. We can see in Fig. \ref{Polygon_Robots_Trajectory}-\ref{Polygon_Robots_Final} that each mobile robot eventually reaches a constant quantized heading angle. In Fig. \ref{Polygon_Robots_Trajectory}, we show the trajectory of a mobile robot to be followed with dynamics (\ref{eq:Revisit_sys1}-\ref{eq:Revisit_sys2}).

\begin{figure}[h]
\begin{center}
   \includegraphics[width=1.0\columnwidth]{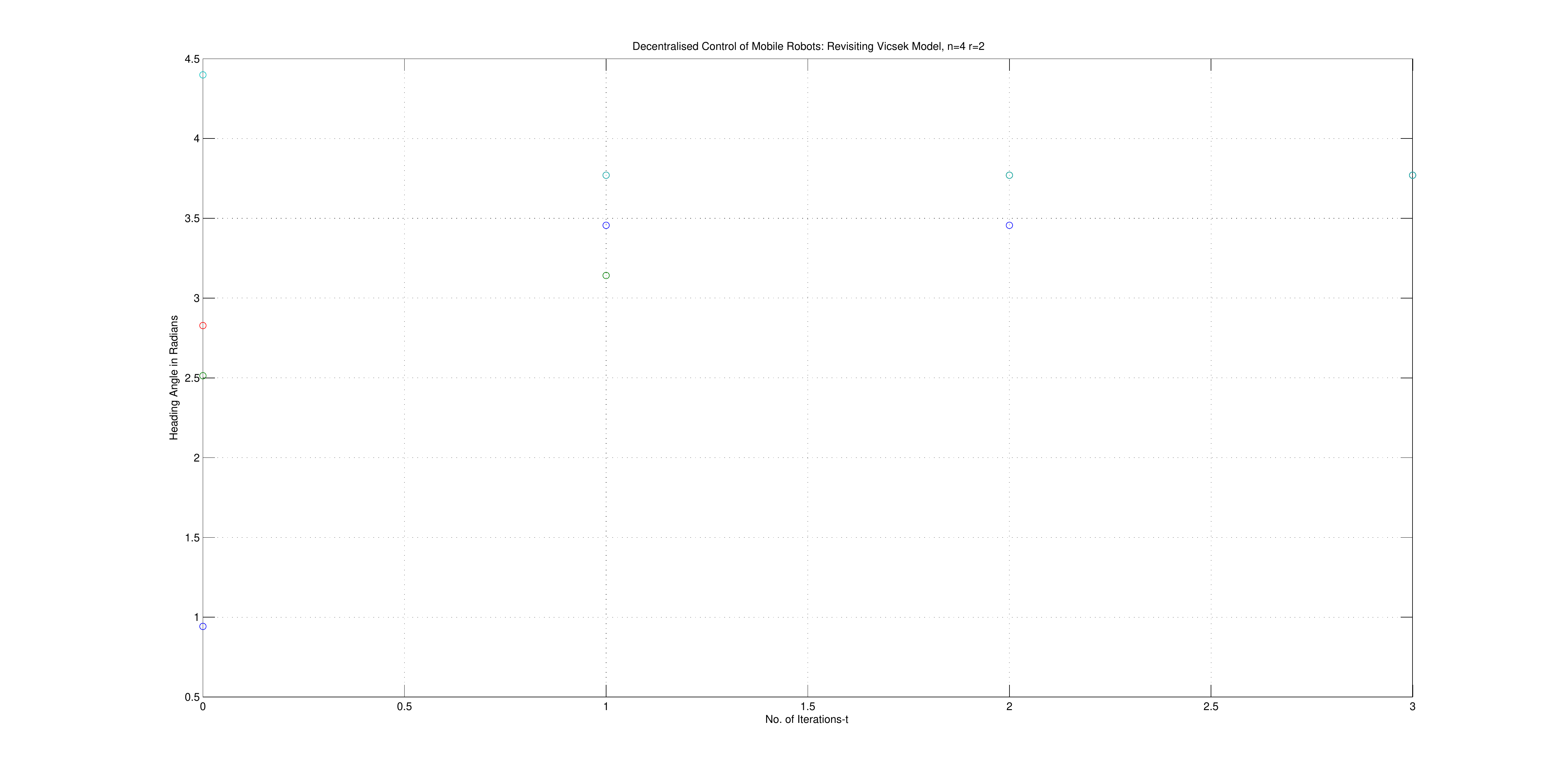}
  % \label{fig:theFig}
     \caption{Mobile Robots on the Vertices of Regular Polygon - Heading Angles Convergence}\label{Polygon_Heading_Convergence}
\end{center}
\end{figure}

\begin{figure}[h]
\begin{center}
   \includegraphics[width=1.0\columnwidth]{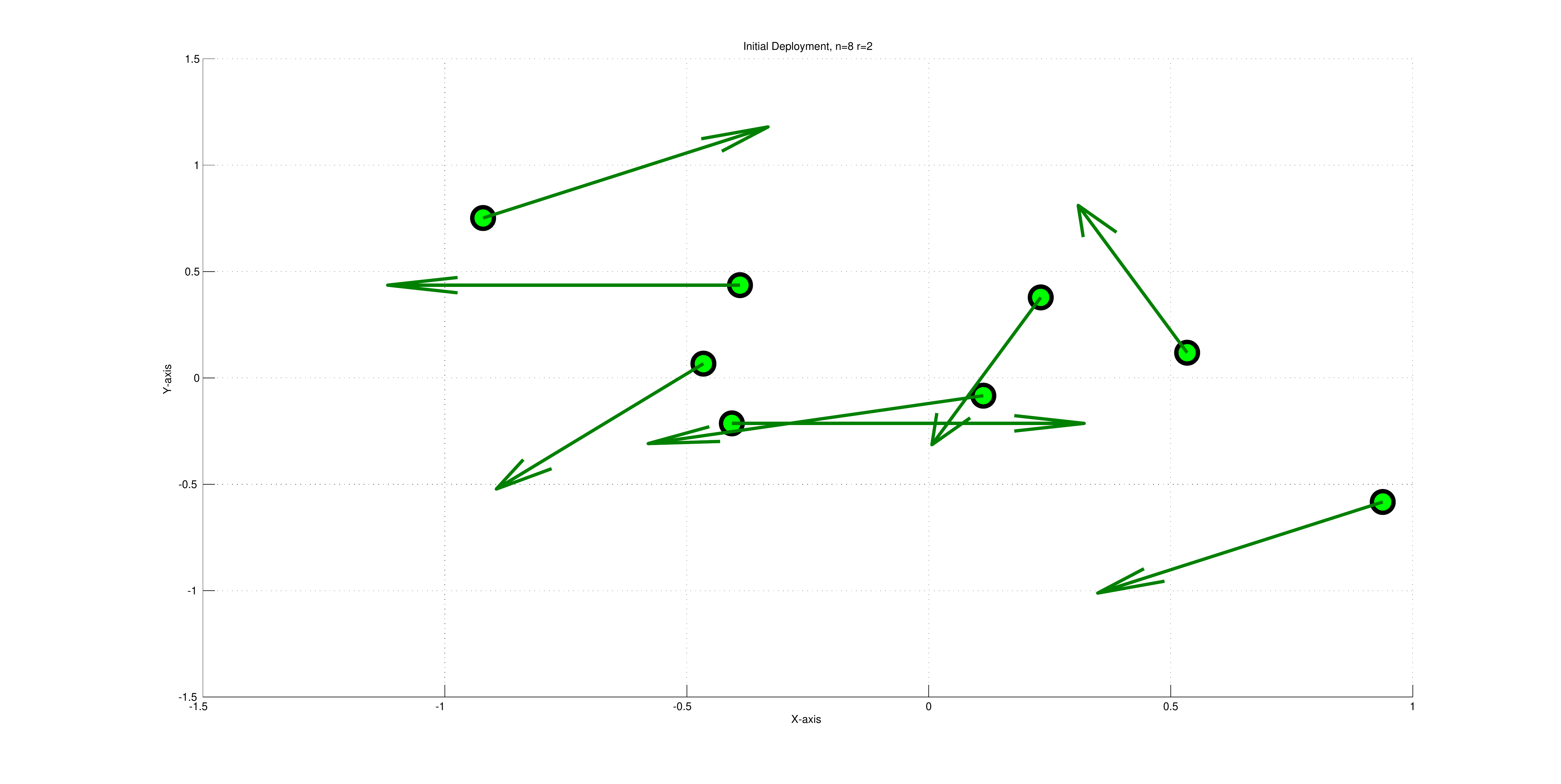}
  % \label{fig:theFig}
    \caption{Randomly Placed Mobile Robots - Initial Random Headings}\label{Random_Initial_Random}
\end{center}
\end{figure}

\begin{figure}[h]
\begin{center}
   \includegraphics[width=1.0\columnwidth]{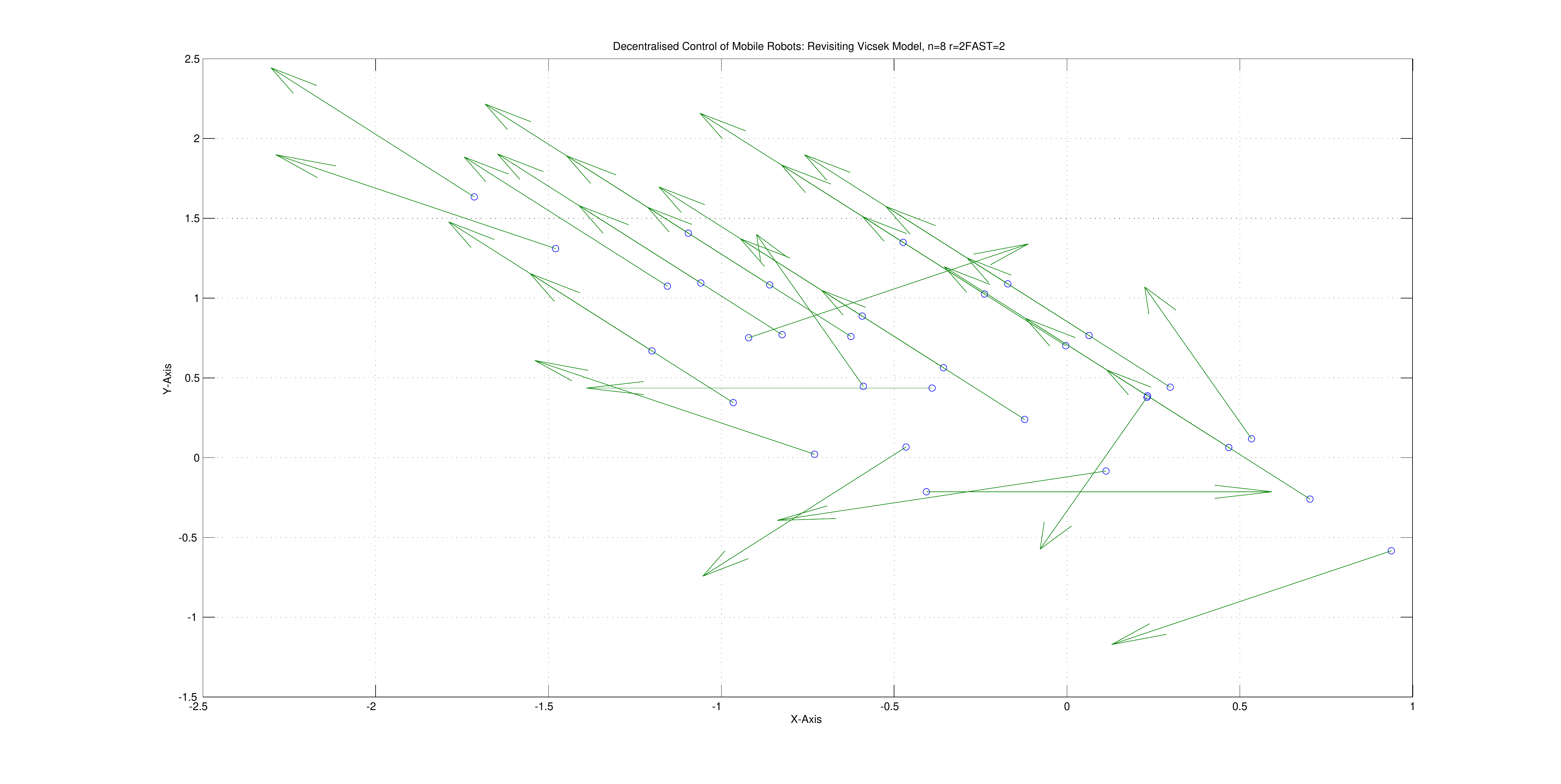}
  % \label{fig:theFig}
   \caption{Randomly Placed Mobile Robots - Mobile Robots Trajectory}\label{Random_Robots_Trajectory}
\end{center}
\end{figure}

\begin{figure}[h]
\begin{center}
   \includegraphics[width=1.0\columnwidth]{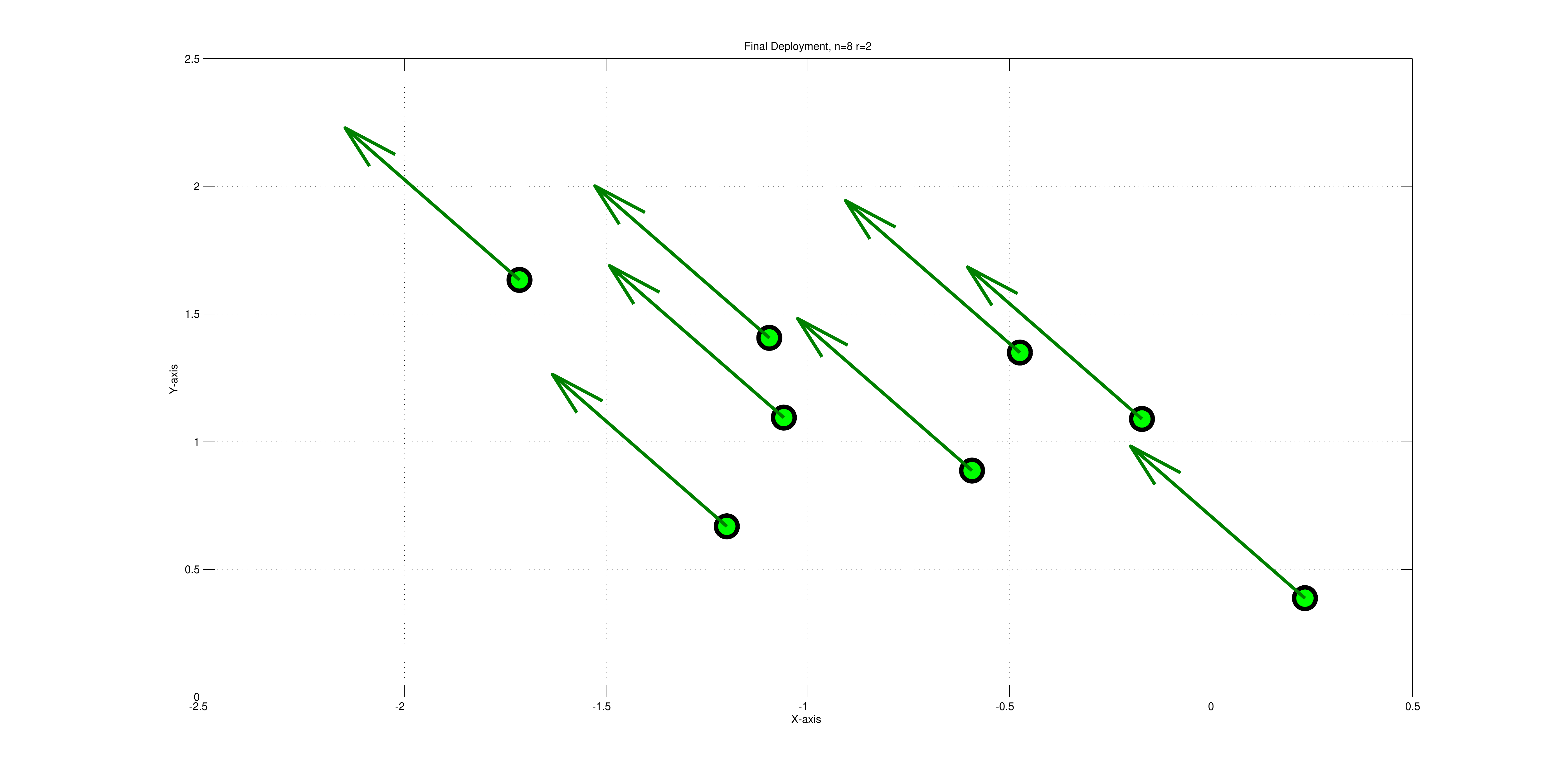}
  % \label{fig:theFig}
    \caption{Randomly Placed Mobile Robots - Mobile Robots Final Heading}\label{Random_Robots_Final}
\end{center}
\end{figure}

\begin{figure}[h]
\begin{center}
   \includegraphics[width=1.0\columnwidth]{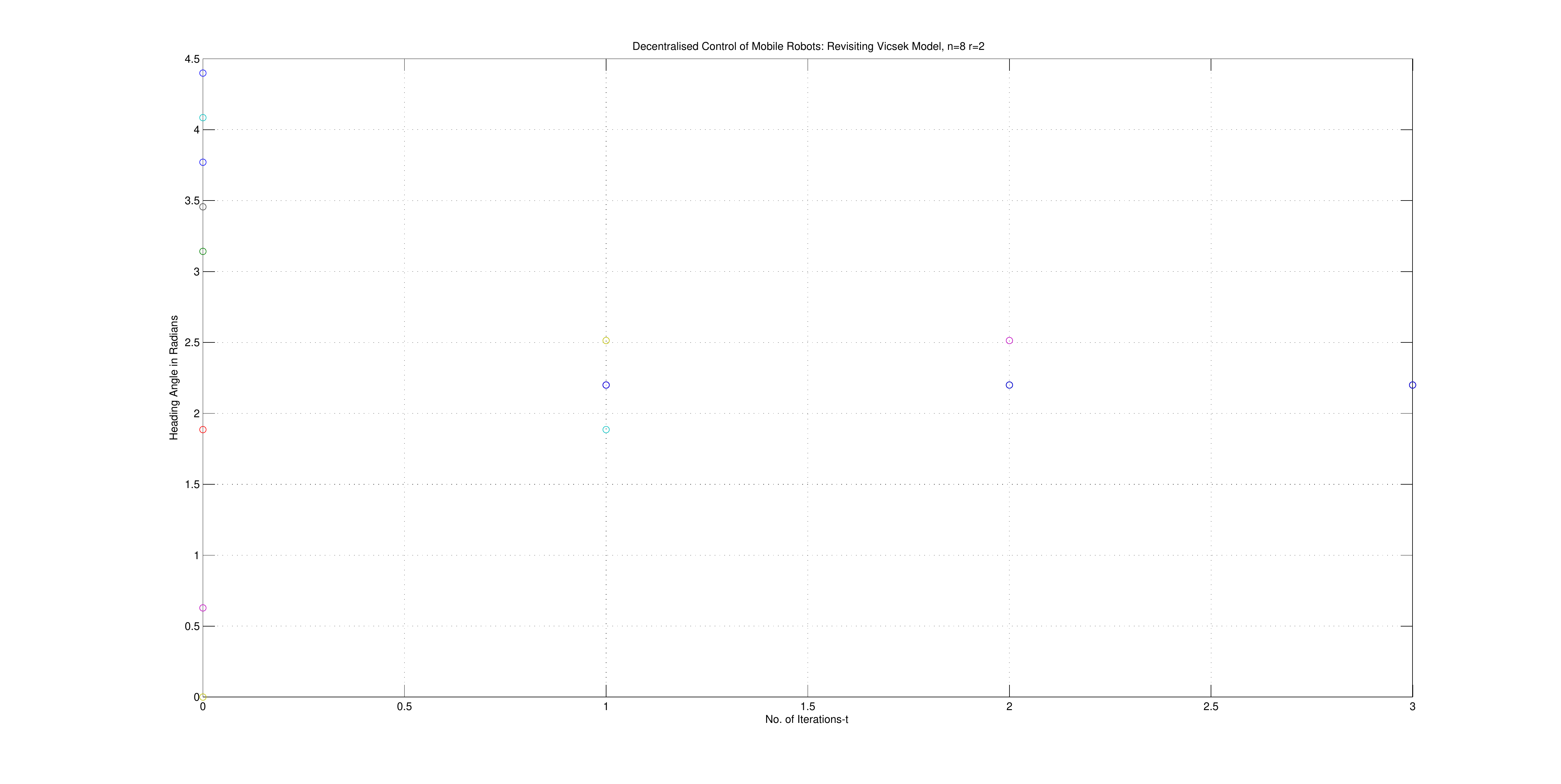}
  % \label{fig:theFig}
    \caption{Randomly Placed Mobile Robots - Heading Angles Convergence}\label{Random_Heading_Convergence}
\end{center}
\end{figure}

In Fig. \ref{Random_Initial_Random}, eight mobile robots with random initial heading angles are deployed. We can notice in Fig. \ref{Random_Robots_Final}-\ref{Random_Heading_Convergence} that each mobile robot converges to a constant quantized heading angle selected from the defined discrete set. In Fig. \ref{Random_Robots_Trajectory}, we show the trajectory of each mobile robot to be followed with dynamics (\ref{eq:Revisit_sys1}-\ref{eq:Revisit_sys2}).

\section{Chapter Conclusions and Future Research Directions}

\subsection{Chapter Conclusions}

The numerical simulations have shown that a cyclic behaviour of discrete time Vicsek's model \cite{vicsek1995novel} can be avoided by using an alternate self-biased and neighbour biased strategy. Our heading update algorithm takes into consideration its real time implementation on the mobile robots. It has been shown that the control strategy provides an alternate for a cyclic behaviour \cite{savkin2008decentralised} to be caused in Vicsek's model \cite{vicsek1995novel} and the result of constant heading angle is extended to any quadrant (not limited to any of two adjacent quadrants as mentioned in \cite{savkin2008decentralised}). Furthermore, the mobile robots reach a constant heading angle within a finite set of discrete time heading angle-which is a realistic consideration for the team of mobile robots to reach a quantized heading angle.\\
In the control algorithm, the main contribution of this chapter is incorporation of a neighbour biased strategy with mathematical operators: $\lfloor.\rfloor$ or $\lceil.\rceil$. However, only mathematical operator $\lfloor.\rfloor$ is used in \cite{savkin2008decentralised} and the operation of each mobile robot with quantized heading angle is limited to a half plane (Assumption 3.1 of \cite{savkin2008decentralised}). But our self-biased and neighbour biased strategy by incorporating a choice of $\lfloor.\rfloor$ or $\lceil.\rceil$ in the algorithm extends the operation of each mobile robot to four quadrants (instead of limiting it to half plane or to two adjacent quadrants only). Hence, our control strategy relaxes assumption 3.1 of \cite{savkin2008decentralised}, which is placing an upper bound on the quantized heading angle of a mobile robot.

\subsection{Future Research Directions}

The control law can be revised to incorporate the collision avoidance technique from a stationary or moving obstacle. Fig. \ref{Polygon_Heading_Convergence} and Fig. \ref{Random_Heading_Convergence} shows how each mobile robot converge its heading angle to a constant value belonging to the defined discrete set. However, a study can be performed to estimate the data lost due to the quantization error. We can find a proven optimal way of linear iterations to maximize the convergence speed for this problem. The minimized number of linear iterations has utmost importance in these sorts of decentralized control algorithms.

%\section*{Acknowledgement}
%This work has been supported in part by the Australian Research Council (ARC) with an allocated grant (DP130103898). The main objective of this grant is to conceptually develop new design rules for the control of mobile networked systems.

\chapter{Nearest Neighbour Rule with a Similarity Measure}
In the previous chapter, we addressed the physical constraint during implementation of nearest neighbour rule based approach. This chapter focuses on the importance of a similarity measure to be considered among the nearest neighbours of a mobile robot. We test our decentralized control algorithm with cosine similarity measure. We validate our work with mathematical arguments and computer simulations.

\section{Introduction}
We investigate the control of a team of randomly heading mobile robots to reach asymptotic consensus for almost sure. In Chapter 4, we saw that Vicsek's model is based on the state of an agent and its neighbours and the inspiration from Vicsek's model leads to a locally computed nearest neighbour rule. In the nearest neighbour rule, a mobile robotic sensor always assigns an equal weighting factor to itself and every nearest neighbour coming under its communication range. We consider Salton's cosine similarity measure (see e.g. \cite{salton1989automatic,salton1991developments}) among the nearest neighbours of a mobile robotic sensor and the nearest neighbours are not always equally weighted. Consequently, some of the nearest neighbour mobile robots become influential and play an effective role in reaching a consensus value. After introducing the similarity measure among the nearest neighbours, we have compared the obtained simulation results with that of simple nearest neighbour rule. We conclude that the algorithm becomes fast after including the similarity measure among the nearest neighbours of a mobile robot.

\section{Related Work}

In Chapter 2, we surveyed that a number of mobile robots cooperating in a team\footnote{The terms, "agent" or "sensor" or "the mobile robotic sensor" or simply, "the mobile robot" will be used throughout this chapter for an autonomous mobile robot having an on-board computation, operation-specific sensing and communication capability.} with limited local information has got different applications like surveillance of an area, reconnaissance, maintenance job and inspection in hazardous areas \cite{gage1995many,gage1992command}. It is normally required that the team of mobile robots could reach a constant heading angle in a decentralized fashion. In Chapter 1, we introduced that this type of decentralized control with local information is inspired from animal aggregation. The mathematical analysis of Vicsek's model \cite{vicsek1995novel} already demonstrates how the decentralized coordination of flock of birds can reach the same direction and we showed the necessary references in Chapter 4. In the previous Chapters, we can remind that an agent updates its state based on the average of its own plus its neighbours \cite{vicsek1995novel,jadbabaie2003coordination} and a number of researchers \cite{cheng2013decentralized,teimoori2010biologically,cheng2009distributed,cheng2011decentralized,savkin2010decentralized,cheng2011decentralizedsweep,bullo2009distributed,savkin2012optimal} have developed decentralized control laws based on Vicsek's model \cite{vicsek1995novel} to update the heading angle of each mobile robot connected as part of a team.

%In real time applications, a team of mobile robots to reach the same heading angle needs a discrete time model addressing the physical constraints. Some researchers \cite{jadbabaie2003coordination,savkin2008decentralised,tahbaz2008necessary,savkin2004coordinated,jadbabaie2003distributed,tahbaz2007recurrence,li2004multi} have shown that all agents reach the same heading provided existence of a mutual linkage. In research work \cite{jadbabaie2003coordination}, a rigorous qualitative analysis of Vicsek's model shows how each mobile robot as part of a team eventually reaches a constant heading angle. A remark regarding counter-intuitive consequences has been given for the use of the averaging rule mentioned in \cite{savkin2008decentralised}. In fact, all the literature \cite{cheng2013decentralized,teimoori2010biologically,cheng2009distributed,cheng2011decentralized,savkin2010decentralized,cheng2011decentralizedsweep,bullo2009distributed,savkin2012optimal} considers the averaging rule mentioned in \cite{vicsek1995novel,jadbabaie2003coordination}.

In this chapter, we extend the concept of simple averaging rule \cite{jadbabaie2003coordination} to a weighted averaging nearest neighbour rule considering a similarity measure among the neighbours of a connected team of mobile robots. The weight factors are calculated based on the similarity measure among the neighbours of a mobile robot. Our work consider the similarity measure specifically for a team of mobile robots, which is different from the Jaccard similarity measure as considered for the consensus problem of \cite{avrachenkov2011local}. The communication problem between robots is not considered in this work. Practically, the necessary information about neighbouring mobile robots can be obtained by the use of Kalman state estimation via limited communication channels (see e.g. \cite{malyavej2005problem,savkin2003set,matveev2005comments,savkin2006analysis}).

In the upcoming sections, we formulate the objective of this chapter along with necessary assumptions to be considered. Then, we write basic steps of our decentralized control algorithm. We also provide the mathematical arguments proving how asymptotic state consensus among the team of mobile robots is reached for almost sure. We present some simulation examples showing how number of linear iterations is reduced if each mobile robot in the team considers the similarity measure in its algorithm. At the end, we give conclusion stating the achievement of this chapter along with some future research directions.

\section{Problem Statement}

Let a mobile robotic sensor $i$ has $x_{i}(\cdot)$ and $y_{i}(\cdot)$ as the Cartesian coordinates (say $s_{i}(\cdot) \in \mathbb{R}^2$), $v$ be its constant speed and $\theta_{i}(\cdot)\in \mathbb{R}$ denotes its heading angle measured counter-clockwise from the $X$-axis.

\begin{equation}\label{eq:sim_sys1}
 x_{i}((k+1)T)= x_{i}(kT) + Tv\cos(\theta_{i}(kT))\\
\end{equation}
\begin{equation}\label{eq:sim_sys2}
 y_{i}((k+1)T)= y_{i}(kT) + Tv\sin(\theta_{i}(kT))\\
\end{equation} where $i=1, 2, \ldots,n$ is number of mobile robots and $k=0, 1, 2, \ldots$ is the discrete time-index.

Our calculated control input in the above mentioned system is the heading angle, $\theta_{i}$.

The main objective is to develop a decentralized control strategy achieving asymptotic state consensus for almost sure. However, we consider the similarity measure among the nearest neighbours of a mobile robot to achieve this objective.

\subsection{Assumptions}\label{sim_Assumptions}

\begin{enumerate}
  \item Each mobile robot is equipped with on-board computation capability.
  \item Each mobile robot has a wireless communication to neighbour(s) within a disk of radius $R_{c}>0$ for $t\in[kT, (k+1)T)$ and $k=0, 1, 2, \ldots$. We mathematically define disk of communication range $R_{c}$ as under:

\begin{equation}
C_{i,R_{c}}(kT):=p\in R^2:||p - s_{i}(kT)|| \leq R_{c}
\end{equation}, where $\|\cdot\|$ denotes the Euclidean norm.
  \item We consider each mobile robot as a vertex of an undirected graph. So, we assume that the graph is connected.
\end{enumerate}

\section{Control Strategy}

The average heading rule is defined after following Vicsek's model \cite{vicsek1995novel}.

\begin{equation}\label{eq:vicsek_model}
\Theta_{i}(kT):=\arctan\left(\dfrac{\sin(\phi_i(kT))+\sum_ {j\in {\cal F}_{i}(kT)}\sin(\phi_{j}(kT))}{\cos(\phi_i(kT))+\sum_ {j\in {\cal F}_{i}(kT)}\cos(\phi_{j}(kT))}\right)
\end{equation}

for $i=1,2,3,\ldots, n$, where ${\cal F}_i(kT)$ denotes the set of neighbours for a mobile robotic sensor $i$ and $\phi_i(kT)$ its heading angle at a particular time $kT$.

The heading angle model (\ref{eq:vicsek_model}) takes values in the continuous set $[0, 2\pi),$ and it is defined as the average of velocity vectors. However, there are some conditions for which the rule (\ref{eq:vicsek_model}) can exhibit a cyclic dynamics under quantized heading angle and it has been clearly pointed out in \cite{savkin2008decentralised}.

Further, the following averaging rule of \cite{jadbabaie2003coordination} is defined as the average of headings and it makes the analysis simple by applying the properties of stochastic matrices.

\begin{equation}\label{eq:sim_jadbabie}
 {\Theta}_{i}(kT) :=\dfrac{1}{|1+{\cal F}_{i}(kT)|}\left(\phi_i(kT)+\sum_ {j\in {\cal F}_{i}(kT)}\phi_{j}(kT)\right)
\end{equation}

In Chapter 4, it has been remarked \cite{jadbabaie2003coordination} that the use of rule (\ref{eq:sim_jadbabie}) can exhibit counter-intuitive consequences under some circumstances. The remark \cite{jadbabaie2003coordination} also states that the counter-intuitive consequences can be avoided by considering the domain of the defined function within the interval $[0, \pi)$.

We consider the following system inspired from weighted average of agent's own state plus its neighbours.

\begin{equation}\label{eq:sim_weighted_avg}
 {\Theta}_{i}(kT) :=w_{ii}\phi_i(kT)+\sum_ {j\in {\cal F}_{i}(kT)}w_{ij}\phi_{j}(kT)
\end{equation}
for $i=1, 2, 3, \ldots, n$. We define the weighting factor, $w_{ii}$, for a mobile robotic sensor itself and a weighting factor, $w_{ij}$, for its neighbour(s) at a particular time $(kT)$ such that the following condition is satisfied:
\begin{equation}
 w_{ii}+\sum_ {j\in {\cal F}_{i}(kT)}w_{ij}=1
\end{equation}

We can note that the rule (\ref{eq:sim_weighted_avg}) becomes exactly equivalent to (\ref{eq:sim_jadbabie}) under the following condition:

\begin{equation}\label{eq:jad_weighted_avg_eq}
 w_{ii}=w_{ij}=\dfrac{1}{|1+{\cal F}_{i}(kT)|}
\end{equation}

Our control strategy is based on incorporating the similarity measure to calculate the weights ($w_{ii},w_{ij}$). Hence, we represent the undirected graph of neighbour mobile robotic sensors in the form of an adjacency matrix, $A$.

\begin{equation}\label{eq:adjacency_matrix}
A := [a_{ij}]
\end{equation}

If $ith, jth$ robots are neighbours then $ith, jth$ entry ($a_{ij}$) of $A$ is represented as 1, otherwise, it is considered as 0. Formally,

\begin{equation}\label{eq:elements_adjacency}
\begin{split}
a_{ij}= \begin{cases}
 1 \text{ if } j\in {\cal F}_{i}(kT) \\
 0 \text{ otherwise }\\
\end{cases}
\end{split}
\end{equation}

Let $a_{i}, a_{j}$ be the vectors representing $ith,jth$ row (or column) of adjacency matrix $A$. Then, we define Salton's cosine similarity measure between these vectors (rows or columns of $A$) \cite{salton1986introduction}.

\begin{equation}\label{eq:cos_sim}
 \cos (v_{ij})= \dfrac{<a_{i}|a_{j}>}{||a_{i}||.||a_{j}||}
\end{equation}

A mobile robot assigns the weighting factor to a neighbour mobile robot based on the dissimilarity between two defined vectors ($a_{i}$ and $a_{j}$). The fundamental statement is that a mobile robot assigns a big weighting factor to a neighbour mobile robot having higher dissimilarity (or lower similarity) measure and a small weighting factor is assigned to a neighbour mobile robot with lower dissimilarity (or higher similarity.)

\begin{equation}\label{eq:neighbour_weight}
 w_{ij}= 1-\cos (v_{ij})= 1-\dfrac{<a_{i}|a_{j}>}{||a_{i}||.||a_{j}||}
\end{equation}

Then, we normalise all the calculated $ w_{ij}$ for ${j\in {\cal F}_{i}(kT)}$.

IF $\text{} \sum_{j\in {\cal F}_{i}(kT)}w_{ij} > 1$ THEN\\
\begin{equation}\label{eq:norm_edge_weight}
w_{ij}=\dfrac{w_{ij}}{\sum_ {j\in {\cal F}_{i}(kT)}w_{ij}}\\
\end{equation}

A mobile robot calculates its self-weight as under:

\begin{equation}\label{eq:self_weight}
 w_{ii}= 1- \sum_{j\in {\cal F}_{i}(kT)}w_{ij}
\end{equation}

Next, a mobile robot updates its heading angle from (\ref{eq:sim_weighted_avg}).

\begin{equation}\label{eq:sim_phi_update}
 \theta_{i}((k+1)T):= \Theta_{i}(kT)
\end{equation}

The vector form of the system (\ref{eq:sim_weighted_avg}) and (\ref{eq:sim_phi_update}) can be represented as under:

\begin{equation}\label{eq:sim_weighted_avg_vector}
  \theta((k+1)T) := W_{k}\phi(kT)
\end{equation}

where $k=0, 1, 2,\ldots$ is the discrete time index and $\theta((k+1)T)\in \mathbb{R}^n$ is the state vector at time $(k+1)T$.

For the discrete time linear dynamical system (\ref{eq:sim_weighted_avg_vector}), the authors in \cite{tahbaz2008necessary} assigns a necessary and sufficient condition by using ergodicity and probabilistic arguments. In the below mentioned theorem, we provide the necessary and sufficient condition similar to \cite{tahbaz2008necessary}.

\subsection{Theorem}

Consider $n$ autonomous mobile robots governed by the dynamics (\ref{eq:sim_sys1}-\ref{eq:sim_sys2}), meeting the assumptions (\ref{sim_Assumptions}) and following the rule of (\ref{eq:sim_weighted_avg_vector}). Then, there exists a heading $\bar{\theta} \in \mathbb{R} $ such that, for all $i=1,2,3,\ldots, n$,

\begin{equation}\label{eq:Theor_update}
 \lim_{k\to \infty }\theta_{i}(kT) \xrightarrow{a.s.} \bar{\theta}
\end{equation}
 (i.e. asymptotic consensus to $\bar{\theta}$ is reached for almost sure) if and only if $\rho (W_{k})$ has exactly one eigen value with unit modulus.

\textbf{Proof:}\\

 We provide the detailed mathematical arguments to prove how our algorithm achieves the consensus.\\
Let $a_{i}, a_{j}$ be the respective vectors representing $ith,jth$ row (or column) of adjacency matrix $A$. As stated in (\ref{eq:cos_sim}), we define cosine similarity measure between these vectors as under:

\begin{equation}
 \cos (v_{ij})= \dfrac{<a_{i}|a_{j}>}{||a_{i}||.||a_{j}||}
\end{equation}

In extreme conditions of this similarity measure, one can notice that $\cos (v_{ij}) \in [0,1]$.
Hence, ($\ref{eq:norm_edge_weight}$) and ($\ref{eq:self_weight}$) is also bounded, i.e. $w_{ii}, w_{ij}\in [0,1]$. This implies,

\begin{equation}\label{Wk_Non_Negative}
W_{k} \geq 0 \text{ $\forall$ $i,j$ $w_{ij}\geq 0$ }
\end{equation}

So, our calculated $W_{k}$ is always non-negative. Further, the sum of weight vector ($[w_{ii}\text{ }w_{ij}]$) is normalized to 1 at each time index $k$. Therefore, the following condition also holds:

\begin{equation}\label{eq:W_stoch_Cond}
 W_{k}\mathbf{1}= \mathbf{1}
\end{equation}
where $\mathbf{1}$ is an $n$ dimensional column vector with all coefficients as 1. We have proved so far:

\begin{enumerate}
  \item (Eq. \ref{Wk_Non_Negative}) $\Rightarrow$ $W_{k}$ is a square matrix with non-negative real entries.
  \item (Eq. \ref{eq:W_stoch_Cond}) $\Rightarrow$ $W_{k}$ is a stochastic matrix.
  \item (Assumption \ref{sim_Assumptions}-3) $\Rightarrow$ $W_{k}$ is an irreducible square matrix.
\end{enumerate}

 As long as assumption (\ref{sim_Assumptions}-3) holds, each mobile robot is connected and we assume that our $G(W_{k})$ is aperiodic. Hence, the associated weight matrix, $W_{k}$, is considered irreducible and aperiodic stochastic matrix. Thus, the condition (\ref{eq:W_stoch_Cond}) on the basis of extended version of Perron-Forbenius theorem ( see e.g. \cite{meyer2000matrix,berman1979nonnegative}) for irreducible and aperiodic stochastic matrices states that:

\begin{equation}\label{eq:W_Spect_Radius}
 \rho(W_{k})= 1
\end{equation} and any other eigen value of $W_{k}$ is strictly smaller than its spectral radius ($\rho(W_{k})$). The converse of the above mentioned result has been proved with a different approach in Chapter 6. So, the product of our irreducible and aperiodic stochastic matrices is asymptotically converged to a rank one matrix for almost sure. A similar result has been stated in ( see e.g. \cite{touri2014product}).\\
Remark: A cyclic behaviour can exist for a connected graph of mobile robots. For example, a weight matrix for two mobile robots is calculated as under:

\[ \scriptscriptstyle
W_{k}  =  \left[
        \begin{array}{cc}
         0&1  \\
         1&0  \\
        \end{array}
      \right]
\]

We can see that this weight matrix is irreducible and it has only two eigen values +1 and -1. So, the absolute of second largest eigen value is not less than 1 and the algorithm can exhibit a cyclic behaviour. Such a periodicity can be avoided by ensuring at least one mobile robot has strictly positive self-weight , i.e. $W_{k}$ should have at least one strictly positive diagonal entry. So, a strongly connected team of mobile robots with at least one strictly positive self-weighting mobile robot will ensure that our connected $W_{k}$ is always irreducible and aperiodic (acyclic).

\section{Simulation}

\begin{figure}[H]
  \centering
  \includegraphics[width=1.0\columnwidth]{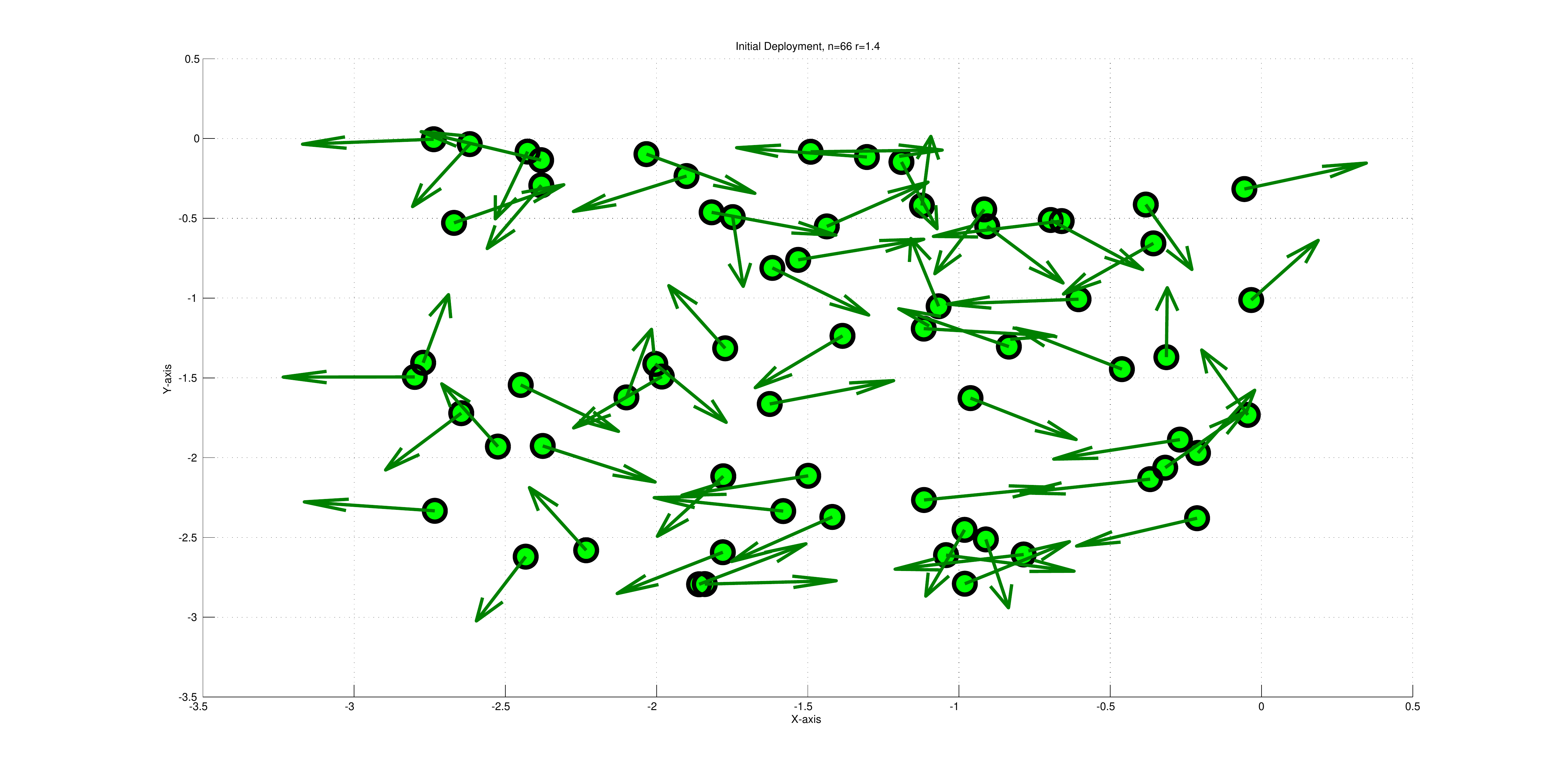}
  \caption{Simple Averaging Function - Initial Random Deployment}\label{SAF_Sim_Initial}
\end{figure}

\begin{figure}[H]
  \centering
  \includegraphics[width=1.0\columnwidth]{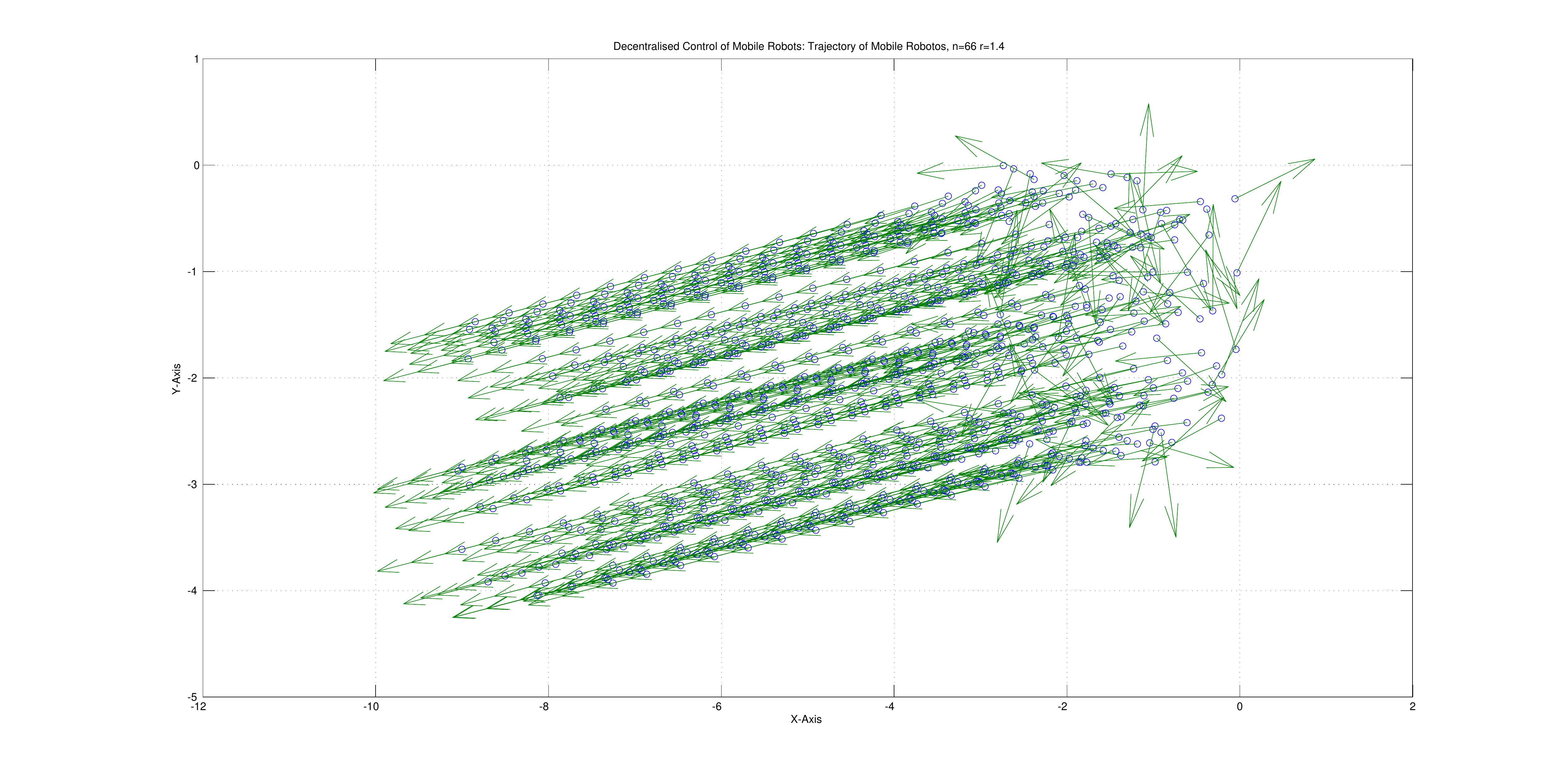}
  \caption{Simple Averaging Function - Mobile Robots Trajectories}\label{SAF_Sim_Trajectories}
\end{figure}

\begin{figure}[H]
  \centering
  \includegraphics[width=1.0\columnwidth]{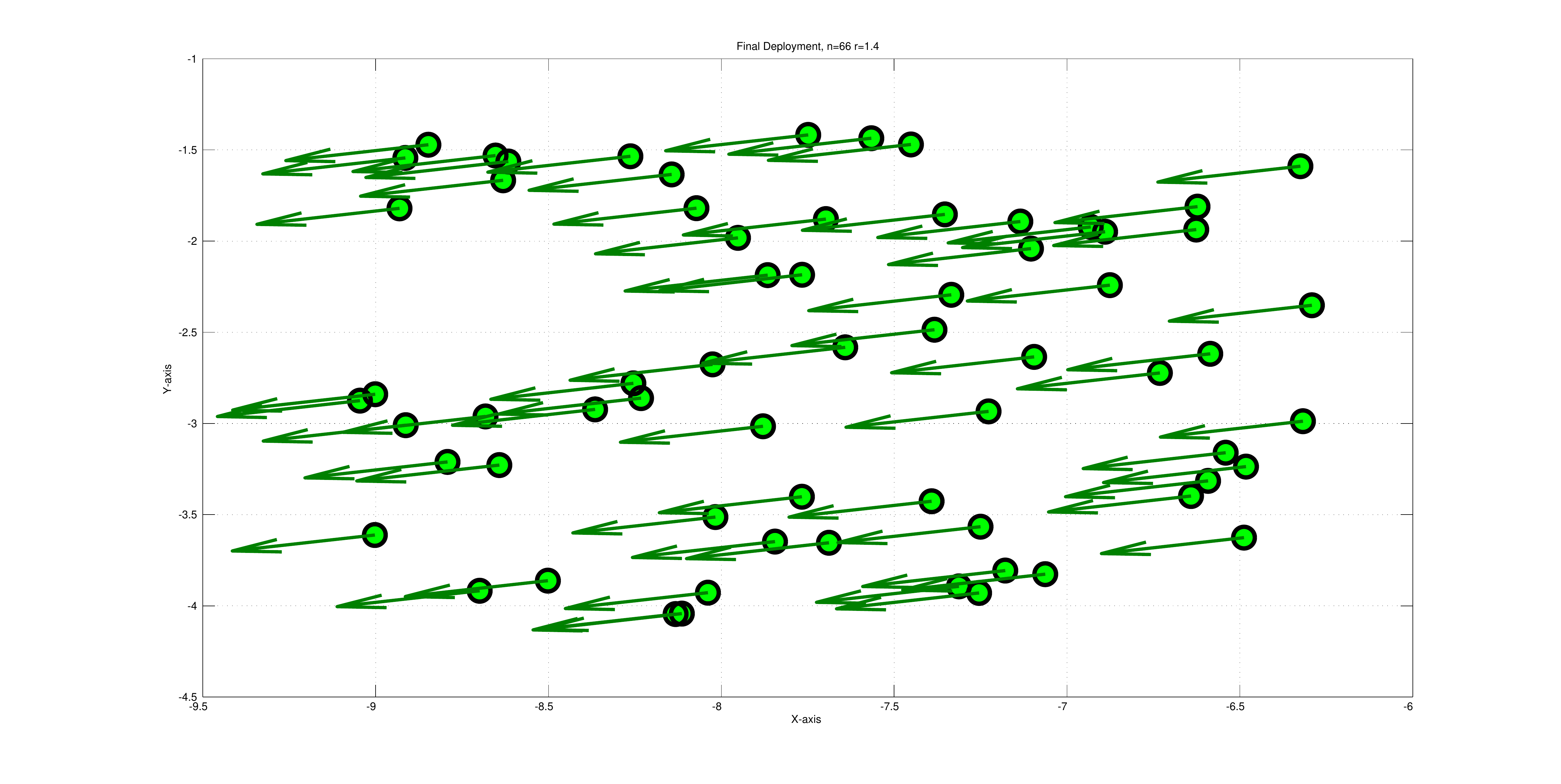}
  \caption{Simple Averaging Function - Final Heading}\label{SAF_Sim_Final}
\end{figure}

\begin{figure}[H]
  \centering
  \includegraphics[width=1.0\columnwidth]{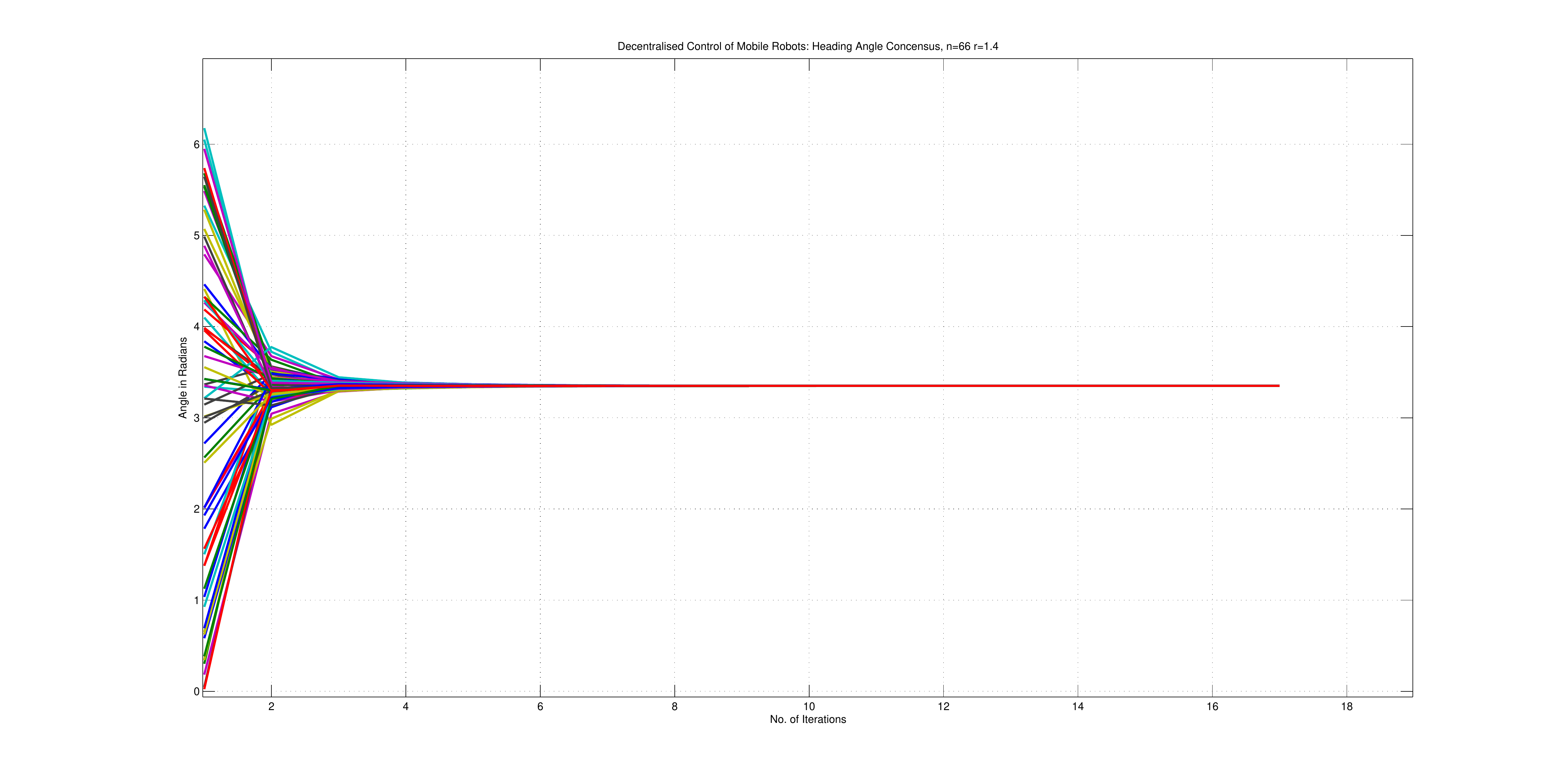}
  \caption{Simple Averaging Function - Heading Angle Convergence}\label{SAF_Sim_Convergence}
\end{figure}

We run these simulations on MATLAB R2012b. The mobile robotic sensor is represented with a green circle and its heading angle with a green arrow. We use a rounding function (up to 4 decimal places) with the following simulation parameters:\\
$R_{c}=1.4, v=0.002, T=0.5$.\\
\textbf{Example 1}:

Fig. \ref{SAF_Sim_Initial}-\ref{SAF_Sim_Convergence} shows a simulation study with the simple nearest neighbour rule (\ref{eq:sim_jadbabie}). In Fig. \ref{SAF_Sim_Initial}, we deploy 66 mobile robots with random initial heading angles and positions. After 16 linear iterations (as shown in Fig. \ref{SAF_Sim_Final}-\ref{SAF_Sim_Convergence}), each mobile robot reaches asymptotic consensus for almost sure. In Fig. \ref{SAF_Sim_Trajectories}, we show the trajectory of each mobile robot governed by dynamics (\ref{eq:sim_sys1}-\ref{eq:sim_sys2}) for the case of rule (\ref{eq:sim_jadbabie}).

\begin{figure}[H]
  \centering
  \includegraphics[width=1.0\columnwidth]{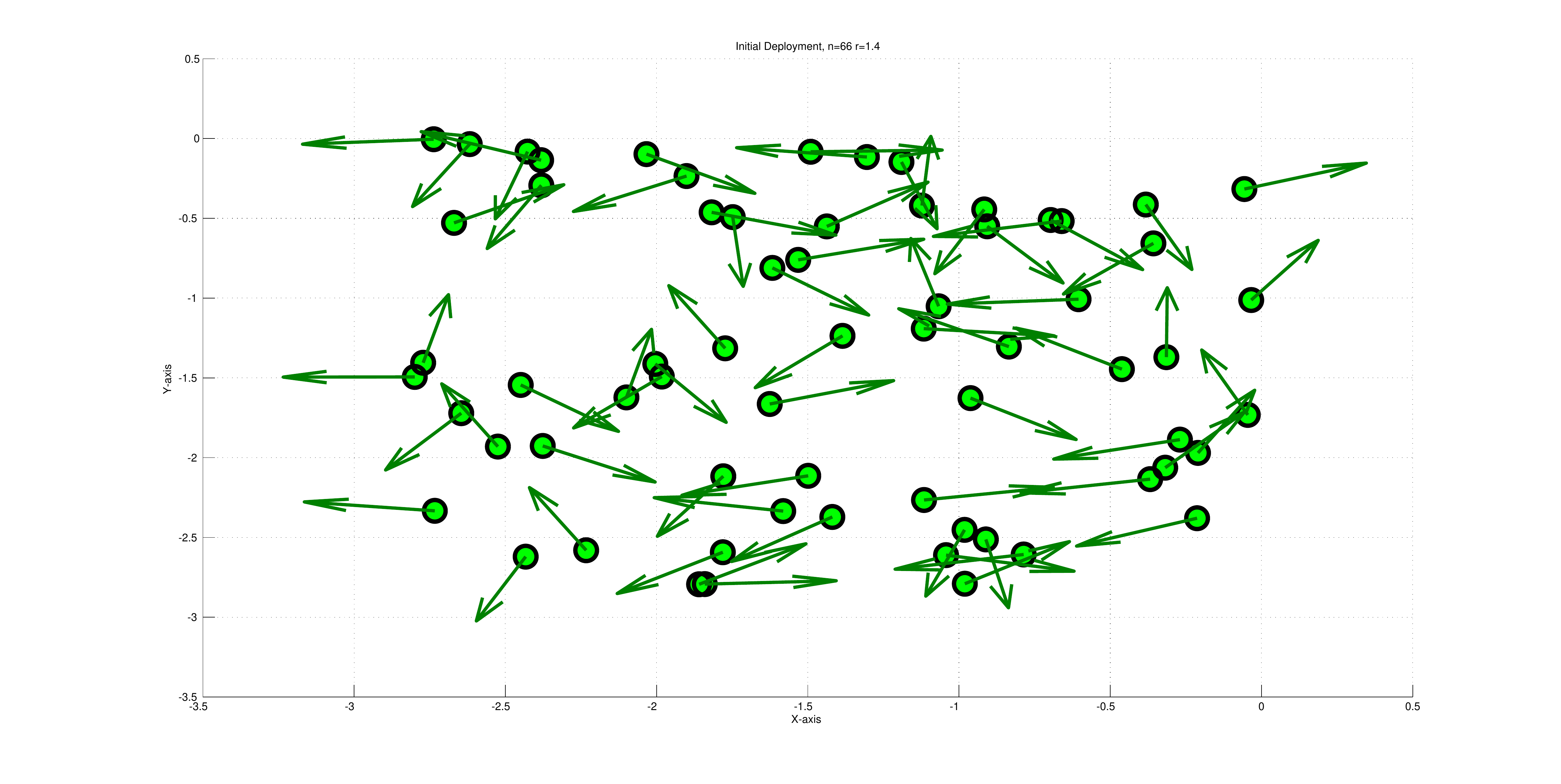}
  \caption{Weighted Averaging Function - Initial Random Deployment}\label{WAF_Sim_Initial}
\end{figure}

\begin{figure}[H]
  \centering
  \includegraphics[width=1.0\columnwidth]{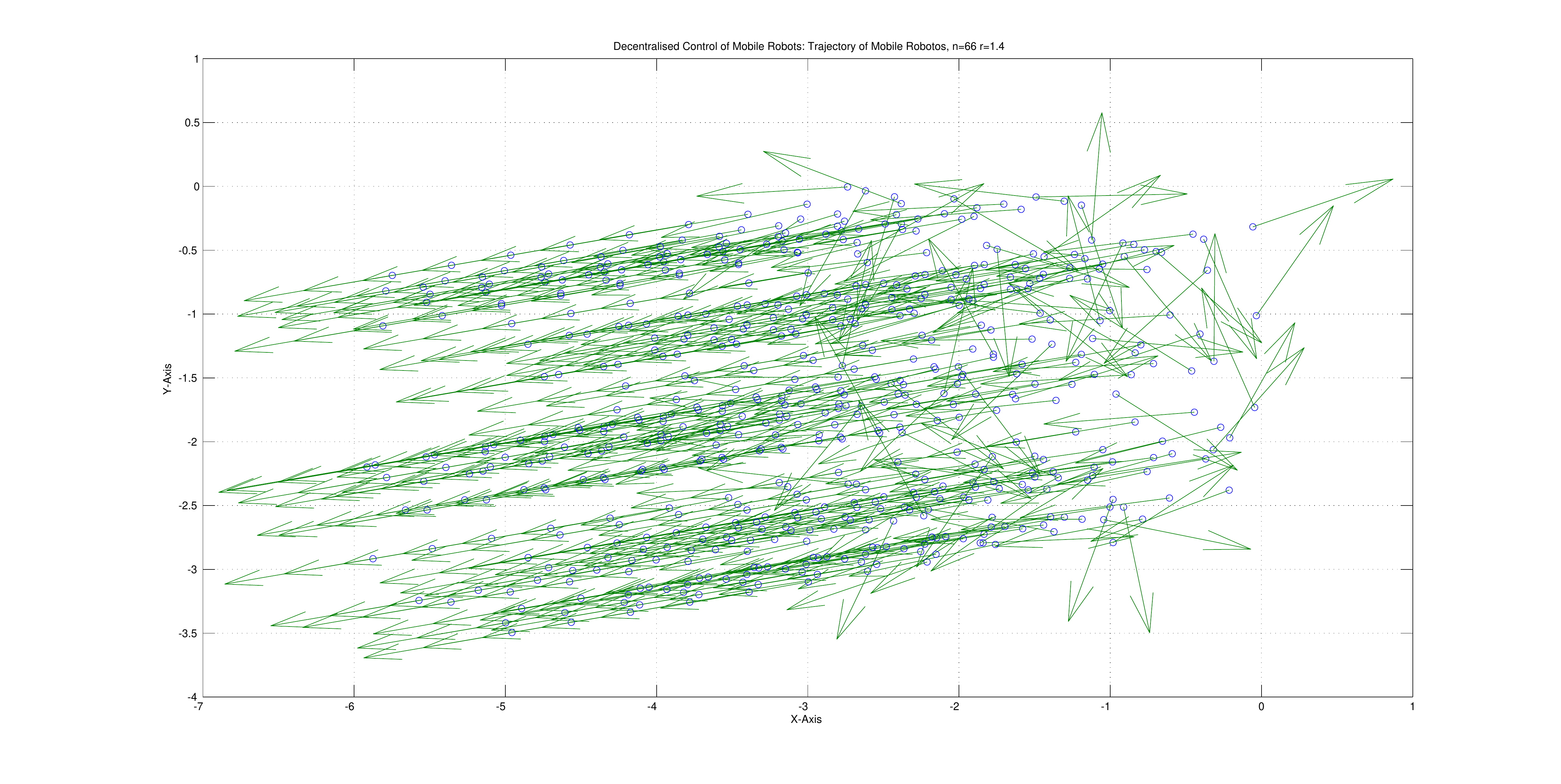}
  \caption{Weighted Averaging Function - Mobile Robots Trajectories}\label{WAF_Sim_Trajectories}
\end{figure}

\begin{figure}[H]
  \centering
  \includegraphics[width=1.0\columnwidth]{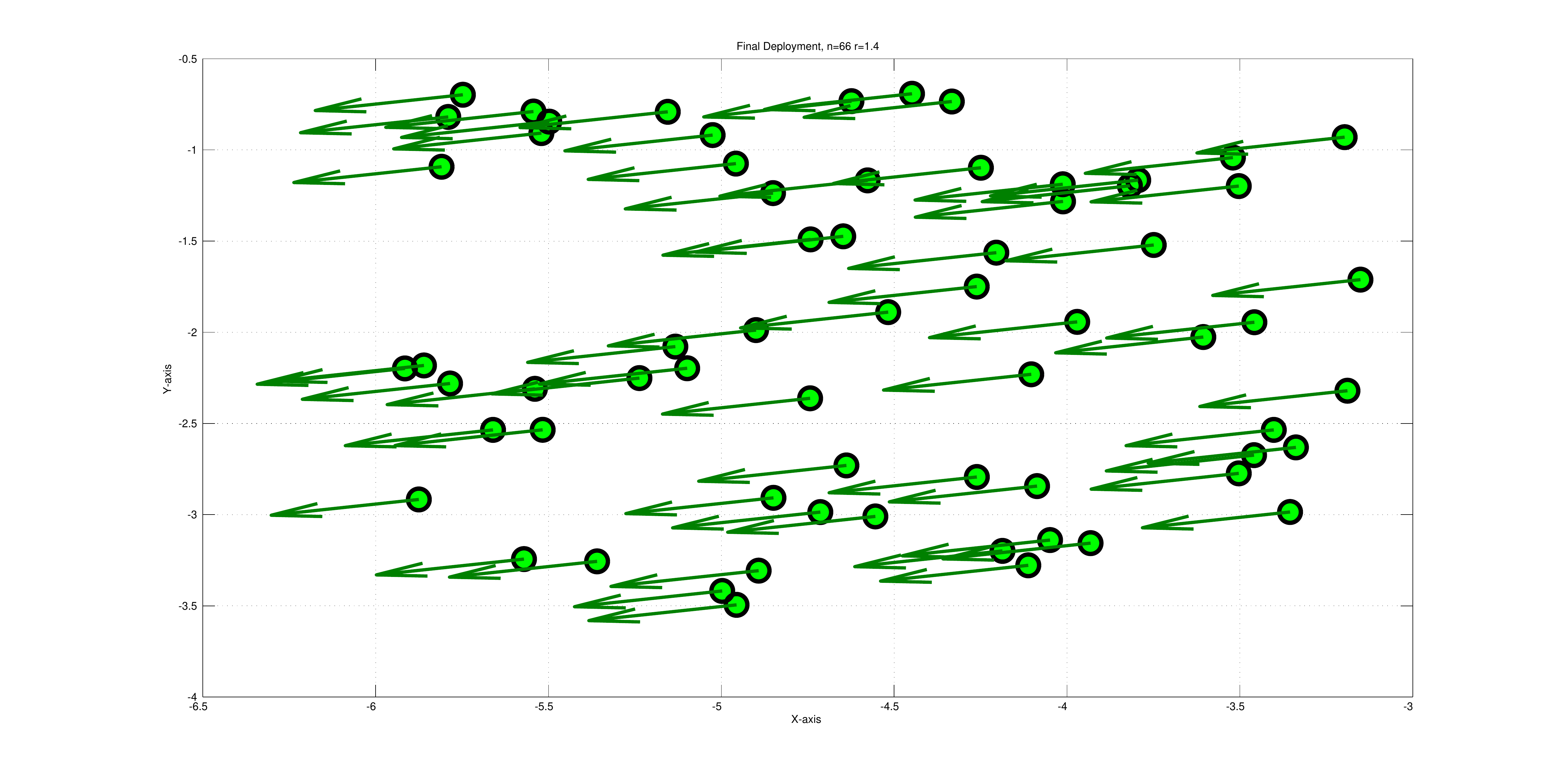}
  \caption{Weighted Averaging Function - Final Heading}\label{WAF_Sim_Final}
\end{figure}

\begin{figure}[H]
  \centering
  \includegraphics[width=1.0\columnwidth]{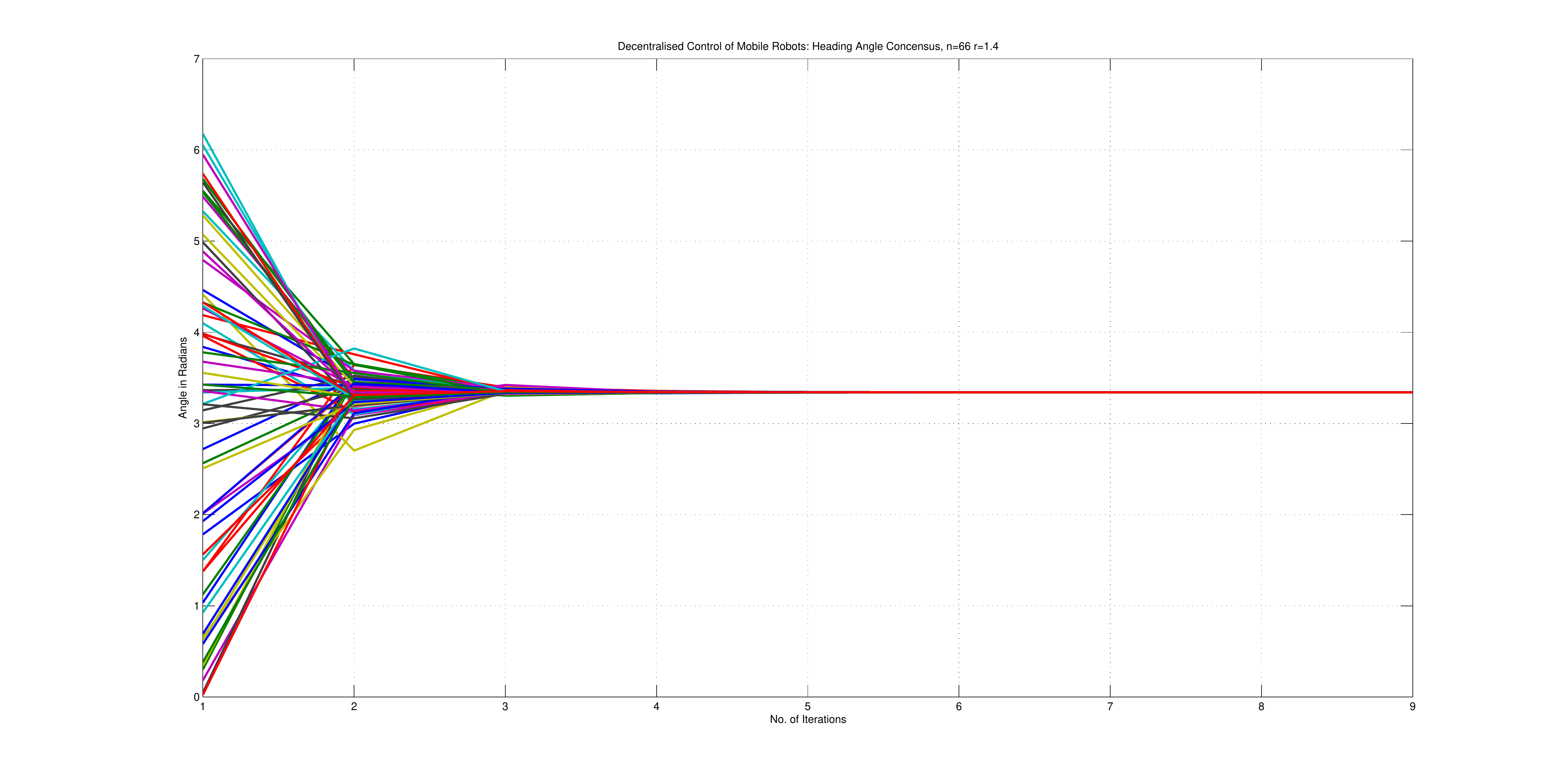}
  \caption{Weighted Averaging Function - Heading Angle Convergence}\label{WAF_Sim_Convergence}
\end{figure}

Fig. \ref{WAF_Sim_Initial}-\ref{WAF_Sim_Convergence} shows a simulation study with the weighted averaging function of nearest neighbour rule (\ref{eq:sim_weighted_avg}). In Fig. \ref{WAF_Sim_Initial}, we deploy 66 mobile robots with random initial heading angles and positions. After 8 linear iterations (as shown in Fig. \ref{WAF_Sim_Final}-\ref{WAF_Sim_Convergence}), each mobile robot reaches asymptotic consensus for almost sure. In case of simple averaging function of nearest neighbour rule ($\ref{eq:sim_jadbabie}$), we notice that all the mobile robots (with the same initial conditions) consume 16 linear iterations. In Fig. \ref{WAF_Sim_Trajectories}, we show the trajectory of each mobile robot governed by dynamics (\ref{eq:sim_sys1}-\ref{eq:sim_sys2}) for the case of rule (\ref{eq:sim_weighted_avg}).

\textbf{Example 2}:
In this example, we consider 12 mobile robots with randomly generated initial vector $\phi(0)$ and connected as per the below mentioned adjacency matrix:

\[ \scriptscriptstyle
A  =  \left[
        \begin{array}{cccccccccccc}
         0&1&1&1&1&1&0&0&0&0&0&0  \\
         1&0&1&1&1&1&0&0&0&0&0&1  \\
         1&1&0&1&1&1&0&0&0&0&0&0  \\
         1&1&1&0&1&1&0&0&0&0&0&0  \\
         1&1&1&1&0&1&0&0&0&0&0&0  \\
         1&1&1&1&1&0&0&0&0&0&0&0  \\
         0&0&0&0&0&0&0&1&1&1&1&1  \\
         0&0&0&0&0&0&1&0&1&1&1&1  \\
         0&0&0&0&0&0&1&1&0&1&1&1  \\
         0&0&0&0&0&0&1&1&1&0&1&1  \\
         0&0&0&0&0&0&1&1&1&1&0&1  \\
         0&1&0&0&0&0&1&1&1&1&1&0  \\
        \end{array}
      \right]
\]
\

\begin{figure}[H]
  \centering
  \includegraphics[width=1.0\columnwidth]{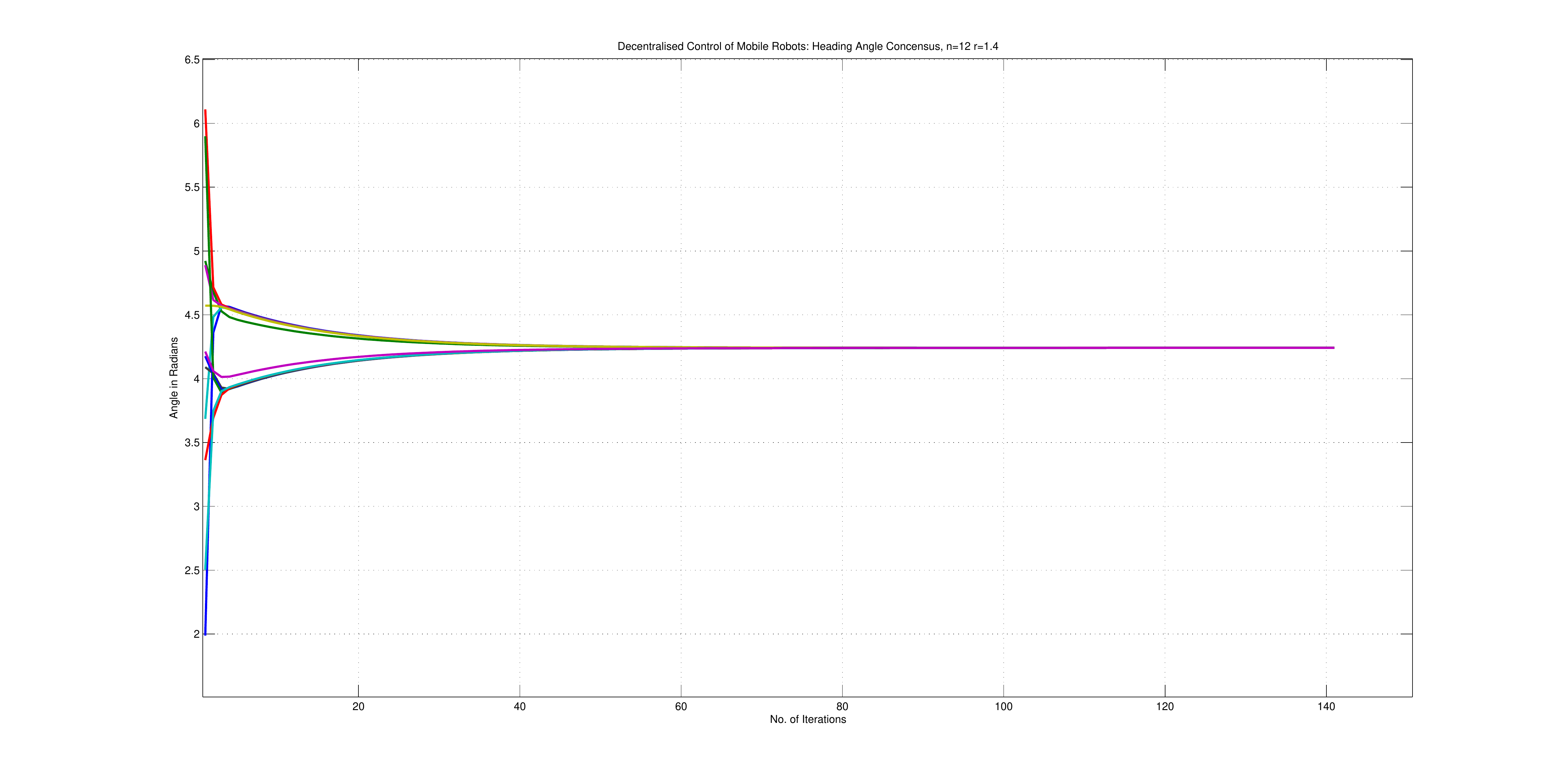}
  \caption{Simple Averaging Function - Mobile Robots Consensus}\label{SAF_Sim_Ex2}
\end{figure}

\begin{figure}[H]
  \centering
  \includegraphics[width=1.0\columnwidth]{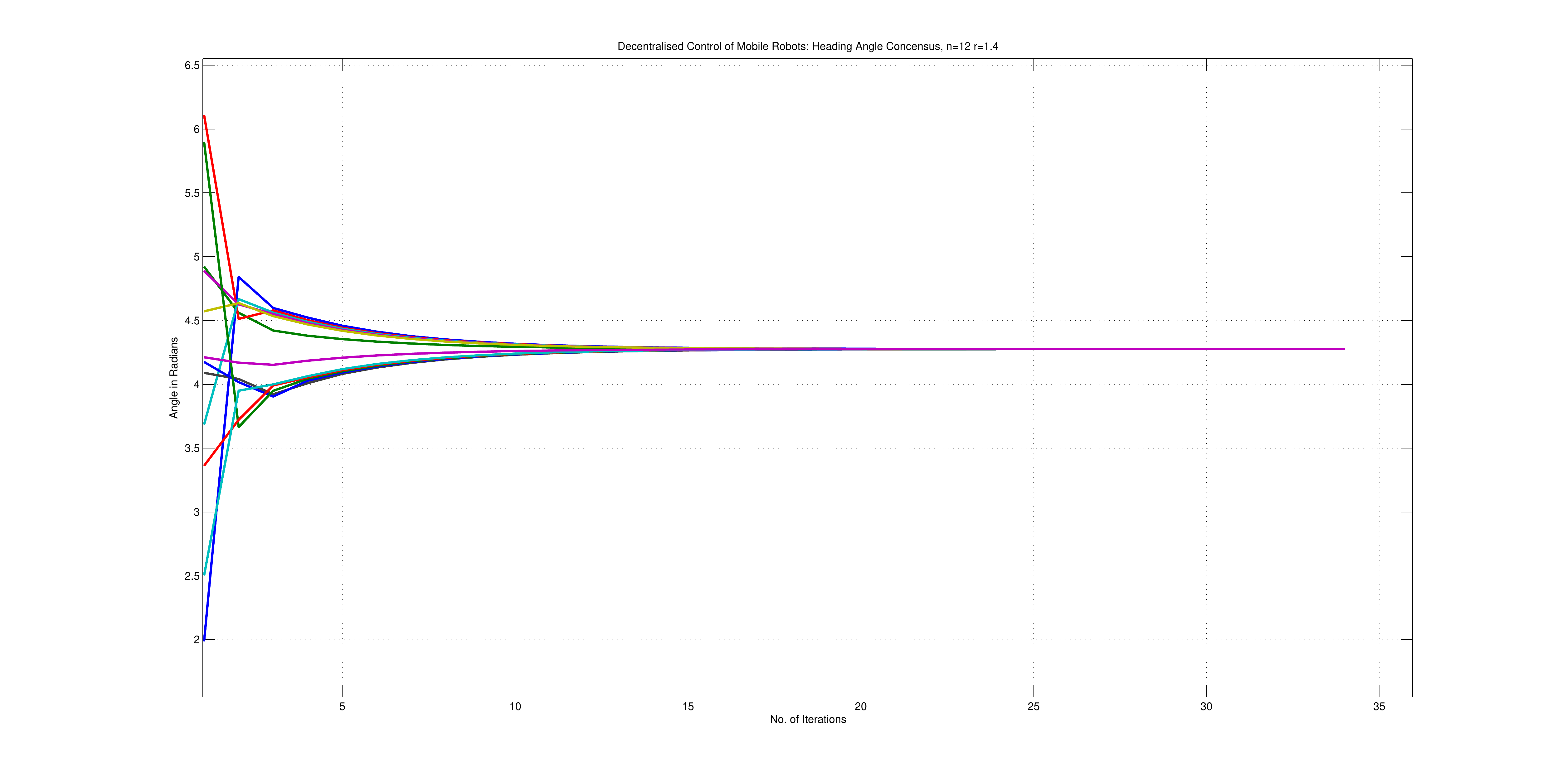}
  \caption{Weighted Averaging Function - Mobile Robots Consensus}\label{WAF_Sim_Ex2}
\end{figure}

Each mobile robot reaches a consensus value after a number of linear iterations recorded as under:\\
No. of Linear Iterations reaching heading consensus (Fig. \ref{SAF_Sim_Ex2}) with simple average function (\ref{eq:sim_jadbabie})=140\\
No. of Linear Iterations reaching heading consensus (Fig. \ref{WAF_Sim_Ex2}) with weighted average function (\ref{eq:sim_weighted_avg})=33\\
We can notice that the rule (\ref{eq:sim_weighted_avg}) with the similarity measure significantly reduces the number of linear iterations.

\textbf{Example 3}:

In this example, we again consider 12 mobile robots with the above mentioned initial vector $\phi(0)$. We consider mobile robots are based on the vertices of a complete graph represented by the adjacency matrix as under:

\[ \scriptscriptstyle
A  =  \left[
        \begin{array}{cccccccccccc}
         0&1&1&1&1&1&1&1&1&1&1&1  \\
         1&0&1&1&1&1&1&1&1&1&1&1  \\
         1&1&0&1&1&1&1&1&1&1&1&1 \\
         1&1&1&0&1&1&1&1&1&1&1&1 \\
         1&1&1&1&0&1&1&1&1&1&1&1  \\
         1&1&1&1&1&0&1&1&1&1&1&1  \\
         1&1&1&1&1&1&0&1&1&1&1&1  \\
         1&1&1&1&1&1&1&0&1&1&1&1  \\
         1&1&1&1&1&1&1&1&0&1&1&1  \\
         1&1&1&1&1&1&1&1&1&0&1&1  \\
         1&1&1&1&1&1&1&1&1&1&0&1  \\
         1&1&1&1&1&1&1&1&1&1&1&0  \\
        \end{array}
      \right]
\]
\

\begin{figure}[H]
  \centering
  \includegraphics[width=1.0\columnwidth]{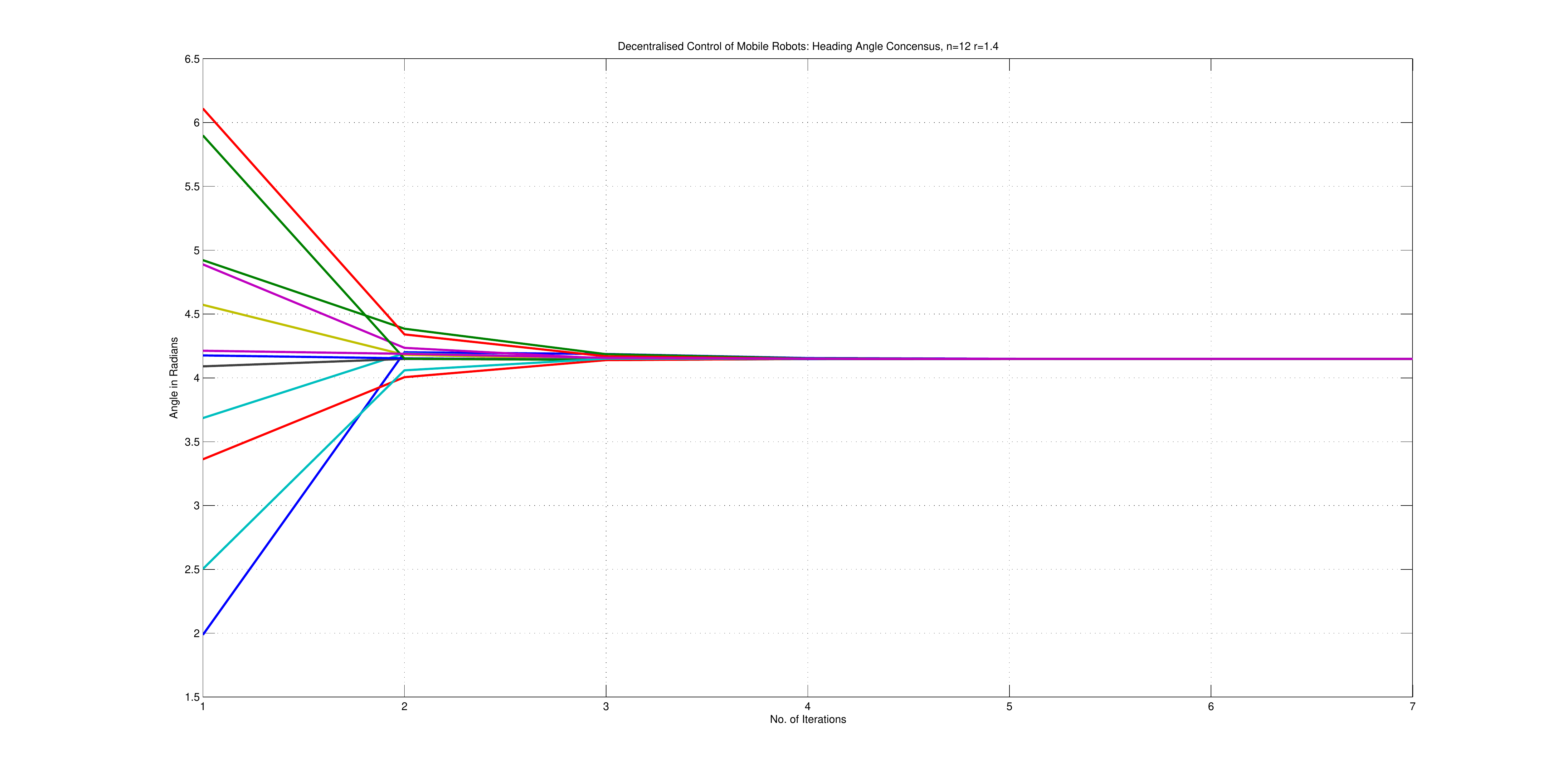}
  \caption{Simple Averaging Function - Mobile Robots Consensus}\label{SAF_Sim_Ex3}
\end{figure}

\begin{figure}[H]
  \centering
  \includegraphics[width=1.0\columnwidth]{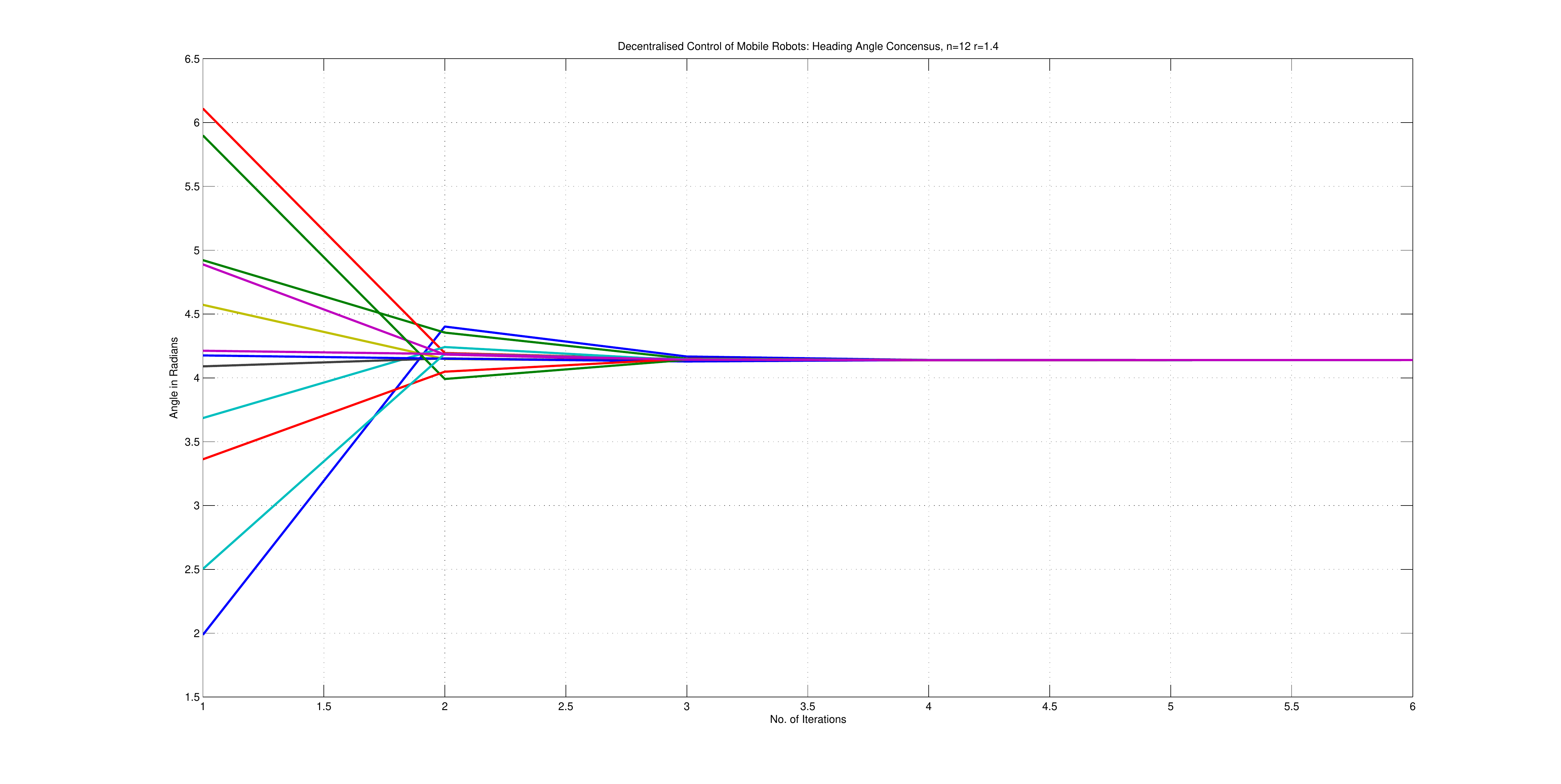}
  \caption{Weighted Averaging Function - Mobile Robots Consensus}\label{WAF_Sim_Ex3}
\end{figure}

Each mobile robot reaches a consensus value after a number of linear iterations recorded as under:\\
No. of Linear Iterations reaching heading consensus (Fig. \ref{SAF_Sim_Ex3}) with simple average function (\ref{eq:sim_jadbabie}) =06\\
No. of Linear Iterations reaching heading consensus (Fig. \ref{WAF_Sim_Ex3}) with weighted average function (\ref{eq:sim_weighted_avg}) =05\\
We can notice that the rule (\ref{eq:sim_weighted_avg}) with the similarity measure still outperforms in case of a complete graph.

\section{Chapter Conclusions and Future Research Directions}

\subsection{Chapter Conclusions}
The numerical simulations have clearly compared the number of linear iterations consumed by simple averaging function ($\ref{eq:sim_jadbabie}$) with that of weighted averaging function ($\ref{eq:sim_weighted_avg}$). Our approach brings a reduction in the number of linear iterations required to reach asymptotic consensus for almost sure.

\subsection{Future Research Directions}

 \begin{enumerate}
   \item There can be some consideration for increment in the size of information message (comprising of heading angle and rows of adjacency matrix) received from the nearest neighbour mobile robot.
   \item We can perform some spectral analysis showing how the similarity measure makes the algorithm fast.
   \item There can be some consideration for an optimisation model to maximize the convergence speed for this problem.
 \end{enumerate}

\chapter{Criteria of Achieving Consensus in Robotic Networks}
In the previous chapter, we addressed the decentralized control based strategy with the similarity measure. The decentralized control was based on the similarity measure calculating weight factors in the algorithm. In this chapter, we perform a rigorous mathematical analysis to address a necessary and sufficient condition in case of assigning these weighting factors to mobile robots. In fact, we establish and prove criteria for achieving consensus on heading angle, velocity, etc.

\section{Introduction}
We address a necessary and sufficient condition to guarantee consensus over randomly connected team of mobile robots having at least one self-weighting mobile robot. A necessary and sufficient condition for consensus over random networks considers a weight matrix with strictly positive diagonal entries over the state vector of mobile robots. We consider the associated non-negative weight matrix with at least one positive diagonal entry, and the condition is based on its spectral radius. Our considered sequence of non-negative weight matrices drives the state dynamics of connected mobile robots to a consensus value. Our mathematical arguments to verify the condition are based on the extended version of Perron-Forbenius theorem applicable to a class of non-negative, irreducible and aperiodic square matrices. Finally, we conclude that our easily verifiable condition does not necessarily require an irreducible weight matrix with all the main diagonal entries to be strictly positive. We also verify the validation of our condition with computer simulations performed for decentralized control of connected mobile robots placed with random initial heading angles.

%
%%%%%%%%%%%%%%%%%%%%%%%%%%%%%%%%%%%%%%%%%%%%%%%%%%%%%%%%%%%%%%%%%%%%%%%%%%%%%%%%%
%
\section{Related Work}

%A team of coordinating mobile robots \footnote{The terms, "sensor" or "agent" or "the mobile robotic" or "the mobile robotic sensor" will be used throughout this chapter for an autonomous mobile robot equipped with an on-board computation, operation-specific sensing and communication capability.} without any global information has different applications like surveillance of an area, reconnaissance, search and rescue operation in hazardous unknown areas, etc. This sort of team of mobile robots is adaptable with local information, and operates with decentralized control. The decentralized control relying on local information is inspired from flock of birds. Vicsek's model \cite{vicsek1995novel} already demonstrates how the decentralized coordination of flock of birds can reach the same direction (i.e. consensus in heading angle). It is believed that a team member is inspired based on the average of its own plus its neighbours \cite{vicsek1995novel}. Consequently the team of mobile robots behaves in a decentralized fashion. This demonstrates that the system must be adaptable based on the local information exchange, and there is a necessary requirement that the team of mobile robots should operate with a decentralized control strategy.

The recent research work \cite{savkin2012optimal,cheng2010decentralizedregion,cheng2009decentralized,cheng2009distributed,cheng2011decentralizedsweep,cheng2011decentralized,teimoori2010biologically,savkin2008decentralised,savkin2010decentralized} demonstrates decentralized control of mobile robots designed for different applications like Blanket, Barrier and Sweep Coverage problems. All the control laws \cite{savkin2012optimal,cheng2010decentralizedregion,cheng2009decentralized,cheng2009distributed,cheng2011decentralizedsweep,cheng2011decentralized,teimoori2010biologically,savkin2008decentralised,savkin2010decentralized} rely on local information obtained from the nearest neighbours of an agent. In these sorts of decentralized control applications, the mobile robots are normally required to reach a consensus in terms of heading angle, position or velocity. This chapter is based on developing a necessary and sufficient condition required for such sort of mobile robots to reach a consensus.

There has been a necessary and sufficient condition \cite{tahbaz2008necessary} to check consensus for discrete time linear dynamical system. This condition is based on the spectral radius of the weight matrix of the system. The decentralized control problems \cite{savkin2012optimal,cheng2010decentralizedregion,cheng2009decentralized,cheng2009distributed,cheng2011decentralizedsweep,cheng2011decentralized,teimoori2010biologically,savkin2008decentralised,savkin2010decentralized} can also be verified with a necessary and sufficient condition based on the spectral radius of the weight matrix. In these problems \cite{savkin2012optimal,cheng2010decentralizedregion,cheng2009decentralized,cheng2009distributed,cheng2011decentralizedsweep,cheng2011decentralized,teimoori2010biologically,savkin2008decentralised,savkin2010decentralized}, the team of mobile robots is normally connected. So, we develop a necessary and sufficient condition similar to \cite{tahbaz2008necessary}, but our system under consideration is based on a team of connected mobile robots with at least one self-weighting mobile robot. The connected team of mobile robots has been considered as a requirement for decentralized control of mobile robots \cite{savkin2012optimal,cheng2010decentralizedregion,cheng2009decentralized,cheng2009distributed,cheng2011decentralizedsweep,cheng2011decentralized,teimoori2010biologically,savkin2008decentralised,savkin2010decentralized}, and the associated weight matrix with non-negative entries is irreducible and row stochastic matrix. However, the condition in \cite{tahbaz2008necessary} considers a system having weight matrix with strictly positive diagonals.

Our necessary and sufficient condition is based on the spectral radius of the weight matrix and can be easily used to analyse the systems like \cite{savkin2012optimal,cheng2010decentralizedregion,cheng2009decentralized,cheng2009distributed,cheng2011decentralizedsweep,cheng2011decentralized,teimoori2010biologically,savkin2008decentralised,savkin2010decentralized}. Our condition is less strict than \cite{tahbaz2008necessary}, and it is easily applicable on decentralized control systems like \cite{savkin2012optimal,cheng2010decentralizedregion,cheng2009decentralized,cheng2009distributed,cheng2011decentralizedsweep,cheng2011decentralized,teimoori2010biologically,savkin2008decentralised,savkin2010decentralized}. In the upcoming sections, we formally define the objective of this chapter along with some necessary assumptions to be considered. Then, we write a step by step proof of the necessary and sufficient condition applicable to our system. We also present some computer simulation examples showing how randomly placed mobile reach asymptotic consensus for almost sure. At the end of this chapter, we conclude our system under consideration and the verifiable condition for such systems to reach a state consensus.
%
%\newpage

\section{Problem Statement}

We define average heading rule after following Vicsek's model \cite{vicsek1995novel}.

\begin{equation}\label{eq:cond_vicsek_model}
 {\Theta}_{i}(kT) :=\arctan\left(\dfrac{\sin(\phi_i(kT))+\sum_ {j\in {\cal N}_{i}(kT)}\sin(\phi_{j}(kT))}{\cos(\phi_i(kT))+\sum_ {j\in {\cal N}_{i}(kT)}\cos(\phi_{j}(kT))}\right)
\end{equation}
for $i=1,2,3,\ldots, n$, where ${\cal N}_i(kT)$ denotes the set of neighbors of the mobile robotic sensor $i$ at a particular time $kT$.

The above model (\ref{eq:cond_vicsek_model}) takes values in the continuous set $[0,2\pi),$ and it is based on the average of velocity vectors. The model (\ref{eq:cond_vicsek_model}) can behave in a cyclic manner under quantized heading angle and it has been shown in \cite{savkin2008decentralised}.

However, the following rule of \cite{jadbabaie2003coordination} is average of headings and it makes the analysis simple by applying the properties of stochastic matrices \cite{savkin2008decentralised}.

\begin{equation}\label{eq:cond_jadbabie}
 {\Theta}_{i}(kT) :=\dfrac{1}{|1+{\cal F}_{i}(kT)|}\left(\phi_i(kT)+\sum_ {j\in {\cal F}_{i}(kT)}\phi_{j}(kT)\right)
\end{equation}

However, the above model (\ref{eq:cond_jadbabie}) can exhibit counter-intuitive consequences under some circumstances as remarked in \cite{savkin2008decentralised}. The counter-intuitive consequences can be avoided by considering the domain of the defined function within the interval $[0,\pi)$ as mentioned in \cite{savkin2008decentralised}.
Our system under consideration is inspired from weighted average of agent's own state plus its neighbours.

\begin{equation}\label{eq:cond_weighted_avg}
 x_{i}(kT) :=w_{ii}x_i(kT)+\sum_ {j\in {\cal N}_{i}(kT)}w_{ij}x_{j}(kT)
\end{equation}
for $i=1,2,3,\ldots, n$ (assuming $n>2$). We have defined a weighting factor, $w_{ii}$, for the mobile robotic sensor itself and a weighting factor, $w_{ij}$, for its neighbour(s) at a particular time $(kT)$ such that the following condition is satisfied:
\begin{equation}
 w_{ii}+\sum_ {j\in {\cal F}_{i}(kT)}w_{ij}=1
\end{equation}

One can note that the rule (\ref{eq:cond_weighted_avg}) becomes exactly equivalent to (\ref{eq:cond_jadbabie}) under the following condition:

\begin{equation}\label{eq:jad_weighted_avg_eq}
 w_{ii}=w_{ij}=\dfrac{1}{|1+{\cal N}_{i}(kT)|}
\end{equation}

Next, a mobile robot updates its state from (\ref{eq:cond_weighted_avg}).

\begin{equation}\label{eq:phi_update}
 x_{i}(k+1)T):= x_{i}(kT)
\end{equation}

The vector form of the system (\ref{eq:cond_weighted_avg}) and (\ref{eq:phi_update}) can be represented as under:

\begin{equation}\label{eq:weighted_avg_vector}
  x((k+1)T) := W_{k}x(kT)
\end{equation}

where $k=0, 1, 2,\ldots$ is the discrete time index and $x((k+1)T)\in \mathbb{R}^n$ is the state vector at time $(k+1)T$.
\subsection{Definitions}
We provide some basic definitions similar to \cite{tahbaz2008necessary} as under:

\begin{enumerate}
  \item Directed Graph: The associated directed graph, $G(W)$, of an $nxn$ weight matrix "$W$" consists of $n$ mobile robots regarded as vertices of graph where an edge leads from robot "$i$" to robot "$j$" if and only if "$w_{ij}\neq0$", i.e. a robot "$i$" is connected to neighbouring robot "$j$" with some influential scalar value.

  \item Strongly Connected Mobile Robots: A directed graph "G" of a team of mobile robots is strongly connected if for any ordered pair of robots ($i$ and $j$) regarded as vertices of "$G$", there exists a sequence of edges (a communication path or information share link) from robot "$i$" to "$j$".
  \item Irreducible Weight Matrix: The weight matrix "$W$" is irreducible if and only if its associated graph, "$G(W)$", is strongly connected.
  \item Stochastic Matrix: A non-negative matrix $W=[w_{ij}], i,j=1,2,...,n$ is called as row stochastic if
  \begin{equation}
  \sum_{j=1}^{n}w_{ij}=1
  \end{equation} for each $i$.

  \item  Weakly Ergodic: Let $W_{k}$ be a chain of stochastic matrices. Then, $W_{k}$ is said to be weakly ergodic if $\lim_{k \to \infty} (W_{ip}(k:t_{0})-W_{jp}(k:t_{0})) = 0$ for any $t_{0} \geq 0$ and $i,j,l\in[m]$.

  \item Coefficient of Ergodicity: A scalar continuous function $\tau(.)$ on the set of $n \times n$ stochastic matrices and satisfying $0\leq \tau(.) \leq 1$ is said to be coefficient of ergodicity. The coefficient of ergodicity is said to be strictly proper if
  \begin{equation}
  \tau(W)=0 \text{ if and only if } W=\mathbf{1}v'
  \end{equation} for each $i$.
  Where $v$ is any probability vector $(v\geq0, v'\mathbf{1}=1)$.

\end{enumerate}

We state below proper coefficient of ergodicity \cite{tahbaz2008necessary}:

  \begin{equation}\label{eq:Erg_Prop}
 \kappa(W)=\frac{1}{2}\max_{i,j} \Sigma_{s=1}^{n}|W_{is}-W_{js}|
  \end{equation}

and improper coefficient of ergodicity \cite{tahbaz2008necessary} as under:

  \begin{equation}\label{eq:Erg_ImProp}
 \nu(W)=1-max_{j}(min_{i}W_{ij})
  \end{equation}
For a stochastic matrix $W$, (\ref{eq:Erg_Prop}) and (\ref{eq:Erg_ImProp}) satisfy \cite{tahbaz2008necessary}:

  \begin{equation}\label{eq:Erg_Coeffs_Relation}
 \kappa(W) \leq \nu(W)
  \end{equation}

\subsection{Assumptions}

\begin{enumerate}
  \item Each mobile robot is equipped with on-board computation capability.
  \item Each mobile robot has a wireless communication to neighbor(s) within a disk of radius $R_{c}>0$ for $t\in[kT, (k+1)T)$ and $k=0, 1, 2,\ldots$. The communication range, $R_{c}$, basically describes the edge of an undirected graph. We mathematically define disk of communication range $R_{c}$ as under:
  \begin{equation}
  C_{i,R_{c}}(kT):=p\in R^2:||p - s_{i}(kT)|| \leq R_{c}
  \end{equation}
  where $\|\cdot\|$ denotes the Euclidean norm.

  \item We consider each mobile robot as a vertex of an undirected graph. We assume that the graph is strongly connected, i.e. there exists an infinite sequence of contiguous, non-empty, bounded, time intervals $[k_{i},k_{i+1}]$  (where $i=0, 1, 2,\ldots,n$ and $k_{0}=0$) such that for all $[k_{i},k_{i+1}]$ the graph from the union of the collection $G(kT)\in{\cal P}$ for all $kT\in [k_{i},k_{i+1}]$ is connected.
\end{enumerate}

\subsection{Cyclic Behaviour of Weight Matrix}

The associated adjacency matrix of our weight matrix, $W_{k}$, is strongly connected. So, our $W_{k}$ is irreducible, but it is not necessarily aperiodic. Our irreducible $W_{k}$ matrix might have more than one unit modulus eigenvalues on the unit circle, and the system would behave in a cyclic manner. \\
In order to ensure a unique unit modulus eigenvalue of $W_{k}$, we can induce the property of aperiodicity. Like, we can introduce self-loop weight to all the vertices of $G(W_{k})$. Otherwise, if at least one of the vertex of $G(W_{k})$ has strictly positive weight (i.e. a self loop weight) then our irreducible $W_{k}$ is aperiodic \cite{meyer2000matrix}. For simplicity, we assume our $W_{k}$ is irreducible aperiodic stochastic matrix.
%%
%//////////////////////////////////Theorem 2////////////////////////////

\subsection{Theorem 1}
We state the theorem (\cite{tahbaz2008necessary}, Theorem 2) as under:

Suppose $\tau(\cdot)$ is proper coefficients of ergodicity and for any $r$ stochastic matrices $W_{i}$,$i = 1, . . . , r$ with each $r \geq 1$.

\begin{equation}\label{eq:FWD_PROD}
\tau(W_{r}...W_{2}W_{1}) \leq \Pi_{i=1}^{r} \tau(W_{i})
\end{equation}

Then, the sequence $\{{W_{i}}\}_{i=1}^\infty$ is weakly ergodic if and only if there exists a strictly increasing sequence of positive integers $k_{s},s=1,2,...$ such that

\begin{equation}\label{eq:WEAK_ERG}
\sum_{s=1}^{\infty} ({1-\tau(W_{k_{s}+1}...W_{k_{s}+1})}) = \infty
\end{equation}

Now, we state the main theorem for our considered $W_{k}$ and provide a step by step proof similar to (\cite{tahbaz2008necessary},Theorem 3).

%//////////////////////////////////THEOREM 4////////////////////////////////////////////////////////////////////////////////
\subsection{Theorem 2}
For a random independent and identically distributed (i.i.d.) sequence $\{ W_{k} \}_{k=0}^\infty = W_{1}, W_{2}, W_{3}, ...$ of irreducible stochastic matrices with at least one positive diagonal entry, the following three statements are equivalent.

\begin{enumerate}
 \item The random sequence $\{ W_{k} \}_{k=0}^ \infty $ is weakly ergodic for almost sure.
  \item The system (\ref{eq:weighted_avg_vector}) reaches asymptotic consensus.
  \item  The second largest eigen value modulus (S.L.E.M.) of $\mathbb{E} W_{k}$ is less than one, i.e. $|\lambda_{2}(\mathbb{E}W_{k})| < 1$.

\end{enumerate}
Proof: The below mentioned proof similar to \cite{tahbaz2008necessary} is for our considered $ W_{k}$ associated with the system (\ref{eq:weighted_avg_vector}).

First, we prove: (1) $\implies$ (2)

If the random sequence $W_{1},W_{2} , . . .$ is weakly ergodic with probability '1' then by definition:

$(U_{i,s}^{(k,p)}-U_{j,s}^{(k,p)} )\rightarrow 0 \text{ a.s.}$

If we follow the dominated convergence theorem \cite{durrett2013probability}, (2) can also be stated as under:

$(\mathbb{E}W)_{is}^{k}-(\mathbb{E}W)_{js}^{k} \rightarrow 0$.

Now, we prove: (2) $\implies$ (3). We prove our claim based on the contradiction. We assume $|\lambda_{2}(\mathbb{E}W_{k})| = 1$.

As our system under consideration has $ W_{k}$ with non-negative entries, $\mathbb{E}W_{k}$ will also have non-negative entries. Thus, it is not always primitive. However, $W_{k}$ is row stochastic, but it cannot always be irreducible. Thus, the standard version of Perron-Forbenius theorem does require $W_{k}$ with strictly positive diagonal entries. Our system under consideration is connected and the associated $W_{k}$ with non-negative entries is irreducible plus aperiodic stochastic matrix. Thus, the extended version of Perron-Forbenius Theorem \cite{berman1979nonnegative} for a class of non-negative and irreducible stochastic matrices applies and it states $|\lambda_{2}(\mathbb{E}W_{k})| < 1$ - which is in contradiction with our assumption because $|\lambda_{2}(\mathbb{E}W_{k})| = 1$ means $\mathbb{E}W_{k}$ is reducible. Therefore, the reducibility of $\mathbb{E}W_{k}$ implies the following block triangular form.

%\begin{equation}\label{eq:Tri_Form}
%\mathbb{E}W_{k} = [Q_11]
%\end{equation}

\begin{equation}\label{eq:Tri_Form}
%A_{m,n} =
\mathbb{E}W_{k}=
 \begin{bmatrix}
  Q_{11} & 0 & \cdots & 0 \\
  Q_{21} & Q_{22} & \cdots & 0\\
  \vdots  & \vdots  & \ddots & \vdots  \\
  Q_{s1} & Q_{s2} & \cdots & Q_{ss}
 \end{bmatrix}
\end{equation}

where each $Q_{ii}$ is an irreducible matrix and represents vertices in the equivalent class $\alpha_{i}$. As $|\lambda_{2}(\mathbb{E}W_{k})| = 1$, submatrices corresponding to at least two of the classes have unit spectral radii. As $Q_{ii}$ is irreducible and aperiodic, it does guarantee the multiplicity of the unit-modulus eigen value to be not more than 1.

Therefore, it implies (\cite{tahbaz2008necessary}, Lemma 2):
 \begin{equation}\label{eq:Ini_Classes}
\exists i \neq j \mid \text{$\alpha_{i}$ and $\alpha_{j}$ are both initial classes.}
 \end{equation}
Equivalently, $Q_{ir}=0 \forall r \neq i$ and $Q_{jl}=0$ $ \forall l \neq j$. It means, the matrix $\mathbb{E}W_{k}$ has two orthogonal rows and as a result (2) cannot hold.
Now, we prove the last implication by assuming (3) holds. Then, $|\lambda_{2}(\mathbb{E}W_{k})| < 1$ which means that $G(\mathbb{E}W_{k})$ has exactly one initial class. As we are considering strongly connected mobile robots, $\mathbb{E}W_{k}$ is irreducible. Our $\mathbb{E}W_{k}$ is also aperiodic (row) stochastic matrix. So,

 \begin{equation}\label{eq:EW_Non-negative}
\exists m \mid {[\mathbb{E}W_{k}]}^m > 0
 \end{equation}
 i.e. the matrix $\mathbb{E}W_{k}$ has strictly positive entries.

 The independence over time implies:

 \begin{equation}\label{eq:EW_Independence}
 \mathbb{E}(W_{m}...W_{2}W_{1})= {[\mathbb{E}W_{k}]}^m > 0
 \end{equation}
  Therefore, for all $i,j=1,...,n$ the $(i,j)$ entry of $U^{(m,0)}=W_{m}...W_{1}$ is positive with a probability, $p_{ij}>0$. As the irreducible aperiodic weight matrices are i.i.d. with non-negative entries, the matrix $W_{n^2m}...W_{2}W_{1}$ in completely entry-wise positive with at least probability $\Pi_{i,j} p_{ij}>0$. If we define $\delta(W)=1-\upsilon(W)=max_{j}(min_{i}W_{ij})$, there exists $\epsilon > 0$ such that\\
  $\mathbb{P}(\delta(W_{n^2m}...W_{2}W_{1})> \epsilon) > 0$.\\
  Hence, the second Borel-Cantelli lemma (\cite{durrett2013probability}, Theorem 2.3.6 on p.64) states: If the events $W_{n}$ are independent then $ \sum   P(W_{n}) = \infty$ implies $P (W_{n} i.o.) = 1$. Similarly, when $r \rightarrow \infty$,

  $\mathbb{P}(\delta(W_{(r+1)n^2m}...W_{rn^2m+1})> \epsilon )=1$.

  If we put $k_{r}=rn^2m$ then

  $\delta(W_{k_{(r+1)}}...W_{k_{(r+1)}})>\epsilon \text{ } i.o. \text{ } a.s.$

If $\tau(.)=\kappa(.)$ then $\eqref{eq:FWD_PROD}$ still holds. This together with (\ref{eq:Erg_Coeffs_Relation}) implies:

\begin{equation}\label{eq:WEAK_ERG_kappa}
\sum_{r=1}^{\infty}( {1-\kappa(W_{k_{r+1}}...W_{k_{r+1}})}) = \infty \text { a.s. }
\end{equation}
 The above condition is similar to the sufficient condition for weak ergodicity (\ref{eq:WEAK_ERG}). Therefore, the sequence is weakly ergodic for sure.

Now, we present the below mentioned corollary similar to (\cite{tahbaz2008necessary}, Corollary 4).

 \subsection{Corollary}
Our considered linear dynamical system (\ref{eq:weighted_avg_vector}) having irreducible stochastic matrix $W_{k}$ with at least one strictly positive diagonal entry reaches consensus almost surely if and only if $|\lambda_{2}(\mathbb{E}W_{k})|<1$.

Proof: We have proved in our last theorem that the second largest eigenvalue ($\lambda_{2}$) less than 1 guarantees weak ergodicity with probability '1', and our system (\ref{eq:weighted_avg_vector}) reaches asymptotic consensus with probability '1'.

Conversely, if $\mathbb{E}W_{k}$ has more than unit modulus eigenvalue then $G(\mathbb{E}W_{k})$ has more than one initial class. $G(\mathbb{E}W_{k})$ with more than one initial class implies that $\mathbb{E}W_{k}$ has two orthogonal rows. As $\Omega_{0}$ is a subset of non-negative matrices, $W_{k}$ has the same type (zero block pattern) as $\mathbb{E}W_{k}$ for all discrete time index $k$ with probability 1. Therefore, $U^{k,0}=W_{k}...W_{2}W_{1}$ has two orthogonal rows almost surely for any $k$. So, our system (\ref{eq:weighted_avg_vector}) reaches a consensus with probability 0. Therefore, the second largest eigen value modulus of $\mathbb{E}W_{k}$ must be less than 1. This proves the necessary and sufficient condition for almost sure asymptotic consensus in our system (\ref{eq:weighted_avg_vector}).
We also validate our condition by the computer simulations. We present the results of a few examples under the following section.
%\addtolength{\textheight}{-3cm}   % This command serves to balance the column lengths
%                                  % on the last page of the document manually. It shortens
%                                  % the textheight of the last page by a suitable amount.
%                                  % This command does not take effect until the next page
%                                  % so it should come on the page before the last. Make
%                                  % sure that you do not shorten the textheight too much.

\section{Computer Simulations}

We run these simulations on MATLAB R2015. The mobile robotic sensor is represented with a green circle and its heading angle with a green arrow. We use a rounding function (up to 4 decimal places) for comparative statements. In the below mentioned simulation examples, we have used the following parameters:\\
$R_{c}=1.4, v=0.8$.

\begin{figure}[H]
  \centering
  \includegraphics[width=1.0\columnwidth]{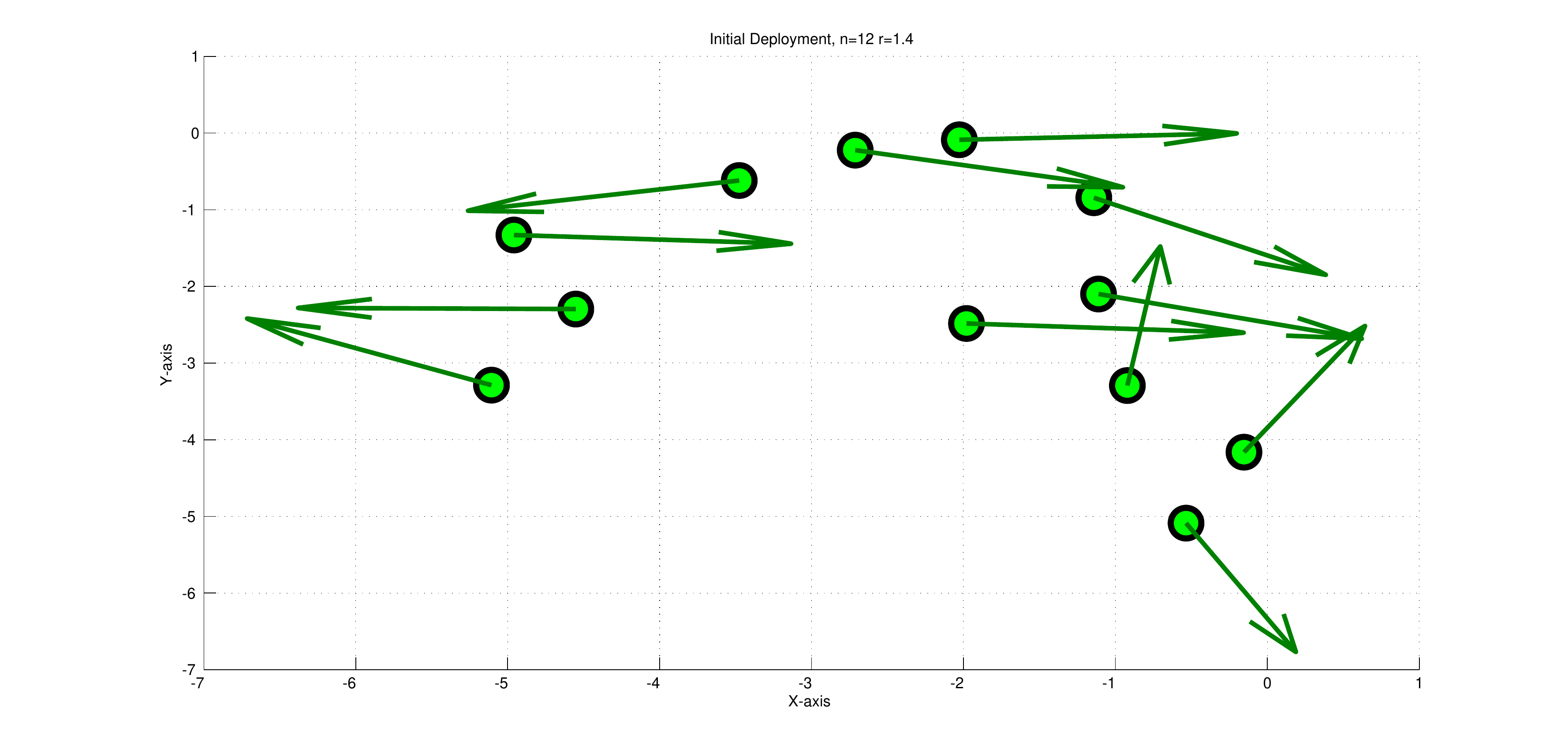}
  \caption{Example 1 - Initial Random Deployment}\label{Ex1_Cond_Initial}
\end{figure}

\begin{figure}[H]
  \centering
  \includegraphics[width=1.0\columnwidth]{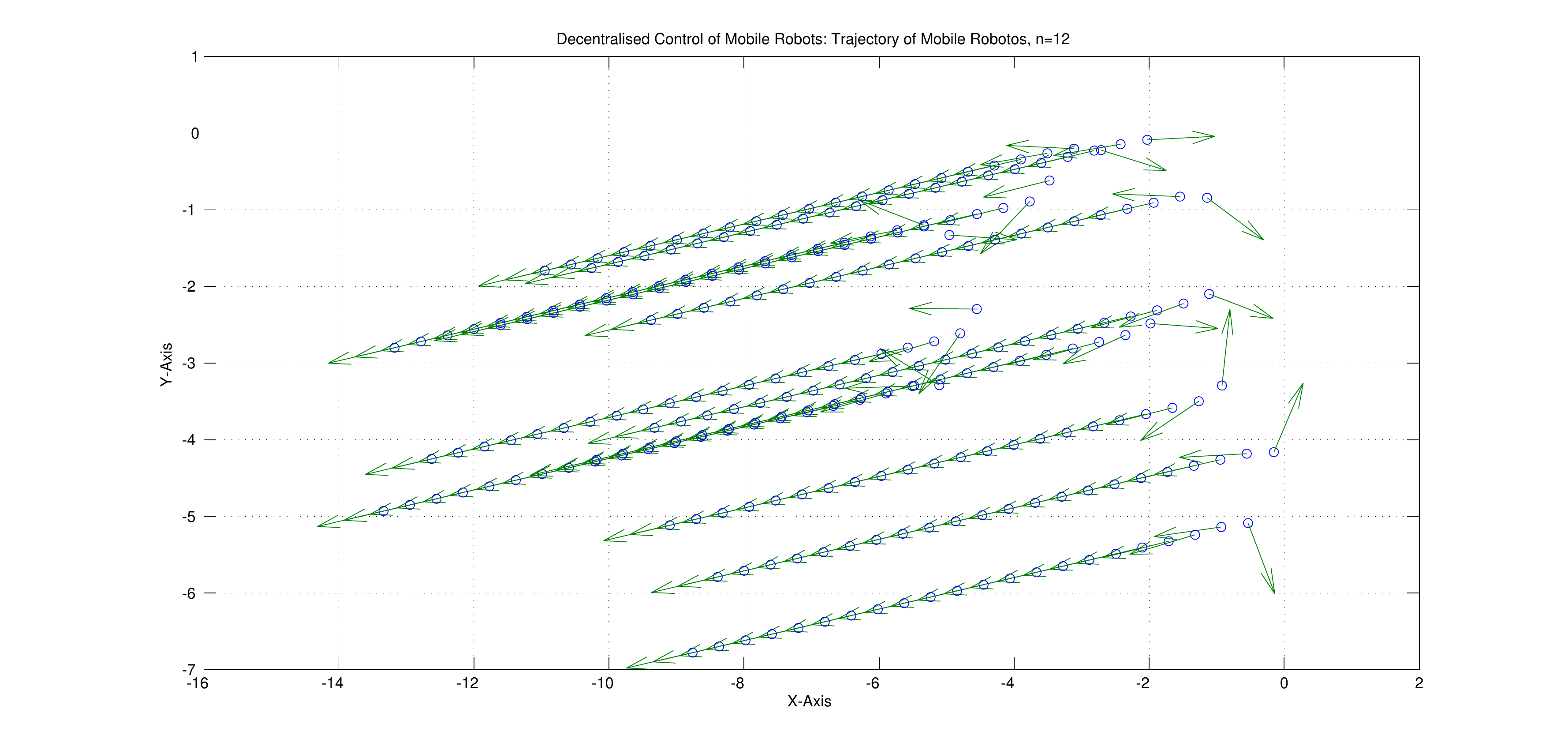}
  \caption{Example 1 - Mobile Robots Trajectories}\label{Ex1_Cond_Trajectories}
\end{figure}

\begin{figure}[H]
  \centering
  \includegraphics[width=1.0\columnwidth]{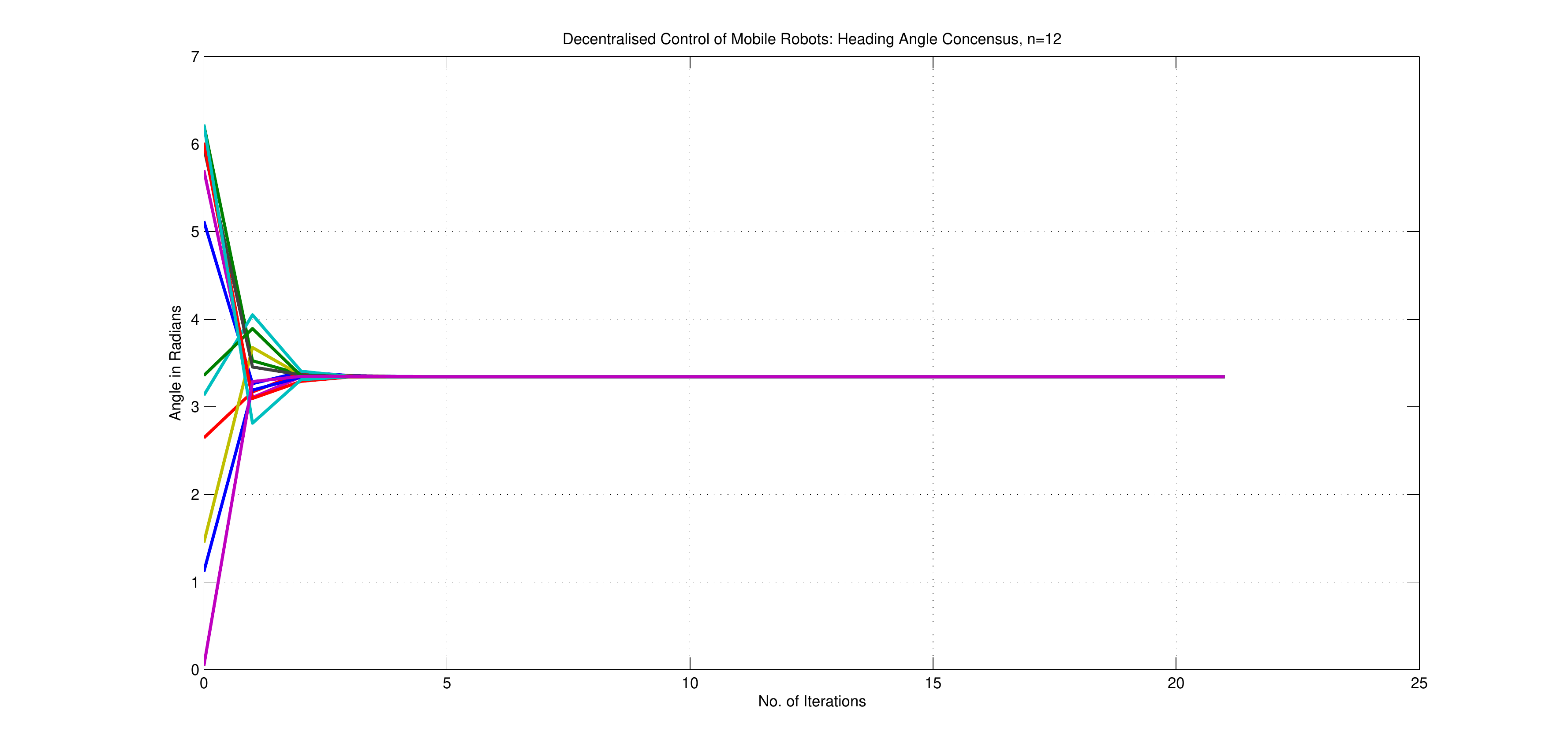}
  \caption{Example 1 - Heading Angle Consensus}\label{Ex1_Cond_Convergence}
\end{figure}

\begin{figure}[H]
  \centering
  \includegraphics[width=1.0\columnwidth]{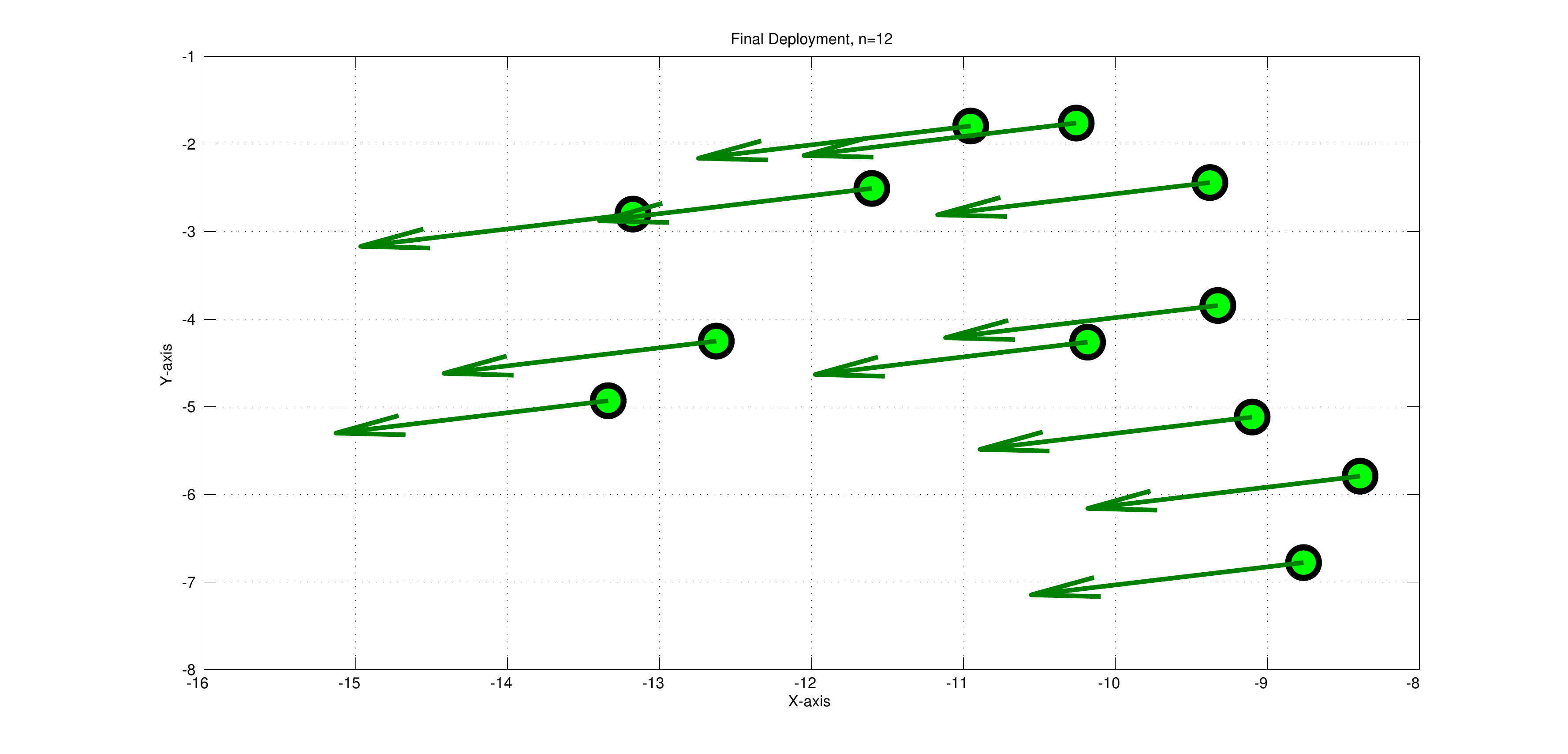}
  \caption{Example 1 - Final Deployment}\label{Ex1_Cond_Final}
\end{figure}

\begin{figure}[H]
  \centering
  \includegraphics[width=1.0\columnwidth]{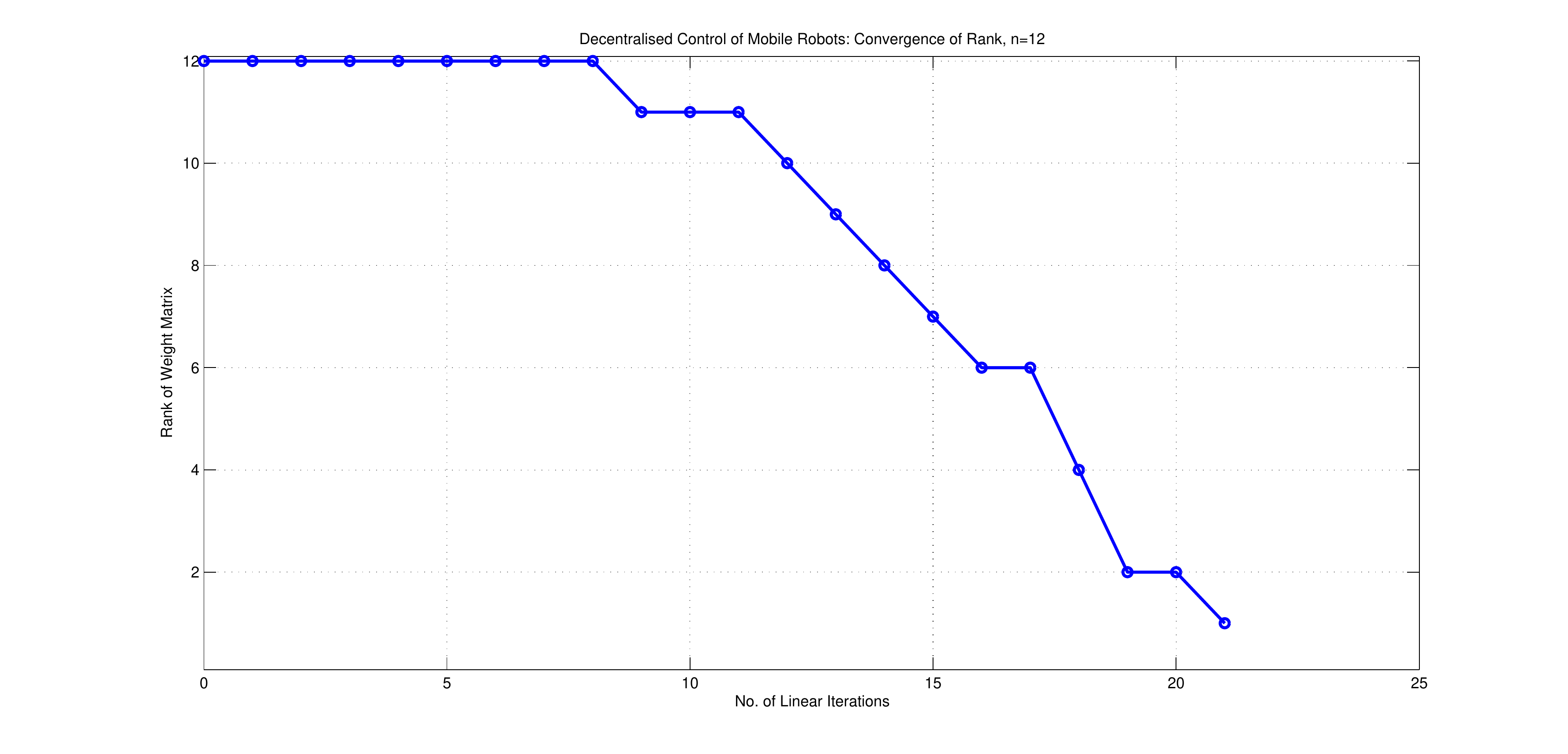}
  \caption{Example 1 - Weight Matrices Rank Convergence to '1'}\label{Ex1_Cond_Rank}
\end{figure}

\begin{figure}[H]
  \centering
  \includegraphics[width=1.0\columnwidth]{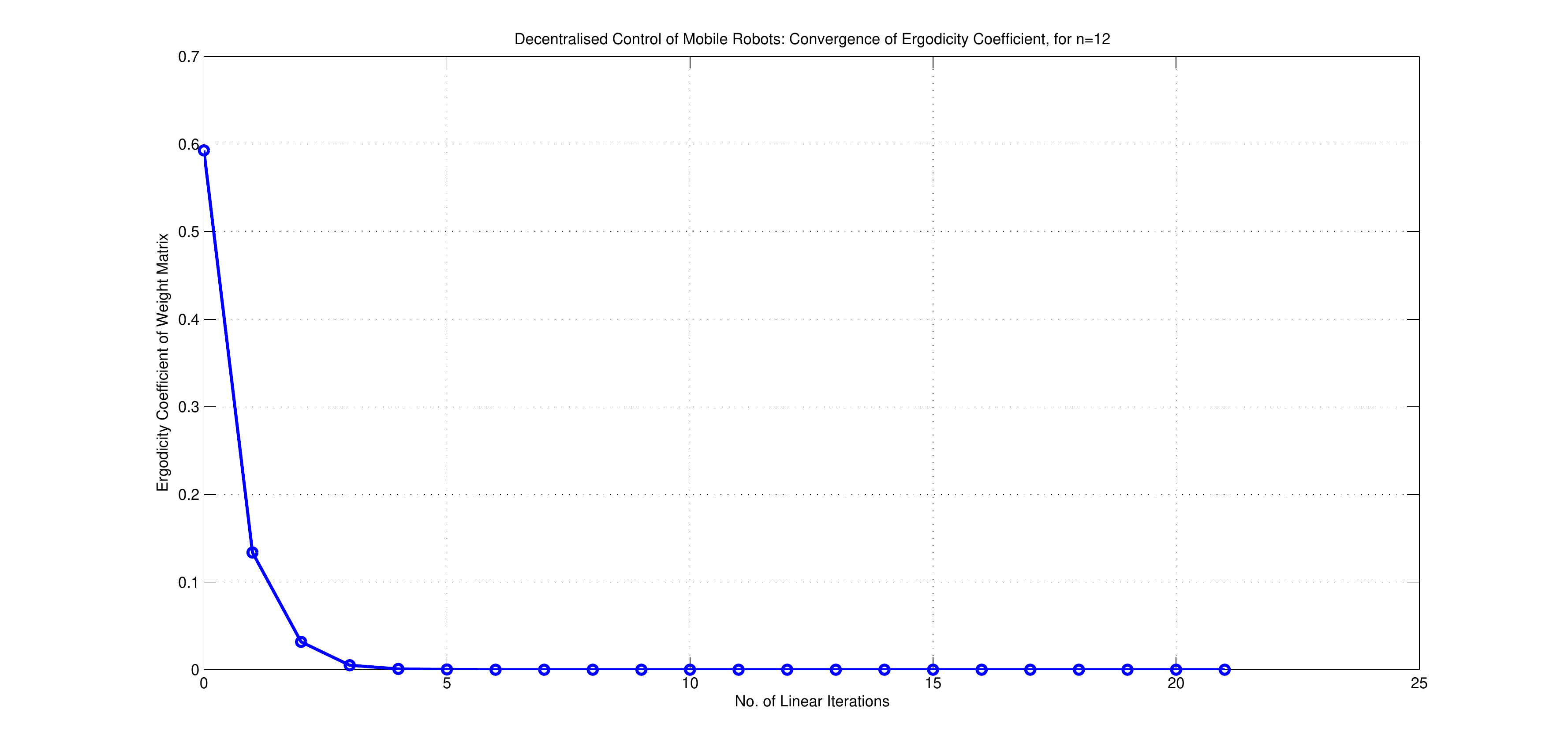}
  \caption{Example 1 - Weight Matrices Ergodicity Coefficient Convergence to '0'}\label{Ex1_Cond_Ergod}
\end{figure}

Example 1: In this example, we consider 12 (Fig. \ref{Ex1_Cond_Initial}) mobile robots with randomly generated initial vector $\phi(0)$. In Fig. \ref{Ex1_Cond_Trajectories}, we show the trajectory of each mobile robot being drawn with the dynamics from Chapter 5 (\ref{eq:sim_sys1}-\ref{eq:sim_sys2}). Fig. \ref{Ex1_Cond_Convergence}-\ref{Ex1_Cond_Final} show the consensus achieved in terms of the heading angle. Finally, Fig. \ref{Ex1_Cond_Rank}-\ref{Ex1_Cond_Ergod} draws $W_{k}$ rank convergence to '1' and ergodicity coefficient convergence to '0', respectively.

\begin{figure}[H]
  \centering
  \includegraphics[width=1.0\columnwidth]{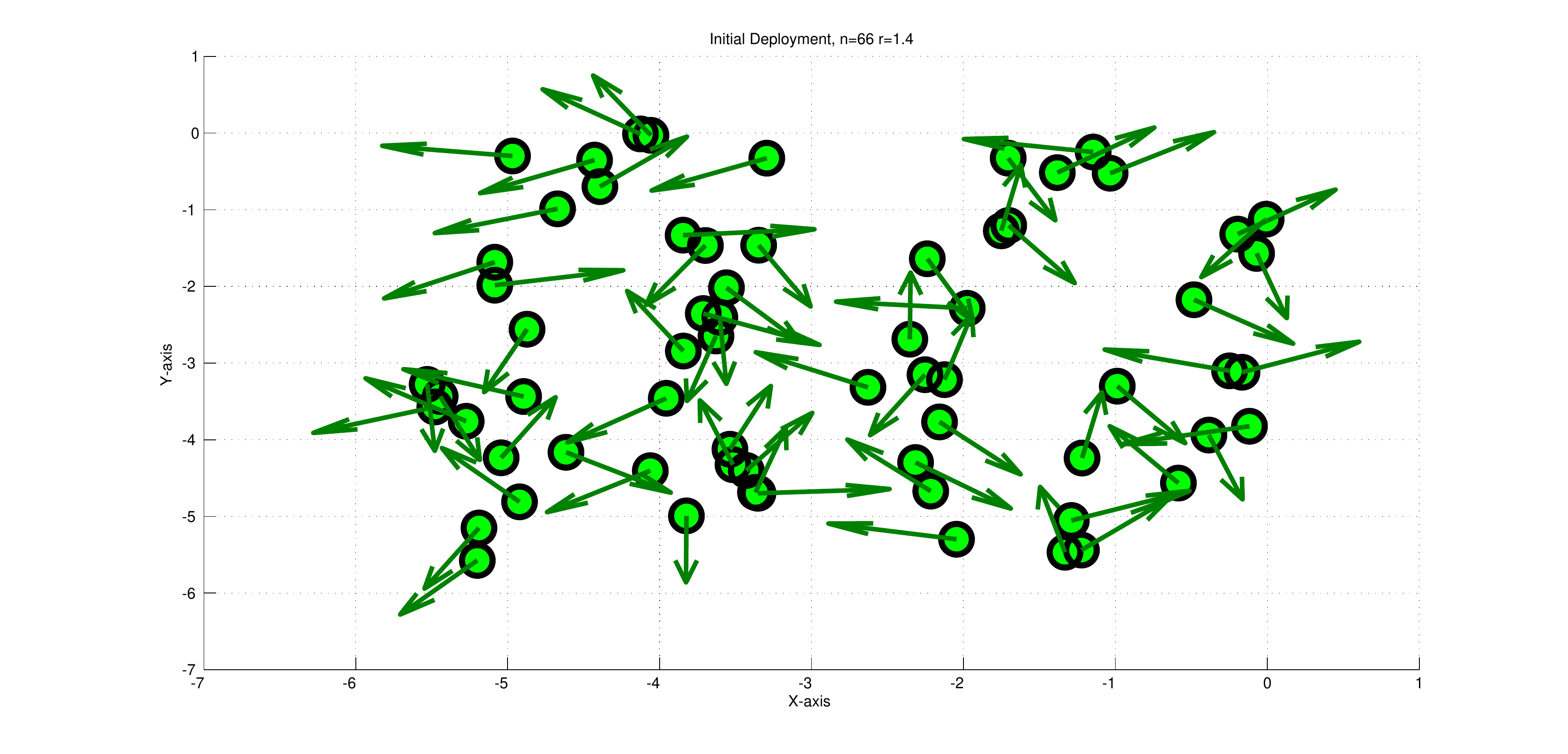}
  \caption{Example 2 - Initial Random Deployment}\label{Ex2_Cond_Initial}
\end{figure}

\begin{figure}[H]
  \centering
  \includegraphics[width=1.0\columnwidth]{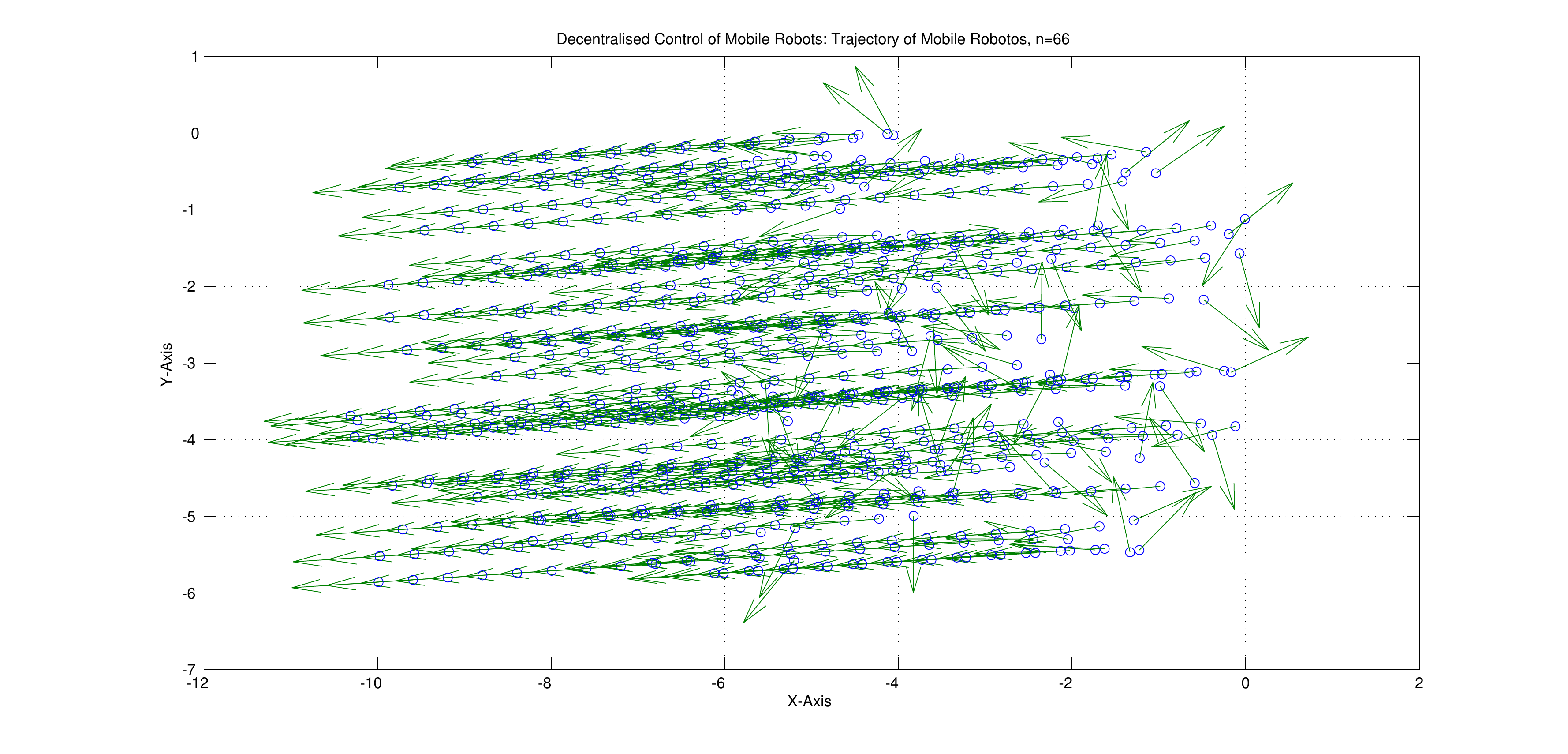}
  \caption{Example 2 - Mobile Robots Trajectories}\label{Ex2_Cond_Trajectories}
\end{figure}

\begin{figure}[H]
  \centering
  \includegraphics[width=1.0\columnwidth]{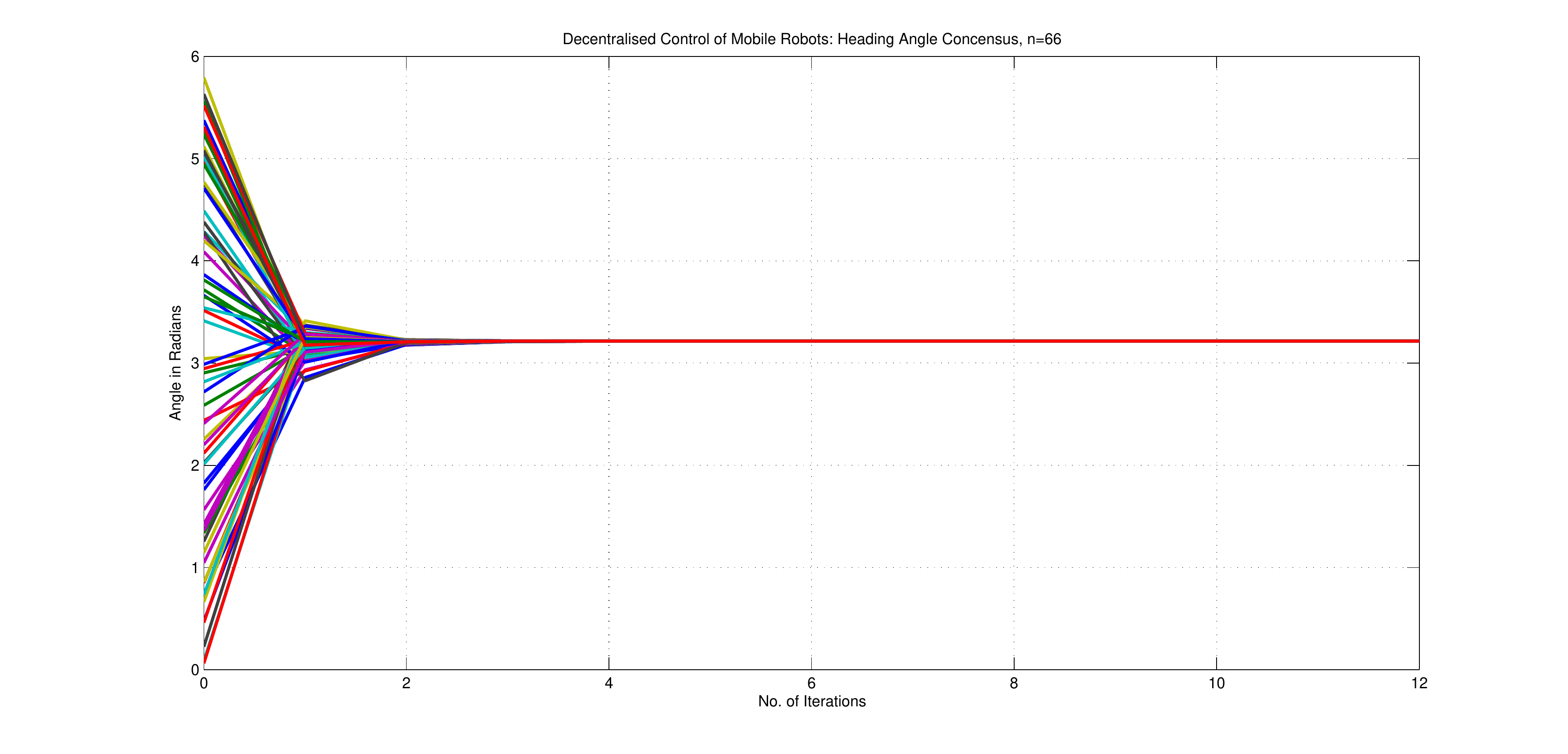}
  \caption{Example 2 - Heading Angle Consensus}\label{Ex2_Cond_Convergence}
\end{figure}

\begin{figure}[H]
  \centering
  \includegraphics[width=1.0\columnwidth]{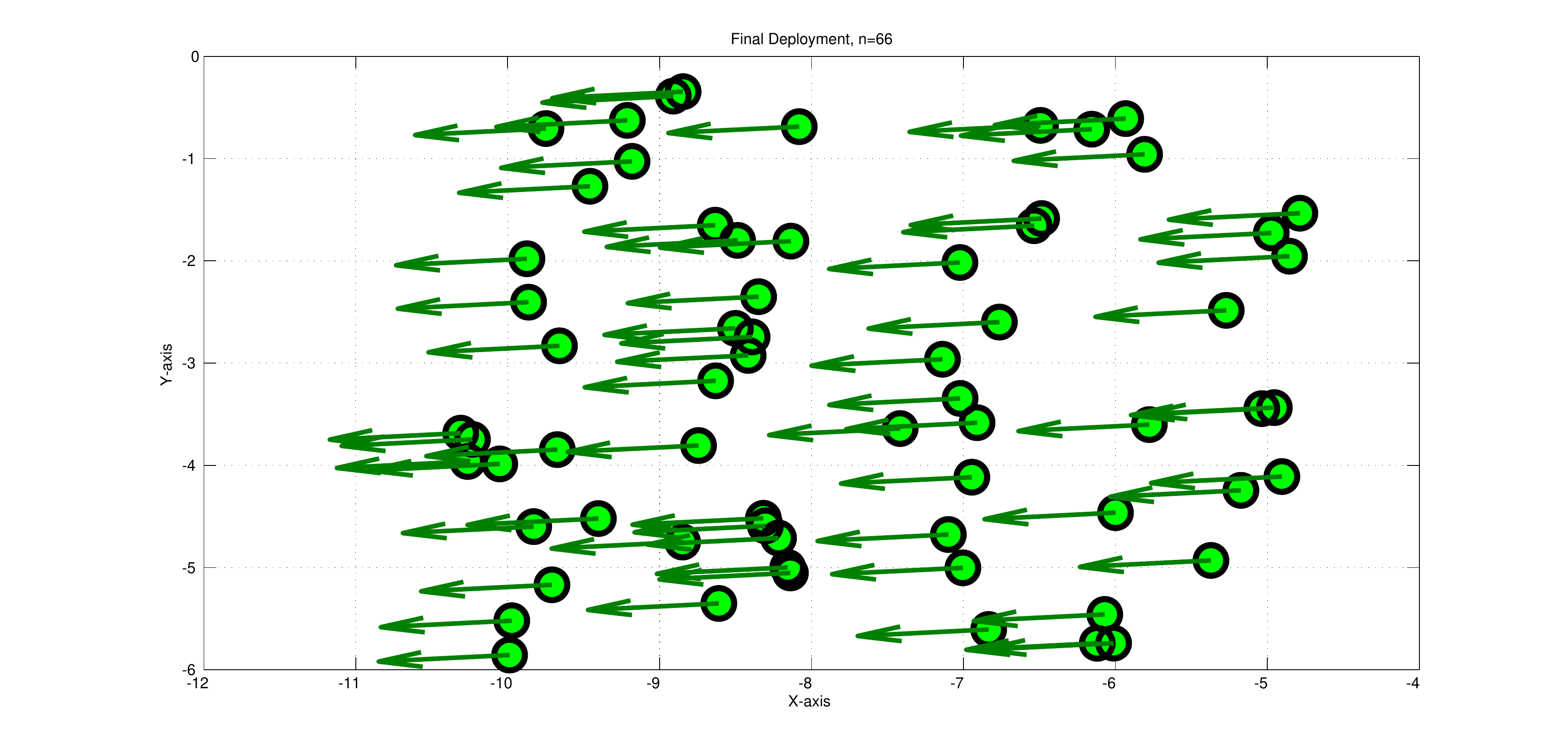}
  \caption{Example 2 - Final Deployment}\label{Ex2_Cond_Final}
\end{figure}

Example 2:
%Fig. 2 shows a simulation study with a weighted averaging function of nearest neighbour rule ($\ref{eq:cond_weighted_avg}$).

In Fig. \ref{Ex2_Cond_Initial}, 66 mobile robots with random initial heading angles are deployed. We can notice in Fig. \ref{Ex2_Cond_Convergence}-\ref{Ex2_Cond_Final} that each mobile robot converges to a constant heading angle. In Fig. \ref{Ex2_Cond_Trajectories}, we show the trajectory of each mobile robot being drawn with the dynamics from Chapter 5 (\ref{eq:sim_sys1}-\ref{eq:sim_sys2}).

%No. of Linear Iterations with simple average function (\ref{eq:cond_jadbabie}) =06\\
%No. of Linear Iterations with weighted average function (\ref{eq:cond_weighted_avg}) =05

\begin{figure}[H]
  \centering
  \includegraphics[width=1.0\columnwidth]{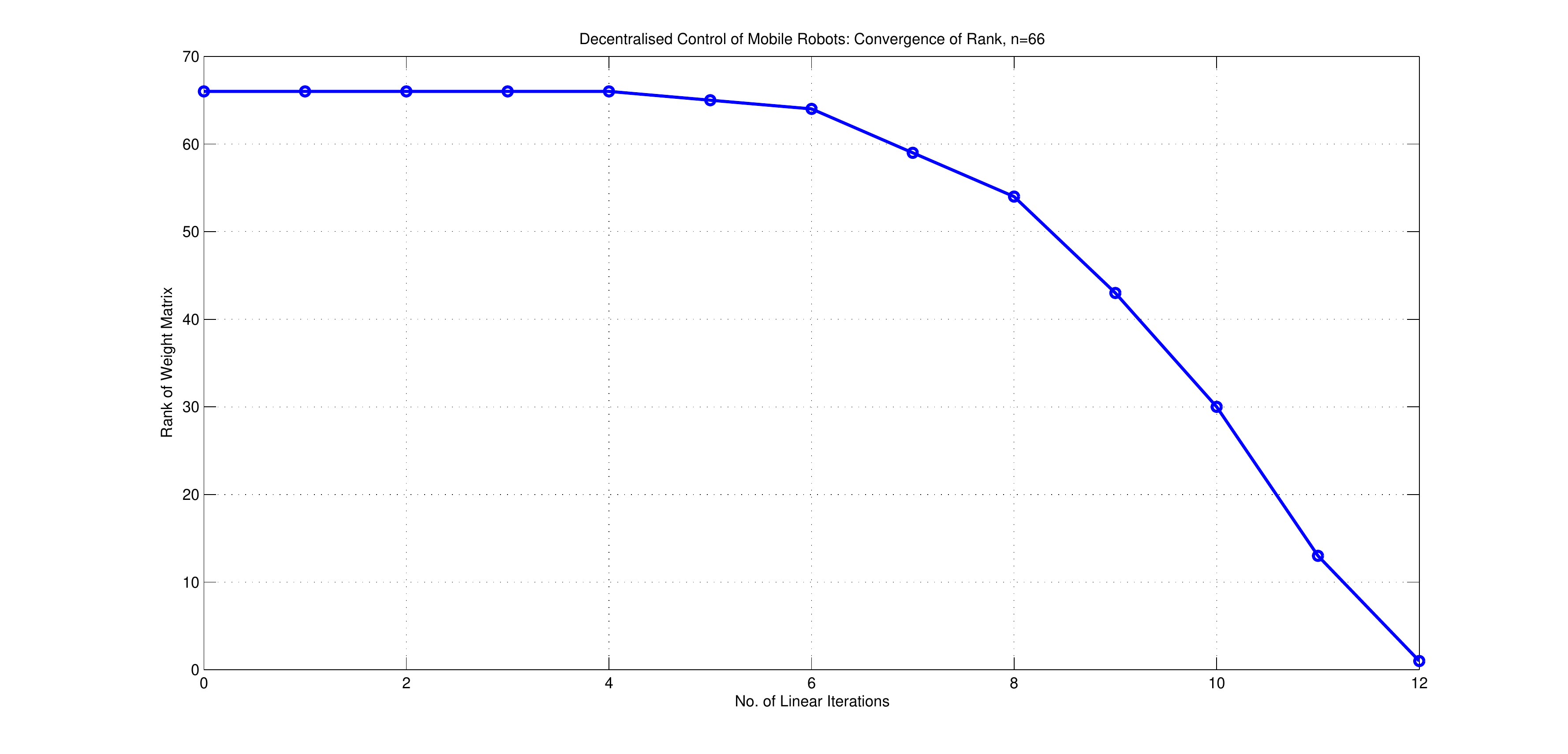}
  \caption{Example 2 - Weight Matrices - Rank Convergence to '1'}\label{Ex2_Cond_Rank}
\end{figure}

\begin{figure}[H]
  \centering
  \includegraphics[width=1.0\columnwidth]{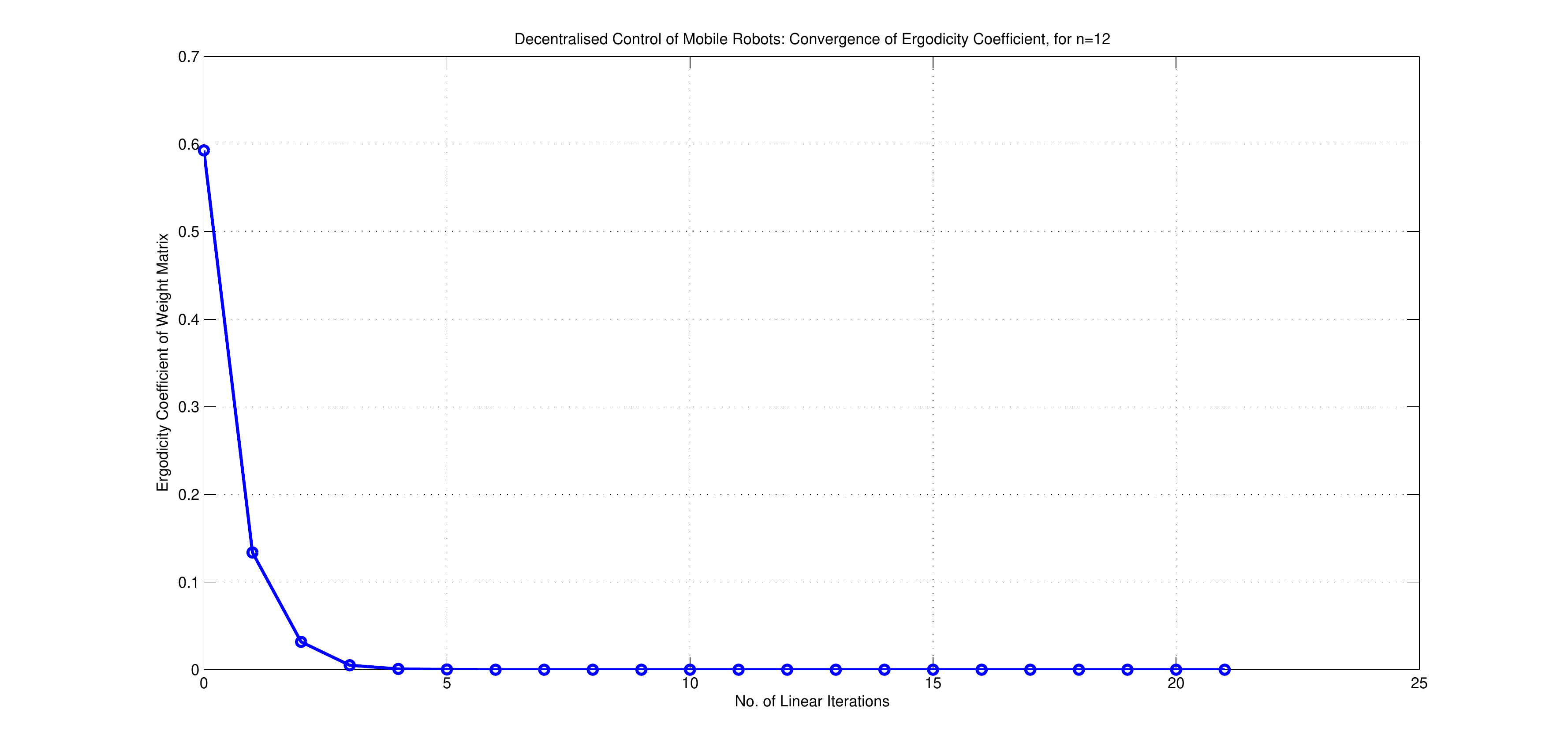}
  \caption{Example 2 - Weight Matrices - Ergodicity Coefficient Convergence to '0'}\label{Ex2_Cond_Ergod}
\end{figure}

\section{Chapter Conclusions and Future Research Directions}

\subsection{Chapter Conclusions}
We have considered a linear dynamical system as under:\\
$x(k)=W_{k}x(k-1)$, where $W_{k}$ is a random irreducible and aperiodic stochastic matrix. Our considered system does not necessarily require $W_{k}$ to be strictly diagonal-positive. In fact, our considered $W_{k}$ can have non-negative diagonal with at least one positive diagonal entry. We have shown the problem of reaching state consensus can still be considered as the problem of weak ergodicity of a sequence of considered weight matrices. We have shown that our considered system reaches a state consensus if and only if $(\mathbb{E}W_{k})$ has a unique eigen value with unit modulus.
We have also performed some computer simulations to validate our mathematical arguments, and we again conclude the validity of necessary and sufficient condition to reach state consensus over strongly connected mobile robots. This relaxed condition could be used to design the aperiodic weight matrices for strongly connected mobile robots. The weighted averaging algorithm is superior to simple averaging algorithms, and we have seen its effectiveness in chapter 5.

\subsection{Future Research Directions}
This condition can be useful for validating decentralized control laws for strongly connected mobile robots. In the future, this condition can be used along with an optimization model for validating decentralized control laws.

\chapter{Conclusions and Future Research Directions}

\subsection{Conclusions}

We justify the decentralized coverage control algorithms inspired from animal aggregation. Our broad survey on decentralized control of mobile robotic sensors for coverage problems provides a literature classification, comparison of different control approaches achieving coverage goals, considerations in the literature and giving direction to formulate future research problems. However, we recommend a future literature review on quantitative grounds.

We make some of the individual mobile robots influential than others. In the existing research on coverage control algorithms, there is always an equal importance of every individual mobile robot to achieve a group level coverage. We consider nearest neighbour rule with weighted average functions and test this approach on the coverage algorithm. The group level algorithm adapts quickly according to local change. In Chapter 3, we have specifically addressed the nearest neighbour rule with weighted average functions and showed how it helps in achieving a smooth manoeuvering.

In Chapter 4, we have addressed a cyclic behaviour arising from the quantized control of mobile robots. Our approach has shown how this cyclic behaviour can be avoided in the decentralized coverage control algorithm. Our quantized control approach is based on a realistic consideration for the team of mobile robots to achieve consensus in the heading angle.

We investigate the consideration of a similarity measure at the individual level of a mobile robot. We prove how a consideration of the similarity measure brings a significant reduction in the number of linear iterations to achieve state consensus for coverage control applications. Our nearest neighbour rules with weighted averaging functions have shown superiority in terms of developing fast algorithms. Finally, we also develop relax criteria to validate the control consensus on heading angle, velocity, etc.

 \subsection{Future Research Directions}

 The communication channels among the multi-robot network has got some limitations. The coverage control approaches mentioned in this research might be combined with the tools for control and estimation via limited capacity communication channels developed in (see e.g. \cite{matveev2003problem,matveev2004problem,matveev2005multirate,matveev2007analogue,savkin2007detectability,matveev2009estimation}).

Another future direction might be to apply the coverage approaches to more complex models including state dynamics. A good example is to combine the decentralized coverage control algorithms with the methods of robust nonlinear control and robust state estimation (see e.g. \cite{krstic1995nonlinear,savkin1994connection,savkin1995minimax,savkin1996robust,savkin2000robust}). Another good example is to combine these coverage control algorithms with the hybrid dynamical system approach (see e.g. \cite{savkin1996hybrid,skafidas1999stability,matveev2012qualitative,savkin2002hybrid}).

The future research in these problems could be conducted by considering an obstacle avoidance strategy (see e.g. \cite{savkin2015safe,hoy2014improved,matveev2011method,matveev2012real,savkin2013simple,savkin2014seeking,matveev2015globally}) in the coverage control algorithms.

\bibliographystyle{IEEEtran}
\bibliography{BReference}

\end{document}